\documentclass[acmsmall]{acmart}

\usepackage{array}
\usepackage[longtable]{multirow}
\usepackage{ragged2e}
\usepackage{colortbl}
\usepackage{longtable}
\usepackage{hhline}
\usepackage{subcaption}
\usepackage{wrapfig}
\AtBeginDocument{%
  \providecommand\BibTeX{{%
    \normalfont B\kern-0.5em{\scshape i\kern-0.25em b}\kern-0.8em\TeX}}}

\newcommand{\etal}{\textit{et al.}}
\newcommand{\ie}{\textit{i.e.,}}
\newcommand{\eg}{\textit{e.g.,}}
\newcommand{\aka}{\textit{a.k.a.,}}
\newcommand{\etc}{\textit{etc.,}}

\begin{document}

    \title{A Survey and Perspective on Artificial Intelligence for Security-Aware Electronic Design Automation}

\author{David Selasi Koblah}
\authornote{Both authors contributed equally to this research.}
\email{dkoblah@ufl.edu}
\orcid{1234-5678-9012}
\author{Rabin Yu Acharya}
\authornotemark[1]
\email{rabin.acharya@ufl.edu}
\author{Daniel Capecci}
\author{Olivia P. Dizon-Paradis}
\affiliation{%
  \institution{Florida Institute for Cybersecurity (FICS) Research, University of Florida}
  \city{Gainesville}
  \state{Florida}
  \country{USA}
  \postcode{32611}
}



\author{Shahin Tajik}
\author{Fatemeh Ganji}
\affiliation{%
 \institution{Worcester Polytechnic Institute}
 \city{Worcester}
 \state{Massachusetts}
 \country{USA}
  \postcode{01609}}


\author{Damon L. Woodard}
\author{Domenic Forte}
\affiliation{%
  \institution{Florida Institute for Cybersecurity (FICS) Research, University of Florida}
  \city{Gainesville}
  \state{Florida}
  \country{USA}
  \postcode{32611}}

\renewcommand{\shortauthors}{Koblah and Acharya, et al.}

\begin{abstract}
Artificial intelligence (AI) and machine learning (ML) techniques have been increasingly used in several fields to improve performance and the level of automation. In recent years, this use has exponentially increased due to the advancement of high-performance computing and the ever increasing size of data. One of such fields is that of hardware design; specifically the design of digital and analog integrated circuits~(ICs), where AI/ ML techniques have been extensively used to address ever-increasing design complexity, aggressive time-to-market, and the growing number of ubiquitous interconnected devices (IoT). However, the security concerns and issues related to IC design have been highly overlooked. In this paper, we summarize the state-of-the-art in AL/ML for circuit design/optimization, security and engineering challenges, research in security-aware CAD/EDA, and future research directions and needs for using AI/ML for security-aware circuit design.. 
\end{abstract}



\maketitle

\section{Introduction}\label{sec:introduction}
With the proliferation of IoT devices, and the ever-increasing design complexity of electronic systems, artificial intelligence (AI) and machine learning (ML) techniques have been increasingly used to optimize electronic design automation~(EDA) frameworks and accelerate the overall process. The physical platform, or hardware, represents the first stage for any layered security approach, and provides the initial protection mechanisms to help ensure that preliminary security controls can be trusted. Hence, it is considered to be the "root-of-trust" of any electronic system. However, research over the last two decades has revealed that the hardware establishing these systems can no longer be considered secure and trusted. Incidents like Meltdown~\cite{lipp2018meltdown} and Spectre~\cite{kocher2019spectre}, the Big Hack of 2018~\cite{robertson2018big}, the 63$\%$ rate of security breaches among organizations due to hardware vulnerabilities in 2019~\cite{bayern_detwiler_academy_eckel_wallen_2019}, and the recent surge of fake components on the market due to the ongoing global chip-shortage~\cite{leprince-ringuet_2021} are some of the most recent examples of hardware security-related issues. They are partly the result of the shift in the supply chain paradigm from a vertical model to a horizontal one. With multiple untrusted entities involved in the design process, and the rich connectivity features of the modern day computing systems, the critical hardware resources and intellectual property~(IP) are left exposed to attackers who in some cases can attack such resources remotely. In addition, the modern complexity of chip designs with billions of transistors and interactions among hundreds of IPs, coupled with the lack of security-aware CAD/EDA tools and short time to market, can allow security weaknesses to be introduced or go unnoticed during design. 

Hardware security threats can arise during various stages of the integrated circuit (IC) design cycle. They are generated from unintentional design flaws, system-based side effects, and intended malicious design modifications. With the advancement in AI and ML, hardware security threats such as physical attacks (invasive, semi-invasive, and non-invasive), and hardware Trojans are evolving too~\cite{krachenfels2021automatic, jin2020recent, timon2019non}. However, the development of AI/ML based models to defend intellectual property (IP) and detect known and potentially unknown attacks has yet to catch up. Nevertheless, in the last couple of years, researchers in both academia and industry have turned to AI to not only speed up chip design but improve upon various figures of merit. 

 A recent article~\cite{fung_2022} from Jason Fung, CWE/CAPEC Board Member and Intel's current Director of Offensive Security Research and Academic Research Engagement, highlights seven essentials for more security-aware electronic design automation (EDA). We firmly believe that AI/ML can play a significant role in meeting each essential:

\begin{enumerate}
    \item \textbf{Guide users to make design tradeoffs by taking both functionality and security into consideration.} Today's EDA tools are primarily driven by traditional metrics, such as area, power, and delay. Security metrics are challenging because oftentimes security cannot be modeled analytically. AI/ML, on the other hand, can learn models from data and learn tradeoffs/policies through trial-and-error actions with an environment.
    
    \item \textbf{Educate users on security best practices when design decisions are being made.} Most design engineers are either untrained in security or at best experts on one particular topic. Security-aware tools would ideally train and re-train engineers to keep up with new attacks and threats. Although AI/ML, especially deep learning~(DL), is often considered as a black-box, interest in explainable AI~\cite{Barredo_Arrieta2020-xai-concepts-taxonomies} is growing and should help fulfill/support this directive.
    
    \item \textbf{Detect security issues in real time when code is being written.} According to the rule-of-ten in VLSI~\cite{bushnell2004essentials}, the cost of dealing with a fault, bug, or security issue grows by a factor of 10 after completing each step within a chip's design and fabrication lifecycle. For example, finding and fixing a bug at the register transfer level (RTL) is 10 times less expensive than finding it after synthesis. At the extreme case, a bug found during pre-silicon phase might be 1000 times less expensive to fix than if the same bug were found within a chip of a fielded system, and so on. Thus, it is essential to find security weaknesses as early as possible. AI has recently found success in the software domain for vulnerability detection~\cite{li2018vuldeepecker}, fuzzing\cite{guan2020survey}, etc. For digital IC design, many of these can be directly transferred to hardware weaknesses at the source code level or RTL. In the domain of analog ICs, there is already progress to speed up performance evaluation of circuits~\cite{pan2019late}.
    
    \item \textbf{Beyond just finding problems, provide reliable mitigation options to address them.} It's not enough to just point out vulnerabilities. Engineers may need guidance on the available mitigation options and how they might impact the rest of the design. Once again, in the software domain, there is evidence that real-time recommendation and completion~\cite{svyatkovskiy2019pythia} are possible. At a minimum, it seems that AI can already provide reasonable high level synthesis (HLS)~\cite{Makrani2019}, place-and-route~\cite{xie2020powernet}, physical implementation~\cite{ding2011aeneid}, predictions for digital and analog circuits in terms of traditional metrics and, in rare cases, security metrics (e.g. SAT attack time~\cite{chen2020}). It may only be a matter of time before AI-based optimizers, especially reinforcement learning, can automatically insert solutions and countermeasures to meet an objective and constraints.
    
    \item \textbf{Seamlessly integrate best-in-class protections.} Aside from tradeoffs with traditional design metrics, metrics for certain  hardware attacks and vulnerabilities may be at odds. For example, in the area of logic locking, the community has often mentioned an unavoidable tradeoff between resistance to SAT attacks and corruptibility~\cite{xu2017novel,liu2020strong}. In such situations, a combination of methods or protections may be the best approach. In the area of AI-based digital circuit design, ``decision-making'' tools have been developed that select the right combination of algorithms and tools~\cite{yu2018developing,Neto2019LSOracleAL} for a given IP to achieve optimal quality of result.
    
    \item \textbf{Recommend the most efficient test strategy for a given coverage guarantee.} Not all security properties can be verified statically. In such cases, hardware simulation, emulation, or formal verification can be used. AI-based prediction and classification can be used to predict performance and avoid simulations entirely or narrow the simulation space down. Further, ``decision-making'' tools can guide the development team's choice of what tests to run and what parameters should be used for a specific IP.
    
    \item \textbf{Learn continuously from users.} The continuous evolution of attacks and new threats implies the need for moving target defenses. Similarly, EDA itself needs to keep up with new challenges facing the industry such as the end of Moore's law, 3D integration, etc. Advancement in AI, especially DL and federated learning~\cite{McMahan2017-federated-first-intro}, is coming at the perfect time to promote continuous learning by crowdsourcing EDA problems and solutions while still maintaining privacy of user inputs and IPs.
\end{enumerate}

In this survey, we summarize both constructive and destructive use of AI/ML techniques in the security-aware design of ICs. In other words, we review the recent works in design/attack of security approaches at every level of the IC design flow and the design/attack of hardware-based security primitives using AI/ML-based approaches. As such this survey is organized as follows: Section~\ref{sec:background} provides a brief background on ML, AI, and DL. Section~\ref{sec:digital} reviews the studies that utilize AI/ML for general and security-aware design of digital ICs. Section~\ref{sec:analog} reviews the AI/ML and optimization techniques involved in the design of analog ICs, analog security primitives, and attacks on analog ICs and security primitives. Section~\ref{sec:needs} discusses the outstanding needs and requirements for a security-aware design of ICs. Finally, Section~\ref{sec:conclusion} concludes the survey by highlighting the future trends and opportunities in the field of security-aware IC design.

\section{Background}\label{sec:background}
\subsection{Machine Learning \& Artificial Intelligence}\label{sec:mlai}
\begin{figure}[t]
\begin{subfigure}{.52\textwidth}
  \centering
  \includegraphics[width=\linewidth]{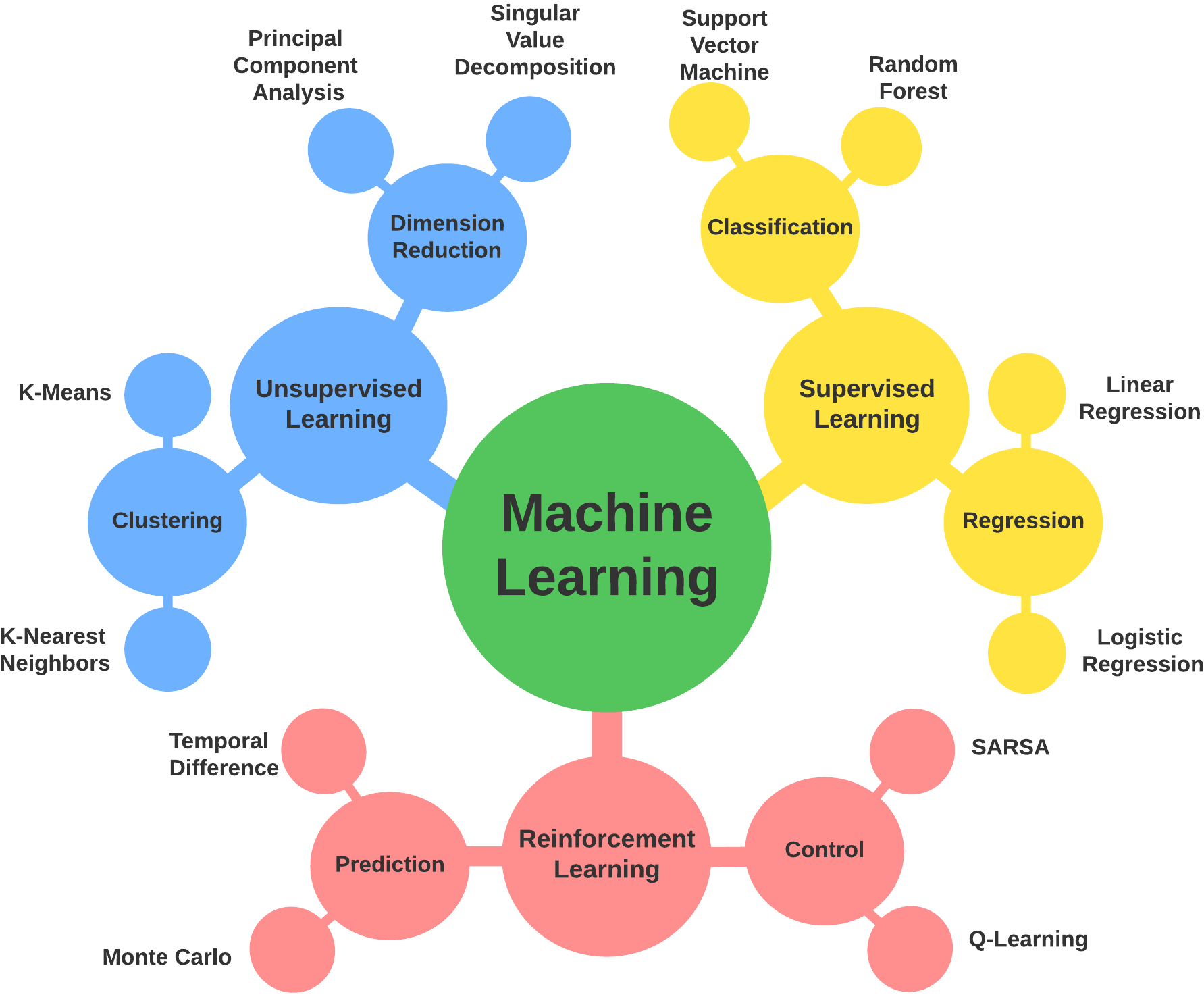} 
  \caption{}
  \label{fig:ml_methods}
\end{subfigure}
\begin{subfigure}{.45\textwidth}
  \centering
  \includegraphics[width=0.9\linewidth]{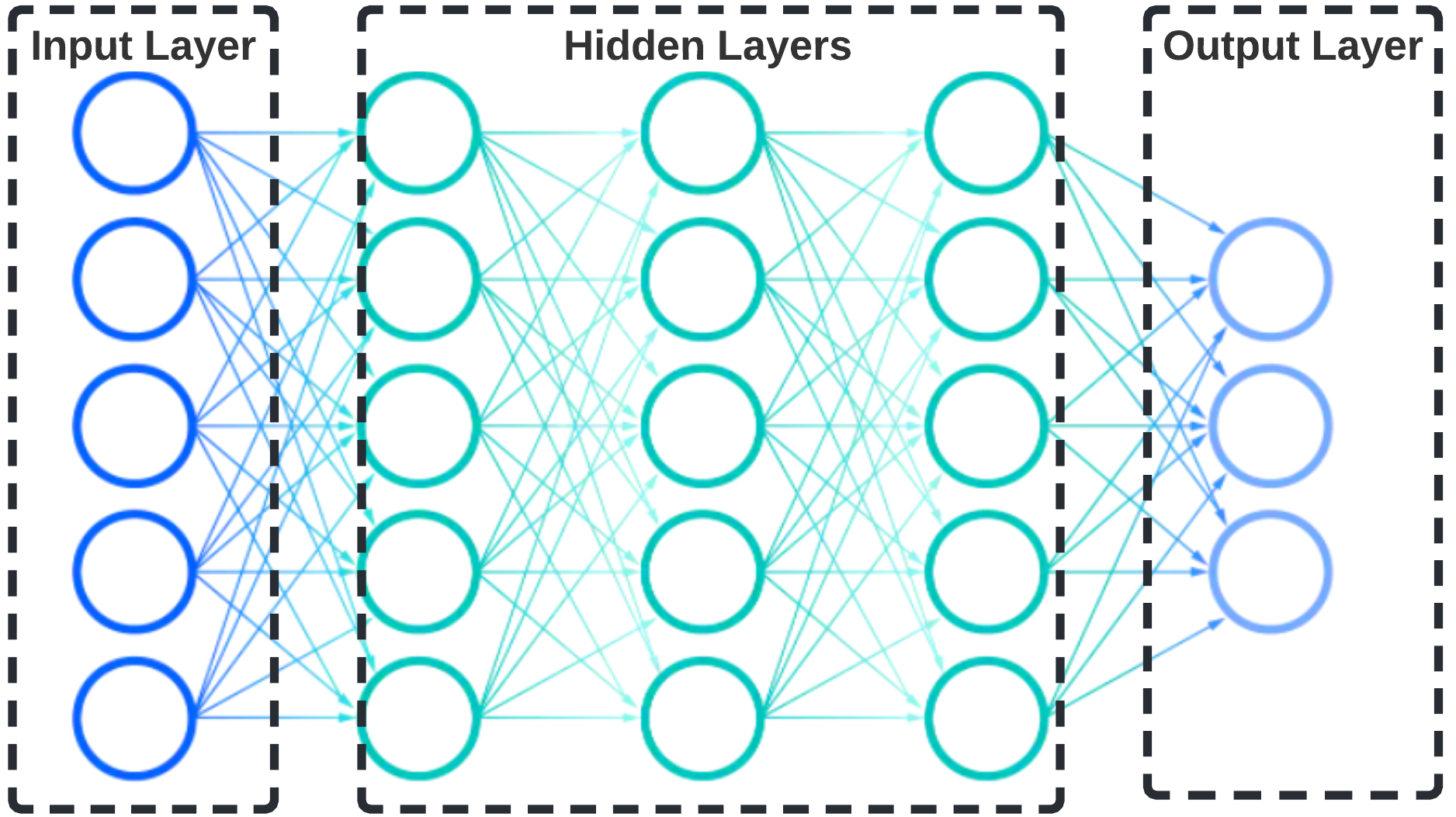}  
  \caption{}
  \label{fig:deep_learning}
\end{subfigure}
\caption{(a) A categorization of major machine learning techniques with relevant examples; (b) An example neural network with three hidden layers.}
\end{figure}

AI is a general term for any machine and/or program that displays some level of intelligence. ML is a sub-field of AI concerned with algorithms that are capable of learning patterns from data without explicitly being told those patterns by a human. Over the years, the availability of large amounts of data and computational resources has led to a rise in the use of ML in a variety of fields. DL is a sub-field of ML that involves the use of artificial neural networks (ANNs), which are organized algorithmic structures that mimic how humans learn as shown in Fig.~\ref{fig:deep_learning}. In recent years, DL has been favored over traditional ML techniques for highly complex tasks with large amounts of complex data.
 
\subsubsection{Types of Machine Learning} There are three main types of ML algorithms: supervised learning, unsupervised learning, and reinforcement learning. An overview of these three main types is provided in the following paragraphs, and example methods are summarized in Fig.~\ref{fig:ml_methods}. 
 
\textbf{Supervised Learning} is a form of task-driven ML that maps inputs to an output, given labeled training examples. Supervised ML is often used for classification and regression tasks. Classification tasks involve mapping inputs to a discrete value (i.e., class label), while regression tasks involve mapping inputs to a continuous value. Note that regression is the general form of classification, since all classification tasks can be conceptualized as predicting the likelihood (which is a continuous value) that an input belongs to each class and returning the class with the highest likelihood. Popular supervised ML methods include support vector machines (SVMs) and random forest for classification tasks, and linear and logistic regression for regression tasks~\cite{alpaydin_2020}.

\textbf{Unsupervised Learning} is a form of data-driven ML that finds patterns in input data, without the need for labeled training examples. Unsupervised ML is often used for clustering and dimensionality reduction tasks. Clustering tasks involve grouping data based on similar underlying structures, while dimensionality reduction tasks involve simplifying data to its most principle (i.e. salient) underlying structures. Common unsupervised ML methods include k-means and k-nearest neighbor (KNN) for clustering, and principle component analysis (PCA) and singular value decomposition (SVD) for dimensionality reduction~\cite{alpaydin_2020}. 

\textbf{Reinforcement Learning (RL)} is a form of feedback-based ML that involves learning from mistakes in a trial-and-error fashion. RL is often used for prediction and control in problem domains where time and event sequences matter, feedback may be delayed, and actions have consequences. In RL, prediction involves predicting the performance of some policy, whereas control involves determining the optimal policy that yields the best performance. Though practical applications are mostly concerned with control, the problem of prediction must oftentimes be solved first. Popular RL methods include Monte Carlo and temporal difference (TD) Learning for prediction, and SARSA and Q learning for control~\cite{alpaydin_2020}.  

Note that there are other, less mainstream types of ML such as semi-supervised learning, self-supervised learning, multiple instance learning (MIL), inductive learning, deductive learning, transductive learning, multi-task learning (MTL), and active learning. In addition, there are other ML tasks including generative modelling and association rule learning.
    

\subsubsection{Machine Learning Pipeline}
\begin{figure*}[t]
     \includegraphics[width=\textwidth]{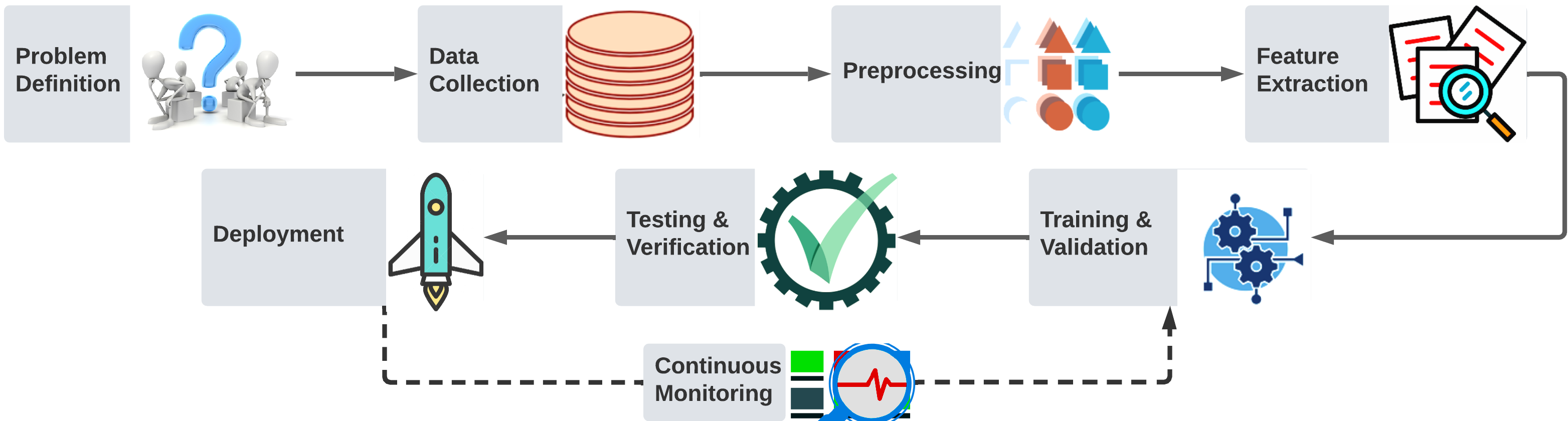}
     \centering
     \caption{General machine learning pipeline.}
     \label{fig:ml_pipeline}
\end{figure*}

Regardless of which type of ML model is used, it is important to carefully plan the ML pipeline for any task. A generalized ML pipeline is described in the following paragraphs, and summarized in Fig.~\ref{fig:ml_pipeline}. 

The first step of the ML pipeline, \textbf{Problem Definition}, involves thoroughly understanding the problem to be solved. This involves determining what the end goal is in terms of desired model outputs, what inputs might be needed for a model to yield such desired outputs, and what existing techniques have already been applied to similar problems and their assumptions, benefits, and limitations. In other words, Problem Definition involves gaining enough domain knowledge to make informed decisions for the remaining steps in the ML pipeline. 

During \textbf{Data Collection}, information relevant to the problem is gathered. Data can be collected historically and/or in an online fashion, and includes organized and unorganized data structures such as text, images, videos, audio clips,~\etc{} In supervised ML, human subject matter experts are often employed to label the data so ML systems have training examples. Ideally, data is sufficient in amount and properly representative of the problem space.

In the \textbf{Preprocessing} stage, the collected data is cleaned and converted to a machine-readable format in preparation for feature extraction. Common preprocessing methods concern handling missing values (\eg{} removing all datapoints with missing values from the study, replacing missing values with a zero), handling categorical data (\eg{} creating dummy variables), and standardizing the data (\eg{} normalizing values from 0 to 1).

After Preprocessing, the data is then transformed in the \textbf{Feature Extraction} stage. The goal of feature extraction is to ensure only the most salient information relevant to the problem will be used for training the ML model. In other words, this step helps ensure that no unnecessary features are used during training. At best, unnecessary features do not degrade ML model performance, at the cost of increasing the amount of data needed (as is characteristic of the curse of dimensionality). At worst, unnecessary features can significantly degrade ML model performance (as is characteristic of irrelevant and/or highly correlated inputs). Common feature extraction techniques include feature selection (\eg{}, linear discriminant analysis or LDA), dimensionality reduction (\eg{}, PCA), and computer vision techniques for input images (\eg{} color, shape, and texture feature extraction methods).

During \textbf{Training and Validation}, the extracted features are used to train the ML model. The type of ML model(s) to use will depend on a variety of factors such as the nature and amount of input data, format of the desired output, and the given problem's associated ML task. Note that in DL, feature extraction and training can occur simultaneously. Ideally, the type of ML model(s) to use were considered at the Problem Definition stage but sometimes it is accomplished by trial-and-error. 

After the ML model is trained and its performance is validated, the \textbf{Testing and Verification} stage ensures the model is ready for deployment. Testing involves evaluating the trained model on unseen data and validation typically involves humans ensuring the model is performing up-to-standard. Validation data is different from testing data in that validation data is used to evaluate the model during the training process, while testing data is completely unseen throughout the training process (i.e. testing data is a separate holdout set). Separate training, validation, and testing data ensures the model is not overtraining (i.e. memorizing rather than actually learning). 

After the ML model achieves desired performance on testing data and has been verified, the model is ready for \textbf{Deployment} in the field. Depending on the nature of the problem, the ML model may require updates. If so, it may be necessary to periodically monitor performance changes over time, collect additional data, and refine or retrain the model. 

\subsubsection{Deployment and Tools}
The popularity of Python for data science applications helped drive the development of AI, ML, and DL frameworks coded in Python. Python packages such as Scikit-Learn~\cite{scikit-learn}, Tensorflow~\cite{tensorflow2015-whitepaper}, and PyTorch~\cite{NEURIPS2019_9015} make AI accessible and have hence garnered the support of a large community of developers and researchers over the years. These tools have become widely-used for experimentation, evaluation, and deployment with the help of application program interfaces (APIs). Advances in hardware and software such as parallel processing, GPUs, and cloud computing help optimize computing resources to be more suitable for training large, complex AI models. Moreover, data is becoming more available than ever for training and testing ML models (\eg{} kaggle, Trust-hub,~\etc{}).

\subsection{Deep Learning}\label{sec:dl}
DL models have become a hot topic again in the last decade because of the availability of huge amount of data and deserve a short description here. They are made up of multiple processing layers that learn representations of training data with multiple levels of abstraction, and are used for speech recognition, object visualization and detection task, among others. DL has helped address complex problems that have persisted in various industries for many years~\cite{lecun2021}. The core of DL is the use of neural networks (NNs), as seen in Fig.~\ref{fig:deep_learning}, which consist of an input layer, one or more hidden layers, and an output layer. 
Most deep neural networks~(DNNs) flow in one direction only from input to output. However, backpropagation is used to change internal parameters or weights that are used to compute the representation between a current and previous layer. And it is very efficient for computing gradients in deep networks~\cite{ibm}.
Some examples of DL models include convolutional neural networks (CNNs), graph neural networks (GNNs) and graph convolutional networks (GCNs).

\subsection{Geometric Deep Learning}\label{sec:gnn}
\begin{wrapfigure}{l}{0.28\textwidth}
    \centering
    \includegraphics[width=0.25\textwidth]{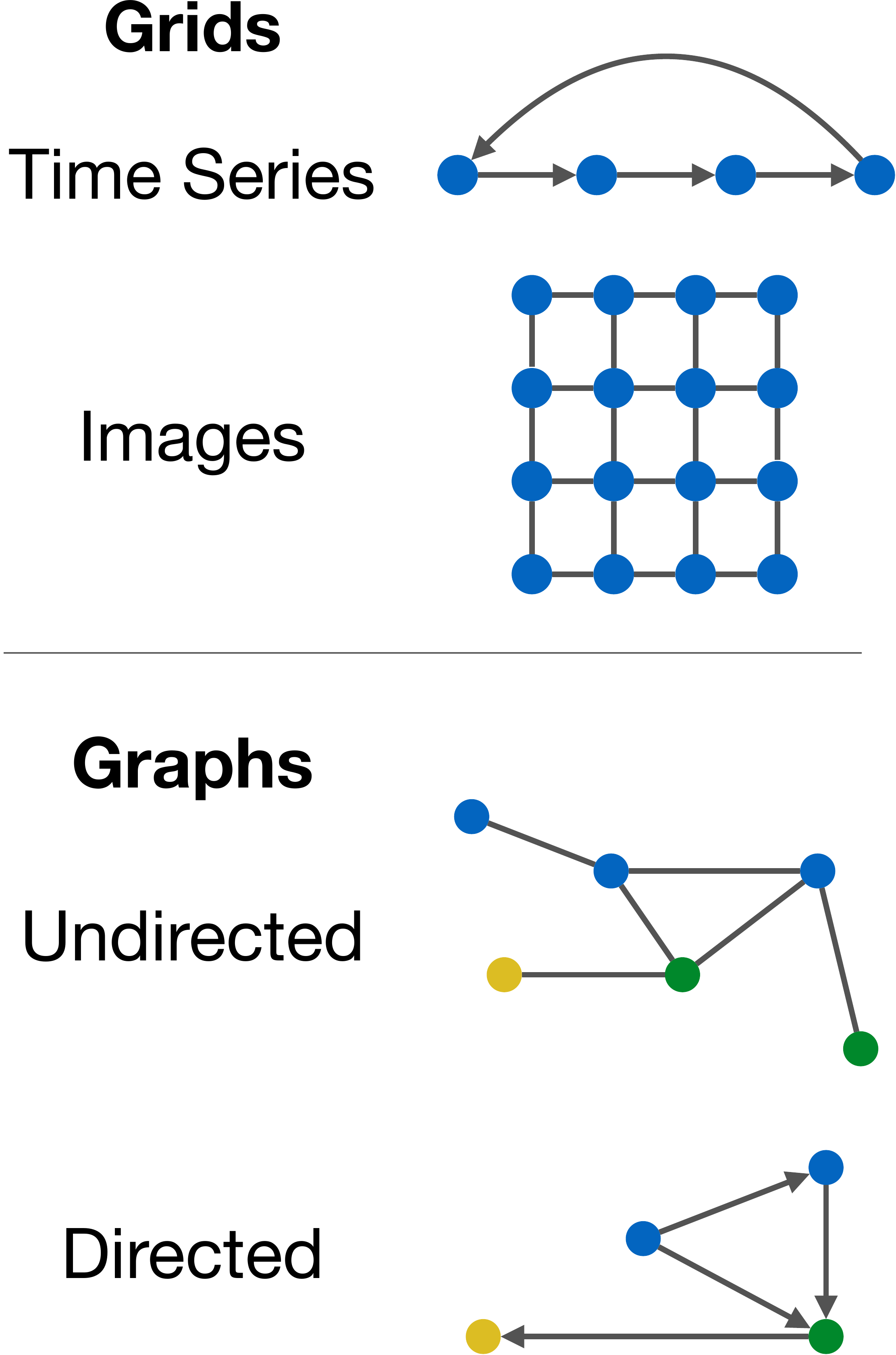}
    \caption{Traditional DL and ML primarily work on Euclidean data or grids, while GDL generalizes to non-Euclidean data, \eg{} graphs.}
    \vspace{-1.5em}
    \label{fig:grids-graphs}
\end{wrapfigure}

Perhaps most relevant to EDA applications is \emph{Geometric Deep Learning} (GDL, which provides a blueprint for generalizing DL to non-Euclidean data (\eg{} circuits)~\cite{Bronstein2017-geometric}.
The success of DL approaches is owed largely to their ability to capture local statistics of the input data.
For example, CNNs have been widely useful in computer vision tasks due to the shift invariant properties of the convolution operator.
Traditional formulations of DL models (CNNs, RNNs \etc{}) rely on Euclidean structures in the data (\emph{grid} structures), \eg{} image, video and speech signals.
Figure~\ref{fig:grids-graphs} shows examples of Euclidean (grid) and non-Euclidean (graph) data.  
The \emph{geometric prior} on grid data is the ordering of the nodes, which is the structure that RNNs and CNNs implicitly take advantage of.
Graph data, on the other hand, requires no explicit ordering on the nodes and operations on the graph are permutation invariant.
Graphs, on the other hand, have permutation invariant structure in that the nodes are not assumed to be in any specific ordering.
\begin{figure*}
    \centering
    \includegraphics[width=0.7\textwidth]{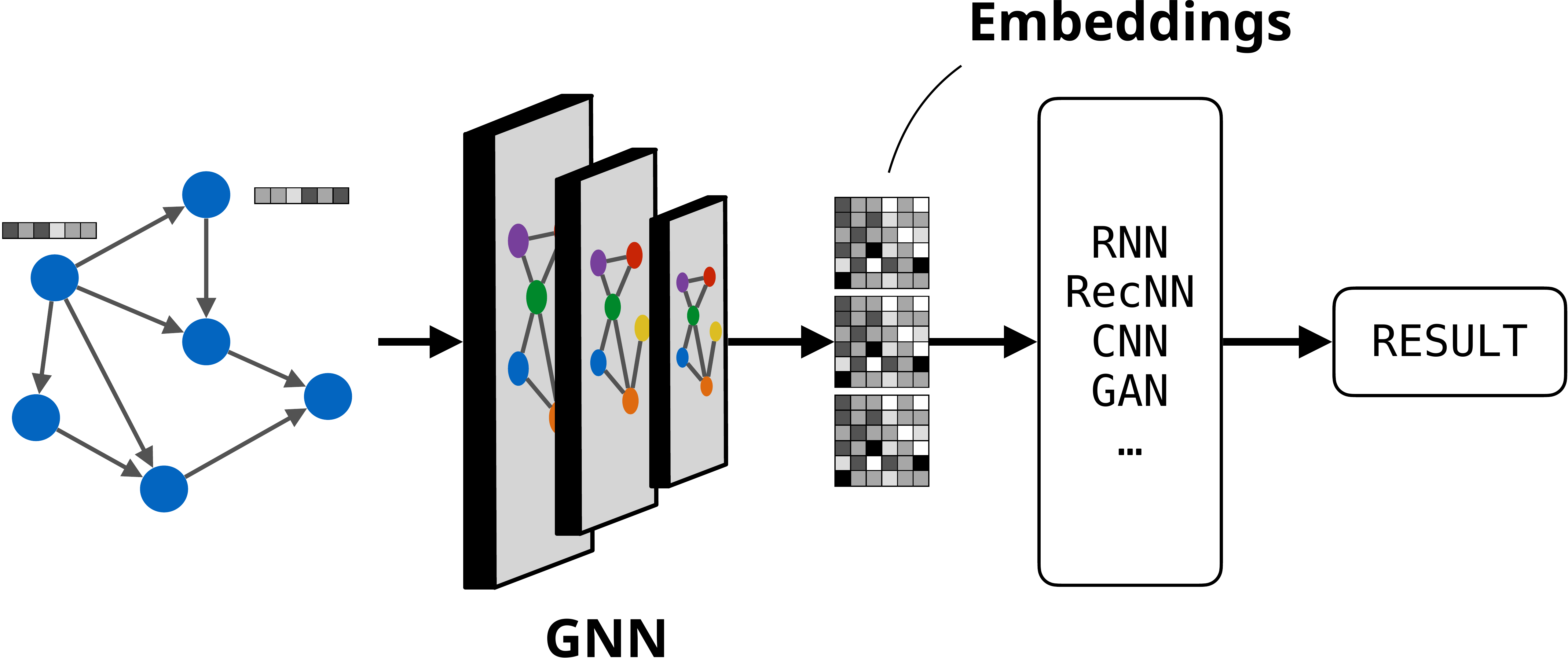}
    \caption{Overview of a GNN architecture.  The GNN layers provide embedded representations of graph input data for other downstream architectures such as RNNs, CNNs, etc.}
    \label{fig:gnn-arch}
\end{figure*}

GNNs and GCNs are examples of GDL models which can learn directly from graph data (this category of approaches is also referred to as \emph{Graph Representation Learning} (GRL) and is a subset of GDL)~\cite{Hamilton2020-grl}.
These models are able to leverage graph structure as well as node and/or edge features to learn a useful representations of data.
GCNs possess the same shift invariance power of traditional CNNs~\cite{xu2018powerful}.

Recent work by Mirhoseini~\etal{} demonstrates the power of combining GRL techniques with RL training algorithms to improve placement in chip layout~\cite{Mirhoseini2021AGP}.
The Edge-GNN in~\cite{Mirhoseini2021AGP} was first trained to create graph embeddings that predict reward labels (in this case a reward derived from wire-length, congestion and density metrics).
Then the Edge-GNN is used as an ``encoder'' for training the CNN-based policy networks in and RL training algorithm.
Figure~\ref{fig:gnn-arch} shows the general structure of GNN layers and demonstrates how the embedded graph vectors can be passed to various down-stream tasks.
The ability to capture relationship and interaction statistics in the circuit data using graph embeddings enabled the to policy network to reduce placement costs.

\section{Security-aware Design of Digital Integrated Circuits}\label{sec:digital}
\subsection{Electronic Design Automation}\label{sec:eda}
The electronic design automation flow for digital ICs is as follows.

 \subsubsection{Design Specification}
    In design specification, the designer must accurately describe the overall design requirements. The system may be proposed by a design team to initiate the production of a chip. The IC designers decide on functionality and plan out verification and testing procedures for the entire process.
    
\subsubsection{High-Level Synthesis}
    High-Level Synthesis (HLS) tools translate a design written in high-level languages such as C/C++/SystemC into a low-level hardware description language (HDL). Hardware components can be modeled at high levels of abstraction, allowing the designer to map both hardware and software components~\cite{martin_scheffer_lavagno_2016}
    
\subsubsection{Logic Synthesis}
     Logic synthesis is a direct translation from the behavioral domain (register transfer level) to the structural domain (gate level)~\cite{JIANG2009299}. The gate-level netlist is mapped to the standard cell library. The optimization part of logic synthesis minimizes hardware by finding equivalent representations of larger blocks. Design constraints used as input may also be met for circuit area, power, and performance.

\subsubsection{Design for Testability (DFT)}
   DFT encompasses the techniques used to generate cost-effective tests for the gate-level netlist obtained from logic synthesis. These techniques are geared towards increasing observability, controllability, and predictability of the design~\cite{martin_scheffer_lavagno_2016}. Ad-hoc methods like partitioning and test point insertion are relatively easier to implement, while structured ones like scan registers, scan chains, scan architectures and algorithms ensure good testability at the expense of additional area and delay overhead. The resulting tests utilize test programs which drive automatic test equipment (ATE). The ATEs also use automatic test pattern generation (ATPG) to identify input sequences which trigger errant circuit behavior arising from manufacturing defects~\cite{HSIAO2006161}.

\subsubsection{Floorplanning}
    Flooplanning focuses on the sizes and arrangement of the physical blocks created during logic synthesis. It affects the optimization results of the increasing physical stages. It is the first major step in physical design. Pins and ports are also assigned a rough location, which can further be refined depending on the placement and routing results~\cite{Lienig2020}.

\subsubsection{Placement-and-Routing}
   After a successfully-proposed floorplan, the place-and-route~(P\&R) addresses the structural design of a circuit. The designer places and connects all of the blocks that make up the chip such that they meet design criteria and constraints. After the initial P\&R attempts, the design's timing constraints are analyzed. If unsatisfactory, the P\&R software in use tries different placements and signal routing to try to meet the designated constraints~\cite{Lienig2020}.

\vspace{1em}

\begin{figure*}[t]
   \includegraphics[width=\textwidth]{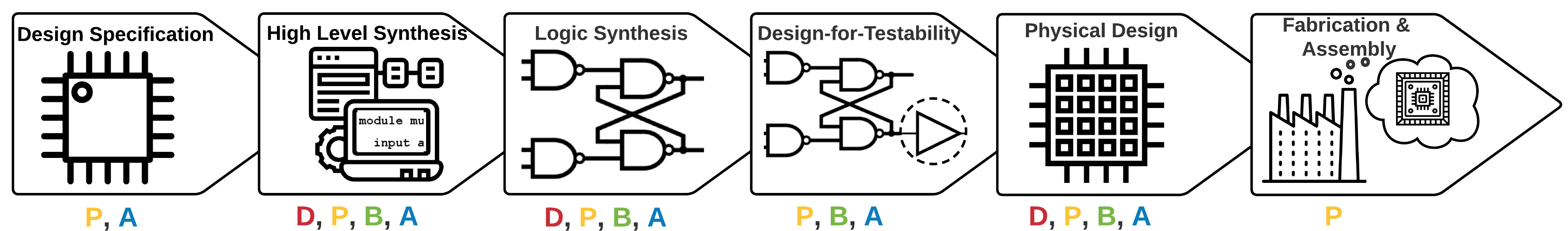}
   \centering
   \caption{General digital IC design pipeline with applicable security topics for each stage. \textbf{D} stands for decision making, \textbf{P} represents performance prediction, while \textbf{B} and \textbf{A} stand for black box optimization or design space exploration and AI-assisted EDA, respectively.}
   \label{fig:icsecurity}
\end{figure*}

ML is already applied to some extent in the digital IC design process, albeit with primary focus on general design prediction and optimization. Some needs tackled at this stage include logic and design optimization~\cite{Haaswijk2018}, low-power approximation~\cite{Pasandi2020} and area estimation~\cite{froemmer2020}. Careful assessment of available publications incorporating learning into digital IC design will reveal an absolute truth; very little work has been done to directly address hardware security vulnerabilities. The use of intelligent models presents the opportunity to not only optimize designs, but to secure them against the array of tools and attacks that malicious parties have at their disposal. This is especially important for some security properties that cannot be modelled analytically and situations where multiple security constraints are at odds. According to~\cite{huang2021machine}, ML approaches in EDA can be broken down into four main directions: \textbf{decision making}, \textbf{performance prediction}, \textbf{black box optimization or design space exploration} and \textbf{AI-assisted EDA}. In the rest of this section, we follow this same organization to discuss security-aware CAD research and opportunities. Fig.~\ref{fig:icsecurity} also highlights the general EDA steps and the applicable directions for each one.


\subsection{\textbf{Decision Making}}\label{subsection: decisionmaking}
\begin{figure*}[t]
   \includegraphics[width=\textwidth]{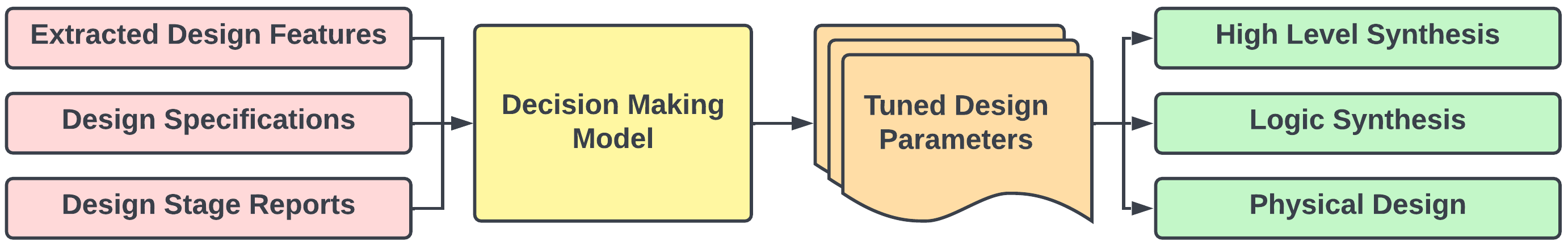}
   \centering
   \caption{General overview of decision making process for security-aware EDA using AI/ML/DL.}
   \label{fig:decisionmaking}
\end{figure*}

The decision-making perspective encapsulates the replacement of traditional, brute-force design-for-security methods with more automated and efficient techniques. In most cases, brute-force techniques involve countless trial-and-error steps until the best-performing configuration is found and used on a design. From a heuristic perspective, the knowledge gained may prove to be invaluable. However, they are only feasible on simpler designs; increased complexity transforms them into cumbersome requirements that most designers would gladly bypass. Further, the tradeoffs between security and traditional metrics as well as between different security metrics might be unknown or too complex for manual tuning. Automation becomes valuable only when it is characterized by efficiency. ML can definitively make decisions on the most suitable settings and parameters for countermeasures with varying levels of human input~(see Fig.~\ref{fig:decisionmaking}).



\subsubsection{High Level Synthesis and/or Register Transfer Level}

\paragraph{Encoding Optimization}\label{para: encoding}
Another aspect that stands to benefit from the addition of ML-powered decision-making is finite state machine (FSM) encoding optimization. The encoding schemes are selected based on design constraints, including power and area. The two most common options are:

\begin{itemize}
\item \textbf{Binary Encoding} -
In binary encoding scheme, states are encoded as a binary sequence where the states are numbered starting from 0 and up. The number of state flip-flops (FFS), $q$, required for binary encoding scheme is given by $q = log2(n)$; where, $n$ is the number of states. The binary encoding scheme is better suited for FSM with a fewer number of states~\cite{lin2012binary}.

\item \textbf{One-hot Encoding} -
This encoding scheme is designed with only one bit of the state variable as ``1'' while all other state bits are ``0''. It requires as many state FFs as the number of states. This creates the need for more state FFs than binary encoding~\cite{cassel2006onehot}.
\end{itemize}

One-hot encoding scheme is less vulnerable to fault injection attacks while binary is more resistant to hardware Trojans that exploit don't care states. This is demonstrated by Nahiyan~\etal{} using the FSM of the SHA-256 digest engine~\cite{nahiyan2019}. With this knowledge, a secure encoding scheme was proposed in~\cite{nahiyan2019} that combines binary and one-hot schemes to achieve the best of both worlds. In future work, AI/ML may be used to determine the best mix given a specific FSM.

\paragraph{Hardware Security Primitive Deployment at Higher Levels of Abstraction}\label{paragraph: aiss}
The Automatic Implementation of Secure Silicon (AISS) program was developed by the Defense Advanced Research Projects Agency (DARPA) to automate the process of incorporating scalable defense mechanisms into chip designs, by allowing designers to explore chip economics versus security trade-offs based on the expected application and intent while maximizing designer productivity~\cite{darpa}. Deployment of security methods pre-silicon are always cost-effective and less time-consuming. With this in mind, AI/ML can be used to make decisions as early as during HLS, or at RTL. Based on the discernible features at these levels of abstraction, a classification-based model can use vector embeddings to suggest suitable hardware security primitives and their associated parameters (e.g., size). For example silicon odometers are inserted into IC designs to measure aging and detect recycled counterfeit chips~\cite{guin2015design}. Their accuracy and yield are determined by various parameters~\cite{shakya2015performance}, which can be optimized through ML/AI. 

\begin{figure*}[t]
   \includegraphics[width=0.8\textwidth]{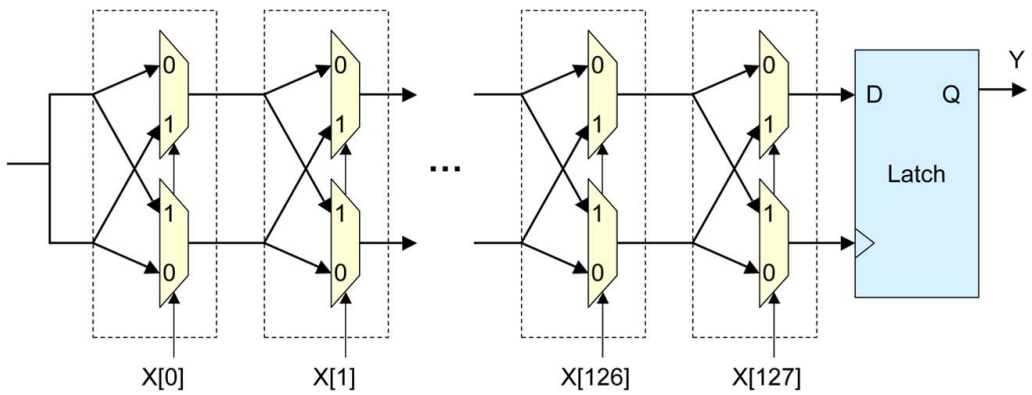}
   \centering
   \caption{Arbiter PUF circuit. The circuit creates two delay paths with the same layout length for each input X, and produces an output Y based on which path is faster. Reprinted from~\cite{herder2014puf}}
   \label{fig:puf}
\end{figure*}

Physically unclonable functions (PUFs)~\cite{gassend2002silicon} provide digital fingerprints using the internal process variations (entropy) of a physical device. An example is arbiter PUF circuit shown in Fig.~\ref{fig:puf}. the The translation of these unique and unclonable manufacturing variations into challenge-response pairs provides unique chip identifiers or authentication mechanisms. One of the main issues with PUFs is their reliability,~\ie{} ability to provide the same output in the presence of environmental variations and aging. To combat this, they often rely on error correcting codes~\cite{hiller2020review}, which can be designed to handle a specific number of bit flips. Another primitive is the true random number generator (TRNG), which uses an entropy source to generate non-deterministic data with a given throughput to seed security algorithms like cipher keys~\cite{gabriel_wittmann_sych_dong_mauerer_andersen_marquardt_leuchs_2010} or masking~\cite{trichina2003combinational}. For any design that contains all three of these primitives, their characteristics can be tuned together for a specific application or to fit within a dedicated silicon area.

\subsubsection{Logic Synthesis}
\paragraph{\textbf{Tuning EDA Parameters and Commands during Logic Synthesis}}\label{subsubsec: dectuninglogic}
Although a constraint set may be part of the design specification, the gate level netlist provides a better understanding of functionality to a designer. Area constraints guide the P\&R tool to locate a specified design block partition, while timing constraints specify path delays~\cite{else2009}. To tune high-level directives, it is important to accurately predict their impact. This is a tall order for most designs due to complex optimizations throughout the design process. These complex optimizations create disparity between desired results at the HLS stage and implemented values post-synthesis. Conventional tools assist with optimization tasks, but machine learning is already being used to mitigate the latter problem. For example, Yu~\etal{}~\cite{yu2018developing} have developed Intellectual Property (IP)-specific synthesis flows (i.e., sequence of synthesis transformations for TCL scripts) using CNNs that improve Quality of Result (QoR). In~\cite{Neto2019LSOracleAL}, DL is used to automatically decide which logic optimizer (And-Inverter Graph (AIG) and Majority-Inverter Graph (MIG)) should handle different portions of the circuit for better performance and lower area. Makrani~\etal{}~\cite{Makrani2019} proposed a Pyramid framework to estimate best performance and resource usage. The input features are derived from the HLS report, and a stacked regression model provides accurate timing, delay and resource values before reaching post-implementation. 

Optimization goals during tuning can be geared towards obtaining a more secure, synthesised design with minimum overhead. Although this would only be evident with the combination of other design parts, early-stage deployment saves time and labor. For example, secure split-test (SST) is a method of securing the manufacturing process by mandating test results to be verified by the IP owner and by using a ``key'' to unlock the IPs correct functionality after tests are passed. For the functional locking block, XOR logic is added in series to non-critical paths. If the two inputs are different, the XOR logic will act as an inverter~\cite{contreras2013} causing any locked (still untested or failed chip) to act deliberately incorrect. The optimal number of XOR gates to insert into the circuit, insertion algorithm~\cite{chakraborty2009harpoon,6616532,lee2015improving}, or partitioning algorithm~\cite{alaql2021saro} to employ for a given IP's function and/or structure can be determined using ML techniques.

\begin{table*}[t!]
\centering
\caption{Security requirements under decision making with applicable AI/ML algorithms. The cited options indicate optimization or security implementations available for reference}
\label{table:decisionmaking}
\arrayrulecolor{black}
\resizebox{\linewidth}{!}{
\begin{tabular}{|c|c|c|} 
\hline
\rowcolor[rgb]{0.361,0.769,0.769} \textbf{Design Stage }   & \textbf{Security Task }     & \textbf{Applicable AI/ML Algorithms}            \\ 
\hline
{\cellcolor[rgb]{1,0.859,0.604}}      & Encoding Optimization       & Polynomial Regression      \\ 
\hhline{|>{\arrayrulecolor[rgb]{1,0.859,0.604}}->{\arrayrulecolor{black}}--|}
\multirow{-2}{*}{{\cellcolor[rgb]{1,0.859,0.604}}\textbf{High Level Synthesis}} & \begin{tabular}[c]{@{}c@{}}Hardware Security Primitive Deployment\\at Higher Levels of Abstraction\end{tabular} & \begin{tabular}[c]{@{}c@{}}Polynomial Regression,\\SVM\end{tabular}  \\ 
\hline
{\cellcolor[rgb]{1,0.859,0.604}}      & \begin{tabular}[c]{@{}c@{}}Tuning EDA Parameters and Commands\\during Logic Synthesis\end{tabular}              & CNN~\cite{yu2018developing}, Stacked Regression~\cite{Makrani2019}     \\ 
\hhline{|>{\arrayrulecolor[rgb]{1,0.859,0.604}}->{\arrayrulecolor{black}}--|}
\multirow{-2}{*}{{\cellcolor[rgb]{1,0.859,0.604}}\textbf{Logic Synthesis }}     & Standard Cell Selection     & SVM   \\ 
\hline
{\cellcolor[rgb]{1,0.859,0.604}}\textbf{Cross-Abstraction} & \begin{tabular}[c]{@{}c@{}}Benchmark Selection for Security-\\Aware EDA Tool Evaluation\end{tabular}            & GNN, GCN                   \\
\hline
\end{tabular}
}
\end{table*}

\paragraph{\textbf{Standard Cell Selection}}\label{subsubsec: standardcell} A specific category of available standard cells may be used in an IC design to address specific security vulnerabilities. As far back as 2004, sleep transistor cells were proposed as a method for sub-threshold leakage current reduction~\cite{babighian2004}. The sleep transistor cells are picked from a library designed to target high layout efficiency. The results were shown to have achieved a minimum reduction percentage of 74\%. A trained support vector machine~(SVM) can be used to decide from a set of available libraries with specific target optimizations. 

\subsubsection{Cross-Abstraction}

\paragraph{\textbf{Benchmark Selection for Security-Aware EDA Tool Evaluation}}\label{subsubsec: benchselection}
One additional optimization activity that may easily be overlooked is the evaluation of EDA tools used during the IC design process. Companies and research groups that develop EDA tools constantly update features to enhance their output. In order to properly target specific problem areas, the testing process must involve specific benchmarks. For example, if a designer intends to assess the scalability of the tool, the most important requirement for the test benchmark would be its size. If more nuanced conditions must be met, an automated method would make this process faster and more efficient. Succinct features of a benchmark and a tool's problem area in a DNN can determine the most suitable sample for testing. 

In the area of hardware security, benchmarking has played a critical role in development and comparison of hardware Trojan detection schemes~\cite{shakya2017benchmarking} and recently in de-obfuscation~\cite{amir2018development,tan2020benchmarking,amir2020adaptable}. Thus far, however, none of these benchmarking activities have explicitly incorporated AL/ML. Table~\ref{table:decisionmaking} provides suggested AI/ML algorithms for future endeavors.


\subsection{\textbf{Performance Prediction}}\label{subsec: performanceprediction}
\begin{figure*}[t]
   \includegraphics[width=\textwidth]{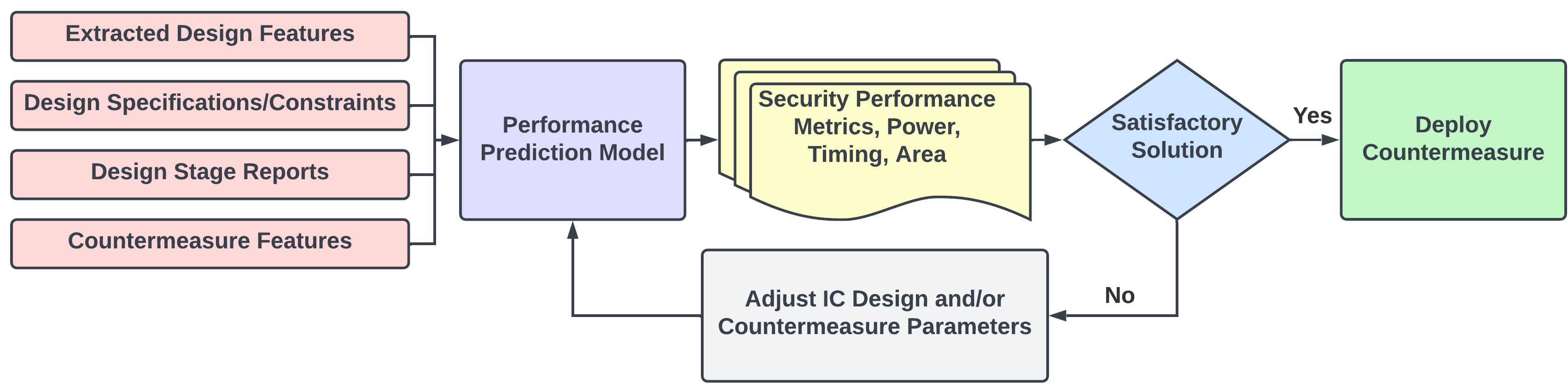}
   \centering
   \caption{General overview of performance prediction process for security-aware EDA using AI/ML/DL.}
   \label{fig:performanceprediction}
\end{figure*}

Performance prediction describes forecasting and modeling trade-offs associated with adding countermeasures, especially at earlier design stages. Although an ideal scenario would involve achieving optimum design goals while guaranteeing near-perfect security awareness, a designer must be aware of the inevitable compromises that a secure design may entail. A more achievable target is finding the balance between optimization and security, as seen in Fig.~\ref{fig:performanceprediction}. This is where AI/ML comes into the picture. The automation of the balance-seeking process will not only speed up the IC design pipeline, but will ensure that the most suitable operating parameters and modifications are available to the designer. From a business standpoint, time-to-market of ICs in production may also decrease.

\subsubsection{Physical Design}

\begin{figure}[t]
\centering
\includegraphics[width=0.5\columnwidth]{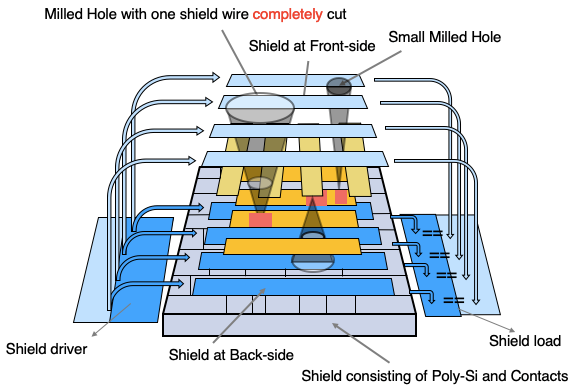}
\caption{iPROBE version 2 with protection for frontside and backside probing attacks~\cite{osti_10174121}. The light blue metal lines and darker blue metal lines form active shields that protect against frontside attacks (through top passivation and metal) and backside (silicon substrate), respectively. The metal lines sandwiched between the shields carry critical information that the attacker wants to probe. A FIB cut though an entire shield line of frontside/backside will be detected because the signal present on it will not match its counterpart in backside/frontside.}
\label{fig:iprobe}
\end{figure}

\paragraph{\textbf{Security-aware Layouts and Tradeoff Modeling}} \label{subsubsec: eaprediction}
Invasive physical attacks are expensive to execute but difficult to protect against. Attackers resort to focused ion beam~(FIB) to extract an asset’s value by milling to its location in the IC layout, creating a metal contact to it, and probing the contact while the chip is in operation. There are two main sets of countermeasures proposed to combat probing attacks; the preventive strategy utilises active or analog shields, such as meshes, while the detection strategy checks for attacks by sensing any probing or editing and alerting the operators~\cite{wang2017}. To improve the effectiveness of active shields, an automated anti-FIB probing flow called iPROBE was proposed~\cite{Wang2019,osti_10174121}, see Fig.~\ref{fig:iprobe}. iPROBE uses ad hoc rules to determine the areas and metal layers where computer-aided design (CAD) tools should place shields to reduce ``exposed area''. For a physical IC, the exposed area (EA) metric (as described in \cite{shi2016layout}) is used to measure the design’s vulnerability to probing attacks. A larger EA means more flexibility with the expense that an attacker can probe without triggering contingencies. While iPROBE did improve EA while reducing overhead as compared to traditional shields, the security and overheads cannot yet be explicitly controlled by the designer. In other words, trial-and-error needs to be performed to make sure that constraints are met. To further boost efficacy of the anti-probing method described above, it may be possible to model the relationship between EA and power, timing and area with a multiple linear regression model or even DL. Then, iPROBE parameters could be chosen with security, area, power,~\etc{} in mind.

\subsubsection{Cross-Abstraction}
\paragraph{\textbf{Modeling Area, Power, Timing, and Security Impacts of IP Protection}} \label{subsubsec: ipprediction}

In general, design optimization problems have always been modeled based on area, power and timing; timing is used here because it directly relates to performance. Even at the HLS stage, ML implementations, like Pyramid~\cite{Makrani2019} have been used to predict optimal timing. There has also been work to predict area at both HLS~\cite{Zennaro2018} and logic synthesis~\cite{froemmer2020}. Deep-PowerX is a DNN-based dynamic power consumption minimization tool which exploits the power-area and delay-area relationships within a netlist~\cite{Pasandi2020}. Even at the placement and routing stage, timing prediction is possible using ML~\cite{Barboza2019}. 

One of the most significant threats to design integrity is IP piracy. The availability of a design to malicious parties gives way to overuse, modification, cloning, overproduction and/or reverse engineering. There are various defense-centered modifications available to an engineer at various stages of the IC design cycle, and ideas from ML-based optimization can be applied to them. 
\subparagraph{IP Watermarking} A watermark is a specific modification of an IP core that allows it to be uniquely identifiable, which is useful for piracy detection. However, IP core watermarking in general must never alter the functionality of the design~\cite{Chang2016watermark}. One popular method of watermarking is the the don't-care condition based technique. It involves adding function blocks with unspecified input combinations to the original design. Their outputs can be forced and used to verify the authenticity of a circuit~\cite{Dunbar2015SatisfiabilityDC}. However, additional logic may mean area, timing and power overhead. To hasten the design cycle, these overhead metrics must be readily available to a designer. If a trained algorithm can accurately model and predict these values, the design house would be able to ascertain the viability of a specific watermarking scheme, like the don't-care condition based method.

\subparagraph{Hardware Obfuscation}\label{par: dectuning, obfus} Hardware obfuscation is used to actively protect IP and has been applied at all major design stages preceding fabrication and assembly. It is intended to obscure an original design to prevent IP piracy and reverse engineering. Hardware obfuscation also addresses IP overuse and hardware Trojan insertion. 

The set of approaches that can be implemented until the DFT stage are typically \textbf{keyed} approaches. These techniques attempt to prevent black box use by key gates/inputs. One common example is \textbf{logic locking}, which hides the functionality and the implementation of a design by inserting additional gates (``key gates'') into the original netlist. The circuit will only function correctly if the acceptable inputs or keys are provided to the key gates~\cite{Yasin2020}. The additional gates must provide low overhead and be usable for larger designs. A significant challenge, however, has been making logic locking scheme secure against the entire suite of oracle-based attacks (e.g., SAT~\cite{subramanyan2015evaluating}, AppSAT~\cite{shamsi2017appsat}, key sensitization~\cite{rajendran2012security},~\etc{}.) and oracle-less attacks (e.g., desynthesis~\cite{massad2017logic}, signal probability skew~\cite{yasin2017security},~\etc{}.). The former use an unlocked chip as an oracle to non-invasively derive its key while the latter operate solely on the netlist. Preventing oracle attacks implies that the time to recover the key should be prohibitively high for an attacker. Recently, ML has been used to predict the obfuscation strength in terms of the predicted time taken to de-obfuscate an obfuscated circuit using the SAT attack. Chen~\etal{}~\cite{chen2020} demonstrate this using a GCN. They combine graph structure with gate feature by using an enhanced graph convolutional operator named ICNet. The features employed are divided into the graph structure, which describes connections, and gate features, including gate mask and gate type. While this approach has only been applied to predict SAT attack time, it can nonetheless be used to predict resistance to other oracle attacks. Recently, oracle-less attacks such as SAIL~\cite{chakraborty2018sail} have employed ML to recover the original (unobfuscated) netlist by reverse engineering logic synthesis rules.

Another key-based countermeasure that can be considered is FSM locking~\cite{chakraborty2009harpoon}. FSM locking adds extra states into the FSM of an IC design. The correct ``key'' is the right sequence of state transitions, which guarantees non-erroneous functionality. The additional states are materialised by FFs with their corresponding combinational logic. Active hardware metering~\cite{alkabani2007active}, a version of FSM locking, creates boosted finite state machines (BFSMs) from the new states. At this stage, it would be imperative to be able to predict the overhead created by the added states. This is because the design house sends the verified netlist to the untrusted foundry to create PUFs (challenge-response pairs) from the fabrication process variation. This may introduce additional overhead that could affect overall functionality. Depending on the complexity of the circuit, either a regression model or a DNN may suffice.

The \textbf{keyless} variety of obfuscation does not require keys, and cannot affect functionality. The desire is to conceal design intent. Without some form of obfuscation, a malicious party can either copy an IP or seek and exploit design vulnerabilities using reverse engineering. The most noteworthy example of keyless obfuscation is \textbf{IC camouflaging}~\cite{chow2012camouflaging}, which replaces original gates with ``camo gates'' which are difficult to discern by pattern recognition. It is worth noting, however, that most camouflaging techniques can be ``mapped'' into logic locked circuits~\cite{yasin2015transforming} where the above oracle attacks are quite effective~\cite{el2015integrated}. Hence, the prediction schemes mentioned above for logic locking may also apply to camouflaging. Further, the replacement of gates with camo gates to facilitate camouflaging will almost always alter power consumption and area. With the right set of features extracted from a report (HLS report), a trained regression model could predict the parameter values with near-perfect accuracy. There is evidence of this in other EDA applications. For example, Zennaro~\etal{} combine an RTL generation framework with ML algorithms to estimate the area of the final design based on the features of an abstract specification. They retrieve the number of configurable logic blocks (CLBs) information from synthesised RTL design reports created from an automation framework. They mainly use a multilayer perceptron (MLP), but also experiment with random forests (RF) and gradient boosting (GB)~\cite{Zennaro2018}.

Another popular keyless approach is \textbf{split manufacturing}~\cite{imeson2013securing} where a design is split, each part is fabricated at one or more untrusted entities, and a trusted entity puts them together. Split manufacturing can have significant overheads and security concerns (e.g., proximity attacks~\cite{magana2017proximity}) if P\&R is performed poorly. Ad hoc and linear programming approaches to perturb P\&R~\cite{wang2018cat} and/or split designs~\cite{shi2018obfuscated} have met with some success, but can likely be improved further through the incorporation of AI-based modeling and prediction.

\paragraph{\textbf{Side Channel Attack Resistance}} \label{subsubsec: sidechannelresist}
Side channel leakage allows an attacker to break cryptographic systems. Signals that a cipher's implementation inadvertently emits allow for systematic extraction of underlying information. Examples include instantaneous power consumption, timing/delay, and electromagnetic emissions. Whether the attacker targets the understanding of device operation (simple) or uses leakage information to correlate data values (differential), there are generic countermeasures to reduce the effect of side channel leakage~\cite{markowitch_2011sca}. A high signal-to-noise ratio~(SNR) implies the availability of more meaningful leakage information. \textbf{Hiding} countermeasures reduce the SNR by either dampening the signal with low power designing and shielding, or flooding the leak-prone areas with noise using noise generators~\cite{levi2020ask}. To measure the value of hiding on a design, a predictive, regression-based model could provide an accurate percentage reduction in side-channel leakage. An obvious feature would be the tuning of the technique used. 

Another available countermeasure is \textbf{masking} which seeks to remove the correlation between input data and side-channel leakage ~\cite{prouff_2013_masking}. Unlike hiding, masking is applied at the algorithmic level with logic gates. Nonetheless, higher order side-channel analysis allows an attacker to bypass the routine masking schemes. A higher order masking may be employed but it may require significant overhead~\cite{prouff2009}. 
In order to evaluate the effectiveness of masking, various types of AI and ML algorithms have been employed to accomplish (a) the assessment of side-channel resiliency against a specific attack~\cite{picek2021sok}, and (b) the full leakage assessment methodology~\cite{moos2021dl}. 
However, both of these evaluation techniques need the actual ASIC or a correctly-programmed FPGA, as well as special equipment to be performed cf.~\cite{arribas2018vermi}. 
To overcome these obstacles, another line of research has been pursued, which is devoted to (1) theoretical conditions for a design to be SCA-resistant~\cite{barthe2015verified,coron2018formal,bloem2018formal} or (2) model the implementation for simulating its behavior~\cite{bertoni2017methodology,reparaz2016detecting} in the pre-silicon steps. 
In the first category, AMASIVE framework~\cite{zohner2012,huss2013amasive} can be mentioned that identifies hypothesis functions for leakage models cf.~\cite{buhan2021sok}. An example from the second class is VerMI, which is a verification tool in the form of a logic simulator that checks the properties of a class of masking methods (threshold implementation)~\cite{arribas2018vermi} at both algorithmic and implementation levels. 
Nevertheless, at this level of abstraction,~\ie{} pre-silicon, AI techniques have not found direct applications yet. Perhaps, the most straightforward way of incorporating such a technique can be imagined to examine whether the conditions for SCA-resiliency are fulfilled (the first category). 

Yet another step further, Ma~\etal{} presented a security-driven placement and routing tool to protect designs against EM side-channel attacks~\cite{CADSec2020} that attempts to break the balance of signal delays by register reallocation under the condition of layout constraints. The CAD-based tool, named CAD4EM-P, uses register reallocation and wire length adjustments to reduce the data dependency of EM leakage with acceptable area and power overheads~\cite{CADSec2020}. Opportunities to improve the effectiveness and efficiency of this approach can rely on more advanced forms of ML-for-prediction. For example, a graph neural network (GNN) can replace the graph-based algorithms that are the basis of the original version.

\paragraph{\textbf{Fault Injection Tolerance}}
In contrast to most SCA attacks, fault injection attacks involve an active adversary.
In other words, the adversary tries to observe a faulty behavior of the target device by forcing it to function outside of its specified operating range or feeding it undefined data. 
For example, an attacker can cause erroneous operation of the target platform by tampering with the supply voltage (\aka{} voltage glitching), modifying the clock signal's frequency (a.k.a glitching), or flipping bits in the memory using a laser beam~\cite{bar2006sorcerer}.
While most fault attacks require physical access to the victim device, recent studies have shown that in certain cases similar fault attacks can also be carried out remotely on a variety of systems~\cite{alam2019ram}.
Cryptographic devices and secure hardware have been the main targets of fault injection attacks. 
For example, an adversary could inject faults into an FSM implemented on a hardware platform to bypass authentication states and obtain unauthorized access to security-sensitive states.
On the other hand, an adversary may be able to extract the secret key by injecting faults into a cryptographic implementation and applying mathematical tools, like Differential Fault Analysis (DFA), on faulty generated ciphertexts.

Several protection/detection-based countermeasures have been proposed to mitigate the vulnerabilities of circuits against fault attacks.
While device-level countermeasures can be effective against fault injection attacks, the cost and extra manufacturing steps make the algorithmic and circuit-based countermeasures more attractive.
EDA tools can be deployed to improve the circuits' resiliency for such countermeasures.
However, conventional countermeasures are generic and create high overhead in terms of area and power.
Several protection/detection-based countermeasures have been proposed to mitigate the vulnerabilities of circuits against laser fault injection~(LFI) attacks.
While physical countermeasures, such as tamper-proof packaging and light sensors, can be effective, the cost and extra manufacturing steps make the algorithmic and circuit-based countermeasures more attractive.
For such countermeasures, EDA tools can be deployed to improve the circuits' resiliency.
However, conventional countermeasures are generic and not tailored to specific fault injection techniques.
For instance, by using triple-modular redundancy~\cite{lyons1962use}, error-detection/correction codes~\cite{karpovsky2004new,sunar2007sequential,tomashevich2014protecting, neumeier2012punctured}, and MAC tags/infective computation~\cite{CAPA,MandM}, the circuit becomes more resilient in general.
In these cases, however, depending on the capabilities of the adversary and how these countermeasures are synthesized, placed, and routed, they can still be bypassed by powerful fault attacks.
AI/ML can support creating circuit netlists and layouts during the design phase that are more resistant to fault injection attacks by taking the physical models of fault attacks into account.
For instance, by considering fault propagation together as a new feature, the impact of fault resilient placement and routing on security and overhead can be predicted. As shown in Table~\ref{table:performanceprediction}, GNNs and SVMs are viable algorithms for use.

\begin{table*}[t!]
\centering
\caption{Security requirements under performance prediction with applicable AI/ML algorithms. The cited options indicate optimization or security implementations available for reference}
\label{table:performanceprediction}
\arrayrulecolor{black}
\resizebox{\linewidth}{!}{
\begin{tabular}{|c|c|c|} 
\hline
\rowcolor[rgb]{0.361,0.769,0.769} \textbf{Design Stage }                     & \textbf{Security Task }   & \multicolumn{1}{c|}{\textbf{Applicable AI/ML Algorithms}}            \\ 
\hline
{\cellcolor[rgb]{1,0.859,0.604}}\textbf{Physical Design}                     & \begin{tabular}[c]{@{}c@{}}Security-aware Layouts and \\Tradeoff Modeling\end{tabular}   & \multicolumn{1}{c|}{Multiple Linear Regression} \\ 
\hline
{\cellcolor[rgb]{1,0.859,0.604}}   & \begin{tabular}[c]{@{}c@{}}Modeling Area, Power, Timing, and \\Security Impacts of IP Protection\end{tabular} & \multicolumn{1}{c|}{DNN~\cite{Pasandi2020}, GCN~\cite{chen2020}, MLP~\cite{Zennaro2018}}  \\ 
\hhline{|>{\arrayrulecolor[rgb]{1,0.859,0.604}}->{\arrayrulecolor{black}}--|}
{\cellcolor[rgb]{1,0.859,0.604}}   & Side Channel Attack Resistance                 & Polynomial Regression, GNN \\ 
\hhline{|>{\arrayrulecolor[rgb]{1,0.859,0.604}}->{\arrayrulecolor{black}}--|}
{\cellcolor[rgb]{1,0.859,0.604}}   & Fault Injection Tolerance & GNN, SVM                   \\ 
\hhline{|>{\arrayrulecolor[rgb]{1,0.859,0.604}}->{\arrayrulecolor{black}}--|}
{\cellcolor[rgb]{1,0.859,0.604}}   & Hardware Trojan Detection Rates                & SVM, Polynomial Regression \\ 
\hhline{|>{\arrayrulecolor[rgb]{1,0.859,0.604}}->{\arrayrulecolor{black}}--|}
\multirow{-5}{*}{{\cellcolor[rgb]{1,0.859,0.604}}\textbf{Cross-Abstraction}} & \begin{tabular}[c]{@{}c@{}}Vulnerabilities Across Design\\~Abstractions~and/or PDKs\end{tabular}              & GNN, RL~\cite{lu2021rl}           \\
\hline
\end{tabular}
}
\end{table*}

\paragraph{\textbf{Hardware Trojan Detection Rates}}
Malicious modifications,~\ie{} \textbf{hardware Trojans}, are possible during the IC design and fabrication process. Hardware Trojan detection has been extensively researched within the hardware security research domain~\cite{tehranipoor2010}. Salmani~\etal{} proposed a method of increasing the probability of Trojan activation using dummy scan FFs~\cite{salmani2012}. They place a dummy scan FF with a net having low transition probability. To test the concept, Trojan samples have to be deployed on benchmarks. The rate of Trojan detection could be derived from the available data that these simulations produce. Likely features that would be valuable for a prediction-based ML model include nets with dummy scan FFs, their fanouts and their corresponding transition probabilities. 

\begin{figure}[t!]
\centering
\includegraphics[width=0.5\columnwidth]{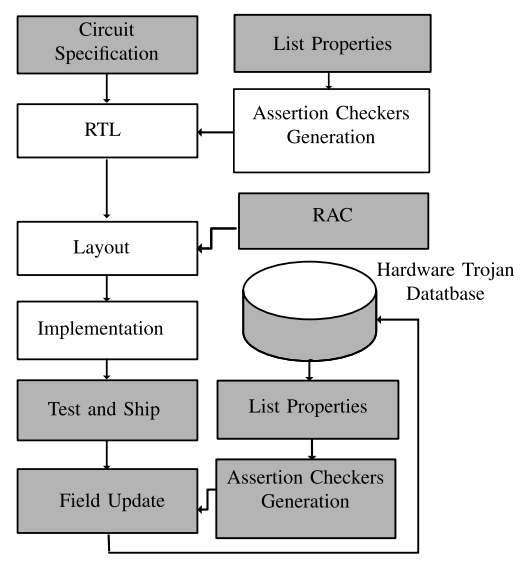}
\caption{Digital IC design flow with inserted assertion checkers. Reprinted from~ \cite{alsaiari2019}.}
\label{fig:assert}
\end{figure}

\textbf{Assertions} are typically used for formal verification of specification patterns. The behavior of circuits determines the correctness of the design. Alsaiari~\etal{} propose reconfigurable assertion checkers which are able to detect hardware Trojan updates on system-on-chips (SoCs)~\cite{alsaiari2019}. The design flow can be seen in Fig.~\ref{fig:assert}. Even with traditional assertion checkers where selection only occurs pre-synthesis, it is possible to ascertain the rate of Trojan detection using the assertion checkers, the Trojan type, as well as the point(s) of insertion.

\textbf{Built-in self-authentication (BISA)} relies on the incorporation of functional filler cells into white spaces at the layout level to form built-in self-test (BIST) circuitry~\cite{xiao2013bisa}. The latter is designed to verify that no BISA cell is tampered with by an untrusted foundry. The possible failure of the BIST circuitry allows inserted Trojans to be detected. To reduce the risk of compromising BISA design during manufacturing, \textbf{split-manufacturing} can also be applied. On its own, split manufacturing cannot be used to detect Trojans, but the technique known as ``obfuscated BISA'' (OBISA)~\cite{shi2018obfuscated} combines both BISA and split-manufacturing. OBISA's ability to detect hardware Trojans can be predicted by training a machine learning model.

\paragraph{\textbf{Vulnerabilities Across Design Abstractions and/or PDKs}}
There is evidence that designs become more prone to certain security vulnerabilities as they move across different levels of abstraction. One example provided in \cite{jiang2018} presented a timing side-channel vulnerability introduced during HLS to produce an optimized RTL representation. Another pointed out fault-injection vulnerabilities after logic synthesis due to FSM encoding styles and newly generated don't-care states~\cite{nahiyan2019}. It is important to have foreknowledge of these potential risks as they develop during the EDA process. This is possible  with predictive algorithms. In one instance, the data flow graph representation of the RTL could be combined with features from the HLS report to produce a GNN capable of identifying less-secure design areas. 
This may also apply to porting designs across different process design kits~(PDKs). The available gate sizes are discrete values which are usually specific to the underlying technology. Process variations may introduce parametric alterations which could make a device more susceptible to malicious parties. For this scenario, PDK electrical and design rules are possible input features for the same GNN described above. As the solution space scales exponentially with respect to the size of the netlist, gate sizing algorithms integrated into EDA tools rely on either heuristics or analytical methods, which leads to sub-optimal sizing solutions. To combat this shortcoming,~\cite{lu2021rl} has demonstrated the feasibility of applying RL algorithms equipped with GNNs that encode design and technology features.


\subsection{\textbf{Black Box Optimization or Design Space Exploration}}\label{subsec: blackbox}
\begin{figure*}[t]
   \includegraphics[width=0.85\textwidth]{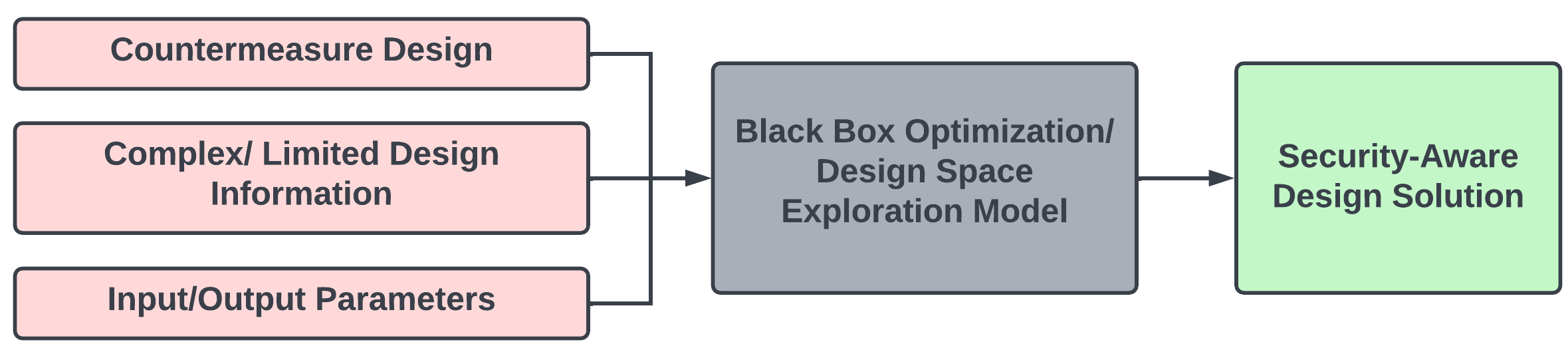}
   \centering
   \caption{General overview of black box optimization and design space exploration process.}
   \label{fig:blackbox}
\end{figure*}

For design space exploration, AI/ML seeks the solution that best meets requirements from available choices with little to no human intervention. The expanse of the search space creates the need for automation to execute selection, generation and evaluation of the solutions. However, the exhaustive option may not always be the optimal one, regardless of the availability of automation techniques. Design space exploration employs the application of different methods: stochastic optimization methods, like random search, evolutionary algorithms and black box optimization~\cite{CARDOSO201799}. Black box optimization uses a set of configurable parameters as inputs instead of results from the design stages, as seen in Fig.~\ref{fig:blackbox}. The circuit may either be too complex to model, or the engineer lacks details of the design to work with. At the training stage, most ML/AI algorithms are being optimized to find the best fit for input parameters, constraints, and design features. Hence, finding the most suitable design based on specifications may be the most intuitive application to the digital IC design.

\subsubsection{High Level Synthesis and/or Register Transfer Level}

\paragraph{\textbf{Crafting Side Channel-resilient Circuits}}
Referring to Section~\ref{subsubsec: sidechannelresist}, masking has been introduced to stop an attacker from mounting side channel analysis. 
AI has become an integral part of this, for instance, RL has been applied to combine a set of countermeasures, which leads to the enhancement of target resilience to, at least, some types of attacks \cite{rijsdijk2021reinforcement}. 
Seen from another perspective, machine learning algorithms, and in particular, NNs, can defeat some countermeasures, including masking~\cite{kim2019make,related_works:automated_hyperparameter_tuning,related_works:SCA_metrics}. 
As an example of such attacks, RL has been employed to determine proper side-channel leakage models and training of neural networks to extract leakage profiles~\cite{DBLP:journals/iacr/RijsdijkWPP21}.
Hence, robustness against such attacks has become one of the objectives of the tests conducted to evaluate the effectiveness of countermeasure, specifically, masking. 

Given the nature of such tests, the security-aware designer may have to deal with the higher levels of abstraction corresponding to a black box leakage model because there is no information, including physical placement or routing signals \cite{buhan2021sok}.
Therefore, masking schemes may be implemented, but there is no guaranteed measure of effectiveness before the RTL stage.
To enhance this process, full access to the design's layout available to the implementing party could be useful and even be bolstered with an understanding of the device's functionality via RTL or gate-level implementation. 
This is in line with the studies focused on modeling the implementation and simulating its behavior as discussed before~\cite{bertoni2017methodology,reparaz2016detecting}. 
In this context, AI has not yet reached its full potential as security-critical implementations are still designed manually.  
The manual design and implementation pose a series of drawbacks stemming from (1) the lack of integration between the various designs, (2) erroneous designs due to human errors, and (3) high consumption of resources, including time (both execution and design times), silicon area and/or power/energy consumption.

\subsubsection{Physical Design}

\paragraph{\textbf{Porting of Cryptographic and Hardware Security Primitives to Different Technology Nodes}}
Security-aware updates to designs are hardly ever direct ``out-of-box'' additions to devices. For integration of an IP or hardware security primitive, it is imperative that the design still fulfils its intended purpose. Selecting the most appropriate parameters for a hardware security primitive is one aspect, but another is porting it to a different technology node~\cite{keating_bricaud_2007}. Although technology nodes no longer categorically correspond to transistor gate length and half pitch, they nonetheless vary considerably with respect to manufacturing processes, design rules, noise sensitivity,~\etc{}~\cite{bagnato_indrusiak_quadri_rossi_2014}. These variations are noteworthy because they may disrupt the transferability of primitives. For example, an optimized PUF applied to an IC manufactured using the 28~nm process is not guaranteed to transfer to the 22~nm process due to factors like geometric variation and transistor type~\cite{keating_bricaud_2007}. The knowledge of process-specific parameters allows for assessment of the hardware primitive porting activity using ML. It is possible to use a classifier model as simple as an SVM to select the most suitable match between circuit and primitive without explicit knowledge of the design's functionality.

\paragraph{\textbf{Anti-Counterfeit and Anti-Tamper Sensor Optimization}}
The scourge of IC counterfeiting continues to plague the semiconductor industry by flooding the consumer market with unauthorized, defective and inferior versions of original chips and designs. From as far back as 2006, electronic companies have missed out on \$100 billion of revenue due to counterfeiting~\cite{pecht2006}. The recent global chip shortage has created the avenue for these illegal devices due to demand exceeding current supply chain capabilities~\cite{leprince-ringuet_2021}. Recycled and remarked counterfeit ICs account for more than 80\% of the counterfeits sold worldwide~\cite{guin2014}. Hence, their detection has become a necessity to tackle IC counterfeiting. Recycled counterfeits are used chips fraudulently sold as new while remarked counterfeits are inferior chips sold as higher grades. 

\begin{figure*}[t]
   \includegraphics[width=0.85\textwidth]{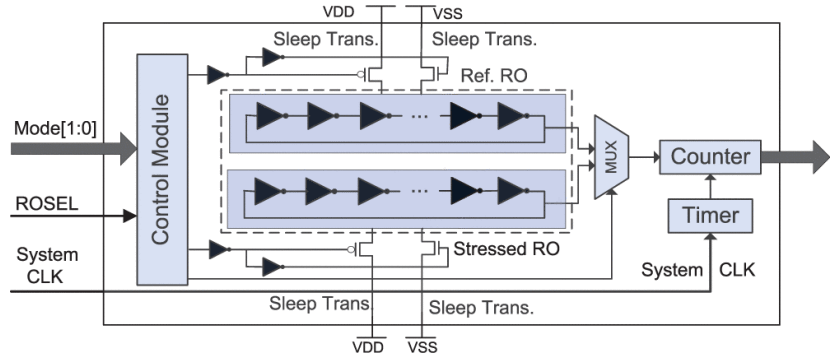}
   \centering
   \caption{Structure of RO-based sensor reprinted from~\cite{zhang2014}}
   \label{fig:rosensor}
\end{figure*}

Different sensors have been designed as a solution to these counterfeits, as seen in the work by Zhang~\etal{}~\cite{zhang2014} and shown in Fig.~\ref{fig:rosensor}. Their first of two proposed techniques uses frequency difference retrieved from paired ring oscillators (ROs) between an original and recycled chip to distinguish the two. RO frequency tends to degrade more with aging because the transistors in the RO gets slower with time~\cite{docking2003ro}. Guin~\etal{} also add anti-fuses and fuses for combating die and IC recycling (CDIR). Their RO-CDIR implementation is negative temperature instability-aware, which is valuable for flagging ICs with shorter usage~\cite{guin2014}. To date, the design of these sensors such as number of inverting stages, choice of threshold voltage, and overall structure are chosen in an ad hoc manner or through empirical means. Further, foundries do not always provide aging models to users. Both of these limits the age estimation of such sensors and their classification accuracy.
AI approaches, such as Bayesian optimization, would be preferable, especially for its ability to optimize black-box functions. Automating the optimization process will ultimately make it less-exhaustive.

Another source of hardware security issues is tampering. Once the device is physically compromised, it may leak on-chip information to the attacker. Hence, insertion of hardware Trojans may also be considered as tampering~\cite{tehranipoor2010}. One common example of tampering is FIB-based circuit edit, which allows an attacker to remove or bypass a security countermeasure~\cite{livengood_tan_hack_kane_greenzweig_2011}. Similar to recycling, sensors have been proposed to combat or detect tampering. A technique proposed by Liu and Kim uses logic-compatible embedded Flash (eFlash)-based tamper sensor built on an exposed floating gate (FG) structure to detect physical attacks~\cite{Liu2019}. Any change in charge stored on the exposed FG can be detected by this sensor. This includes humidity, high temperature, dust particles, chemicals, and electrostatic charges. A regression-based model or a neural network could fine-tune the configuration of such tamper sensors.

\paragraph{\textbf{Hiding Countermeasure Optimization}}
If an attacker is aware that a sensor has been deployed to detect a certain attack modality, the attacker may start by circumventing the sensor. For example, the attacker may apply the minimum possible mechanical force on the metal wires, or use a low intensity light source for a longer duration of time. Assuming maximum throughput possible, it is crucial to supplement successful optimization with robust defense. Hiding can dampen the signal accessible to the attacker. With limited knowledge of the sensor-IC design and operation, the designer could possibly rely on ensemble methods to find the most suitable hiding framework. The same sensors used to protect against tampering and detect counterfeits may also be vulnerable to side channel attacks. Since they are sensitive to changing parameters like voltage and frequency, they are prime targets for malicious parties. Sugarawa~\etal{} show that an attacker can reveal the internal state of a chip by observing how a sensor reacts to laser fault injection. The leakage leads to a feasible, non-invasive, probing attack. The sensor-type, the bit-flip detector, detects a short-circuit current induced by a laser fault injection~\cite{SUGAWARA201963}. Thus, it is advisable to supersede sensor optimization with robust hiding schemes; the ideal framework could combine both activities to boost resistance to side channel attacks. For example, if a noise generator is used as the default hiding mechanism, a simple regression model could predict the amount of noise required to achieve a lower signal-to-noise ratio (SNR). A more versatile system could have options like shielding in addition to noise generators, with decisions and settings also based on countermeasure optimization.

\begin{table*}[t!]
\centering
\caption{Security requirements under black box optimization or design space exploration with applicable AI/ML algorithms. The cited options indicate optimization or security implementations available for reference}
\label{table:blackbox}
\arrayrulecolor{black}
\resizebox{\linewidth}{!}{
\begin{tabular}{|c|c|c|} 
\hline
\rowcolor[rgb]{0.361,0.769,0.769} \textbf{Design Stage }             & \textbf{Security Task }  & \textbf{Applicable AI/ML Algorithms}                     \\ 
\hline
\multicolumn{1}{|l|}{{\cellcolor[rgb]{1,0.859,0.604}}\begin{tabular}[c]{@{}>{\cellcolor[rgb]{1,0.859,0.604}}l@{}}\textbf{High Level Synthesis and}\\\textbf{Register Transfer Level}\end{tabular}} & Crafting Side Channel-resilient Circuits      & RL~\cite{DBLP:journals/iacr/RijsdijkWPP21}, ANN                 \\ 
\hline
{\cellcolor[rgb]{1,0.859,0.604}}                & \begin{tabular}[c]{@{}c@{}}Porting of Cryptographic and\\Hardware~Security~Primitives to\\Different Technology~Nodes\end{tabular} & SVM            \\ 
\hhline{|>{\arrayrulecolor[rgb]{1,0.859,0.604}}->{\arrayrulecolor{black}}--|}
{\cellcolor[rgb]{1,0.859,0.604}}                & \begin{tabular}[c]{@{}c@{}}Anti-Counterfeit and Anti-Tamper \\Sensor Optimization\end{tabular}               & \begin{tabular}[c]{@{}c@{}}Bayesian Optimization,~ ANN, \\Multiple Linear Regression\end{tabular}  \\ 
\hhline{|>{\arrayrulecolor[rgb]{1,0.859,0.604}}->{\arrayrulecolor{black}}--|}
\multirow{-3}{*}{{\cellcolor[rgb]{1,0.859,0.604}}\textbf{Physical Design}}                & \multicolumn{1}{l|}{Hiding Countermeasure Optimization}            & \begin{tabular}[c]{@{}c@{}}Bayesian Optimization,~ \\Ensemble Learning\end{tabular}                \\ 
\hline
{\cellcolor[rgb]{1,0.859,0.604}}\textbf{Cross-Abstraction}           & Obfuscation for Black Box Designs             & Genetic Algorithms~\cite{chen2019genunlock}, ViT                    \\
\hline
\end{tabular}
}
\end{table*}

\subsubsection{Cross-Abstraction}

\paragraph{\textbf{Obfuscation for Black Box Designs}}
Obfuscation is best implemented with complete structural and functional understanding of a circuit, but different levels of abstraction may present the challenge of having only one available. At the gate level, logic locking, as described in Section~\ref{par: dectuning, obfus}, may be viable, but the implementing party may also be restricted by their limited knowledge. Here again, a GNN seems capable of modeling the complex connections within the design, the competing needs to resist attacks, and narrowing down the most suitable implementation for balancing security and overhead. Ideally, the complete framework should provide locking options for each design and generate keys for a user discretely. Genetic algorithms have been applied to logic locking attacks like GenUnlock~\cite{chen2019genunlock}, where the evolutionary process is a black-box procedure overseen by an objective function. Conversely, this concept may be applicable to not only logic locking, but general obfuscation.

The physical design stage also presents opportunities for obfuscation to the designer with black box access using ML/DL. The current state-of-the-art in camouflaging is the covert gate approach~\cite{shakya2019covert}. Covert gates are inserted randomly into the design/layout, but a CNN or vision-based transformer (ViT) can be incorporated to quantify the impact of insertion and ultimately identify the most viable gates to replace. Table~\ref{table:blackbox} provides possible algorithms for the security tasks provided in this subsection.


\subsection{\textbf{Assimilation of Security Rules for Electronic Design Automation}}\label{aieda}
The incorporation of ML/AI into EDA describes replacing traditional heuristics with deep learning and AI-based decision-making policies that integrate the above steps (decision making, performance prediction, and optimization) into a single EDA framework. This will ultimately lead to a custom-based EDA pipeline based on the specific design and/or security application rather than relying on generic, somewhat-archaic EDA tools. Unlike  Section~\ref{subsec: blackbox}, the design space may stretch beyond what a designer has at their disposal. A valuable point to make is the possibility of utilizing a trainable decision-making policy in contrast to exploring the available decisions. The use of AI/ML almost guarantees more efficient results with shorter runtimes. There is budding research within the IC design community to incorporate learning-based tools into the EDA pipeline. 
The OpenROAD project is a DARPA-funded initiative which seeks to solve EDA cost and turnaround time. The goal is to complete the design process without performace, power or quality tradeoffs using ML, cloud-based optimization and other techniques~\cite{ajayi2019openroad,connor_2018}. Their work so far has incorporated problem-specific tools like ioPlacer for fast and scalable input/output pin assignment~\cite{bandeira2020ioPlacer}, TritonCTS for clock tree synthesis~\cite{Han2020OptimalGH} and RTL-MP for macro placement using RTL information and clustering~\cite{kahng_varadarajan_wang_2022}. A team of researchers at Google also presented a floorplanning tool for the physical design stage that can successfully place macro blocks using deep RL~\cite{Mirhoseini2021AGP}. Its results demonstrated improvement over traditional manual efforts across performance, power consumption and area, and allowed its application to a recent tensor processing unit (TPU) product.

With AI/ML firmly established in the pipeline, design optimization and security could be combined. Design-specific EDA will make extraneous steps obsolete and ultimately reduce the time-to-market for ICs. Table~\ref{tab: aieda} highlights the areas of interest for each major design stage.

\begin{figure*}[t]
   \includegraphics[width=\textwidth]{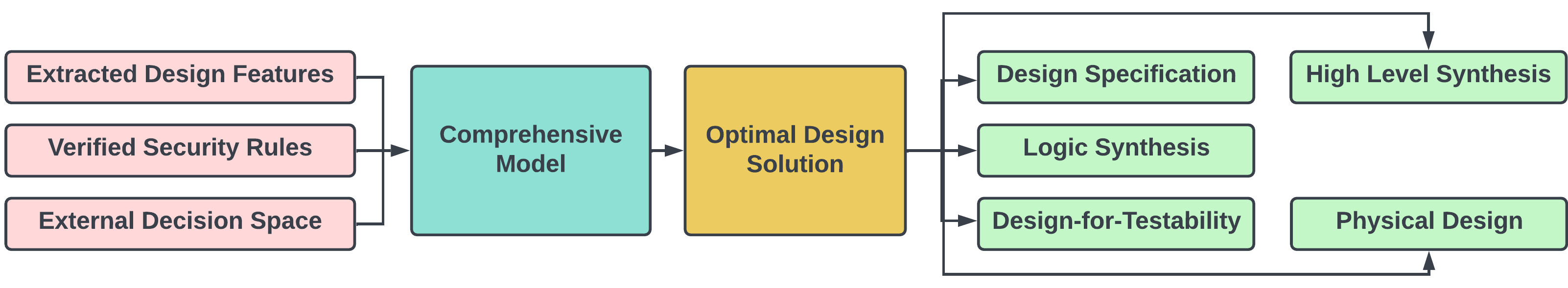}
   \centering
   \caption{General overview of AI-assisted EDA. The emphasis is on direct application to EDA.}
   \label{fig:aieda}
\end{figure*}



The manufacture of secure ICs must begin with embedding of security rules at major steps of the design process. This encompasses the acts of identifying, assessing and incorporating prerogatives and properties within specific stages. All design steps follow defined rules during their series of actions on the input design~\cite{khwaja1997edarule}. For example, during logic synthesis, minimization may occur by combining gate terms where applicable to reduce circuit size~\cite{brayton1990logicsyn}. Even design constraints can be described as rules because they define the behavior of the design at each point of usage. In the same vein, it is necessary to apply security rules as rigorous requirements throughout the entire design sequence. 

Efforts to support such assimilation are starting to grow. For example, the Hardware Common Weakness Enumeration Special Interest Group (HW CWE SIG)~\cite{CWE_MITRE} consists of researchers and representatives from organizations operating in hardware design, manufacturing, and security who interact, share opinions and expertise, and leverage each other’s experiences in defining hardware security weaknesses and identifying them with analysis tools. However, the current initiative is manual and quite subjective. Data-driven and AI-based approaches could make this less tedious and more objective in nature.



\subsubsection{High Level Synthesis and Register Transfer Level}
The overall design of a circuit depends on the requirements stated in its specifications. Traditional execution of HLS results in an RTL description of the design, which is a better hardware-based representation of the intended IC. During the design process, it is difficult to ascertain certain features and how they may impact the entire circuit, but significant work has been done to mitigate this using ML~\cite{Makrani2019,Zennaro2018, Dai2018}. One noteworthy ML-based optimization framework combines a GNN with RL to predict design results and obtain optimal solutions under user-specified constraints and objectives like area and latency~\cite{wu2021ironman}. The security-aware designer can also leverage ML to anticipate and solve security problems that an IC is vulnerable to. Security rules at this early stage of the design process are less-tedious and reasonably-inexpensive to enforce.

\paragraph{Obfuscated Design} Pre-synthesis obfuscation is applied to a design prior to the logic synthesis stage. Although the most prevalent techniques are used post-synthesis, there has been work done on this early stage method of obfuscation. Obfuscation can be implemented at the high level stage before conversion into RTL. One group uses MUXes driven by a locked controller units to create decoy connections~\cite{rajendran2015}. Exploiting the MUX's use of inputs and a control signal, only the correct control signals can unlock the design. The authors of \cite{Chakraborty2010} convert RTL code into control and data flow graph (CDFG) form, apply locking by superimposing an authentication FSM, and then return to RTL. Both applications may be deployed as security rules with deep learning-based execution. The CDFG form could be implemented as a GNN for optimum modeling of a design's complexity.

\paragraph{Masking} Masking to mitigate the effectiveness of differential power analysis (DPA) during a side channel attack is also possible at this level of abstraction. Konigsmark~\etal{} create an HLS flow which addresses side-channel leakage by inferring security critical operations from user-specified confidential variables in HLS input. Their modified framework rectifies imbalanced branches that pose easy attack targets, and changes functional units based on leakage potential~\cite{konigsmark2017}. This experiment could be the foundation for learning-based masking solutions to prevent SCA; it stands to be improved by feature-based rectification. 

\paragraph{Information Flow Security} It is important to verify that systems adhere to information flow security (IFS) policies, especially during the design phase. All assets flowing through the system must be identified; this is known as information flow tracking (IFT)~\cite{hu2021}. Achieving this at higher levels of abstraction reduces the effort required to rectify potential problems. One publication presents RTLIFT as a means of measuring all digital flows through RTL designs to formally prove security properties related to integrity, confidentiality and logic side channels~\cite{Ardeshiricham2017}. Perhaps its most impressive property is the ease of integration into the design flow. Intelligent decisions surrounding security policies means adaptability to different designs trade-offs such as the RL-based policy iteration to calculate digital flows and possibly flag irregularities~\cite{Ardeshiricham2017}.

\subsubsection{Logic Synthesis}
In general, logic synthesis has been extensively-researched under AI/ML~\cite{Neto2019LSOracleAL, pasandi2019approximate,Pasandi2020}. The RTL-to-gate level process provides viable data to train models that have mostly been used for design optimization. The conversion from behavioral level to the gate level is the bridge between a design idea and its generation because the specific logic functions are implemented from combinations of gates selected in a given cell library.


\paragraph{Obfuscated Design} 
At the gate-level, the first form of obfuscation which may come to the security-aware designer's mind is logic locking. One proposed work that could be the baseline for future AI/ML exploration is LeGO~\cite{alaql2021}. The authors point out that it can iteratively harden a design against a set of attacks. The framework begins by performing a simple key-based locking and continues with an iterative feedback process which considers a set of possible attacks to defend against~\cite{alaql2021}. The attacks are provided with corresponding countermeasures for integration. The rule selection algorithm, which uses prior knowledge from a database of mitigation rules, identifies the best protection to be applied for each key-bit against a specific attack. The "convergence point" occurs when all security issues are addressed~\cite{alaql2021}. The learning design could be improved with an RL-based policy iteration framework. With a suitable representation of the gate-level netlist and its features, there are many possibilities.


\paragraph{Masking} 
Similarly, in the quest for side-channel resistant designs, gate-level masking has been researched since the early 2000s. Masking is most ideal for implementation at this stage, because of the gate design's correlation to the physical design's functionality. In 2005, masked dual-rail pre-charge logic (MDPL) was proposed as a promising DPA-resistant logic style~\cite{Popp2005}. MDPL circuits are based on a standard cell libraries and have no routing constraints concerning the balancing of complementary wires. The dual-rail pre-charge property of MDPL specifically prevents glitches in the circuit. A more recent implementation uses a technique that combats the phenomena of glitches and early propagation using only cell-level “don’t touch” constraints~\cite{Leierson2014}. Their framework, a derivative of MDPL called LUT-Masked Dual-rail with Precharge Logic (LMDPL), can works for both FPGA or and ASIC designs. MDPL may be used as a rule-based template for masking gate-level designs. A GCN can be used to identify sections of the circuit that are vulnerable and would require MDPL implementation.

\begin{figure*}[t] 
   \includegraphics[width=0.9\textwidth]{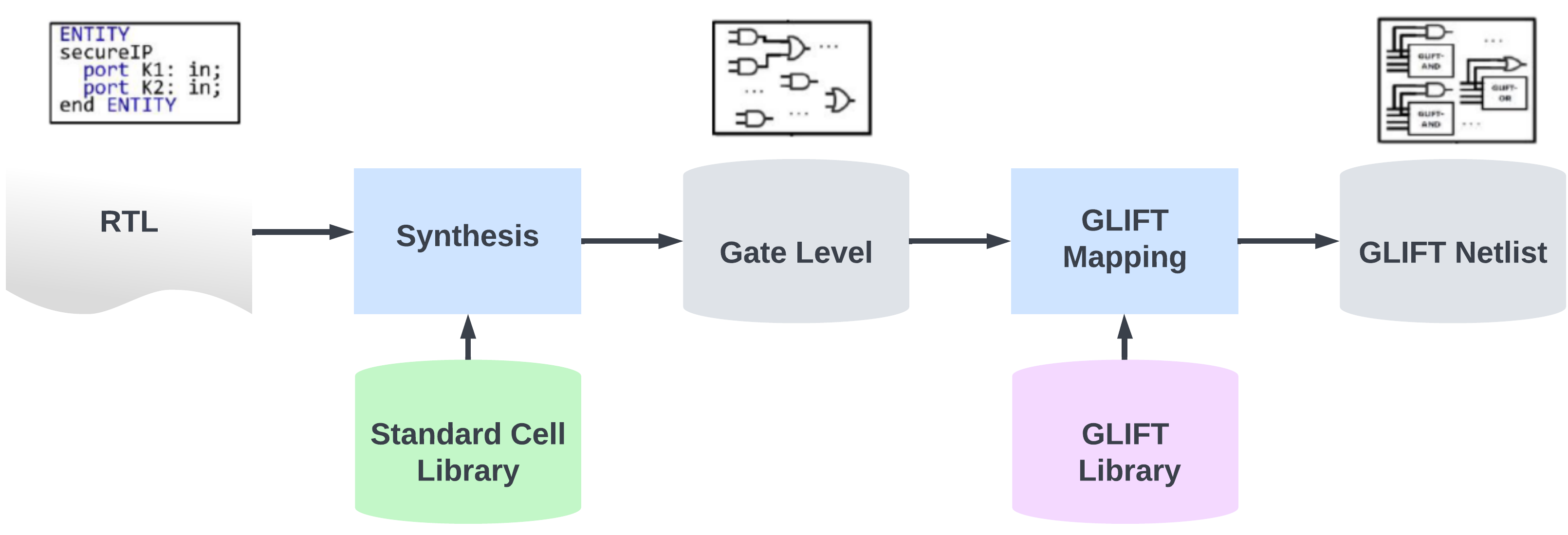}
   \centering
   \caption{GLIFT workflow with standard cell and GLIFT libraries.}
   \label{fig:glift}
\end{figure*}

\paragraph{Information Flow Security} 
Information flow tracking at the gate level also serves to check adherence to security policies. The well-defined gate level information flow tracking (GLIFT) technique has been widely used to GLIFT used to design secure hardware architectures and detect security violations from visible timing flows. Hardware trojan dection has even been accomplished based on harmful information flows that point to sensitive information leakage and data integrity violation~\cite{nahiyan2017trojandetection}. From Fig.~\ref{fig:glift}, the RTL synthesis which produces the gate level representation is followed by standard cell augmentation with shadow logic gate. Design stage verification is the most suitable application of GLIFT to the EDA pipeline. The technique only checks for violation of IFS policies without addition of logic to the fabricated design. Once again, a GCN representation with sufficient features may be used to supplement, or even replace GLIFT.

\paragraph{Side Channel Resistance}


\subsubsection{Device Sizing and/or Technology Mapping}

Modern physical design (PD) flow includes gate sizing for PPA optimization, used in different steps ranging from synthesis to signoff. 
This process is algorithmic, assigning a proper size (gate type) to each optimizable design instance from a set of equivalent standard cell libraries for a different process, voltage, and temperature (PVT) corners. 
For each of the instances, the available gate sizes are discrete values specific to the underlying technology. 
As the solution space scales exponentially with respect to the size of the netlist, gate sizing algorithms integrated into EDA tools rely on either heuristics or analytical methods, which leads to sub-optimal sizing solutions. 
To combat this shortcoming,~\cite{lu2021rl} has demonstrated the feasibility of applying RL algorithms equipped with graph neural networks (GNNs) that encode design and technology features. 
Specifically, the problem of gate sizing for timing optimization at the post-route stage has been the main focus of the study. 
Although the proposed sizer adopts a more global optimization approach, it does not always outperform the commercial tool. 
Nevertheless, it has the great advantage that is automatic gate sizing for timing optimization without any human intervention. 

The technology-mapping step converts internal representation and the optimized logic based on the designated technology library~\cite{kaufmann2008techmap}.
In order to improve the effectiveness of this process in terms of energy consumption, delay, and area overhead, approximate logic synthesis (ALS) is discussed in the literature, where the accuracy of mapping is compromised to meet the design requirements. 
However, techniques for ALS suffer from long execution time, which has been addressed in~\cite{pasandi2019approximate} by adopting RL-based strategies. 
Academic state-of-the-art ALS tools are considered to evaluate the performance of the proposed RL-based ALS. 
The results have demonstrated a great deal of improvement with regard to run-time and area and delay when the  RL-based ALS is applied against academic and industrial benchmarks. 
Security-based optimization can also be added as a design goal for this automatic process.


\paragraph{Fault Tolerance} Critical assets in devices may be compromised by taking advantage of errors from fault injection. Single event upsets (SEUs) affect sequential elements, while single event transients (SETs) arise from combinational logic~\cite{BOGAERTS2014250}. An SEU may lead to a single event latchup (SEL), which results in a high operating current, or a short circuit. Fault tolerance reduces the sensitivity of devices to particle or laser attacks that cause these events. Glitches can be used to progressively break cryptography schemes or disable internal protection mechanisms~\cite{giraud2010}. Redundant computation is one way of making designs impervious to upsets. It uses double computation (in time or space) with a final comparison; if the results from redundant computations do not match, the output of the encryption function is suppressed/randomized. Redundancy may cause increased area and power consumption, but methods like the one presented in~\cite{petrovic2012} show high tolerance with moderate overhead is possible. The optimization procedure can be refined using a Gaussian process to find the best trade-off value.


\subsubsection{Floorplanning and Physical Design}
The floorplanning step formalizes the high level design established at the beginning of the process, and is followed by P\&R to connect all blocks such that they meet design criteria and constraints. After initial P\&R attempts, the design’s timing constraints are analyzed~\cite{martin_scheffer_lavagno_2016}. If unsatisfactory, the P\&R software in use tries different placements and signal routing to try to meet the designated constraints. Even outside academia, work has been done on automating floorplanning to benefit the physical layout. Researchers at Google modeled chip floorplanning as a reinforcement learning (RL) problem, and used a GCN representation of the design to optimize power, area and performance~\cite{Mirhoseini2021AGP}. 

In order to place Application-Specific Integrated Circuits (ASICs), designers must deal with millions or even billions of gate-level instances to be placed on constrained physical layouts. 
The placement directly impacts the quality of the final full-chip design, supposed to be assured by employing commercial EDA tools and spending a significant amount of time in optimization iterations cf.~\cite{lu2020vlsi}. 
In a series of work, the problem of providing automated and accurate placement guidance has been tackled through machine learning methods~\cite{lu2021law,lu2020vlsi,mirhoseini2020chip,fogacca2019finding}. 
Among these proposals,~\cite{mirhoseini2020chip} has reintroduced reinforcement learning into the chip placement domain of study. 
\cite{goldie2020placement} provides roadmaps for placing an ASIC or FPGA netlist onto a grid, whereas~\cite{mirhoseini2020chip} gives more concrete answers to the problem of placing a netlist graph of macros (e.g., SRAMs) and standard cells (logic gates, such as NAND, NOR, and XOR) onto a chip canvas. 
The goal of these studies is to perform chip placement such that power, performance, and area are optimized according to the constraints on placement density and routing congestion. 

Global routing is performed on a coarse grid map to find an approximate routing of all nets that is the basis for the detailed routing to specify the routing of each net and satisfy all design rules. 
Another pressing issue with electronic systems design is automatic routing considering wire-length, crosstalk, via, and layer selection. 
Maze routers are traditionally used for this purpose, although they have two major limitations, namely (1) being not smart enough to perform detailed routing considering signal integrity~\cite{hart1968formal}, and (2) needs for additional constraints and optimization variables to offer global routing. 
RL has been proposed to resolve these by integrating the concept of SI~\cite{kim2020reinforcement}. 
For this, Kim et al. have designed an RL-agent including an encoder-decoder model that is an effective tool for solving combinatorial optimization problems~\cite{kim2020reinforcement}. 
Similarly, deep Q-network (DQN) agents are trained in the RL framework to encounter the global routing problem~\cite{liao2020deep}, where a single DQN conjointly routs the nets and pins in an IC. 
In this process, the DQN acts as an agent and interacts with the environment. 
Upon receiving the state information from the environment, the Q-values of all the potential next states are evaluated, an action is taken, and executed. Consequently, the environment is updated.

These optimization-based techniques provide possibilities for applications towards more security-aware designs.

\paragraph{Crosstalk Mitigation} Design scaling from a technology process may reduce spacing between adjacent interconnects; the resulting increased coupling capacitance between wires causes crosstalk~\cite{zhou1998crosstalk}. The affected signal's value may either change, or have delayed signal transition. False clocking, where a clock changes states unintentionally, may also occur~\cite{terapasirdsin2010crosstalk}. Good signal integrity is necessary for optimum functionality of an IC~\cite{green1999}. Large signal-integrity problems may cause intermittent success or complete system failure. To address crosstalk, the design tool must be able to measure its value. The authors of \cite{chang2006} find a closed form expression of crosstalk using a number of wires. With accurate metrics, a linear regression model coupled with the RL method applied in~\cite{Mirhoseini2021AGP} could address crosstalk expeditiously.

\paragraph{Power Distribution Networks for Anti-Side Channel and Fault Injection} Side channel leakage is one of the results of the nature of an IC's power distribution network (PDN). The PDN is responsible for the supply of stable voltage and power to each functional component of the circuit~\cite{martin_scheffer_lavagno_2016}, but it may also inadvertently introduce side-channel leakages throughout the design~\cite{zhang2020pdnsca}. The growing complexity of devices introduces sophisticated PDNs with the potential to allow successful side-channel attacks or fault injection. To improve the security of a design, it is important to use ML to build on existing work on this topic. PowerScout is a framework geared towards evaluating side-channel vulnerabilities by modeling possible attacks to improve PDN designs~\cite{zhu2020}. It can effectively predict leakage strength and also identify fault injection soft spots. The side channel and fault injection attacks are modeled using non-learning based design space exploration. A decision tree, or an MLP could predict these with the same parameters being extracted. There may also be promise in modeling the PDN using a graph-based network and optimizing leakage-prone nodes for the physical design.

\paragraph{Anti-Tamper Layouts} Anti-tamper designs are meant to deny an adversary’s opportunity to monitor or affect the correct operations of an IC. At the layout level, one CAD assessment technique allows a designer to gauge a designs susceptibility to frontside probing attacks~\cite{Wang2019}. The assessment is performed by varying attack parameters (FIB aspect ratios and angles), shield parameters (layers and shape) and the assets under attack (keys, buses,~\etc{}.). Two of the most useful aspects of the paper are the exposed area (EA), which is free to probe without impacting transmission, and shield structure taxonomy, which is a set of documented shield patterns~\cite{Wang2019}. Based on the attack parameters, EA and the shield taxonomy, an RL model could help recommend the best shield design for a layout.

\paragraph{Anti-Reverse Engineering Layouts} At the layout level, IC camouflaging is a EDA engineer's most suitable option to prevent successful RE. The camouflaged layout is a functional replica of the original design with camo gates (See ~\ref{subsection: decisionmaking}. A novel framework for secure layout generation could focus on generating camo gates using GANs. Recommendations for camouflage points on an IC can be modeled using a GCN or a CNN.

\paragraph{Custom eFPGA fabrics} The embedded FPGA (eFPGA) is an IP core integrated into an ASIC that offers the reprogramming flexibility of FPGAs without their overhead costs. They are added to systems which require frequent updates; by adding their fabric to an SoC or ASIC, teams are able to modify applications seamlessly. One security evaluation in \cite{Bhandari2021} points out that eFPGAs can contribute to SAT attack resilience. The eFPGAs have combinational loops that a specific cyclic SAT solver like CycSAT~\cite{zhou2017} cannot solve. Increasing attack difficulty can be considered when creating eFPGA fabrics at the physical design stage. The incorporation of a policy-based model that incorporates this into an ASIC design could improve security considerably.

\begin{table}
\centering
\caption{Major IC design stages in EDA with security needs and possible applications of AI/ML.}
\arrayrulecolor{black}
\resizebox{\linewidth}{!}{
\begin{tabular}{|c|c|c|c|c|} 
\hline
\rowcolor[rgb]{0.361,0.769,0.769} {\cellcolor[rgb]{0.361,0.769,0.769}}        & {\cellcolor[rgb]{0.361,0.769,0.769}}              & {\cellcolor[rgb]{0.361,0.769,0.769}} & \multicolumn{2}{c|}{\textbf{AI/ML Algorithms Applicable}}            \\ 
\hhline{|>{\arrayrulecolor[rgb]{0.361,0.769,0.769}}--->{\arrayrulecolor{black}}--|}
\rowcolor[rgb]{0.361,0.769,0.769} \multirow{-2}{*}{{\cellcolor[rgb]{0.361,0.769,0.769}}\begin{tabular}[c]{@{}>{\cellcolor[rgb]{0.361,0.769,0.769}}c@{}}\\ \textbf{Design Stage}\end{tabular}}   & \multirow{-2}{*}{{\cellcolor[rgb]{0.361,0.769,0.769}}\textbf{Security Task}}            & \multirow{-2}{*}{{\cellcolor[rgb]{0.361,0.769,0.769}}\textbf{Design Type}} & \textbf{Suggested}         & \textbf{Implemented} \\ 
\hline
{\cellcolor[rgb]{1,0.859,0.604}}        & Obfuscated Design              & \multirow{3}{*}{ASIC, FPGA}          & RL      & GNN~\cite{rozhin_2021}   \\ 
\hhline{|>{\arrayrulecolor[rgb]{1,0.859,0.604}}->{\arrayrulecolor{black}}-~--|}
{\cellcolor[rgb]{1,0.859,0.604}}        & Masking     &                   & GNN, GCN, RF               &   \\ 
\hhline{|>{\arrayrulecolor[rgb]{1,0.859,0.604}}->{\arrayrulecolor{black}}-~--|}
\multirow{-3}{*}{{\cellcolor[rgb]{1,0.859,0.604}}\begin{tabular}[c]{@{}>{\cellcolor[rgb]{1,0.859,0.604}}c@{}}\textbf{High Level Synthesis and}\\\textbf{~Register Transfer Level}\end{tabular}} & Information Flow Security      &                   & RL      &   \\ 
\hline
{\cellcolor[rgb]{1,0.859,0.604}}        & Obfuscated Design              & \multirow{4}{*}{ASIC, FPGA}          & GCN     & ~ ~ANN~\cite{Sisejkovic2021-challenge-logic-locking}            \\ 
\hhline{|>{\arrayrulecolor[rgb]{1,0.859,0.604}}->{\arrayrulecolor{black}}-~--|}
{\cellcolor[rgb]{1,0.859,0.604}}        & Masking     &                   & RL, GNN &   \\ 
\hhline{|>{\arrayrulecolor[rgb]{1,0.859,0.604}}->{\arrayrulecolor{black}}-~--|}
{\cellcolor[rgb]{1,0.859,0.604}}        & Information Flow Security      &                   & GCN     &   \\ 
\hhline{|>{\arrayrulecolor[rgb]{1,0.859,0.604}}->{\arrayrulecolor{black}}-~--|}
\multirow{-4}{*}{{\cellcolor[rgb]{1,0.859,0.604}}\textbf{Logic Synthesis }}   & Side Channel Resistance        &                   & GNN     &   \\ 
\hline
{\cellcolor[rgb]{1,0.859,0.604}}\begin{tabular}[c]{@{}>{\cellcolor[rgb]{1,0.859,0.604}}c@{}}\textbf{Device Sizing/ }\\\textbf{Technology Mapping}\end{tabular}               & Fault Tolerance                & ASIC              & Bayesian Optimization, DNN &   \\ 
\hline
{\cellcolor[rgb]{1,0.859,0.604}}        & Crosstalk Mitigation           & \multirow{5}{*}{ASIC, FPGA}          &         & \begin{tabular}[c]{@{}c@{}}Logistic Regression~, \\XGboost~\cite{liang2020crosstalk} ~\end{tabular}  \\ 
\hhline{|>{\arrayrulecolor[rgb]{1,0.859,0.604}}->{\arrayrulecolor{black}}-~--|}
{\cellcolor[rgb]{1,0.859,0.604}}        & Anti-tamper layouts            &                   & GCN, RL &   \\ 
\hhline{|>{\arrayrulecolor[rgb]{1,0.859,0.604}}->{\arrayrulecolor{black}}-~--|}
{\cellcolor[rgb]{1,0.859,0.604}}        & \begin{tabular}[c]{@{}c@{}}PDNs for anti-side channel \\and fault injection\end{tabular}                   &                   & GNN     &      Linear Regression~\cite{utyamishev2020power}                \\ 
\hhline{|>{\arrayrulecolor[rgb]{1,0.859,0.604}}->{\arrayrulecolor{black}}-~--|}
{\cellcolor[rgb]{1,0.859,0.604}}        & Anti-RE layouts                &                   & GCN                   &       CNN~\cite{zargari2021captive}               \\ 
\hhline{|>{\arrayrulecolor[rgb]{1,0.859,0.604}}->{\arrayrulecolor{black}}-~--|}
\multirow{-5}{*}{{\cellcolor[rgb]{1,0.859,0.604}}\begin{tabular}[c]{@{}>{\cellcolor[rgb]{1,0.859,0.604}}c@{}}\textbf{Floorplanning and}\\\textbf{~Physical Design }\end{tabular}}               & Custom eFPGA fabrics           &                   & RL      &   \\ 
\hline
{\cellcolor[rgb]{1,0.859,0.604}}        & \begin{tabular}[c]{@{}c@{}}Test point insertion and pattern \\generation for Trojan detection\end{tabular} & \multirow{3}{*}{ASIC, FPGA}          & RF, GB  & RL\cite{pan2021atpg}     \\ 
\hhline{|>{\arrayrulecolor[rgb]{1,0.859,0.604}}->{\arrayrulecolor{black}}-~--|}
{\cellcolor[rgb]{1,0.859,0.604}}        & Information flow security      &                   & GNN, DNN,  RL    &   \\ 
\hhline{|>{\arrayrulecolor[rgb]{1,0.859,0.604}}->{\arrayrulecolor{black}}-~--|}
\multirow{-3}{*}{{\cellcolor[rgb]{1,0.859,0.604}}\textbf{Testing and Verification }}             & \begin{tabular}[c]{@{}c@{}}Comparison of original \\specification to final design\end{tabular}             &                   & GNN     & NLP\cite{yu2021hw2vec}   \\
\hline
\end{tabular}
}
\label{tab: aieda}
\end{table}

\subsubsection{Testing and Verification}

As an IC reaches its physical state, the designers must implement design testing and verification. It is important to point out that most major design stages reach completion via some form of verification step~\cite{wang2009electronic}. For example, the logic synthesis step utilizes an RTL and gate level netlist verification procedure to guarantee functionality retention of the gate level design~\cite{martin_scheffer_lavagno_2016}. Testing and verification combined ensure that an IC has maintained original specifications and functionality throughout the design process. For physical testing, the two most significant options are wafer level testing and package level testing~\cite{perry_2007}.


\paragraph{Test Point Insertion and Pattern Generation for Trojan Detection} Design for Testability (DFT) is a technique employed in designing ICs for the purposes of reducing test costs and associated time. A very common technique for DFT is scan-chain insertion; regular FFs in the design are replaced with scan FFs. ATPG is a test methodology used to identify faulty behavior(s) in circuits due to design defects. The goal of ATPG is to create a set of test patterns that achieve a desired test coverage, TC, and fault coverage, FC, through fault simulation~\cite{hardware2011}. ATPG and formal methods have been researched for Trojan detection, but most do not consider partial-scan instances of a third-party IP (3PIP). The authors of~\cite{cruz2018} generate a set of constraints using the model checker to facilitate directed test generation using ATPG tool. A constraint generation procedure uses model checking to produce a set of signal traces. A test vector generation method then uses ATPG with the design, rare nodes and signal traces to produce a set of test vectors for activating each rare node~\cite{cruz2018}.

\paragraph{Information Flow Security} At the verification stage of the design flow, IFT can be executed using static or dynamic techniques. Static techniques assess the design's compliance with security policies using simulation, formal verification, emulation or virtual prototyping. They are applicable within the EDA pipeline, and are removed when verification is completed~\cite{hu2021}. Due to the unique goals of each specific static technique, AI/ML can be added to decrease resource usage, and improve results. For hardware security simulation, levels of abstraction like gate level and RTL are viable. As described in logic synthesis, graph-based networks are ideal solutions. The simulation process links security labels to signals. Formal verification guarantees proof of security for the desired properties, but it is known to have scaling issues~\cite{hu2021}. The large design state space of complex systems requires a policy-based learning solution which RL provides. Emulation can perform hardware-software co-verification, using FPGA emulation servers. Learning-based models could make these servers more robust to even detect software vulnerabilities that impact hardware. The abstract models of hardware components created by virtual prototyping can be tested a different RL-based approach. If the policy used is geared towards creating security problems while verifying software, the designer becomes aware of impending problems. Dynamic techniques use dedicated hardware to track information flow during runtime. A neural network could validate results to predicting flow inconsistencies.

\paragraph{Comparison of Original Specification to Final Design} Furthermore, Natural language processing (NLP) can be applied to circuit designs at different levels of abstraction. One of the most promising bridges to this is HW2VEC, a graph learning tool to extract graph representations from a hardware design in either RTL or gate level~\cite{yu2021hw2vec}. It is divided into two main parts: a HW2GRAPH stage which converts the design to a graph while retaining structural information, and GRAPH2VEC which converts the graph into Euclidean-based embeddings. GRAPH2VEC is just one of a number of graph-to-vector-embedding algorithms that utilizes concepts from the skipgram word embedding model~\cite{narayanan2017graph2vec}. The resulting vector representation may be applied with measures of similarity, like cosine similarity, to verify designs. If HW2WEC can be leveraged at physical design stages, the final IC product can be verified for structural authenticity.
 
\begin{figure*}[t] 
   \includegraphics[width=0.9\textwidth]{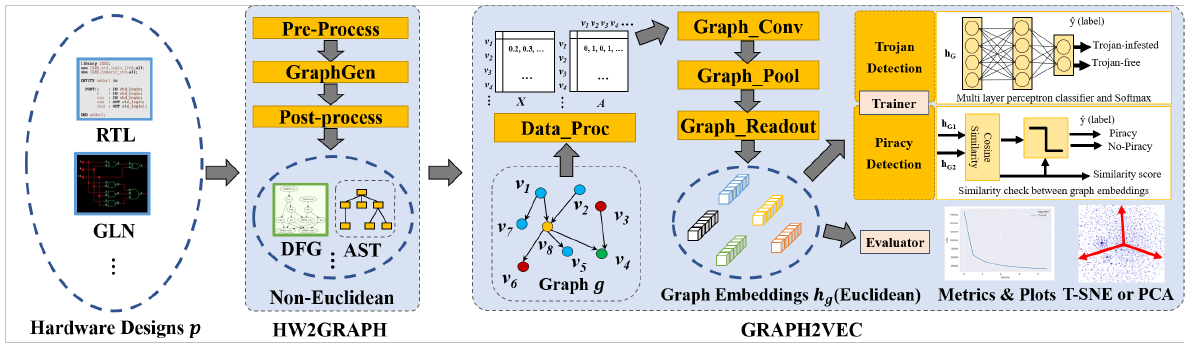}
   \centering
   \caption{Overview of HW2VEC from hardware design (RTL or gate level), through extraction of graph representations to node embedding (X) and adjacency (A) matrices. The graphrepresentations use GRAPH2VEC to acquire the graph embeddings for hardware security applications. Reprinted from~\cite{yu2021hw2vec}}
   \label{fig:hw2vec}
\end{figure*}

\section{AI for security-aware design of Analog Integrated circuits}\label{sec:analog}
Analog ICs are the fundamental components that interface with the real-world signals which are continuous and arbitrary in nature. Hence, the design of analog circuits are considered the foundation for all IC design. In this regards, analog IC design comprises of design of analog, radio frequency (RF), and mixed-signal ICs such as amplifiers, oscillators, filters, power regulators, data converters,~\etc{} Analog IC design traditionally involves a lot of manual work and hence relies heavily on expert knowledge and intuition. This is the case because unlike their digital counterparts, analog design steps typically overlap and have a higher number of complex design constraints and specifications that need to be met simultaneously. Thus, the automation of design of analog circuits and consequently analog security primitives has not gathered a lot of traction compared to digital circuits. Nevertheless, it has become paramount that the design of analog circuits be automated with the inclusion of security-aware features because of the following reasons:
\begin{itemize}
    \item \textbf{Bottleneck of the design process:} Design of analog ICs involves a lot of manual steps as explained above and are hence considered the bottleneck in any electronic system design~\cite{swings1990intelligent}. Since analog ICs are prevalent in almost all ICs, it is imperative that they be designed as swiftly as their digital counterparts to meet the current demands of low power, high performance, and quick time-to-market. Furthermore, this increasing demand coupled with error-prone manual process demands an inclusion of security-aware features in the design.
    \item \textbf{High susceptibility to reverse engineering:} Analog ICs are generally fabricated in higher or older technology nodes compared to digital ICs. They also have fewer number of transistors (in order of hundreds compared to billions) and have a lower current density, which makes it easier for the attacker to figure out the underlying intellectual property (IP) using the process of reverse engineering~\cite{alam2018challenges}. 
    \item \textbf{Lack of cryptomodules, error tolerance, and security primitives:} There's a severe lack of security features and primitives applicable for analog ICs as a large amount of research has focused on digital ICs~\cite{alam2018challenges}.
    \item \textbf{Impact of process variation:} Robust analog design has become ever so challenging with the continuous advancement in technology node and device scaling because the analog parameters are affected significantly more across process, voltage, and temperature (PVT) variations. This has made the design of reliable analog circuits very strenuous~\cite{razavi2000design}. 
    \item \textbf{Lack of comprehensive descriptiveness of design and security issues in conventional CAD approaches:} Depending on the type of analog IC, the specifications and requirements vary quite a lot. Hence, one unified framework or CAD approach is not applicable to the design of analog ICs. Hence, there is a severe lack of comprehensive description of design and security issues in traditional CAD approaches~\cite{scheible2015automation}. 
    \item \textbf{Inapplicability of most digital based security solutions:} The design flow of digital ICs is significantly different compared to analog ICs. Hence, the design and security approaches proposed for digital ICs are not applicable for analog and even mixed-signal ICs. Furthermore, the design of these security features require design and cost overheads which are only expendable in the case of large digital ICs. 
\end{itemize}
Henceforth, in this section, we focus on the recent developments of the use of automated techniques including AI based approaches and optimization techniques in the design of analog ICs and security primitives. 

\subsection{Optimization Techniques for EDA}
Machine learning models instead of pure optimization algorithms have been extensively used to automate the design of digital circuits which has significantly reduced the time-complexity associated with designing a robust digital IC. Furthermore, they are able to learn from the previously explored designs and datasets~\cite{huang2021machine}. However, training machine learning models is a challenge specifically in the analog domain compared to their digital counterpart because of the following reasons.
\begin{itemize}
    \item The specifications of analog design are variable for different applications. Thus, it becomes difficult to construct a uniform framework to evaluate and optimize different analog designs~\cite{mina2022review}.
    \item It is challenging to design a well-performing model because analog parameters are highly non-linear and susceptible to noise and process-voltage-temperature (PVT) variations~\cite{gielen2000computer}.
    \item The models trained for design automation have worked well for digital circuits compared to analog because analog designs suffer more in terms of area, power, and reliability when \textbf{porting} from a higher node to a lower technology node.
    \item The search space for analog design is significantly larger compared to digital IC design. There is variety of circuit schematics and device sizes that a designer can consider to get a reasonable performance. Plus there are more number of specifications to meet for an analog IC compared to their digital counterparts. This makes it computationally expensive and time-consuming to train a decent model. 
    \item Analog design is less systematic and more heuristic and intuitive in nature~\cite{gielen2000computer}. Thus there has been a substantial amount of automated design software tools available for digital IC design which makes it easier to generate data and train models to optimize the design process.
\end{itemize}

Because of the reasons outlined above, there are an abundance of ``automation'' frameworks which use one or more optimization algorithms in conjunction with the simulation software to optimize the design of a specific analog circuit. Many situations call for multi-objective optimization problems as is the case when designing security primitives as they have additional challenges and specifications to meet depending on the type of analog circuit at hand. Nevertheless, depending on the search space and the problem-at-hand, optimization techniques can be divided into three categories:
\begin{enumerate}
    \item \textbf{Random or brute-force search} which involves randomly generating a solution until the optimization criteria is satisfied. These are time-consuming. 
    \item \textbf{Gradient or derivative based optimization} utilizes gradient or derivatives within the data or the objective function to guide the search for the optimal solution~\cite{sun2019survey}. Gradient descent is one of the most widely used techniques specifically in ML algorithms. They are computationally efficient and quick, but prone to getting stuck at a local minimum rather than a global optimum if not designed properly. Further, they require functions that are differentiable.
    \item \textbf{Heuristic derivative-free optimization} methods utilize heuristic techniques characterized by empirical rules to guide the search process to find an optimal solution. The most popular algorithms are evolutionary algorithms (EA) (and simulated annealing (SA) which are inspired by Charles Darwin's theory of evolution and the annealing techniques in metallurgy, respectively. They are highly effective especially when derivatives or gradients do not exist and the search space is highly non-linear or noisy. However, they are quite time-consuming and the quality of solution is highly dependent on the initial configuration and learning rules.
\end{enumerate}
However, with the progress in high-performance computing hardware and ability to model non-linear problems, DL and RL methods have gained a lot of popularity in recent years. Specifically, RL have gained a lot of traction in analog design and optimization where their counterparts, evolutionary algorithms, have been applied for more than a decade~\cite{zebulum2018evolutionary}.

\begin{figure}[t]
\centering
\includegraphics[width=1\columnwidth]{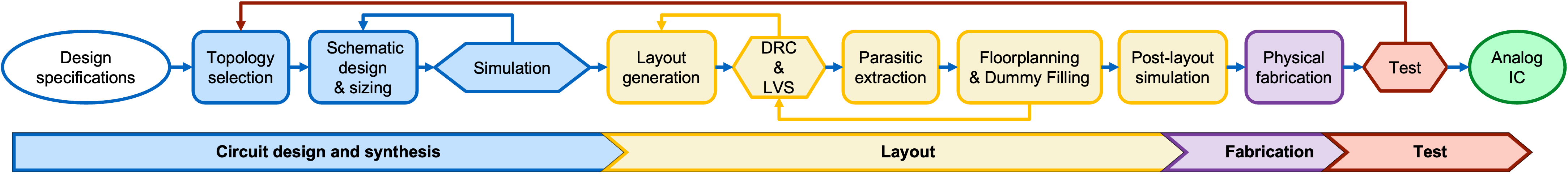} 
\caption{General analog IC design flow.}
\label{fig:analog_design_flow}
\end{figure}
\subsection{Analog IC Design Flow and AI/ ML Techniques}\label{sec:analog_design_flow}
Fig.~\ref{fig:analog_design_flow} provides a general overview of analog IC design which is drastically different compared to the digital circuit flow shown in Fig.~\ref{fig:icsecurity}. This is a top-down design approach as described by Gielen et. al.~\cite{gielen2000computer}. First, the analog IC designer chooses a topology and designs the schematic such that a given set of design constraints and specifications are fulfilled. Then, the transistors and other analog elements are sized very carefully and simulated using a circuit simulator. Next, the layout engineer generates the physical implementation of the IC based on the schematic, performs physical verification (DRC and LVS), extracts the parasitics, optimizes the location of circuit elements to fulfill the area and power specifications, and performs the post-layout simulation. This process of layout and simulation continues until the specified physical constraints are met after which the IC is ready to be fabricated. The design process is completed by testing the analog IC against the design specifications. If the IC fails the test, then the analog designer has to repeat the whole process and either resize the design or choose a different topology altogether. This process usually goes for several iterations until the IC is ready to be fabricated at a large scale. 

\begin{table}
\centering
\caption{AI/ ML techniques used in every design stage of the analog IC design process.}
\label{tab:design_analog}
\arrayrulecolor{black}
\resizebox{\linewidth}{!}{%
\begin{tabular}{|c|c|c|} 
\hline
\rowcolor[rgb]{0.361,0.769,0.769} \textbf{Design stage} & \textbf{Techniques} & \textbf{AI/ ML algorithms used} \\ 
\hline
{\cellcolor[rgb]{1,0.859,0.604}} & Selection & CNN~\cite{matsuba2018inference} , Polynomial regression~\cite{kaya2018analog}  \\ 
\hhline{|>{\arrayrulecolor[rgb]{1,0.859,0.604}}->{\arrayrulecolor{black}}--|}
{\cellcolor[rgb]{1,0.859,0.604}} & Feature
  extraction & Supervised/
  unsupervised learning~\cite{li2016analog} , GNN~\cite{kunal2020general}  \\ 
\hhline{|>{\arrayrulecolor[rgb]{1,0.859,0.604}}->{\arrayrulecolor{black}}--|}
\multirow{-3}{*}{{\cellcolor[rgb]{1,0.859,0.604}}\begin{tabular}[c]{@{}>{\cellcolor[rgb]{1,0.859,0.604}}c@{}}\textbf{ Topology}\\\textbf{~selection }\end{tabular}} & Generation & RNN~\cite{rotman2020electric} \\ 
\hline
{\cellcolor[rgb]{1,0.859,0.604}} & Model-based & \begin{tabular}[c]{@{}c@{}}SVM~\cite{ding2005active}, RBF/ MLP~\cite{passos2006rbf}, Bayesian NN~\cite{gao2019efficient}, \\ANN with sparse regression~\cite{tao2015harvesting}, DNN~\cite{takai2017prediction}, ANN~\cite{lourencco2018exploration, dumesnil2015rf}, \\Gaussian-based regression model~\cite{sanabria2020gaussian}, SVM~\cite{wang2018rfic}, BO~\cite{huang2021bayesian}, RL~\cite{zhao2020deep}\end{tabular} \\ 
\hhline{|>{\arrayrulecolor[rgb]{1,0.859,0.604}}->{\arrayrulecolor{black}}--|}
\multirow{-2}{*}{{\cellcolor[rgb]{1,0.859,0.604}}\begin{tabular}[c]{@{}>{\cellcolor[rgb]{1,0.859,0.604}}c@{}}\textbf{ Circuit sizing }\\\textbf{and biasing }\end{tabular}} & Simulation-based & \begin{tabular}[c]{@{}c@{}}ANN~\cite{islamouglu2019artificial, li2019artificial, wolfe2003extraction, bhatia2016modelling, jafari2010design}, RL~\cite{wang2018learning, settaluri2020autockt},~ GA~\cite{harsha2018integrated, das2007automated}, \\Hybrid EA~\cite{liu2009analog}, (RNN, DNN)~\cite{wang2018application}\\~GCN~\cite{wang2020gcn}, ANN with Polynomial regression~\cite{lourencco2019using}\end{tabular} \\ 
\hline
{\cellcolor[rgb]{1,0.859,0.604}}\textbf{ Simulation } & \begin{tabular}[c]{@{}c@{}}Behavioral\\modeling\end{tabular} & \begin{tabular}[c]{@{}c@{}}TDNN~\cite{grabmann2019power}, ANN~\cite{zhang2003artificial}, RNN~\cite{murphy2021automated}, \\Bayesian linear regression and SVM~\cite{pan2019late}, \\ Augmented NN~\cite{yu2017method},~RNN~\cite{chen2017verilog}, \\ NN~\cite{hasani2017compositional}, ANN with GA~\cite{dumesnil2014rf}\end{tabular} \\ 
\hline
{\cellcolor[rgb]{1,0.859,0.604}} & Layout & ML~\cite{kunal2019align} ,
  Spectral graph analysis~\cite{chen2020magical}  \\ 
\hhline{|>{\arrayrulecolor[rgb]{1,0.859,0.604}}->{\arrayrulecolor{black}}--|}
{\cellcolor[rgb]{1,0.859,0.604}} & Placement & ANN
  ~\cite{ guerra2019artificial, gusmao2020semi} , GNN~\cite{li2020customized} , (SVM, NN, Random forest)~\cite{li2020exploring} , LP~\cite{chen2020magical}  \\ 
\hhline{|>{\arrayrulecolor[rgb]{1,0.859,0.604}}->{\arrayrulecolor{black}}--|}
\multirow{-3}{*}{{\cellcolor[rgb]{1,0.859,0.604}}\textbf{ Layout }} & Routing  & Generative NN~\cite{ zhu2019geniusroute},
  grid-based A*~\cite{chen2020magical} \\ 
\hline
{\cellcolor[rgb]{1,0.859,0.604}} & Performance
  modeling & \begin{tabular}[c]{@{}c@{}}DNN with ES~\cite{hakhamaneshi2019bagnet}, \\ANN with Polynomial regression~\cite{lourencco2019using}, \\(SVM, NN, Random forest)~\cite{li2020exploring}, \\CNN with transfer learning~\cite{liu2020towards}\end{tabular} \\ 
\hhline{|>{\arrayrulecolor[rgb]{1,0.859,0.604}}->{\arrayrulecolor{black}}--|}
\multirow{-2}{*}{{\cellcolor[rgb]{1,0.859,0.604}}\begin{tabular}[c]{@{}>{\cellcolor[rgb]{1,0.859,0.604}}c@{}}\textbf{ Post-layout}\\\textbf{and test}\end{tabular}} & Fault
  diagnosis & Decision
  tree~\cite{stratigopoulos2014fast}, RideNN~\cite{binu2018ridenn} , GB-DBN~\cite{liu2017capturing}, DBN~\cite{zhao2018novel} , BMF~\cite{wang2015bayesian}  \\
\hline
\end{tabular}
}
\end{table}

\subsubsection{Topology selection}
Topology design and selection is the first step in the design of analog ICs. This step is one of the most critical and time-consuming step due to its impact on the circuit performance. Traditionally, it is carried out manually by designers with expert knowledge and intuition. With demands for high performing analog circuits increasing, researchers as shown in Table~\ref{tab:design_analog} begun to explore ML techniques that speed up the topology design process. Most of the early work focused on small functional units such as amplifiers and filters using simple search-based algorithms. Recently, Matsuba~\etal{} in~\cite{matsuba2018inference} attempted to use CNN to select from only four topologies of an amplifier with 13 different characteristics. Similarly, Kaya~\etal{} in~\cite{kaya2018analog} have used design space exploration to generate different \textbf{Pareto-optimal fronts (POFs)} in the context of balancing design specifications across different topologies. These POF based topologies are then fitted into a polynomial regression model for the final selection; however, the performances of these techniques are highly dependent on the created datasets. Researchers in ~\cite{li2016analog, kunal2020general} have instead used feature extraction techniques to ensure that the complex relationships between various topological components are well understood by the algorithm. Li~\etal{}~\cite{li2016analog} uses both supervised and unsupervised methods to extract features of sub-blocks and connections within a topology while Kunal~\etal{}~\cite{kunal2020general} uses graph neural networks~(GNN) to extract symmetry constraints among different topologies. However, none of these techniques generate a topology automatically. One of the recent works in~\cite{rotman2020electric} uses recurrent neural network~(RNN) and hypernetworks to generate a topology for a two-port circuit.

\subsubsection{Circuit sizing and biasing}
After a topology is selected for a specific design, the next step is to size the components within the circuit and bias the design appropriately to meet the desired specifications. This step is also referred to as \textbf{analog synthesis}. Depending on the number of components, this step is one of the most tedious ones as the design space is fairly large. As shown in Table~\ref{tab:design_analog}, there have been even more ML and optimization techniques to automate this step. These works can be briefly categorized into two different techniques: \textit{model-based} and \textit{simulation-based}.

\paragraph{Model-based techniques} These use regression models or surrogate models to represent the performance of the circuit. They are thus quicker than the simulation-based approaches which rely on circuit simulators to continuously optimize the parameters of the circuit. Most of the model-based works have either used neural networks~\cite{passos2006rbf, gao2019efficient, takai2017prediction, lourencco2018exploration, dumesnil2015rf} or regression models~\cite{tao2015harvesting, sanabria2020gaussian} all in the attempt to speed-up the circuit optimization and evaluation process. 
\begin{wrapfigure}[17]{R}{0.35\textwidth}
\centering
\includegraphics[width=0.35\textwidth]{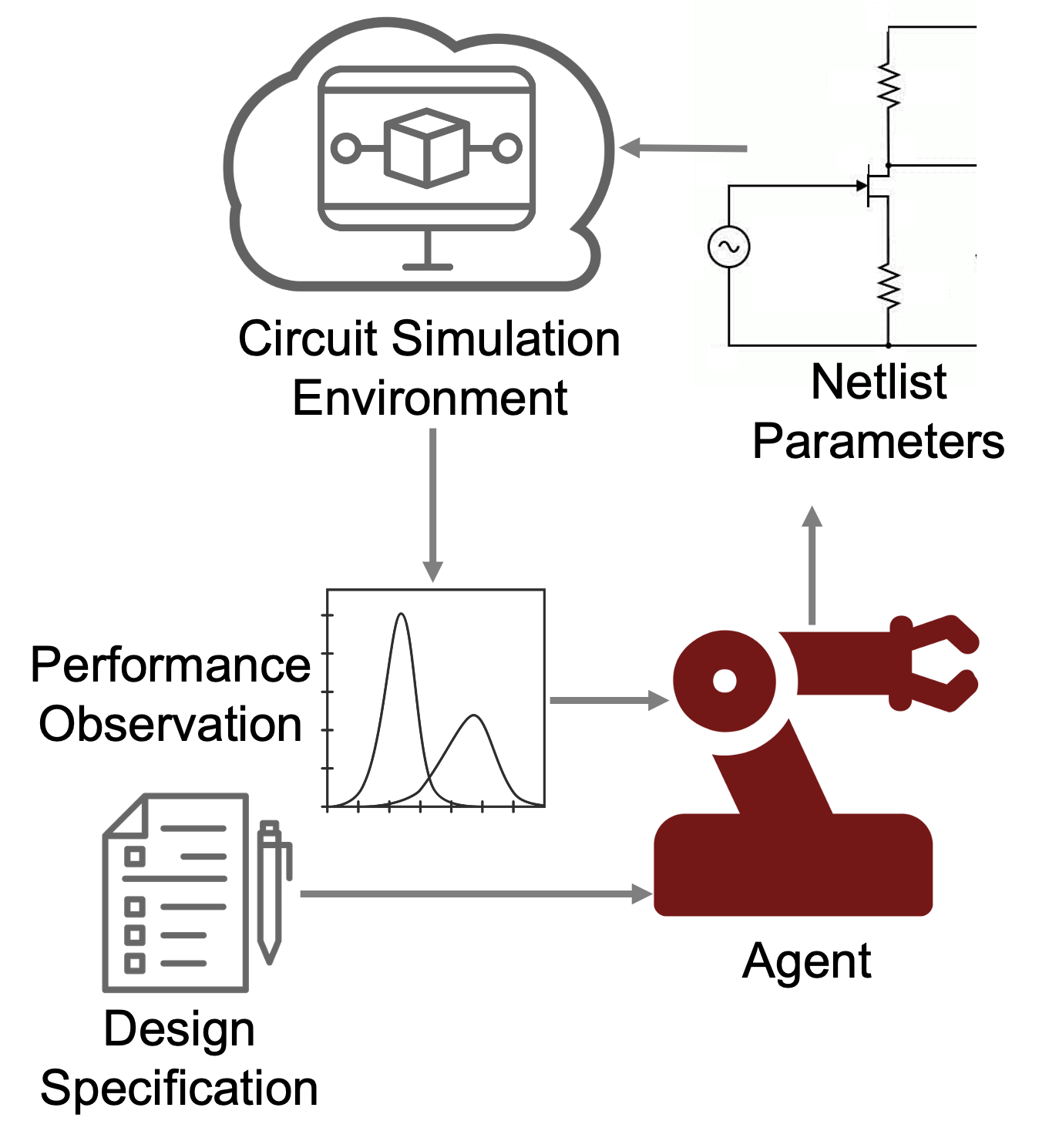}
\caption{Overview of the circuit design framework as proposed in~\cite{settaluri2020autockt}.}
\label{fig:RL_analogcircuit}
\end{wrapfigure}

\paragraph{Simulation-based techniques} These are considered superior since they are able to closely match real world (or post-silicon) circuit performance. Traditionally, simulation-based techniques have used evolutionary approaches~(EA) such as genetic algorithm~(GA) as seen in~\cite{das2007automated, harsha2018integrated, liu2009analog} to synthesize a wide-range of analog circuits. There have been a lot of works using neural networks~\cite{islamouglu2019artificial, li2019artificial, wolfe2003extraction, bhatia2016modelling, jafari2010design} to help find appropriate circuit parameters and biases; however, these techniques only work for a specific circuit such as amplifier or a filter. The goal has always been to either reduce human effort or mimic human intuition in designing these analog circuits. In this regard, the works in~\cite{settaluri2020autockt, wang2018learning, zhao2020deep} have used reinforcement learning~(RL) to design common analog circuits as much as 25 times faster than traditional optimization algorithms. Zhao~\etal{}~\cite{zhao2020deep} used the RL framework along with a symbolic analysis based approach that rapidly evaluated the circuit output without invoking a simulator. 

One of the reasons that RL-based approaches are gaining more traction these days is because of their ability to quickly learn and find the balance across different specifications. The other reason is their ability to transfer their knowledge, which experts expect will enable \textbf{IP reuse}, rapid automation, and \textbf{porting} between different technology nodes. Inspired by this ability, Wang~\etal{}~\cite{wang2020gcn} used graph convolutional networks~(GCN) to extract features of components and connections within the circuit and the RL agent to transfer these features whenever possible (\eg{}, features between a two-stage and a three-stage amplifier) and tune the circuit parameters quickly. In both cases of transfer learning,~\ie{}, between technology nodes and circuit topologies, the proposed RL-based sizing achieved the highest Figures of Merit (FoM) when compared with conventional black-box optimization methods and human expert designs. Similarly, the work in~\cite{settaluri2020autockt} uses NN as their RL agent (see Fig.~\ref{fig:RL_analogcircuit}). Transfer learning is also enabled here since the RL agent trained by running inexpensive schematic simulations can transfer its knowledge to a different environment,~\ie{}, the design including layout parasitics. The results for post-layout simulations provided in~\cite{settaluri2020autockt} demonstrate that it can converge faster than the traditional EA-based approaches.

\subsubsection{Simulation}
The next step in the design process is simulation to check whether the synthesized circuit satisfies the design specifications or not. Analog designers heavily rely on circuit simulators such as HSPICE and Cadence Spectre to assess the performance of their designs. However, behavioral modeling based techniques as shown in Table~\ref{tab:design_analog} have been proposed to speed up the simulation process and depending on the circuit application tune these models to assess application-specific performance metrics with more accuracy. In this regards, researchers have turned to neural network-based learning and modeling~\cite{grabmann2019power, zhang2003artificial, murphy2021automated, pan2019late}. For instance, in~\cite{grabmann2019power}, Grabmann~\etal{} used time-delay neural network~(TDNN) to generate energy-aware AMS IP cores. Similarly, ML algorithms along with optimization techniques have been used to find optimal solutions across a wider search range effectively and efficiently. In this regards, work in~\cite{pan2019late} first creates the data by exploring different device behaviors across different technologies, which is then fitted to a Bayesian regression model. Finally, SVM along with GA is employed in parallel to reduce the runtime compared to a pure EA-based optimization framework. 

\subsubsection{Layout}
ML and AI techniques have also been employed to automate the layout of analog circuits. The work in ~\cite{kunal2019align} summarizes early efforts in putting together an open-source quick layout generation flow for analog circuits that leverages template-driven design and machine learning techniques without human designers in the loop. In this regard, the recent work MAGICAL~\cite{chen2020magical} takes an unannotated netlist and design rules as inputs to create a complete GDSII using an automation framework. They use spectral graph analysis for layout, linear programming~(LP) for placement, and grid-based A* algorithm for efficient optimized routing. Recently, graph based machine learning networks such as GNN have been used to create models that predict the performance for a given placement and allows knowledge transfer between different analog circuits~\cite{li2020customized} which achieves better performance than CNN and regression based plug-in approaches. Similarly, Geniusroute~\cite{zhu2019geniusroute} uses variational autoencoders to extract latent layout strategies based on human expertise and then trains a generative NN based model to guide the automatic routing. 

\subsubsection{Post-layout and Test}
Post-layout simulation involves numerous iterations of verification and re-designs until the desired specifications are met. This is mainly due to the domination of layout parasitics. Performance modeling techniques such as BagNet~\cite{hakhamaneshi2019bagnet} use DNNs along with cross-entropy and canonical evolutionary strategies~(ES) to reduce this number of iterations significantly by leveraging past information and improving the optimizer's sample efficiency. Similarly, the recent work by Liu~\etal{}~\cite{liu2020towards} uses CNN and transfer learning to enable early design pruning for effective design space exploration and ensure robust well-performing placement features.
Further, the difficulty in obtaining accurate performance models of analog circuits have led to data-driven fault-diagnosis models for correctly assessing the analog circuit design. However, because of the highly non-linear and complex nature of analog performance faults, deep learning based techniques~\cite{binu2018ridenn, liu2017capturing, zhao2018novel} have fared better compared to shallow networks~\cite{stratigopoulos2014fast, wang2015bayesian}. Specifically, the use of deep belief networks~(DBN) have proven to be most effective as evident in~\cite{liu2017capturing, zhao2018novel} because of their ability to use raw time domain signals as inputs and employ an adaptive feature extraction and fault classification techniques.

\subsection{Security Issues in the AMS IC Design and Fabrication Flow} \label{sec:analog_security_issues_design}

\begin{figure}[t]
\centering
\includegraphics[width=1\columnwidth]{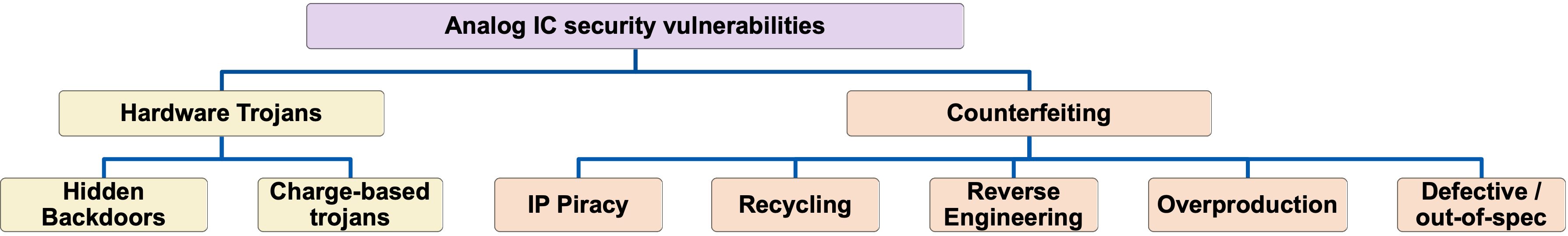}
\caption{Analog IC design security vulnerabilities.}
\label{fig:analog_sec_issues}
\end{figure}

The security issues pertaining to analog ICs are summarized by the supply-chain vulnerabilities which have increased within the last decade or so as a result of globalization. These issues as shown in Fig.~\ref{fig:analog_sec_issues} can be categorized into hardware Trojans and IC counterfeiting. 
\subsubsection{Hardware Trojans}
As discussed previously for digital ICs, Trojans are intentionally created anomalies which remain hidden and difficult to detect because they are triggered by rare conditions. Analog Trojans can be further categorized into charge-based Trojans~\cite{kison2019security} and hidden backdoors such as covert channels established by exploiting process variations~\cite{subramani2017ace}. Charge-based Trojans either use switched capacitors to design a trigger circuit or exploits capacitive couplings to design Trojans with no area overhead~\cite{kison2019security}. Hidden backdoors in the case of analog ICs mainly pertain to wireless RF networks where a Trojan circuit is designed to establish a previously unknown channel by exploiting process variation margins of the target circuit~\cite{subramani2017ace}. Although it is difficult to design purely analog Trojans (analog payload with analog trigger), they are more attractive than their digital counterparts because of their smaller footprint and their incompatibility/ invisibility to digital functional testing and verification techniques. As such there are works in AMS ICs where the trigger is hidden in the digital portion of the IP and its payload is transferred to the analog portion via test access mechanism~\cite{elshamy2020hardware}. With the advent of analog EDA tools and ML-based approaches to circuit design, it opens up more opportunities to exploit such tools and approaches to design analog hardware Trojans efficiently.

\subsubsection{IC Counterfeiting}
Analog ICs are one of the most widely reported counterfeited parts amongst all types of ICs~\cite{IHS_data}. The major reasons for analog counterfeiting are two-fold: First, they are easy to reverse engineer because of low transistor count and large transistor footprint. Second, analog IP design is cumbersome, manual labor-intensive, and highly dependent on knowledge expertise; hence, their value is significantly higher in the black market~\cite{rutenbar2006design}. IC counterfeiting can be briefly categorized into IP Piracy, IC recycling, reverse engineering, overproduction, and sale of defective ICs as shown in Fig.~\ref{fig:analog_sec_issues}. 
As these are supply chain related issues, the use of automation tools and AI/ ML techniques does not change the degree of vulnerability of the analog IPs. However, the use of such techniques does give rise to more opportunities in terms of security-aware design and security primitive optimization. The next subsection describes some of the security primitives which can be designed and exploited at different design stages.

\subsection{Design of Analog Security Primitives Within the Design Flow} \label{sec:analog_security_design}

Table~\ref{tab:security_analog} summarizes analog security primitives that employ either ML algorithms or an automation framework based on popular optimization techniques at various stages of the analog IC design flow. The table also mentions the associated security issues that are tackled by these proposed security primitives. 
\subsubsection{Physically Unclonable Functions (PUFs)}
As discussed earlier, PUFs are hardware security primitives which can either be used to create hardware authentication mechanisms or generate unique keys for cryptographic schemes. PUFs are devised by exploiting the inherent \textbf{process variation} or uncontrollable randomness present in every IC from manufacturing. This capability makes them a low-overhead, volatile security primitive that is more resistant to physical, side-channel, and software-based attacks. PUF design usually involves utilizing a specific circuit structure present inside an IC during the design step to function as a PUF. For example, the analog PUF proposed in~\cite{alvarez201514} leverages the threshold voltage mismatch between transistors in a cascode current mirror. With the size of ICs shrinking over the last few years, the limited number of input/output (I/O) pins and the difficulty of meeting both analog and PUF based specifications has made it challenging to devise a PUF suitable for analog ICs. In recent years, a few researchers have leveraged circuit sizing based automation and optimization tools to design robust PUFs that can produce unique yet random and reliable output bit suitable for device authentication.

The work in~\cite{chowdhury2020weak} uses GA and linear programming (LP) to size the transistors of an asynchronous null conventional logic (NCL) based gate (such as NCL TH22 gate) to devise a PUF circuit. This PUF utilizes the startup characteristics of cross-coupled inverters to produce a metastable output. The circuit is sized such that this output is only affected the inherent process variations and mismatch of the inverters inside the NCL structure. Because of the asynchronous nature of the circuit, this PUF is applicable for AMS circuits. Similarly, the recent work in~\cite{shah2021introducing} demonstrates a way to convert any circuit to a recurrent PUF (Rec-PUF). This is accomplished by utilizing a RNN-inspired technique involving combination of recurrence and XOR to generate a response (PUF output) which has very little correlation to its corresponding challenge (PUF input). They use the current mirror array (CMA) architecture proposed in~\cite{wang2017current} and improve its output reliability and its resiliency against ML-based attacks. In CMA-PUF~\cite{wang2017current}, transistors are sized small and are operated in sub-threshold region to maximize the mismatch. Further, the PUF utilizes a current controlled oscillator~(CCO) to digitize the current values of each column in the CMA. In~\cite{shah2021introducing}, the authors propose a Rec-CMAPUF where they replace CCO with a shared comparator to eliminate each column CCO's bias. They then feed the comparator's digitized output back to the input and perform the XOR operation to get the final output. This feedback or recurrence operation introduces non-linearity in the challenge-response-pair (CRP) space and increases the resistance to ML attacks. As analog PUFs are becoming more attractive, especially in the context of cryptographic ICs, the deep learning based sizing tools like RL based tools will surely be used to explore the design space and devise more reliable PUFs. For example, the sizing tools in~\cite{settaluri2020autockt,wang2020gcn,wang2018learning} can easily be used to include the PUF specifications as the additional design requirement and size the circuit to design a  PUF robust across all possible challenge-response pairs.

\begin{table}
\centering
\caption{AI/ ML based techniques and primitives to tackle security issues in analog ICs.}
\label{tab:security_analog}
\arrayrulecolor{black}
\resizebox{\linewidth}{!}{%
\begin{tabular}{|c|c|c|c|} 
\hline
\rowcolor[rgb]{0.361,0.765,0.773} \begin{tabular}[c]{@{}>{\cellcolor[rgb]{0.361,0.765,0.773}}c@{}}\\ \textbf{Design stage}\end{tabular} & \textbf{Techniques} & \textbf{AI/ ML algorithm} & \textbf{Security Issues tackled} \\ 
\hline
{\cellcolor[rgb]{1,0.855,0.604}} & PUF~\cite{shah2021introducing, chowdhury2020weak} & RNN~\cite{shah2021introducing},  (GA, LP)~\cite{chowdhury2020weak} & IP  Piracy and Counterfeiting \\ 
\hhline{|>{\arrayrulecolor[rgb]{1,0.855,0.604}}->{\arrayrulecolor{black}}---|}
\multirow{-2}{*}{{\cellcolor[rgb]{1,0.855,0.604}}\textbf{ Circuit Design }} & Logic Locking~\cite{volanis2019analog, gummidipoondijayasankaran2020towards} & ANN~\cite{volanis2019analog},  Simulated annealing~\cite{gummidipoondijayasankaran2020towards} & IP Piracy~\cite{volanis2019analog}, Overproduction~\cite{gummidipoondijayasankaran2020towards} \\ 
\hline
{\cellcolor[rgb]{1,0.855,0.604}}\textbf{ Layout } & ObfusX~\cite{zeng2021obfusx} & Explainability with RepTree & IP  Piracy, Hardware Trojans \\ 
\hline
{\cellcolor[rgb]{1,0.855,0.604}}\textbf{Simulation} & Aging simulation & \begin{tabular}[c]{@{}c@{}} RNN~\cite{rosenbaum2020machine}, \\(K-means, NN, SVM)~\cite{santhana2021recycled}\end{tabular} & IC Recycling \\ 
\hline
{\cellcolor[rgb]{1,0.855,0.604}}\textbf{ Fabrication} & \begin{tabular}[c]{@{}c@{}}Split Chip Design~\cite{pagliarini2020split}, \\Fab-of-origin~\cite{ahmadi2016machine}\end{tabular} & \begin{tabular}[c]{@{}c@{}}Branch-and-bound algorithm~\cite{pagliarini2020split},\\(PCA, DNN,~\etc{})~\cite{ahmadi2016machine}~\end{tabular} & IP  Piracy \\ 
\hline
{\cellcolor[rgb]{1,0.855,0.604}}\textbf{Test} & \begin{tabular}[c]{@{}c@{}}Side-channel fingerprinting~ \\and Authentication \\mechanisms\end{tabular} & \begin{tabular}[c]{@{}c@{}}ANN~\cite{liu2015concurrent}, (MLP, CNN)]~\cite{hanna2019deep}, PCA~\cite{casto2018multi},\\ SVM~\cite{chang2014approximating, acharyaldo}, (K-means, KNN, SVM)~\cite{chowdhury2020recycled}\end{tabular} & \begin{tabular}[c]{@{}c@{}}Hardware Trojan~\cite{liu2015concurrent},\\(Counterfeiting, Cloning)~\cite{casto2018multi},\\~IC Recycling~\cite{chowdhury2020recycled, chang2014approximating,acharyaldo, huang2012parametric}\end{tabular} \\
\hline
\end{tabular}
}
\end{table}

\subsubsection{Analog Locking Techniques}
There have been several attempts to expand logic locking to protect analog ICs from piracy and counterfeiting. These techniques either try to obfuscate the circuit current or the voltage biasing to hide the correct functionality of an analog IC. However, most of these techniques are susceptible to logic removal and model approximation attacks~\cite{9000113,acharya2020attack} because of low transistor count, large transistor size, and the fact that the key is implemented in digital domain. To improve the resiliency against such attacks, logic locking techniques utilizing ML algorithms have been proposed. Volanis~\etal{} in~\cite{volanis2019analog} proposed the use of an on-die analog neural network where the trained ANN acts as a lock and its analog input acts as a key. This is achieved by eliminating direct access to the biasing inputs of the analog IC by utilizing the ANN for biasing the IC to its operating point. The ANN is made programmable by the use of floating-gate transistors which serves as a permanent storage for the synapse weights. Furthermore, the keys are in analog domain and are continuous in nature allowing for a large number of key options making this locking technique resilient against brute-force and logic removal attacks. However, because of the very same fact, there does exist a number of possible keys that can give one similar performance as the oracle output. A possible way to ensure that does not happen is to find a global solution using methods such as the one described in~\cite{li2019artificial} that utilizes GA-based global search and ANN-based local minima search techniques. 

Similarly,~\cite{gummidipoondijayasankaran2020towards} attempts to tackle the issue of overproduction of an IC by a foundry. This technique, however, implements the lock in the digital section of an AMS circuit to minimize the effects of process variation by implementing a tuning knob to optimize the values of passive components in an analog circuit. The lock essentially controls the tuning knob based on an on-chip simulated annealing~(SA) based digital optimizer~\cite{wang2015built} to help nullify process variation-related effects. This optimizer hence acts as an post-processing tool that helps obtain a robust output. However, the area overhead of implementing the self-optimization architecture and the lock is too high for a typical AMS circuit such as band-pass filter, LNA, and a low-dropout oscillator~(LDO) as proposed in the paper. This could be overcome by designing robust analog models using learning based approaches~\cite{hasani2017compositional, pan2019late} to find optimal design parameters without the use of additional circuitry. Moreover, the RL-based techniques~\cite{wang2020gcn,settaluri2020autockt} can tune the design to be robust against process variations by using multiple design runs and leveraging its transfer learning abilities.

\subsubsection{Layout Obfuscation}
\cite{zeng2021obfusx} proposed an obfuscator for split manufacturing that relied on \textit{explainability} of ML-based attacks. As discussed earlier, split manufacturing is a technique where the untrusted foundry only receives and fabricates a partial layout to prevent the attacker from extracting the full design or IP of an IC. However, this still does not prevent ML-based attacks if the layout is either not obfuscated properly or is not obfuscated at all. The method in~\cite{zeng2021obfusx} implements different routing-based obfuscation techniques such as blockage insertion, routing perturbation, and wire-lifting based on an explanatory metric devised by leveraging ML attack models. The metric, shapely additive explanation~(SHAP) value essentially helps identify the most vulnerable connections within a layout and helps the obfuscator redo the obfuscation until the desired SHAP value is reached. This obfuscator can be easily incorporated with automated layout tools such as MAGICAL~\cite{chen2020magical} to perform highly secure layouts of analog ICs.

\subsubsection{Aging Simulation}
Incorporating aging analysis and simulation in the design step helps the designer or the IP owner characterize the performance of the circuit according to its life-cycle or usage. Such characterization can also help classify if the chip has been recycled or not. There are commercially available tools such as RelXpert by Cadence~\cite{schaldenbrand2019analog} and MOSRA by Synopsys~\cite{tudor2011mos} which calculate the damage to transistors at incremental simulation times. There are also behavioral models utilizing RNNs to model aged circuits as RNNs are compatible with transient current simulation. The trained RNN models can effectively predict the behavior of a ``fresh'' circuit if it is aged and inversely can help predict the age of an used or aged chip~\cite{rosenbaum2020machine}. In this regard, the works in~\cite{rosenbaum2020machine, santhana2021recycled, chowdhury2020recycled, chang2014approximating} utilize different ML algorithms to predict the age of an used IC. The recent work in~\cite{santhana2021recycled} can predict as low as just 1-day of aging. However, the success of this classification is highly dependent on the aging model developed during the design step which uses the model library, the aging models, and Monte Carlo simulations to help distinguish between process variation related changes and aging related effects. Incorporating circuit specific behavioral models developed using ML techniques such as the operational amplifier model in~\cite{murphy2021automated}, LNA model in~\cite{dumesnil2014rf}, and oscillator model in~\cite{yu2017method} can help design more accurate and robust aging models for various analog circuits. These aging characteristics can be further used after the chip is fabricated by comparing the output performance of fresh, unused ICs to a used one. There are several such techniques proposed which will be explained in Section~\ref{sec:authenticate}.

\subsubsection{Split Chip Design and Fab-of-origin}
Split chip design as proposed in~\cite{pagliarini2020split} differs from its previous counterpart split chip manufacturing proposed in~\cite{vaidyanathan2014building}. The split-chip design consists of dividing the entire chip design into two parts such that they can later be integrated to produce the intended functionality. By doing this, split-chip design avoids technological complexity and logistical challenges of split manufacturing. The design partitioning into multiple sections is then formulated as an optimization problem which is solved using a branch-and-bound algorithm. The objective function is devised based on the vulnerability metric which is essentially the product of exposure (effectiveness of a particular configuration) and criticality (importance of individual models)~\cite{pagliarini2020split}. Split chip design has been described as being effective in tackling the problems of IP theft and piracy within the IC supply chain. However, the chip is split into multiple parts after the design is finalized meaning the appropriate topology is selected and the circuit is already sized and biased appropriately. The circuit designer can easily incorporate this vulnerability metric based optimization with the automated circuit sizing tools such as the one shown in Table~\ref{tab:design_analog} to design even more secure analog circuits.
Similarly, Ahmadi~\etal{}~\cite{ahmadi2016machine} proposed a fab-of-origin attestation technique that leverages wafer-level parametric measurements without any information of the design underneath the die to distinguish between chips fabricated in different facilities. The authors proposed solutions for different scenarios related to this attestation problem and relied on several ML techniques such as PCA and DNN. 

\subsubsection{Side-channel based Fingerprinting and Authentication Mechanisms}\label{sec:authenticate}
The side-channel parameters such as power, delay, EM,~\etc{} emitted by an electronic device can be used as a fingerprint of such device to statistically assess if the device is authentic or not. There are several works which make use of behavioral models developed during the design stage to help characterize a fabricated chip as authentic or not. In this regards, the concurrent hardware Trojan detection~(CHTD) technique proposed in~\cite{liu2015concurrent} helps detect Trojans in a wireless RFICs specifically in cryptographic ICs by continuously extracting side-channel fingerprints and evaluating them on a trained on-chip neural classifier. This classifier essentially looks for systematic variations introduced by the hardware Trojan in the transient supply current to expose the malicious operation.

Similarly, the work in~\cite{hanna2019deep} trains behavioral models~(MLP and CNN) of RF transmitters to devise an authentication mechanism. Essentially these models exploit the non-linear characteristics exhibited by the power amplifier to uniquely identify each RF transmitter deployed in the field. The authors evaluated these models under various factors concerning the effects of signal quality, packet length, channel model and modulation type to demonstrate its effectiveness. Similarly in~\cite{casto2018multi}, the author proposes to use Process Specific Function (PSFs) for authenticating AMS ICs such as data converters. The harmonic frequencies and their corresponding magnitude levels are used to develop the PSF-based device identification model. The presented results showed the detection rate of $90\%$ using PCA and the PSF-based model. These side-channel based fingerprints however may not be as effective once the circuit starts to age. Hence, integrating aging models during the design step such as~\cite{rosenbaum2020machine} can help design robust fingerprinting mechanisms. 
Similarly, there are several techniques such as~\cite{chang2014approximating, chowdhury2020recycled, huang2012parametric, chowdhury2020low} who instead exploit aging characteristics without the behavioral models to train a classifier that can successfully distinguish between an aged and an authentic IC. However, these techniques require the use of a golden IC to train an effective model. 

One of the recent works in~\cite{acharyaldo} which is basically the extension of the work in~\cite{chowdhury2020recycled} instead devises an odometer by re-purposing the existing LDO structure present inside almost every IC. This LDO-odometer basically contains two different paths, a reference path which is not aged even when its send out to the field, and a normal path which is activated when out in the field. This self-referencing approach and a trained one-class SVM is then exploited to help characterize if the chip is recycled or not~\cite{acharyaldo} without the use of a golden IC. These aging-based classifiers can be improved by developing behavioral models at the design step which can help better characterize the profiles of these analog devices. For example, the learning methodology to develop behavioral models for complex MIMO system in~\cite{hasani2017compositional} can be used to characterize the LDO in~\cite{chowdhury2020recycled} and the oscillator and LNA in~\cite{chang2014approximating} at the design stage which helps improve the robustness of these aging-based authentication mechanisms. Furthermore, the automated synthesis techniques such as~\cite{settaluri2020autockt, wang2020gcn, wang2018learning} can be used to size the odometer in~\cite{acharyaldo} without incurring extra area and performance overhead. 

\subsection{Security Issues for Analog Security Primitives} \label{sec:analog_security_issues_primitives}

\begin{table}
\centering
\caption{Security vulnerabilities faced by analog ICs.}
\label{tab:security_issue_analog}
\resizebox{\linewidth}{!}{%
\begin{tabular}{|c|c|c|} 
\hline
\rowcolor[rgb]{0.361,0.769,0.769} \textbf{Attacks on} & \textbf{Technique} & \textbf{AI/ ML alg} \\ 
\hline
{\cellcolor[rgb]{1,0.859,0.604}}\textbf{Analog design} & Reverse engineering~\cite{cornforth2014reverse} & EC~\cite{cornforth2014reverse} \\ 
\hline
{\cellcolor[rgb]{1,0.859,0.604}}\textbf{PUFs} & \begin{tabular}[c]{@{}c@{}}Poster attack~\cite{ye2016poster}, \\Ensemble meta-algorithms~\cite{vijayakumar2016machine}\end{tabular} & \begin{tabular}[c]{@{}c@{}}Simulated annealing and EA~\cite{ye2016poster}, \\Bagging and Gradient Boosting~\cite{vijayakumar2016machine}\end{tabular} \\ 
\hline
{\cellcolor[rgb]{1,0.859,0.604}}\textbf{Logic locking techniques} & Attack of the genes~\cite{acharya2020attack} & GA~\cite{acharya2020attack} \\
\hline
\end{tabular}
}
\end{table}

Table~\ref{tab:security_issue_analog} summarizes some of the techniques that utilizes ML and optimization algorithms to execute attacks on an analog IC. 
\subsubsection{Attacks on Analog Design}
The reverse engineering technique proposed in~\cite{cornforth2014reverse} was one of the earliest works that used evolutionary computation~(EC) to reverse engineer the netlist of a simple non-linear analog IP by treating the circuit as a ''black box''. Although the reverse engineered netlist might differ from the actual netlist, the performance of both of these circuits matched rendering the success of such an approach. The recent automated design approaches specially the ones using DL and RL-based techniques may be used to reverse engineer an analog circuit. For example, the automated CNN-based topology selection technique in~\cite{matsuba2018inference} and the RL-based circuit sizing tool in~\cite{wang2018learning} can figure out the netlist with comparable performance within a day. Furthermore, MAGICAL~\cite{chen2020magical}, the automated layout technique can perform an optimized layout that might even outperform the actual design. 
\subsubsection{Attacks on PUFs}
Although PUFs could be resilient to physical as well as ML-based attacks, there are several works which leverages ML techniques to attack a PUF. The work proposed in~\cite{ye2016poster} was one of the first works that used a compound heuristics of evolutionary strategy~(ES), SA, and ant-colony to efficiently attack current mirror PUF~\cite{kumar2014design} and the voltage transfer PUF~\cite{vijayakumar2015novel} with an accuracy of $99\%$. Similarly, the work in~\cite{vijayakumar2016machine} use bagged trees and boosting technique to attack the non-linear analog PUFs. Bagged trees basically aggregates predictions from different instances of decision trees which are fed with disjoint subsets of training set to obtain the final response. While boosting ensemble of meta-algorithms is used to reduce the inherent bias present in DTs by increasing the model complexity. Nonetheless, the authors in~\cite{vijayakumar2016machine} elaborate on some analytical techniques based on the principal of functional composition to evaluate a PUF on its ML-attack resistance. These techniques can be used as security constraints during the process of analog synthesis using automated sizing frameworks like~\cite{settaluri2020autockt, wang2018learning} to size the PUF accordingly and design a ML-attack resistant PUF.
\subsubsection{Attacks on logic locking techniques}
There are several methods proposed to attack analog logic locking techniques such as~\cite{jayasankaran2019breaking, acharya2020attack}. Both of these works assume that the attacker has access to the obfuscated netlist and knows where in the IC the key is being applied. The work in~\cite{jayasankaran2019breaking} uses SMT-solver to extract the correct key locking the analog circuit. While the work in~\cite{acharya2020attack} uses GA to find the obfuscated parameter and the secret key. For this purpose, they devise an objective function based on the difference of the oracle output and the simulated output where the simulated output represents the output obtained by applying a random key generated using GA. Then given given the oracle output and the obfuscated netlist, they demonstrate how the attacker can easily estimate the value of the obfuscated parameter to reduce the search space for finding the correct key. 


\section{Gaps and Opportunities in AI-based Security-aware EDA}\label{sec:needs}



\subsection{Ecosystem Essentials for Accelerating Research and Development}

\subsubsection{Open-Source Benchmarks, Benchmark Generation Tools, and Data Augmentation}
The academic community is plagued by a lack of benchmarks to facilitate research and experimentation. For pre-silicon tool development, most researchers rely on common sets such as the ISCAS ('85, '89, '99) and OpenCores benchmarks~\cite{jenihhin}. Although these have varying degrees of complexity, the available designs do not accurately capture the entire space of today's IC industry. Companies are particularly hesitant to make even their most outdated designs available due to confidentiality of IPs. The result is a dissonance between academic research and industrial development, which is even more problematic for AI-based EDA. To develop effective AI/ML models, vast amounts of quality data are required. This requires a concerted effort from industry giants, irrespective of their own internal research and development investments. 

Hardware security-focused benchmarks like the ones available on Trust Hub seek to push research in the field. This online resource provides a few hundred examples of board level Trojans~\cite{HARRISON202112}, chip level Trojans~\cite{salmani2012, shakya2017benchmarking} and obfuscation~\cite{amir2018development} benchmarks.
Without benchmark generation tools, the samples become tedious to create. It is imperative that such tools are developed and made open-source, especially for AI training and validation that require large datasets. For this, Amir and Forte~\cite{amir2020adaptable} propose a technique that leverages linear optimization to generate synthetic combinational benchmarks that are adaptable to user input constraints and structurally different from input reference benchmarks. This can be viewed as a form of data augmentation to increase variation of features seen by AI/ML/DL algorithms, which should make them learn better and avoid overfitting. Similarly, behavioral models of common analog circuits can act as benchmarks for analog ICs which designers can easily evaluate and integrate security-related constraints and metrics within those models.

\subsubsection{Standardized Domain Knowledge}
As the academic community needs vast amounts of data, researchers also needs standardized domain knowledge. 
Domain knowledge helps standardize data collection processes, so data collected from different labs is consistent. It also guides researchers to identify relevant features for their specific applications; such prior knowledge means more informed \textbf{feature selection} and \textbf{dimensionality reduction}. Both allow practitioners to reduce the number of features in a dataset by only focusing on important ones. The EDA pipeline produces reports and tunable parameters that may be useful to a model. Without that distinction, an ML model may not achieve its improved automation and efficiency goals. The general benefits include requiring less data for high-accuracy models, creating more robust models, and utilizing interpretable features that help advance explainable AI (XAI) in EDA.

\subsubsection{Open-source EDA Tools and PDKs}
As this survey demonstrates, designing ICs requires many stages; IP cores must be verified and tested for synthesis to different FPGA architectures and various standard cell libraries. Tools like ones available on OpenCores.org are suitable for less-complex demands, and foster collaboration among users. However, they are not industry standard, hence they differ in functionality from commercial tools provided by Synopsys, Cadence, and Xilinx. They are also less-reliable and may have trouble scaling up to modern designs.

Some tools have been provided through extensive research within the academic and industrial communities. The OpenROAD initiative seeks to reduce effort, cost and time for hardware designs by providing 24-hour layout implementation~\cite{ajayi2019openroad}. Their GitHub repository also allows designers with available PDKs to add to the ones available. Similarly, the ALIGN initiative consists of a joint academic/industry team to develop open-source software for analog/mixed-signal circuit layout to translate
a netlist into a physical layout, with 24-hour turnaround and no
human in the loop~\cite{kunal2019align}. The development of more security-aware designs needs a similar effort from interested parties. With the same emphasis on automation, AI/ML requires working EDA tools to build from.

\subsection{Advanced AI Techniques to Explore and Benefits }

\subsubsection{Transfer Learning (TL)}
\begin{figure}[t]
\centering
\includegraphics[width=0.9\columnwidth]{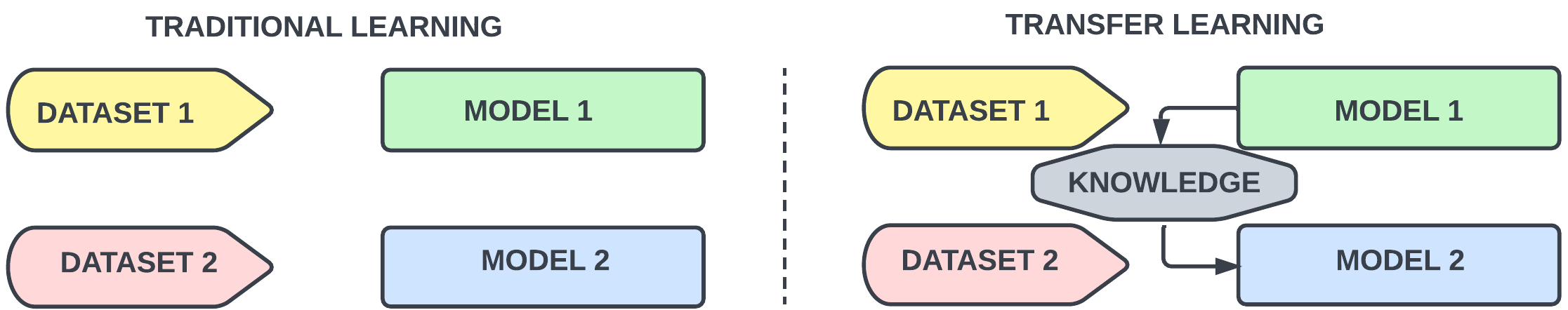}
\caption{High level description of transfer learning: the applied knowledge from pre-training using Model 1 helps reduce data required for Model 2.}
\label{fig:transferlearning}
\end{figure}

Transfer learning (TL) is built on the knowledge of previously-learned tasks; the knowledge gained from solving one machine learning problem can be used to address another one (see Fig.~\ref{fig:transferlearning}). The benefits of utilizing transfer learning are faster training times, increased accuracy, and eliminating the need for large amounts of data. Pan and Yang~\cite{pan2010transfer} use domain, task, and marginal probabilities to describe transfer learning. Take a domain $D$, a two-element tuple with feature space $\chi$ and marginal probability $P(X)$. $X$ is a sample data point, and the entire domain is represented as
\begin{align}
   D = \{\chi, P(X)\}
\end{align}
Now consider a task $T$, another two-element tuple of label space $\gamma$ and predictive function $\eta$. The objective function can be denoted as P($\gamma \vert$ X). Also, $Y$ is a corresponding label point for $X$. Therefore,
\begin{align}
   {T = \{\gamma, P(Y \vert X)\} = \{\gamma, \eta\}}
\end{align}
A source domain can be represented as $D_s$, source task as $T_s$, target domain as $D_t$, and target task as $T_t$. Hence, the aim of transfer learning is to learn the target conditional probability function $\eta_t$ or P($Y_t \vert X_t$) in $D_t$ with information gained from $D_s$ and $T_s$.

TL is ideal for applications within the IC design domain because general design rules are used globally. Design libraries and technology nodes may differ, but pre-trained deep learning models can be applied to various optimization and security-aware problems. This is especially the case in analog IP design where the lack of portability and re-usability across different circuits, design stages, and technology nodes has acted as impediments to the advancement of analog EDA tools. TL has become common in the community for its ability to help train models and quickly deploy models in applications where data is not readily available, as is often the case in the hardware assurance domain. 

\subsubsection{Federated Learning for Collaboration Without Hardware IP Compromise}\label{subsec:federated}


Federated learning (FL) or collaborative learning involves training an algorithm in a decentralized manner across multiple edge devices or servers, which each possess their own local data. FL, first coined by McMahan and Ramage~\cite{McMahan2017-federated-first-intro}, combines machine learning with edge computing for training (\textit{and} data collection). Each endpoint, a low-powered computation device, has a copy of the model, improves the model locally, summarizes changes and sends the updates to the cloud where the global model is updated, updates are then aggregated in the cloud and updates sent back to edge devices. Li et al. summarize the problem as:
\begin{equation}
    \min_w F(w),\text{ where } F(w) \coloneqq \sum_k^N p_k F_k(w) \label{eq:federated}
\end{equation}
where, $N$ is the number of total devices, $p_k\in[0,1]$ and $\sum_k p_k=1$; $F_k$ is the local objective function for the $k$th device~\cite{Li2020-federated-challenges}.





Ideally, FL allows for smarter models, reduced latency and power consumption, all while securing privacy~\cite{McMahan2017-federated-first-intro}. Distributed data resources also eliminates the need for a central massive data store. 

A limiting factor for AI-based IC design is the \textbf{confidentiality of IPs}. Large datasets are required from many existing AI-based design approaches as well as updates over those designs to generate labels. However, companies and governments are typically reluctant to share their IPs, which might include proprietary algorithms, novel design techniques, or in-house countermeasures. The lack of data stifles any one organization's ability to build sufficient datasets and robust models. For instance, a single
company's data might provide a lot of data on a particular corner case or
vulnerability but have little or no data on others resulting in class
imbalance. One solution to this data sharing problem is differential
privacy-based FL, which allows datasets to be computed over at each edge-point \textit{without compromising IP}. Note however that for hardware security applications, this architecture may only be meaningful for cases where the model provides benefits to ALL participating organizations, such as piracy detection~\cite{Yasaei2021-gnn-piracy} and security rule checks~\cite{xiao2016security}.

\subsubsection{Explainable AI and Engineer Education}\label{subsec:xai}
Explainable AI (XAI) is a field of AI concerned with making ML models understandable by humans.
For perspective, take an ANN and attempts to understand the decisions it makes.
ANNs are opaque (\aka{} black-box) predictors and their experimental performance is largely due to their large parametric space, efficient learning algorithms and computation advances in hardware.
Transparent (\aka{} white-box) models expose directly how the model/mechanism works.
Decision tree models (and other rule based models) are largely considered an example of an easily interpretable model for both local and global explanations.
Other transparent models include $k$-Nearest Neighbors, generative additive models and bayesian models~\cite{Barredo_Arrieta2020-xai-concepts-taxonomies}.

Due to their successes, ANNs are increasingly being deployed for high-stakes decision-making.
The opaque nature of ANNs raises critical issues because without an understanding of how the mechanism is predicting, there is no way of justifying the rationale behind any prediction.
The complex latent manifolds within which ANNs operate are difficult to visualize and reason around.
Further complicating things, explainability is defined not only by the details and reasons behind its processes, but also the audience doing the observing~\cite{Barredo_Arrieta2020-xai-concepts-taxonomies}.
Essentially, explanation methods are intended to be a useful interface/translation between a complex model and a stakeholder.

\begin{figure*}[t]
    \centering
    \includegraphics[width=0.9\linewidth]{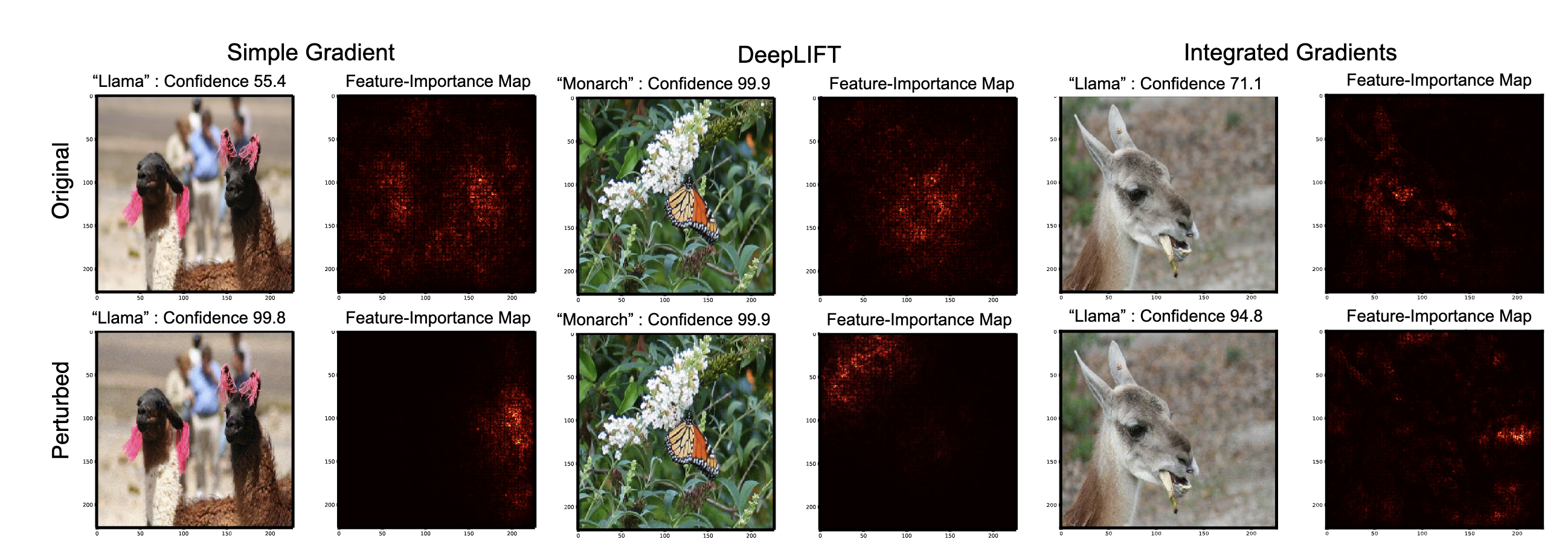}
    \caption{Examples illustrating how small perturbations in input do not change predicitons, but create very different attribution maps. Reprinted from~\cite{Ghorbani2019-fragile-interp}.}
    \label{fig:xai-fragile}
\end{figure*}

A variety of post-hoc methods that attempt to explain opaque predictors have been proposed that can be broadly characterized into model-agnostic and model-specific techniques.
Explainers can be considered global (\aka{} the model explanation problem) or local (\aka{} the prediction explanation problem).
Global explainers attempt to approximate the original model but also remain understandable.
Often global explanation approaches represent the final explanation as a decision tree.
Local explanations focus on creating explanations for a particular input, \ie{} for a given input, explain a specific prediction.
In the model-agnostic case, this often results in explanations that quantify the contribution of each feature to the prediction.
However, explainers themselves tend to be fragile.
As shown in Figure~\ref{fig:xai-fragile}, explanations can vary widely due to small changes in the input.
Or more problematically, they can be equivalent for very different predictions.
The landscape of explainable AI is still evolving and there is much work still needed to be done.
Explanation tooling can provide some guidance as to ``where the model is looking'' but there is also a need for designing interpretable models~\cite{Rudin2019-stop}.

In the semiconductor industry, many design engineers do not have the experience and background to deal with hardware security issues.
Even those that do may have very specialized experience, \eg{} experts in applied cryptography but novices in supply chain security.
Further, security is a domain in constant state of flux with new attack vectors popping up each year.
Explainable AI is an important research area and could provide feedback to organizations and help them train or re-train their design engineers to keep up with new attacks and vulnerabilities.
The lessons learned from explainable AI both in its limitations and possibilities can provide useful insight to the appropriate application, or more importantly interpretation, of ML models in security applications.

\section{Conclusion}\label{sec:conclusion}
In this survey, we summarized the state-of-the-art AI/ ML-based EDA techniques, their contemporary and potential use in security-aware design and provide a general perspective on needs and requirements for an automated security-aware IC design process. Our literature review showed that ML/AI is remarkably suitable for tuning the design to incorporate security constraints without deteriorating performance, develop reliable and robust security primitives to tackle supply chain related vulnerabilities, and make the designs more resilient against both invasive and non-invasive, side-channel, and fault-injection attacks. In that regard, we have extensively covered the intersection between learning and design with the objective of protecting electronic systems against malicious parties, whether it be by AI-based integration of multiple primitives and protections, guidance and optimization of design tradeoffs, or real-time feedback/training for users. We also discussed problems and solutions within this domain, such as scarcity of data, confidentiality, and open-source tools, and highlighted opportunities that will make the AI-based design more formidable and help solidify hardware as the root-of-trust for industry and government.

\bibliographystyle{ACM-Reference-Format}
\bibliography{references.bib}


\begin{thebibliography}{289}


\ifx \showCODEN    \undefined \def \showCODEN     #1{\unskip}     \fi
\ifx \showDOI      \undefined \def \showDOI       #1{#1}\fi
\ifx \showISBNx    \undefined \def \showISBNx     #1{\unskip}     \fi
\ifx \showISBNxiii \undefined \def \showISBNxiii  #1{\unskip}     \fi
\ifx \showISSN     \undefined \def \showISSN      #1{\unskip}     \fi
\ifx \showLCCN     \undefined \def \showLCCN      #1{\unskip}     \fi
\ifx \shownote     \undefined \def \shownote      #1{#1}          \fi
\ifx \showarticletitle \undefined \def \showarticletitle #1{#1}   \fi
\ifx \showURL      \undefined \def \showURL       {\relax}        \fi
\providecommand\bibfield[2]{#2}
\providecommand\bibinfo[2]{#2}
\providecommand\natexlab[1]{#1}
\providecommand\showeprint[2][]{arXiv:#2}

\bibitem[\protect\citeauthoryear{??}{els}{2009}]%
        {else2009}
 \bibinfo{year}{2009}\natexlab{}.
\newblock
\newblock
\urldef\tempurl%
\url{https://booksite.elsevier.com/samplechapters/9780750678667/Sample_Chapter.pdf}
\showURL{%
\tempurl}


\bibitem[\protect\citeauthoryear{??}{IHS}{2012}]%
        {IHS_data}
 \bibinfo{year}{2012}\natexlab{}.
\newblock \bibinfo{title}{Top 5 Most Counterfeited Parts Represent a 169
  Billion Potential Challenge for Global Semiconductor Market}.
\newblock
\newblock
\urldef\tempurl%
\url{https://technology.ihs.com/405654/top-5-most-counterfeited-parts-represent-a-169-billion-potential-challenge-for-global-semiconductor-market}
\showURL{%
\tempurl}


\bibitem[\protect\citeauthoryear{??}{Bar}{2019}]%
        {Barboza2019}
 \bibinfo{year}{2019}\natexlab{}.
\newblock \showarticletitle{Machine learning-based pre-routing timing
  prediction with reduced pessimism}.
\newblock \bibinfo{journal}{\emph{Proceedings - Design Automation Conference}}.
\newblock
\showISBNx{9781450367257}
\showISSN{0738100X}
\urldef\tempurl%
\url{https://doi.org/10.1145/3316781.3317857}
\showDOI{\tempurl}


\bibitem[\protect\citeauthoryear{Abadi, Agarwal, Barham, Brevdo, Chen, Citro,
  Corrado, Davis, Dean, Devin, Ghemawat, Goodfellow, Harp, Irving, Isard, Jia,
  Jozefowicz, Kaiser, Kudlur, Levenberg, Man\'{e}, Monga, Moore, Murray, Olah,
  Schuster, Shlens, Steiner, Sutskever, Talwar, Tucker, Vanhoucke, Vasudevan,
  Vi\'{e}gas, Vinyals, Warden, Wattenberg, Wicke, Yu, and Zheng}{Abadi
  et~al\mbox{.}}{2015}]%
        {tensorflow2015-whitepaper}
\bibfield{author}{\bibinfo{person}{Mart\'{i}n Abadi}, \bibinfo{person}{Ashish
  Agarwal}, \bibinfo{person}{Paul Barham}, \bibinfo{person}{Eugene Brevdo},
  \bibinfo{person}{Zhifeng Chen}, \bibinfo{person}{Craig Citro},
  \bibinfo{person}{Greg~S. Corrado}, \bibinfo{person}{Andy Davis},
  \bibinfo{person}{Jeffrey Dean}, \bibinfo{person}{Matthieu Devin},
  \bibinfo{person}{Sanjay Ghemawat}, \bibinfo{person}{Ian Goodfellow},
  \bibinfo{person}{Andrew Harp}, \bibinfo{person}{Geoffrey Irving},
  \bibinfo{person}{Michael Isard}, \bibinfo{person}{Yangqing Jia},
  \bibinfo{person}{Rafal Jozefowicz}, \bibinfo{person}{Lukasz Kaiser},
  \bibinfo{person}{Manjunath Kudlur}, \bibinfo{person}{Josh Levenberg},
  \bibinfo{person}{Dandelion Man\'{e}}, \bibinfo{person}{Rajat Monga},
  \bibinfo{person}{Sherry Moore}, \bibinfo{person}{Derek Murray},
  \bibinfo{person}{Chris Olah}, \bibinfo{person}{Mike Schuster},
  \bibinfo{person}{Jonathon Shlens}, \bibinfo{person}{Benoit Steiner},
  \bibinfo{person}{Ilya Sutskever}, \bibinfo{person}{Kunal Talwar},
  \bibinfo{person}{Paul Tucker}, \bibinfo{person}{Vincent Vanhoucke},
  \bibinfo{person}{Vijay Vasudevan}, \bibinfo{person}{Fernanda Vi\'{e}gas},
  \bibinfo{person}{Oriol Vinyals}, \bibinfo{person}{Pete Warden},
  \bibinfo{person}{Martin Wattenberg}, \bibinfo{person}{Martin Wicke},
  \bibinfo{person}{Yuan Yu}, {and} \bibinfo{person}{Xiaoqiang Zheng}.}
  \bibinfo{year}{2015}\natexlab{}.
\newblock \bibinfo{title}{{TensorFlow}: Large-Scale Machine Learning on
  Heterogeneous Systems}.
\newblock
\newblock
\urldef\tempurl%
\url{https://www.tensorflow.org/}
\showURL{%
\tempurl}
\newblock
\shownote{Software available from tensorflow.org.}


\bibitem[\protect\citeauthoryear{Acharya, Chowdhury, Ganji, and Forte}{Acharya
  et~al\mbox{.}}{2020}]%
        {acharya2020attack}
\bibfield{author}{\bibinfo{person}{Rabin~Yu Acharya}, \bibinfo{person}{Sreeja
  Chowdhury}, \bibinfo{person}{Fatemeh Ganji}, {and} \bibinfo{person}{Domenic
  Forte}.} \bibinfo{year}{2020}\natexlab{}.
\newblock \showarticletitle{Attack of the genes: Finding keys and parameters of
  locked analog ICs using genetic algorithm}. In \bibinfo{booktitle}{\emph{2020
  IEEE International Symposium on Hardware Oriented Security and Trust
  (HOST)}}. IEEE, \bibinfo{pages}{284--294}.
\newblock


\bibitem[\protect\citeauthoryear{Acharya, Levin, and Forte}{Acharya
  et~al\mbox{.}}{2021}]%
        {acharyaldo}
\bibfield{author}{\bibinfo{person}{Rabin~Yu Acharya},
  \bibinfo{person}{Michael~Valentin Levin}, {and} \bibinfo{person}{Domenic
  Forte}.} \bibinfo{year}{2021}\natexlab{}.
\newblock \showarticletitle{LDO-based Odometer to Combat IC Recycling}. In
  \bibinfo{booktitle}{\emph{2021 IEEE 34th International System-on-Chip
  Conference (SOCC)}}. IEEE, \bibinfo{pages}{206--211}.
\newblock


\bibitem[\protect\citeauthoryear{Ahmadi, Bidmeshki, Nahar, Orr, Pas, and
  Makris}{Ahmadi et~al\mbox{.}}{2016}]%
        {ahmadi2016machine}
\bibfield{author}{\bibinfo{person}{Ali Ahmadi}, \bibinfo{person}{Mohammad-Mahdi
  Bidmeshki}, \bibinfo{person}{Amit Nahar}, \bibinfo{person}{Bob Orr},
  \bibinfo{person}{Michael Pas}, {and} \bibinfo{person}{Yiorgos Makris}.}
  \bibinfo{year}{2016}\natexlab{}.
\newblock \showarticletitle{A machine learning approach to fab-of-origin
  attestation}. In \bibinfo{booktitle}{\emph{Proceedings of the 35th
  International Conference on Computer-Aided Design}}. \bibinfo{pages}{1--6}.
\newblock


\bibitem[\protect\citeauthoryear{Ajayi, Blaauw, Chan, Cheng, Chhabria, Choo,
  Coltella, Dobre, Dreslinski, Foga{\c{c}}a, et~al\mbox{.}}{Ajayi
  et~al\mbox{.}}{2019}]%
        {ajayi2019openroad}
\bibfield{author}{\bibinfo{person}{T Ajayi}, \bibinfo{person}{D Blaauw},
  \bibinfo{person}{TB Chan}, \bibinfo{person}{CK Cheng}, \bibinfo{person}{VA
  Chhabria}, \bibinfo{person}{DK Choo}, \bibinfo{person}{M Coltella},
  \bibinfo{person}{S Dobre}, \bibinfo{person}{R Dreslinski}, \bibinfo{person}{M
  Foga{\c{c}}a}, {et~al\mbox{.}}} \bibinfo{year}{2019}\natexlab{}.
\newblock \showarticletitle{OpenROAD: Toward a Self-Driving, Open-Source
  Digital Layout Implementation Tool Chain}.
\newblock \bibinfo{journal}{\emph{Proc. GOMACTECH}} (\bibinfo{year}{2019}),
  \bibinfo{pages}{1105--1110}.
\newblock


\bibitem[\protect\citeauthoryear{Alam, Chowdhury, Park, Munzer, Maghari,
  Tehranipoor, and Forte}{Alam et~al\mbox{.}}{2018}]%
        {alam2018challenges}
\bibfield{author}{\bibinfo{person}{Md~Mahbub Alam}, \bibinfo{person}{Sreeja
  Chowdhury}, \bibinfo{person}{Beomsoo Park}, \bibinfo{person}{David Munzer},
  \bibinfo{person}{Nima Maghari}, \bibinfo{person}{Mark Tehranipoor}, {and}
  \bibinfo{person}{Domenic Forte}.} \bibinfo{year}{2018}\natexlab{}.
\newblock \showarticletitle{Challenges and opportunities in analog and mixed
  signal (AMS) integrated circuit (IC) security}.
\newblock \bibinfo{journal}{\emph{Journal of Hardware and Systems Security}}
  \bibinfo{volume}{2}, \bibinfo{number}{1} (\bibinfo{year}{2018}),
  \bibinfo{pages}{15--32}.
\newblock


\bibitem[\protect\citeauthoryear{Alam, Tajik, Ganji, Tehranipoor, and
  Forte}{Alam et~al\mbox{.}}{2019}]%
        {alam2019ram}
\bibfield{author}{\bibinfo{person}{Md~Mahbub Alam}, \bibinfo{person}{Shahin
  Tajik}, \bibinfo{person}{Fatemeh Ganji}, \bibinfo{person}{Mark Tehranipoor},
  {and} \bibinfo{person}{Domenic Forte}.} \bibinfo{year}{2019}\natexlab{}.
\newblock \showarticletitle{RAM-Jam: Remote temperature and voltage fault
  attack on FPGAs using memory collisions}. In \bibinfo{booktitle}{\emph{2019
  Workshop on Fault Diagnosis and Tolerance in Cryptography (FDTC)}}. IEEE,
  \bibinfo{pages}{48--55}.
\newblock


\bibitem[\protect\citeauthoryear{Alaql and Bhunia}{Alaql and Bhunia}{2021}]%
        {alaql2021saro}
\bibfield{author}{\bibinfo{person}{Abdulrahman Alaql} {and}
  \bibinfo{person}{Swarup Bhunia}.} \bibinfo{year}{2021}\natexlab{}.
\newblock \showarticletitle{SARO: Scalable Attack-Resistant Logic Locking}.
\newblock \bibinfo{journal}{\emph{IEEE Transactions on Information Forensics
  and Security}}  \bibinfo{volume}{16} (\bibinfo{year}{2021}),
  \bibinfo{pages}{3724--3739}.
\newblock


\bibitem[\protect\citeauthoryear{Alaql, Chattopadhyay, Chakraborty, Hoque, and
  Bhunia}{Alaql et~al\mbox{.}}{2021}]%
        {alaql2021}
\bibfield{author}{\bibinfo{person}{Abdulrahman Alaql}, \bibinfo{person}{Saranyu
  Chattopadhyay}, \bibinfo{person}{Prabuddha Chakraborty},
  \bibinfo{person}{Tamzidul Hoque}, {and} \bibinfo{person}{Swarup Bhunia}.}
  \bibinfo{year}{2021}\natexlab{}.
\newblock \showarticletitle{LeGO: A Learning-Guided Obfuscation Framework for
  Hardware IP Protection}.
\newblock \bibinfo{journal}{\emph{IEEE Transactions on Computer-Aided Design of
  Integrated Circuits and Systems}} (\bibinfo{year}{2021}),
  \bibinfo{pages}{1--1}.
\newblock
\urldef\tempurl%
\url{https://doi.org/10.1109/TCAD.2021.3075939}
\showDOI{\tempurl}


\bibitem[\protect\citeauthoryear{Alkabani and Koushanfar}{Alkabani and
  Koushanfar}{2007}]%
        {alkabani2007active}
\bibfield{author}{\bibinfo{person}{Yousra Alkabani} {and}
  \bibinfo{person}{Farinaz Koushanfar}.} \bibinfo{year}{2007}\natexlab{}.
\newblock \showarticletitle{Active Hardware Metering for Intellectual Property
  Protection and Security.}. In \bibinfo{booktitle}{\emph{USENIX security
  symposium}}, Vol.~\bibinfo{volume}{20}. \bibinfo{pages}{1--20}.
\newblock


\bibitem[\protect\citeauthoryear{ALPAYDIN}{ALPAYDIN}{2020}]%
        {alpaydin_2020}
\bibfield{author}{\bibinfo{person}{ETHEM ALPAYDIN}.}
  \bibinfo{year}{2020}\natexlab{}.
\newblock \bibinfo{booktitle}{\emph{Introduction to machine learning}}.
\newblock \bibinfo{publisher}{MIT Press}.
\newblock


\bibitem[\protect\citeauthoryear{Alsaiari and Gebali}{Alsaiari and
  Gebali}{2019}]%
        {alsaiari2019}
\bibfield{author}{\bibinfo{person}{Uthman Alsaiari} {and}
  \bibinfo{person}{Fayez Gebali}.} \bibinfo{year}{2019}\natexlab{}.
\newblock \showarticletitle{Hardware Trojan Detection Using Reconfigurable
  Assertion Checkers}.
\newblock \bibinfo{journal}{\emph{IEEE Transactions on Very Large Scale
  Integration (VLSI) Systems}} \bibinfo{volume}{27}, \bibinfo{number}{7}
  (\bibinfo{year}{2019}), \bibinfo{pages}{1575--1586}.
\newblock
\urldef\tempurl%
\url{https://doi.org/10.1109/TVLSI.2019.2908964}
\showDOI{\tempurl}


\bibitem[\protect\citeauthoryear{Alvarez, Zhao, and Alioto}{Alvarez
  et~al\mbox{.}}{2015}]%
        {alvarez201514}
\bibfield{author}{\bibinfo{person}{Anastacia Alvarez}, \bibinfo{person}{Wenfeng
  Zhao}, {and} \bibinfo{person}{Massimo Alioto}.}
  \bibinfo{year}{2015}\natexlab{}.
\newblock \showarticletitle{14.3 15fJ/b static physically unclonable functions
  for secure chip identification with< 2\% native bit instability and
  140$\times$ Inter/Intra PUF hamming distance separation in 65nm}. In
  \bibinfo{booktitle}{\emph{2015 IEEE International Solid-State Circuits
  Conference-(ISSCC) Digest of Technical Papers}}. IEEE, \bibinfo{pages}{1--3}.
\newblock


\bibitem[\protect\citeauthoryear{Amir and Forte}{Amir and Forte}{2020}]%
        {amir2020adaptable}
\bibfield{author}{\bibinfo{person}{Sarah Amir} {and} \bibinfo{person}{Domenic
  Forte}.} \bibinfo{year}{2020}\natexlab{}.
\newblock \showarticletitle{Adaptable and divergent synthetic benchmark
  generation for hardware security}. In \bibinfo{booktitle}{\emph{Proceedings
  of the 39th International Conference on Computer-Aided Design}}.
  \bibinfo{pages}{1--9}.
\newblock


\bibitem[\protect\citeauthoryear{Amir, Shakya, Xu, Jin, Bhunia, Tehranipoor,
  and Forte}{Amir et~al\mbox{.}}{2018}]%
        {amir2018development}
\bibfield{author}{\bibinfo{person}{Sarah Amir}, \bibinfo{person}{Bicky Shakya},
  \bibinfo{person}{Xiaolin Xu}, \bibinfo{person}{Yier Jin},
  \bibinfo{person}{Swarup Bhunia}, \bibinfo{person}{Mark Tehranipoor}, {and}
  \bibinfo{person}{Domenic Forte}.} \bibinfo{year}{2018}\natexlab{}.
\newblock \showarticletitle{Development and evaluation of hardware obfuscation
  benchmarks}.
\newblock \bibinfo{journal}{\emph{Journal of Hardware and Systems Security}}
  \bibinfo{volume}{2}, \bibinfo{number}{2} (\bibinfo{year}{2018}),
  \bibinfo{pages}{142--161}.
\newblock


\bibitem[\protect\citeauthoryear{Ardeshiricham, Hu, Marxen, and
  Kastner}{Ardeshiricham et~al\mbox{.}}{2017}]%
        {Ardeshiricham2017}
\bibfield{author}{\bibinfo{person}{Armaiti Ardeshiricham}, \bibinfo{person}{Wei
  Hu}, \bibinfo{person}{Joshua Marxen}, {and} \bibinfo{person}{Ryan Kastner}.}
  \bibinfo{year}{2017}\natexlab{}.
\newblock \showarticletitle{Register transfer level information flow tracking
  for provably secure hardware design}. In \bibinfo{booktitle}{\emph{Design,
  Automation Test in Europe Conference Exhibition (DATE), 2017}}.
  \bibinfo{pages}{1691--1696}.
\newblock
\urldef\tempurl%
\url{https://doi.org/10.23919/DATE.2017.7927266}
\showDOI{\tempurl}


\bibitem[\protect\citeauthoryear{Arribas, Nikova, and Rijmen}{Arribas
  et~al\mbox{.}}{2018}]%
        {arribas2018vermi}
\bibfield{author}{\bibinfo{person}{Victor Arribas}, \bibinfo{person}{Svetla
  Nikova}, {and} \bibinfo{person}{Vincent Rijmen}.}
  \bibinfo{year}{2018}\natexlab{}.
\newblock \showarticletitle{{VerMI}: Verification tool for masked
  implementations}. In \bibinfo{booktitle}{\emph{2018 25th IEEE International
  Conference on Electronics, Circuits and Systems (ICECS)}}. IEEE,
  \bibinfo{pages}{381--384}.
\newblock


\bibitem[\protect\citeauthoryear{Babighian, Benini, Macii, and Macii}{Babighian
  et~al\mbox{.}}{2004}]%
        {babighian2004}
\bibfield{author}{\bibinfo{person}{Pietro Babighian}, \bibinfo{person}{Luca
  Benini}, \bibinfo{person}{Alberto Macii}, {and} \bibinfo{person}{Enrico
  Macii}.} \bibinfo{year}{2004}\natexlab{}.
\newblock \showarticletitle{Post-Layout Leakage Power Minimization Based on
  Distributed Sleep Transistor Insertion}. In
  \bibinfo{booktitle}{\emph{Proceedings of the 2004 International Symposium on
  Low Power Electronics and Design}} (Newport Beach, California, USA)
  \emph{(\bibinfo{series}{ISLPED '04})}. \bibinfo{publisher}{Association for
  Computing Machinery}, \bibinfo{address}{New York, NY, USA},
  \bibinfo{pages}{138–143}.
\newblock
\showISBNx{1581139292}
\urldef\tempurl%
\url{https://doi.org/10.1145/1013235.1013275}
\showDOI{\tempurl}


\bibitem[\protect\citeauthoryear{Bagnato, Indrusiak, Quadri, and Rossi}{Bagnato
  et~al\mbox{.}}{2014}]%
        {bagnato_indrusiak_quadri_rossi_2014}
\bibfield{author}{\bibinfo{person}{Alessandra Bagnato},
  \bibinfo{person}{Leandro~Soares Indrusiak}, \bibinfo{person}{Imran~Rafiq
  Quadri}, {and} \bibinfo{person}{Matteo Rossi}.}
  \bibinfo{year}{2014}\natexlab{}.
\newblock \bibinfo{booktitle}{\emph{Handbook of Research on Embedded Systems
  Design}}.
\newblock \bibinfo{publisher}{Information Science Reference}.
\newblock


\bibitem[\protect\citeauthoryear{Bandeira, Fogaça, Monteiro, Oliveira, Woo,
  and Reis}{Bandeira et~al\mbox{.}}{2020}]%
        {bandeira2020ioPlacer}
\bibfield{author}{\bibinfo{person}{Vitor Bandeira}, \bibinfo{person}{Mateus
  Fogaça}, \bibinfo{person}{Eder~Matheus Monteiro}, \bibinfo{person}{Isadora
  Oliveira}, \bibinfo{person}{Mingyu Woo}, {and} \bibinfo{person}{Ricardo
  Reis}.} \bibinfo{year}{2020}\natexlab{}.
\newblock \showarticletitle{Fast and Scalable I/O Pin Assignment with
  Divide-and-Conquer and Hungarian Matching}. In \bibinfo{booktitle}{\emph{2020
  18th IEEE International New Circuits and Systems Conference (NEWCAS)}}.
  \bibinfo{pages}{74--77}.
\newblock
\urldef\tempurl%
\url{https://doi.org/10.1109/NEWCAS49341.2020.9159791}
\showDOI{\tempurl}


\bibitem[\protect\citeauthoryear{Bar-El, Choukri, Naccache, Tunstall, and
  Whelan}{Bar-El et~al\mbox{.}}{2006}]%
        {bar2006sorcerer}
\bibfield{author}{\bibinfo{person}{Hagai Bar-El}, \bibinfo{person}{Hamid
  Choukri}, \bibinfo{person}{David Naccache}, \bibinfo{person}{Michael
  Tunstall}, {and} \bibinfo{person}{Claire Whelan}.}
  \bibinfo{year}{2006}\natexlab{}.
\newblock \showarticletitle{The sorcerer's apprentice guide to fault attacks}.
\newblock \bibinfo{journal}{\emph{Proc. IEEE}} \bibinfo{volume}{94},
  \bibinfo{number}{2} (\bibinfo{year}{2006}), \bibinfo{pages}{370--382}.
\newblock


\bibitem[\protect\citeauthoryear{Barredo~Arrieta, D{\'\i}az-Rodr{\'\i}guez,
  Del~Ser, Bennetot, Tabik, Barbado, Garcia, Gil-Lopez, Molina, Benjamins,
  Chatila, and Herrera}{Barredo~Arrieta et~al\mbox{.}}{2020}]%
        {Barredo_Arrieta2020-xai-concepts-taxonomies}
\bibfield{author}{\bibinfo{person}{Alejandro Barredo~Arrieta},
  \bibinfo{person}{Natalia D{\'\i}az-Rodr{\'\i}guez}, \bibinfo{person}{Javier
  Del~Ser}, \bibinfo{person}{Adrien Bennetot}, \bibinfo{person}{Siham Tabik},
  \bibinfo{person}{Alberto Barbado}, \bibinfo{person}{Salvador Garcia},
  \bibinfo{person}{Sergio Gil-Lopez}, \bibinfo{person}{Daniel Molina},
  \bibinfo{person}{Richard Benjamins}, \bibinfo{person}{Raja Chatila}, {and}
  \bibinfo{person}{Francisco Herrera}.} \bibinfo{year}{2020}\natexlab{}.
\newblock \showarticletitle{{Explainable Artificial Intelligence (XAI):
  Concepts, taxonomies, opportunities and challenges toward responsible AI}}.
\newblock \bibinfo{journal}{\emph{Inf. Fusion}}  \bibinfo{volume}{58}
  (\bibinfo{date}{June} \bibinfo{year}{2020}), \bibinfo{pages}{82--115}.
\newblock
\showISSN{1566-2535}
\urldef\tempurl%
\url{https://doi.org/10.1016/j.inffus.2019.12.012}
\showDOI{\tempurl}


\bibitem[\protect\citeauthoryear{Barthe, Bela{\"\i}d, Dupressoir, Fouque,
  Gr{\'e}goire, and Strub}{Barthe et~al\mbox{.}}{2015}]%
        {barthe2015verified}
\bibfield{author}{\bibinfo{person}{Gilles Barthe}, \bibinfo{person}{Sonia
  Bela{\"\i}d}, \bibinfo{person}{Fran{\c{c}}ois Dupressoir},
  \bibinfo{person}{Pierre-Alain Fouque}, \bibinfo{person}{Benjamin
  Gr{\'e}goire}, {and} \bibinfo{person}{Pierre-Yves Strub}.}
  \bibinfo{year}{2015}\natexlab{}.
\newblock \showarticletitle{Verified proofs of higher-order masking}. In
  \bibinfo{booktitle}{\emph{Annual International Conference on the Theory and
  Applications of Cryptographic Techniques}}. Springer,
  \bibinfo{pages}{457--485}.
\newblock


\bibitem[\protect\citeauthoryear{Bayern, Detwiler, Academy, Eckel, and
  Wallen}{Bayern et~al\mbox{.}}{2019}]%
        {bayern_detwiler_academy_eckel_wallen_2019}
\bibfield{author}{\bibinfo{person}{Macy Bayern}, \bibinfo{person}{Bill
  Detwiler}, \bibinfo{person}{TechRepublic Academy}, \bibinfo{person}{Erik
  Eckel}, {and} \bibinfo{person}{Jack Wallen}.}
  \bibinfo{year}{2019}\natexlab{}.
\newblock \bibinfo{title}{63$\%$ of organizations face security breaches due to
  hardware vulnerabilities}.
\newblock
\newblock
\urldef\tempurl%
\url{https://www.techrepublic.com/article/63-of-organizations-face-security-breaches-due-to-hardware-vulnerabilities/}
\showURL{%
\tempurl}


\bibitem[\protect\citeauthoryear{Bertoni, Martinoli, and Molteni}{Bertoni
  et~al\mbox{.}}{2017}]%
        {bertoni2017methodology}
\bibfield{author}{\bibinfo{person}{Guido Bertoni}, \bibinfo{person}{Marco
  Martinoli}, {and} \bibinfo{person}{Maria~Chiara Molteni}.}
  \bibinfo{year}{2017}\natexlab{}.
\newblock \showarticletitle{A methodology for the characterisation of leakages
  in combinatorial logic}.
\newblock \bibinfo{journal}{\emph{Journal of Hardware and Systems Security}}
  \bibinfo{volume}{1}, \bibinfo{number}{3} (\bibinfo{year}{2017}),
  \bibinfo{pages}{269--281}.
\newblock


\bibitem[\protect\citeauthoryear{Bhandari, Thalakkattu~Moosa, Tan, Pilato,
  Gore, Tang, Temple, Gaillardon, and Karri}{Bhandari et~al\mbox{.}}{2021}]%
        {Bhandari2021}
\bibfield{author}{\bibinfo{person}{Jitendra Bhandari},
  \bibinfo{person}{Abdul~Khader Thalakkattu~Moosa}, \bibinfo{person}{Benjamin
  Tan}, \bibinfo{person}{Christian Pilato}, \bibinfo{person}{Ganesh Gore},
  \bibinfo{person}{Xifan Tang}, \bibinfo{person}{Scott Temple},
  \bibinfo{person}{Pierre-Emmanuel Gaillardon}, {and} \bibinfo{person}{Ramesh
  Karri}.} \bibinfo{year}{2021}\natexlab{}.
\newblock \showarticletitle{Exploring eFPGA-based Redaction for IP Protection}.
  In \bibinfo{booktitle}{\emph{2021 IEEE/ACM International Conference On
  Computer Aided Design (ICCAD)}}. \bibinfo{pages}{1--9}.
\newblock
\urldef\tempurl%
\url{https://doi.org/10.1109/ICCAD51958.2021.9643548}
\showDOI{\tempurl}


\bibitem[\protect\citeauthoryear{Bhatia, Pandey, and Bhattacharyya}{Bhatia
  et~al\mbox{.}}{2016}]%
        {bhatia2016modelling}
\bibfield{author}{\bibinfo{person}{Veepsa Bhatia}, \bibinfo{person}{Neeta
  Pandey}, {and} \bibinfo{person}{Asok Bhattacharyya}.}
  \bibinfo{year}{2016}\natexlab{}.
\newblock \showarticletitle{Modelling and Design of Inverter Threshold
  Quantization Based Current Comparator using Artificial Neural Networks.}
\newblock \bibinfo{journal}{\emph{International Journal of Electrical \&
  Computer Engineering (2088-8708)}} \bibinfo{volume}{6}, \bibinfo{number}{1}
  (\bibinfo{year}{2016}).
\newblock


\bibitem[\protect\citeauthoryear{Binu and Kariyappa}{Binu and
  Kariyappa}{2018}]%
        {binu2018ridenn}
\bibfield{author}{\bibinfo{person}{D Binu} {and} \bibinfo{person}{BS
  Kariyappa}.} \bibinfo{year}{2018}\natexlab{}.
\newblock \showarticletitle{RideNN: A new rider optimization algorithm-based
  neural network for fault diagnosis in analog circuits}.
\newblock \bibinfo{journal}{\emph{IEEE Transactions on Instrumentation and
  Measurement}} \bibinfo{volume}{68}, \bibinfo{number}{1}
  (\bibinfo{year}{2018}), \bibinfo{pages}{2--26}.
\newblock


\bibitem[\protect\citeauthoryear{Bloem, Gro{\ss}, Iusupov, K{\"o}nighofer,
  Mangard, and Winter}{Bloem et~al\mbox{.}}{2018}]%
        {bloem2018formal}
\bibfield{author}{\bibinfo{person}{Roderick Bloem}, \bibinfo{person}{Hannes
  Gro{\ss}}, \bibinfo{person}{Rinat Iusupov}, \bibinfo{person}{Bettina
  K{\"o}nighofer}, \bibinfo{person}{Stefan Mangard}, {and}
  \bibinfo{person}{Johannes Winter}.} \bibinfo{year}{2018}\natexlab{}.
\newblock \showarticletitle{Formal verification of masked hardware
  implementations in the presence of glitches}. In
  \bibinfo{booktitle}{\emph{Annual International Conference on the Theory and
  Applications of Cryptographic Techniques}}. Springer,
  \bibinfo{pages}{321--353}.
\newblock


\bibitem[\protect\citeauthoryear{Bogaerts}{Bogaerts}{2014}]%
        {BOGAERTS2014250}
\bibfield{author}{\bibinfo{person}{J. Bogaerts}.}
  \bibinfo{year}{2014}\natexlab{}.
\newblock \showarticletitle{9 - Complementary metal-oxide-semiconductor (CMOS)
  image sensors for use in space}.
\newblock In \bibinfo{booktitle}{\emph{High Performance Silicon Imaging}},
  \bibfield{editor}{\bibinfo{person}{Daniel Durini}} (Ed.).
  \bibinfo{publisher}{Woodhead Publishing}, \bibinfo{pages}{250--280}.
\newblock
\showISBNx{978-0-85709-598-5}
\urldef\tempurl%
\url{https://doi.org/10.1533/9780857097521.2.250}
\showDOI{\tempurl}


\bibitem[\protect\citeauthoryear{Brayton, Hachtel, and
  Sangiovanni-Vincentelli}{Brayton et~al\mbox{.}}{1990}]%
        {brayton1990logicsyn}
\bibfield{author}{\bibinfo{person}{R.K. Brayton}, \bibinfo{person}{G.D.
  Hachtel}, {and} \bibinfo{person}{A.L. Sangiovanni-Vincentelli}.}
  \bibinfo{year}{1990}\natexlab{}.
\newblock \showarticletitle{Multilevel logic synthesis}.
\newblock \bibinfo{journal}{\emph{Proc. IEEE}} \bibinfo{volume}{78},
  \bibinfo{number}{2} (\bibinfo{year}{1990}), \bibinfo{pages}{264--300}.
\newblock
\urldef\tempurl%
\url{https://doi.org/10.1109/5.52213}
\showDOI{\tempurl}


\bibitem[\protect\citeauthoryear{Bronstein, Bruna, LeCun, Szlam, and
  Vandergheynst}{Bronstein et~al\mbox{.}}{2017}]%
        {Bronstein2017-geometric}
\bibfield{author}{\bibinfo{person}{Michael~M Bronstein}, \bibinfo{person}{Joan
  Bruna}, \bibinfo{person}{Yann LeCun}, \bibinfo{person}{Arthur Szlam}, {and}
  \bibinfo{person}{Pierre Vandergheynst}.} \bibinfo{year}{2017}\natexlab{}.
\newblock \showarticletitle{{Geometric Deep Learning: Going beyond Euclidean
  data}}.
\newblock \bibinfo{journal}{\emph{IEEE Signal Process. Mag.}}
  \bibinfo{volume}{34}, \bibinfo{number}{4} (\bibinfo{date}{July}
  \bibinfo{year}{2017}), \bibinfo{pages}{18--42}.
\newblock
\showISSN{1558-0792, 1558-0792}
\urldef\tempurl%
\url{https://doi.org/10.1109/MSP.2017.2693418}
\showDOI{\tempurl}


\bibitem[\protect\citeauthoryear{Buhan, Batina, Yarom, and Schaumont}{Buhan
  et~al\mbox{.}}{2021}]%
        {buhan2021sok}
\bibfield{author}{\bibinfo{person}{Ileana Buhan}, \bibinfo{person}{Lejla
  Batina}, \bibinfo{person}{Yuval Yarom}, {and} \bibinfo{person}{Patrick
  Schaumont}.} \bibinfo{year}{2021}\natexlab{}.
\newblock \bibinfo{title}{SoK: Design Tools for Side-Channel-Aware
  Implementations}.
\newblock
\newblock
\showeprint[arxiv]{2104.08593}~[cs.CR]


\bibitem[\protect\citeauthoryear{Bushnell and Agrawal}{Bushnell and
  Agrawal}{2004}]%
        {bushnell2004essentials}
\bibfield{author}{\bibinfo{person}{Michael Bushnell} {and}
  \bibinfo{person}{Vishwani Agrawal}.} \bibinfo{year}{2004}\natexlab{}.
\newblock \bibinfo{booktitle}{\emph{Essentials of electronic testing for
  digital, memory and mixed-signal VLSI circuits}}. Vol.~\bibinfo{volume}{17}.
\newblock \bibinfo{publisher}{Springer Science \& Business Media}.
\newblock


\bibitem[\protect\citeauthoryear{Cardoso, Coutinho, and Diniz}{Cardoso
  et~al\mbox{.}}{2017}]%
        {CARDOSO201799}
\bibfield{author}{\bibinfo{person}{João~M.P. Cardoso}, \bibinfo{person}{José
  Gabriel~F. Coutinho}, {and} \bibinfo{person}{Pedro~C. Diniz}.}
  \bibinfo{year}{2017}\natexlab{}.
\newblock \showarticletitle{Chapter 4 - Source code analysis and
  instrumentation}.
\newblock In \bibinfo{booktitle}{\emph{Embedded Computing for High
  Performance}}, \bibfield{editor}{\bibinfo{person}{João~M.P. Cardoso},
  \bibinfo{person}{José Gabriel~F. Coutinho}, {and} \bibinfo{person}{Pedro~C.
  Diniz}} (Eds.). \bibinfo{publisher}{Morgan Kaufmann},
  \bibinfo{address}{Boston}, \bibinfo{pages}{99--135}.
\newblock
\showISBNx{978-0-12-804189-5}
\urldef\tempurl%
\url{https://doi.org/10.1016/B978-0-12-804189-5.00004-1}
\showDOI{\tempurl}


\bibitem[\protect\citeauthoryear{Cassel and Lima}{Cassel and Lima}{2006}]%
        {cassel2006onehot}
\bibfield{author}{\bibinfo{person}{M. Cassel} {and} \bibinfo{person}{F. Lima}.}
  \bibinfo{year}{2006}\natexlab{}.
\newblock \showarticletitle{Evaluating one-hot encoding finite state machines
  for SEU reliability in SRAM-based FPGAs}. In \bibinfo{booktitle}{\emph{12th
  IEEE International On-Line Testing Symposium (IOLTS'06)}}. \bibinfo{pages}{6
  pp.--}.
\newblock
\urldef\tempurl%
\url{https://doi.org/10.1109/IOLTS.2006.32}
\showDOI{\tempurl}


\bibitem[\protect\citeauthoryear{Casto}{Casto}{2018}]%
        {casto2018multi}
\bibfield{author}{\bibinfo{person}{Matthew~James Casto}.}
  \bibinfo{year}{2018}\natexlab{}.
\newblock \emph{\bibinfo{title}{Multi-Attribute Design for Authentication and
  Reliability (MADAR)}}.
\newblock \bibinfo{thesistype}{Ph.D. Dissertation}. \bibinfo{school}{The Ohio
  State University}.
\newblock


\bibitem[\protect\citeauthoryear{Chakraborty, Cruz, and Bhunia}{Chakraborty
  et~al\mbox{.}}{2018}]%
        {chakraborty2018sail}
\bibfield{author}{\bibinfo{person}{Prabuddha Chakraborty},
  \bibinfo{person}{Jonathan Cruz}, {and} \bibinfo{person}{Swarup Bhunia}.}
  \bibinfo{year}{2018}\natexlab{}.
\newblock \showarticletitle{SAIL: Machine learning guided structural analysis
  attack on hardware obfuscation}. In \bibinfo{booktitle}{\emph{2018 Asian
  Hardware Oriented Security and Trust Symposium (AsianHOST)}}. IEEE,
  \bibinfo{pages}{56--61}.
\newblock


\bibitem[\protect\citeauthoryear{Chakraborty and Bhunia}{Chakraborty and
  Bhunia}{2009}]%
        {chakraborty2009harpoon}
\bibfield{author}{\bibinfo{person}{Rajat~Subhra Chakraborty} {and}
  \bibinfo{person}{Swarup Bhunia}.} \bibinfo{year}{2009}\natexlab{}.
\newblock \showarticletitle{HARPOON: An obfuscation-based SoC design
  methodology for hardware protection}.
\newblock \bibinfo{journal}{\emph{IEEE Transactions on Computer-Aided Design of
  Integrated Circuits and Systems}} \bibinfo{volume}{28}, \bibinfo{number}{10}
  (\bibinfo{year}{2009}), \bibinfo{pages}{1493--1502}.
\newblock


\bibitem[\protect\citeauthoryear{Chakraborty and Bhunia}{Chakraborty and
  Bhunia}{2010}]%
        {Chakraborty2010}
\bibfield{author}{\bibinfo{person}{Rajat~Subhra Chakraborty} {and}
  \bibinfo{person}{Swarup Bhunia}.} \bibinfo{year}{2010}\natexlab{}.
\newblock \showarticletitle{RTL Hardware IP Protection Using Key-Based Control
  and Data Flow Obfuscation}. In \bibinfo{booktitle}{\emph{2010 23rd
  International Conference on VLSI Design}}. \bibinfo{pages}{405--410}.
\newblock
\urldef\tempurl%
\url{https://doi.org/10.1109/VLSI.Design.2010.54}
\showDOI{\tempurl}


\bibitem[\protect\citeauthoryear{Chang, Potkonjak, and Zhang}{Chang
  et~al\mbox{.}}{2016}]%
        {Chang2016watermark}
\bibfield{author}{\bibinfo{person}{Chip-Hong Chang}, \bibinfo{person}{Miodrag
  Potkonjak}, {and} \bibinfo{person}{Li Zhang}.}
  \bibinfo{year}{2016}\natexlab{}.
\newblock \bibinfo{booktitle}{\emph{Hardware IP Watermarking and
  Fingerprinting}}.
\newblock \bibinfo{publisher}{Springer International Publishing},
  \bibinfo{address}{Cham}, \bibinfo{pages}{329--368}.
\newblock
\showISBNx{978-3-319-14971-4}
\urldef\tempurl%
\url{https://doi.org/10.1007/978-3-319-14971-4_10}
\showDOI{\tempurl}


\bibitem[\protect\citeauthoryear{Chang, Ozev, Sinanoglu, and Karri}{Chang
  et~al\mbox{.}}{2014}]%
        {chang2014approximating}
\bibfield{author}{\bibinfo{person}{Doohwang Chang}, \bibinfo{person}{Sule
  Ozev}, \bibinfo{person}{Ozgur Sinanoglu}, {and} \bibinfo{person}{Ramesh
  Karri}.} \bibinfo{year}{2014}\natexlab{}.
\newblock \showarticletitle{Approximating the age of RF/analog circuits through
  re-characterization and statistical estimation}. In
  \bibinfo{booktitle}{\emph{2014 Design, Automation \& Test in Europe
  Conference \& Exhibition (DATE)}}. IEEE, \bibinfo{pages}{1--4}.
\newblock


\bibitem[\protect\citeauthoryear{Chang, Chen, and Shen}{Chang
  et~al\mbox{.}}{2006}]%
        {chang2006}
\bibfield{author}{\bibinfo{person}{Po-Hao Chang}, \bibinfo{person}{Jia-Ming
  Chen}, {and} \bibinfo{person}{Chao-Ying Shen}.}
  \bibinfo{year}{2006}\natexlab{}.
\newblock \showarticletitle{On an Efficient Closed Form Expression to Estimate
  the Crosstalk Noise in the Circuit with Multiple Wires}. In
  \bibinfo{booktitle}{\emph{APCCAS 2006 - 2006 IEEE Asia Pacific Conference on
  Circuits and Systems}}. \bibinfo{pages}{1329--1332}.
\newblock
\urldef\tempurl%
\url{https://doi.org/10.1109/APCCAS.2006.342429}
\showDOI{\tempurl}


\bibitem[\protect\citeauthoryear{Chen, Fu, Zhao, and Koushanfar}{Chen
  et~al\mbox{.}}{2019}]%
        {chen2019genunlock}
\bibfield{author}{\bibinfo{person}{Huili Chen}, \bibinfo{person}{Cheng Fu},
  \bibinfo{person}{Jishen Zhao}, {and} \bibinfo{person}{Farinaz Koushanfar}.}
  \bibinfo{year}{2019}\natexlab{}.
\newblock \showarticletitle{GenUnlock: An Automated Genetic Algorithm Framework
  for Unlocking Logic Encryption}. In \bibinfo{booktitle}{\emph{2019 IEEE/ACM
  International Conference on Computer-Aided Design (ICCAD)}}.
  \bibinfo{pages}{1--8}.
\newblock
\urldef\tempurl%
\url{https://doi.org/10.1109/ICCAD45719.2019.8942134}
\showDOI{\tempurl}


\bibitem[\protect\citeauthoryear{Chen, Liu, Xu, Zhu, Tang, Li, Lin, Sun, and
  Pan}{Chen et~al\mbox{.}}{2020b}]%
        {chen2020magical}
\bibfield{author}{\bibinfo{person}{Hao Chen}, \bibinfo{person}{Mingjie Liu},
  \bibinfo{person}{Biying Xu}, \bibinfo{person}{Keren Zhu},
  \bibinfo{person}{Xiyuan Tang}, \bibinfo{person}{Shaolan Li},
  \bibinfo{person}{Yibo Lin}, \bibinfo{person}{Nan Sun}, {and}
  \bibinfo{person}{David~Z Pan}.} \bibinfo{year}{2020}\natexlab{b}.
\newblock \showarticletitle{MAGICAL: An open-source fully automated analog IC
  layout system from netlist to GDSII}.
\newblock \bibinfo{journal}{\emph{IEEE Design \& Test}} \bibinfo{volume}{38},
  \bibinfo{number}{2} (\bibinfo{year}{2020}), \bibinfo{pages}{19--26}.
\newblock


\bibitem[\protect\citeauthoryear{Chen, Kolhe, Rafatirad, Lu, Manoj~P.D.,
  Homayoun, and Zhao}{Chen et~al\mbox{.}}{2020a}]%
        {chen2020}
\bibfield{author}{\bibinfo{person}{Zhiqian Chen}, \bibinfo{person}{Gaurav
  Kolhe}, \bibinfo{person}{Setareh Rafatirad}, \bibinfo{person}{Chang-Tien Lu},
  \bibinfo{person}{Sai Manoj~P.D.}, \bibinfo{person}{Houman Homayoun}, {and}
  \bibinfo{person}{Liang Zhao}.} \bibinfo{year}{2020}\natexlab{a}.
\newblock \showarticletitle{Estimating the Circuit De-obfuscation Runtime based
  on Graph Deep Learning}. In \bibinfo{booktitle}{\emph{2020 Design, Automation
  Test in Europe Conference Exhibition (DATE)}}. \bibinfo{pages}{358--363}.
\newblock
\urldef\tempurl%
\url{https://doi.org/10.23919/DATE48585.2020.9116544}
\showDOI{\tempurl}


\bibitem[\protect\citeauthoryear{Chen, Raginsky, and Rosenbaum}{Chen
  et~al\mbox{.}}{2017}]%
        {chen2017verilog}
\bibfield{author}{\bibinfo{person}{Zaichen Chen}, \bibinfo{person}{Maxim
  Raginsky}, {and} \bibinfo{person}{Elyse Rosenbaum}.}
  \bibinfo{year}{2017}\natexlab{}.
\newblock \showarticletitle{Verilog-A compatible recurrent neural network model
  for transient circuit simulation}. In \bibinfo{booktitle}{\emph{2017 IEEE
  26th Conference on Electrical Performance of Electronic Packaging and Systems
  (EPEPS)}}. IEEE, \bibinfo{pages}{1--3}.
\newblock


\bibitem[\protect\citeauthoryear{Chow, Baukus, Wang, and Cocchi}{Chow
  et~al\mbox{.}}{2012}]%
        {chow2012camouflaging}
\bibfield{author}{\bibinfo{person}{Lap~Wai Chow}, \bibinfo{person}{James~P
  Baukus}, \bibinfo{person}{Bryan~J Wang}, {and} \bibinfo{person}{Ronald~P
  Cocchi}.} \bibinfo{year}{2012}\natexlab{}.
\newblock \bibinfo{title}{Camouflaging a standard cell based integrated
  circuit}.
\newblock
\newblock
\newblock
\shownote{US Patent 8,151,235.}


\bibitem[\protect\citeauthoryear{Chowdhury, Acharya, Boullion, Felder, Howard,
  Di, and Forte}{Chowdhury et~al\mbox{.}}{2020a}]%
        {chowdhury2020weak}
\bibfield{author}{\bibinfo{person}{Sreeja Chowdhury}, \bibinfo{person}{Rabin
  Acharya}, \bibinfo{person}{William Boullion}, \bibinfo{person}{Andrew
  Felder}, \bibinfo{person}{Mark Howard}, \bibinfo{person}{Jia Di}, {and}
  \bibinfo{person}{Domenic Forte}.} \bibinfo{year}{2020}\natexlab{a}.
\newblock \showarticletitle{A weak asynchronous reset (ares) puf using start-up
  characteristics of null conventional logic gates}. In
  \bibinfo{booktitle}{\emph{2020 IEEE International Test Conference (ITC)}}.
  IEEE, \bibinfo{pages}{1--10}.
\newblock


\bibitem[\protect\citeauthoryear{Chowdhury, Ganji, and Forte}{Chowdhury
  et~al\mbox{.}}{2020b}]%
        {chowdhury2020low}
\bibfield{author}{\bibinfo{person}{Sreeja Chowdhury}, \bibinfo{person}{Fatehmeh
  Ganji}, {and} \bibinfo{person}{Domenic Forte}.}
  \bibinfo{year}{2020}\natexlab{b}.
\newblock \showarticletitle{Low-cost remarked counterfeit IC detection using
  LDO regulators}. In \bibinfo{booktitle}{\emph{2020 IEEE International
  Symposium on Circuits and Systems (ISCAS)}}. IEEE, \bibinfo{pages}{1--5}.
\newblock


\bibitem[\protect\citeauthoryear{Chowdhury, Ganji, and Forte}{Chowdhury
  et~al\mbox{.}}{2020c}]%
        {chowdhury2020recycled}
\bibfield{author}{\bibinfo{person}{Sreeja Chowdhury}, \bibinfo{person}{Fatemeh
  Ganji}, {and} \bibinfo{person}{Domenic Forte}.}
  \bibinfo{year}{2020}\natexlab{c}.
\newblock \showarticletitle{Recycled SoC Detection Using LDO Degradation}.
\newblock \bibinfo{journal}{\emph{SN Computer Science}} \bibinfo{volume}{1},
  \bibinfo{number}{6} (\bibinfo{year}{2020}), \bibinfo{pages}{1--21}.
\newblock


\bibitem[\protect\citeauthoryear{Cong and Pan}{Cong and Pan}{2008}]%
        {kaufmann2008techmap}
\bibfield{author}{\bibinfo{person}{Jason Cong} {and} \bibinfo{person}{Peichen
  Pan}.} \bibinfo{year}{2008}\natexlab{}.
\newblock \showarticletitle{Chapter 13 - Technology Mapping}.
\newblock In \bibinfo{booktitle}{\emph{Reconfigurable Computing}},
  \bibfield{editor}{\bibinfo{person}{Scott Hauck} {and} \bibinfo{person}{André
  Dehon}} (Eds.). \bibinfo{publisher}{Morgan Kaufmann},
  \bibinfo{address}{Burlington}, \bibinfo{pages}{277--296}.
\newblock
\showISSN{18759661}
\urldef\tempurl%
\url{https://doi.org/10.1016/B978-012370522-8.50019-4}
\showDOI{\tempurl}


\bibitem[\protect\citeauthoryear{Connor}{Connor}{2018}]%
        {connor_2018}
\bibfield{author}{\bibinfo{person}{Katherine Connor}.}
  \bibinfo{year}{2018}\natexlab{}.
\newblock \bibinfo{title}{UC San Diego selected to lead development of
  open-source tools for hardware design automation}.
\newblock
\newblock
\urldef\tempurl%
\url{https://ucsdnews.ucsd.edu/pressrelease/uc_san_diego_selected_to_lead_development_of_open_source_tools_for_hardware_design_automation}
\showURL{%
\tempurl}


\bibitem[\protect\citeauthoryear{Contreras, Rahman, and Tehranipoor}{Contreras
  et~al\mbox{.}}{2013}]%
        {contreras2013}
\bibfield{author}{\bibinfo{person}{Gustavo~K. Contreras},
  \bibinfo{person}{Md.~Tauhidur Rahman}, {and} \bibinfo{person}{Mohammad
  Tehranipoor}.} \bibinfo{year}{2013}\natexlab{}.
\newblock \showarticletitle{Secure Split-Test for preventing IC piracy by
  untrusted foundry and assembly}. In \bibinfo{booktitle}{\emph{2013 IEEE
  International Symposium on Defect and Fault Tolerance in VLSI and
  Nanotechnology Systems (DFTS)}}. \bibinfo{pages}{196--203}.
\newblock
\urldef\tempurl%
\url{https://doi.org/10.1109/DFT.2013.6653606}
\showDOI{\tempurl}


\bibitem[\protect\citeauthoryear{Cornforth and Lipson}{Cornforth and
  Lipson}{2014}]%
        {cornforth2014reverse}
\bibfield{author}{\bibinfo{person}{Theodore~W Cornforth} {and}
  \bibinfo{person}{Hod Lipson}.} \bibinfo{year}{2014}\natexlab{}.
\newblock \showarticletitle{Reverse-Engineering Nonlinear Analog Circuits with
  Evolutionary Computation}. In \bibinfo{booktitle}{\emph{International
  Conference on Unconventional Computation and Natural Computation}}. Springer,
  \bibinfo{pages}{105--116}.
\newblock


\bibitem[\protect\citeauthoryear{Coron}{Coron}{2018}]%
        {coron2018formal}
\bibfield{author}{\bibinfo{person}{Jean-S{\'e}bastien Coron}.}
  \bibinfo{year}{2018}\natexlab{}.
\newblock \showarticletitle{Formal verification of side-channel countermeasures
  via elementary circuit transformations}. In
  \bibinfo{booktitle}{\emph{International Conference on Applied Cryptography
  and Network Security}}. Springer, \bibinfo{pages}{65--82}.
\newblock


\bibitem[\protect\citeauthoryear{Cruz, Farahmandi, Ahmed, and Mishra}{Cruz
  et~al\mbox{.}}{2018}]%
        {cruz2018}
\bibfield{author}{\bibinfo{person}{Jonathan Cruz}, \bibinfo{person}{Farimah
  Farahmandi}, \bibinfo{person}{Alif Ahmed}, {and} \bibinfo{person}{Prabhat
  Mishra}.} \bibinfo{year}{2018}\natexlab{}.
\newblock \showarticletitle{Hardware Trojan Detection Using ATPG and Model
  Checking}. In \bibinfo{booktitle}{\emph{2018 31st International Conference on
  VLSI Design and 2018 17th International Conference on Embedded Systems
  (VLSID)}}. \bibinfo{pages}{91--96}.
\newblock
\urldef\tempurl%
\url{https://doi.org/10.1109/VLSID.2018.43}
\showDOI{\tempurl}


\bibitem[\protect\citeauthoryear{cwe@mitre.org}{cwe@mitre.org}{2022}]%
        {CWE_MITRE}
\bibfield{author}{\bibinfo{person}{cwe@mitre.org}.}
  \bibinfo{year}{2022}\natexlab{}.
\newblock \bibinfo{title}{HW CWE SIG}.
\newblock
\newblock
\newblock
\shownote{\url{https://cwe.mitre.org/documents/HW_CWE_SIG.pdf}.}


\bibitem[\protect\citeauthoryear{Dai, Zhou, Zhang, Ustun, Young, and Zhang}{Dai
  et~al\mbox{.}}{2018}]%
        {Dai2018}
\bibfield{author}{\bibinfo{person}{Steve Dai}, \bibinfo{person}{Yuan Zhou},
  \bibinfo{person}{Hang Zhang}, \bibinfo{person}{Ecenur Ustun},
  \bibinfo{person}{Evangeline~F.Y. Young}, {and} \bibinfo{person}{Zhiru
  Zhang}.} \bibinfo{year}{2018}\natexlab{}.
\newblock \showarticletitle{Fast and Accurate Estimation of Quality of Results
  in High-Level Synthesis with Machine Learning}.
\newblock \bibinfo{journal}{\emph{Proceedings - 26th IEEE International
  Symposium on Field-Programmable Custom Computing Machines, FCCM 2018}},
  \bibinfo{pages}{129--132}.
\newblock
\showISBNx{9781538655221}
\urldef\tempurl%
\url{https://doi.org/10.1109/FCCM.2018.00029}
\showDOI{\tempurl}


\bibitem[\protect\citeauthoryear{Das and Vemuri}{Das and Vemuri}{2007}]%
        {das2007automated}
\bibfield{author}{\bibinfo{person}{Angan Das} {and} \bibinfo{person}{Ranga
  Vemuri}.} \bibinfo{year}{2007}\natexlab{}.
\newblock \showarticletitle{An automated passive analog circuit synthesis
  framework using genetic algorithms}. In \bibinfo{booktitle}{\emph{IEEE
  computer society annual symposium on VLSI (ISVLSI'07)}}. IEEE,
  \bibinfo{pages}{145--152}.
\newblock


\bibitem[\protect\citeauthoryear{De~Meyer, Arribas~Abril, Nikova, Nikov, and
  Rijmen}{De~Meyer et~al\mbox{.}}{2018}]%
        {MandM}
\bibfield{author}{\bibinfo{person}{Lauren De~Meyer}, \bibinfo{person}{Victor
  Arribas~Abril}, \bibinfo{person}{Svetla Nikova}, \bibinfo{person}{Ventzislav
  Nikov}, {and} \bibinfo{person}{Vincent Rijmen}.}
  \bibinfo{year}{2018}\natexlab{}.
\newblock \showarticletitle{M\&m: Masks and macs against physical attacks}.
\newblock \bibinfo{journal}{\emph{IACR Transactions on Cryptographic Hardware
  and Embedded Systems}} \bibinfo{volume}{2019}, \bibinfo{number}{1}
  (\bibinfo{year}{2018}), \bibinfo{pages}{25--50}.
\newblock


\bibitem[\protect\citeauthoryear{Ding, Gao, Yuan, and Pan}{Ding
  et~al\mbox{.}}{2011}]%
        {ding2011aeneid}
\bibfield{author}{\bibinfo{person}{Duo Ding}, \bibinfo{person}{Jhih-Rong Gao},
  \bibinfo{person}{Kun Yuan}, {and} \bibinfo{person}{David~Z Pan}.}
  \bibinfo{year}{2011}\natexlab{}.
\newblock \showarticletitle{AENEID: a generic lithography-friendly detailed
  router based on post-RET data learning and hotspot detection}. In
  \bibinfo{booktitle}{\emph{Proceedings of the 48th Design Automation
  Conference}}. \bibinfo{pages}{795--800}.
\newblock


\bibitem[\protect\citeauthoryear{Ding and Vemur}{Ding and Vemur}{2005}]%
        {ding2005active}
\bibfield{author}{\bibinfo{person}{Mengmeng Ding} {and} \bibinfo{person}{RI
  Vemur}.} \bibinfo{year}{2005}\natexlab{}.
\newblock \showarticletitle{An active learning scheme using support vector
  machines for analog circuit feasibility classification}. In
  \bibinfo{booktitle}{\emph{18th International Conference on VLSI Design held
  jointly with 4th International Conference on Embedded Systems Design}}. IEEE,
  \bibinfo{pages}{528--534}.
\newblock


\bibitem[\protect\citeauthoryear{Docking and Sachdev}{Docking and
  Sachdev}{2003}]%
        {docking2003ro}
\bibfield{author}{\bibinfo{person}{S. Docking} {and} \bibinfo{person}{M.
  Sachdev}.} \bibinfo{year}{2003}\natexlab{}.
\newblock \showarticletitle{A method to derive an equation for the oscillation
  frequency of a ring oscillator}.
\newblock \bibinfo{journal}{\emph{IEEE Transactions on Circuits and Systems I:
  Fundamental Theory and Applications}} \bibinfo{volume}{50},
  \bibinfo{number}{2} (\bibinfo{year}{2003}), \bibinfo{pages}{259--264}.
\newblock
\urldef\tempurl%
\url{https://doi.org/10.1109/TCSI.2002.808235}
\showDOI{\tempurl}


\bibitem[\protect\citeauthoryear{Dumesnil, Nabki, and Boukadoum}{Dumesnil
  et~al\mbox{.}}{2014}]%
        {dumesnil2014rf}
\bibfield{author}{\bibinfo{person}{Etienne Dumesnil}, \bibinfo{person}{Frederic
  Nabki}, {and} \bibinfo{person}{Mounir Boukadoum}.}
  \bibinfo{year}{2014}\natexlab{}.
\newblock \showarticletitle{RF-LNA circuit synthesis by genetic
  algorithm-specified artificial neural network}. In
  \bibinfo{booktitle}{\emph{2014 21st IEEE International Conference on
  Electronics, Circuits and Systems (ICECS)}}. IEEE, \bibinfo{pages}{758--761}.
\newblock


\bibitem[\protect\citeauthoryear{Dumesnil, Nabki, and Boukadoum}{Dumesnil
  et~al\mbox{.}}{2015}]%
        {dumesnil2015rf}
\bibfield{author}{\bibinfo{person}{Etienne Dumesnil}, \bibinfo{person}{Frederic
  Nabki}, {and} \bibinfo{person}{Mounir Boukadoum}.}
  \bibinfo{year}{2015}\natexlab{}.
\newblock \showarticletitle{RF-LNA circuit synthesis using an array of
  artificial neural networks with constrained inputs}. In
  \bibinfo{booktitle}{\emph{2015 IEEE International Symposium on Circuits and
  Systems (ISCAS)}}. IEEE, \bibinfo{pages}{573--576}.
\newblock


\bibitem[\protect\citeauthoryear{Dunbar and Qu}{Dunbar and Qu}{2015}]%
        {Dunbar2015SatisfiabilityDC}
\bibfield{author}{\bibinfo{person}{Carson Dunbar} {and} \bibinfo{person}{Gang
  Qu}.} \bibinfo{year}{2015}\natexlab{}.
\newblock \showarticletitle{Satisfiability Don't Care condition based circuit
  fingerprinting techniques}.
\newblock \bibinfo{journal}{\emph{The 20th Asia and South Pacific Design
  Automation Conference}} (\bibinfo{year}{2015}), \bibinfo{pages}{815--820}.
\newblock


\bibitem[\protect\citeauthoryear{Education}{Education}{[n.d.]}]%
        {ibm}
\bibfield{author}{\bibinfo{person}{By: IBM~Cloud Education}.}
  \bibinfo{year}{[n.d.]}\natexlab{}.
\newblock \bibinfo{title}{What is deep learning?}
\newblock
\newblock
\urldef\tempurl%
\url{https://www.ibm.com/cloud/learn/deep-learning}
\showURL{%
\tempurl}


\bibitem[\protect\citeauthoryear{El~Massad, Garg, and Tripunitara}{El~Massad
  et~al\mbox{.}}{2015}]%
        {el2015integrated}
\bibfield{author}{\bibinfo{person}{Mohamed El~Massad},
  \bibinfo{person}{Siddharth Garg}, {and} \bibinfo{person}{Mahesh~V
  Tripunitara}.} \bibinfo{year}{2015}\natexlab{}.
\newblock \showarticletitle{Integrated Circuit (IC) Decamouflaging: Reverse
  Engineering Camouflaged ICs within Minutes.}. In
  \bibinfo{booktitle}{\emph{NDSS}}. \bibinfo{pages}{1--14}.
\newblock


\bibitem[\protect\citeauthoryear{Elshamy, Di~Natale, Pavlidis, Lou{\"e}rat, and
  Stratigopoulos}{Elshamy et~al\mbox{.}}{2020}]%
        {elshamy2020hardware}
\bibfield{author}{\bibinfo{person}{Mohamed Elshamy}, \bibinfo{person}{Giorgio
  Di~Natale}, \bibinfo{person}{Antonios Pavlidis},
  \bibinfo{person}{Marie-Minerve Lou{\"e}rat}, {and}
  \bibinfo{person}{Haralampos-G Stratigopoulos}.}
  \bibinfo{year}{2020}\natexlab{}.
\newblock \showarticletitle{Hardware trojan attacks in analog/mixed-signal ICs
  via the test access mechanism}. In \bibinfo{booktitle}{\emph{2020 IEEE
  European Test Symposium (ETS)}}. IEEE, \bibinfo{pages}{1--6}.
\newblock


\bibitem[\protect\citeauthoryear{Foga{\c{c}}a, Kahng, Reis, and
  Wang}{Foga{\c{c}}a et~al\mbox{.}}{2019}]%
        {fogacca2019finding}
\bibfield{author}{\bibinfo{person}{Mateus Foga{\c{c}}a},
  \bibinfo{person}{Andrew~B Kahng}, \bibinfo{person}{Ricardo Reis}, {and}
  \bibinfo{person}{Lutong Wang}.} \bibinfo{year}{2019}\natexlab{}.
\newblock \showarticletitle{Finding placement-relevant clusters with fast
  modularity-based clustering}. In \bibinfo{booktitle}{\emph{Proceedings of the
  24th Asia and South Pacific Design Automation Conference}}.
  \bibinfo{pages}{569--576}.
\newblock


\bibitem[\protect\citeauthoryear{Froemmer, Gowayed, Bannow, Kunz, Grimm, and
  Schneider}{Froemmer et~al\mbox{.}}{2020}]%
        {froemmer2020}
\bibfield{author}{\bibinfo{person}{Jens Froemmer}, \bibinfo{person}{Yara
  Gowayed}, \bibinfo{person}{Nico Bannow}, \bibinfo{person}{Wolfgang Kunz},
  \bibinfo{person}{Christoph Grimm}, {and} \bibinfo{person}{Klaus Schneider}.}
  \bibinfo{year}{2020}\natexlab{}.
\newblock \showarticletitle{Area Estimation Framework for Digital Hardware
  Design using Machine Learning}. In \bibinfo{booktitle}{\emph{MBMV 2020 -
  Methods and Description Languages for Modelling and Verification of Circuits
  and Systems; GMM/ITG/GI-Workshop}}. \bibinfo{pages}{1--10}.
\newblock


\bibitem[\protect\citeauthoryear{Fung}{Fung}{2022}]%
        {fung_2022}
\bibfield{author}{\bibinfo{person}{Jason Fung}.}
  \bibinfo{year}{2022}\natexlab{}.
\newblock \bibinfo{title}{7 essentials for more security-aware design
  automation}.
\newblock
\newblock
\newblock
\shownote{\url{https://www.darkreading.com/vulnerabilities-threats/7-essentials-for-more-security-aware-design-automation}.}


\bibitem[\protect\citeauthoryear{Gabriel, Wittmann, Sych, Dong, Mauerer,
  Andersen, Marquardt, and Leuchs}{Gabriel et~al\mbox{.}}{2010}]%
        {gabriel_wittmann_sych_dong_mauerer_andersen_marquardt_leuchs_2010}
\bibfield{author}{\bibinfo{person}{Christian Gabriel},
  \bibinfo{person}{Christoffer Wittmann}, \bibinfo{person}{Denis Sych},
  \bibinfo{person}{Ruifang Dong}, \bibinfo{person}{Wolfgang Mauerer},
  \bibinfo{person}{Ulrik~L. Andersen}, \bibinfo{person}{Christoph Marquardt},
  {and} \bibinfo{person}{Gerd Leuchs}.} \bibinfo{year}{2010}\natexlab{}.
\newblock \showarticletitle{A generator for unique quantum random numbers based
  on vacuum states}.
\newblock \bibinfo{journal}{\emph{Nature Photonics}} \bibinfo{volume}{4},
  \bibinfo{number}{10} (\bibinfo{year}{2010}), \bibinfo{pages}{711–715}.
\newblock
\urldef\tempurl%
\url{https://doi.org/10.1038/nphoton.2010.197}
\showDOI{\tempurl}


\bibitem[\protect\citeauthoryear{Gao, Wang, Tehranipoor, and Forte}{Gao
  et~al\mbox{.}}{[n.d.]}]%
        {osti_10174121}
\bibfield{author}{\bibinfo{person}{Minyan Gao}, \bibinfo{person}{Huanyu Wang},
  \bibinfo{person}{Mark~M. Tehranipoor}, {and} \bibinfo{person}{Domenic
  Forte}.} \bibinfo{year}{[n.d.]}\natexlab{}.
\newblock \showarticletitle{iPROBE V2: Internal Shielding-based Countermeasures
  against Both Back-side and Front-side Probing Attacks}.
\newblock \bibinfo{journal}{\emph{SRC TECHCON}} (\bibinfo{year}{[n.\,d.]}).
\newblock
\urldef\tempurl%
\url{https://par.nsf.gov/biblio/10174121}
\showURL{%
\tempurl}


\bibitem[\protect\citeauthoryear{Gao, Tao, Yang, Su, Zhou, and Zeng}{Gao
  et~al\mbox{.}}{2019}]%
        {gao2019efficient}
\bibfield{author}{\bibinfo{person}{Zhengqi Gao}, \bibinfo{person}{Jun Tao},
  \bibinfo{person}{Fan Yang}, \bibinfo{person}{Yangfeng Su},
  \bibinfo{person}{Dian Zhou}, {and} \bibinfo{person}{Xuan Zeng}.}
  \bibinfo{year}{2019}\natexlab{}.
\newblock \showarticletitle{Efficient performance trade-off modeling for analog
  circuit based on Bayesian neural network}. In \bibinfo{booktitle}{\emph{2019
  IEEE/ACM International Conference on Computer-Aided Design (ICCAD)}}. IEEE,
  \bibinfo{pages}{1--8}.
\newblock


\bibitem[\protect\citeauthoryear{Gassend, Clarke, Van~Dijk, and
  Devadas}{Gassend et~al\mbox{.}}{2002}]%
        {gassend2002silicon}
\bibfield{author}{\bibinfo{person}{Blaise Gassend}, \bibinfo{person}{Dwaine
  Clarke}, \bibinfo{person}{Marten Van~Dijk}, {and} \bibinfo{person}{Srinivas
  Devadas}.} \bibinfo{year}{2002}\natexlab{}.
\newblock \showarticletitle{Silicon physical random functions}. In
  \bibinfo{booktitle}{\emph{Proceedings of the 9th ACM Conference on Computer
  and Communications Security}}. \bibinfo{pages}{148--160}.
\newblock


\bibitem[\protect\citeauthoryear{Ghorbani, Abid, and Zou}{Ghorbani
  et~al\mbox{.}}{2019}]%
        {Ghorbani2019-fragile-interp}
\bibfield{author}{\bibinfo{person}{Amirata Ghorbani}, \bibinfo{person}{Abubakar
  Abid}, {and} \bibinfo{person}{James Zou}.} \bibinfo{year}{2019}\natexlab{}.
\newblock \showarticletitle{{Interpretation of Neural Networks Is Fragile}}.
\newblock \bibinfo{journal}{\emph{AAAI}} \bibinfo{volume}{33},
  \bibinfo{number}{01} (\bibinfo{date}{July} \bibinfo{year}{2019}),
  \bibinfo{pages}{3681--3688}.
\newblock
\showISSN{2374-3468, 2374-3468}
\urldef\tempurl%
\url{https://doi.org/10.1609/aaai.v33i01.33013681}
\showDOI{\tempurl}


\bibitem[\protect\citeauthoryear{Gielen and Rutenbar}{Gielen and
  Rutenbar}{2000}]%
        {gielen2000computer}
\bibfield{author}{\bibinfo{person}{Georges~GE Gielen} {and}
  \bibinfo{person}{Rob~A Rutenbar}.} \bibinfo{year}{2000}\natexlab{}.
\newblock \showarticletitle{Computer-aided design of analog and mixed-signal
  integrated circuits}.
\newblock \bibinfo{journal}{\emph{Proc. IEEE}} \bibinfo{volume}{88},
  \bibinfo{number}{12} (\bibinfo{year}{2000}), \bibinfo{pages}{1825--1854}.
\newblock


\bibitem[\protect\citeauthoryear{Giraud and Thillard}{Giraud and
  Thillard}{2010}]%
        {giraud2010}
\bibfield{author}{\bibinfo{person}{Christophe Giraud} {and}
  \bibinfo{person}{Adrian Thillard}.} \bibinfo{year}{2010}\natexlab{}.
\newblock \showarticletitle{Piret and Quisquater's DFA on AES Revisited.}
\newblock \bibinfo{journal}{\emph{IACR Cryptology ePrint Archive}}
  \bibinfo{volume}{2010} (\bibinfo{date}{01} \bibinfo{year}{2010}),
  \bibinfo{pages}{440}.
\newblock


\bibitem[\protect\citeauthoryear{Goldie and Mirhoseini}{Goldie and
  Mirhoseini}{2020}]%
        {goldie2020placement}
\bibfield{author}{\bibinfo{person}{Anna Goldie} {and} \bibinfo{person}{Azalia
  Mirhoseini}.} \bibinfo{year}{2020}\natexlab{}.
\newblock \showarticletitle{Placement optimization with deep reinforcement
  learning}. In \bibinfo{booktitle}{\emph{Proceedings of the 2020 International
  Symposium on Physical Design}}. \bibinfo{pages}{3--7}.
\newblock


\bibitem[\protect\citeauthoryear{Grabmann, Feldhoff, and Gl{\"a}ser}{Grabmann
  et~al\mbox{.}}{2019}]%
        {grabmann2019power}
\bibfield{author}{\bibinfo{person}{Martin Grabmann}, \bibinfo{person}{Frank
  Feldhoff}, {and} \bibinfo{person}{Georg Gl{\"a}ser}.}
  \bibinfo{year}{2019}\natexlab{}.
\newblock \showarticletitle{Power to the model: Generating energy-aware
  mixed-signal models using machine learning}. In
  \bibinfo{booktitle}{\emph{2019 16th International Conference on Synthesis,
  Modeling, Analysis and Simulation Methods and Applications to Circuit Design
  (SMACD)}}. IEEE, \bibinfo{pages}{5--8}.
\newblock


\bibitem[\protect\citeauthoryear{Green}{Green}{1999}]%
        {green1999}
\bibfield{author}{\bibinfo{person}{L. Green}.} \bibinfo{year}{1999}\natexlab{}.
\newblock \showarticletitle{Understanding the importance of signal integrity}.
\newblock \bibinfo{journal}{\emph{IEEE Circuits and Devices Magazine}}
  \bibinfo{volume}{15}, \bibinfo{number}{6} (\bibinfo{year}{1999}),
  \bibinfo{pages}{7--10}.
\newblock
\urldef\tempurl%
\url{https://doi.org/10.1109/101.808850}
\showDOI{\tempurl}


\bibitem[\protect\citeauthoryear{Guan, Wang, Xin, Wang, and Zhang}{Guan
  et~al\mbox{.}}{2020}]%
        {guan2020survey}
\bibfield{author}{\bibinfo{person}{Zhibin Guan}, \bibinfo{person}{Xiaomeng
  Wang}, \bibinfo{person}{Wei Xin}, \bibinfo{person}{Jiajie Wang}, {and}
  \bibinfo{person}{Li Zhang}.} \bibinfo{year}{2020}\natexlab{}.
\newblock \showarticletitle{A survey on deep learning-based source code defect
  analysis}. In \bibinfo{booktitle}{\emph{2020 5th International Conference on
  Computer and Communication Systems (ICCCS)}}. IEEE,
  \bibinfo{pages}{167--171}.
\newblock


\bibitem[\protect\citeauthoryear{Guerra, Canelas, P{\'o}voa, Horta,
  Louren{\c{c}}o, and Martins}{Guerra et~al\mbox{.}}{2019}]%
        {guerra2019artificial}
\bibfield{author}{\bibinfo{person}{Daniel Guerra}, \bibinfo{person}{Ant{\'o}nio
  Canelas}, \bibinfo{person}{Ricardo P{\'o}voa}, \bibinfo{person}{Nuno Horta},
  \bibinfo{person}{Nuno Louren{\c{c}}o}, {and} \bibinfo{person}{Ricardo
  Martins}.} \bibinfo{year}{2019}\natexlab{}.
\newblock \showarticletitle{Artificial neural networks as an alternative for
  automatic analog IC placement}. In \bibinfo{booktitle}{\emph{2019 16th
  International Conference on Synthesis, Modeling, Analysis and Simulation
  Methods and Applications to Circuit Design (SMACD)}}. IEEE,
  \bibinfo{pages}{1--4}.
\newblock


\bibitem[\protect\citeauthoryear{Guin, Forte, and Tehranipoor}{Guin
  et~al\mbox{.}}{2015}]%
        {guin2015design}
\bibfield{author}{\bibinfo{person}{Ujjwal Guin}, \bibinfo{person}{Domenic
  Forte}, {and} \bibinfo{person}{Mark Tehranipoor}.}
  \bibinfo{year}{2015}\natexlab{}.
\newblock \showarticletitle{Design of accurate low-cost on-chip structures for
  protecting integrated circuits against recycling}.
\newblock \bibinfo{journal}{\emph{IEEE Transactions on Very Large Scale
  Integration (VLSI) Systems}} \bibinfo{volume}{24}, \bibinfo{number}{4}
  (\bibinfo{year}{2015}), \bibinfo{pages}{1233--1246}.
\newblock


\bibitem[\protect\citeauthoryear{Guin, Zhang, Forte, and Tehranipoor}{Guin
  et~al\mbox{.}}{2014}]%
        {guin2014}
\bibfield{author}{\bibinfo{person}{Ujjwal Guin}, \bibinfo{person}{Xuehui
  Zhang}, \bibinfo{person}{Domenic Forte}, {and} \bibinfo{person}{Mohammad
  Tehranipoor}.} \bibinfo{year}{2014}\natexlab{}.
\newblock \showarticletitle{Low-cost on-chip structures for combating die and
  IC recycling}. In \bibinfo{booktitle}{\emph{2014 51st ACM/EDAC/IEEE Design
  Automation Conference (DAC)}}. \bibinfo{pages}{1--6}.
\newblock
\urldef\tempurl%
\url{https://doi.org/10.1145/2593069.2593157}
\showDOI{\tempurl}


\bibitem[\protect\citeauthoryear{GummidipoondiJayasankaran, Borbon, Sinencio,
  Hu, and Rajendran}{GummidipoondiJayasankaran et~al\mbox{.}}{2020}]%
        {gummidipoondijayasankaran2020towards}
\bibfield{author}{\bibinfo{person}{Nithyashankari GummidipoondiJayasankaran},
  \bibinfo{person}{Adriana~Sanabria Borbon}, \bibinfo{person}{Edgar~Sanchez
  Sinencio}, \bibinfo{person}{Jiang Hu}, {and} \bibinfo{person}{Jeyavijayan
  Rajendran}.} \bibinfo{year}{2020}\natexlab{}.
\newblock \showarticletitle{Towards Provably-Secure Analog and Mixed-Signal
  Locking Against Overproduction}.
\newblock \bibinfo{journal}{\emph{IEEE Transactions on Emerging Topics in
  Computing}} (\bibinfo{year}{2020}).
\newblock


\bibitem[\protect\citeauthoryear{Gusm{\~a}o, Passos, P{\'o}voa, Horta,
  Louren{\c{c}}o, and Martins}{Gusm{\~a}o et~al\mbox{.}}{2020}]%
        {gusmao2020semi}
\bibfield{author}{\bibinfo{person}{Ant{\'o}nio Gusm{\~a}o},
  \bibinfo{person}{F{\'a}bio Passos}, \bibinfo{person}{Ricardo P{\'o}voa},
  \bibinfo{person}{Nuno Horta}, \bibinfo{person}{Nuno Louren{\c{c}}o}, {and}
  \bibinfo{person}{Ricardo Martins}.} \bibinfo{year}{2020}\natexlab{}.
\newblock \showarticletitle{Semi-supervised artificial neural networks towards
  analog IC placement recommender}. In \bibinfo{booktitle}{\emph{2020 IEEE
  International Symposium on Circuits and Systems (ISCAS)}}. IEEE,
  \bibinfo{pages}{1--5}.
\newblock


\bibitem[\protect\citeauthoryear{Haaswijk, Collins, Seguin, Soeken, Kaplan,
  Süsstrunk, and De~Micheli}{Haaswijk et~al\mbox{.}}{2018}]%
        {Haaswijk2018}
\bibfield{author}{\bibinfo{person}{Winston Haaswijk}, \bibinfo{person}{Edo
  Collins}, \bibinfo{person}{Benoit Seguin}, \bibinfo{person}{Mathias Soeken},
  \bibinfo{person}{Frédéric Kaplan}, \bibinfo{person}{Sabine Süsstrunk},
  {and} \bibinfo{person}{Giovanni De~Micheli}.}
  \bibinfo{year}{2018}\natexlab{}.
\newblock \showarticletitle{Deep Learning for Logic Optimization Algorithms}.
  In \bibinfo{booktitle}{\emph{2018 IEEE International Symposium on Circuits
  and Systems (ISCAS)}}. \bibinfo{pages}{1--4}.
\newblock
\urldef\tempurl%
\url{https://doi.org/10.1109/ISCAS.2018.8351885}
\showDOI{\tempurl}


\bibitem[\protect\citeauthoryear{Hakhamaneshi, Werblun, Abbeel, and
  Stojanovi{\'c}}{Hakhamaneshi et~al\mbox{.}}{2019}]%
        {hakhamaneshi2019bagnet}
\bibfield{author}{\bibinfo{person}{Kourosh Hakhamaneshi}, \bibinfo{person}{Nick
  Werblun}, \bibinfo{person}{Pieter Abbeel}, {and} \bibinfo{person}{Vladimir
  Stojanovi{\'c}}.} \bibinfo{year}{2019}\natexlab{}.
\newblock \showarticletitle{BagNet: Berkeley analog generator with layout
  optimizer boosted with deep neural networks}. In
  \bibinfo{booktitle}{\emph{2019 IEEE/ACM International Conference on
  Computer-Aided Design (ICCAD)}}. IEEE, \bibinfo{pages}{1--8}.
\newblock


\bibitem[\protect\citeauthoryear{Hamilton}{Hamilton}{2020}]%
        {Hamilton2020-grl}
\bibfield{author}{\bibinfo{person}{William~L Hamilton}.}
  \bibinfo{year}{2020}\natexlab{}.
\newblock \showarticletitle{{Graph Representation Learning}}.
\newblock \bibinfo{journal}{\emph{Synthesis Lectures on Artificial Intelligence
  and Machine Learning}} \bibinfo{volume}{14}, \bibinfo{number}{3}
  (\bibinfo{date}{Sept.} \bibinfo{year}{2020}), \bibinfo{pages}{1--159}.
\newblock
\showISSN{1939-4608}
\urldef\tempurl%
\url{https://doi.org/10.2200/S01045ED1V01Y202009AIM046}
\showDOI{\tempurl}


\bibitem[\protect\citeauthoryear{Han, Kahng, and Li}{Han et~al\mbox{.}}{2020}]%
        {Han2020OptimalGH}
\bibfield{author}{\bibinfo{person}{Kwangsoo Han}, \bibinfo{person}{Andrew~B.
  Kahng}, {and} \bibinfo{person}{Jiajia Li}.} \bibinfo{year}{2020}\natexlab{}.
\newblock \showarticletitle{Optimal Generalized H-Tree Topology and Buffering
  for High-Performance and Low-Power Clock Distribution}.
\newblock \bibinfo{journal}{\emph{IEEE Transactions on Computer-Aided Design of
  Integrated Circuits and Systems}}  \bibinfo{volume}{39}
  (\bibinfo{year}{2020}), \bibinfo{pages}{478--491}.
\newblock


\bibitem[\protect\citeauthoryear{Hanna and Cabric}{Hanna and Cabric}{2019}]%
        {hanna2019deep}
\bibfield{author}{\bibinfo{person}{Samer~S Hanna} {and}
  \bibinfo{person}{Danijela Cabric}.} \bibinfo{year}{2019}\natexlab{}.
\newblock \showarticletitle{Deep learning based transmitter identification
  using power amplifier nonlinearity}. In \bibinfo{booktitle}{\emph{2019
  International Conference on Computing, Networking and Communications
  (ICNC)}}. IEEE, \bibinfo{pages}{674--680}.
\newblock


\bibitem[\protect\citeauthoryear{Harrison, Asadizanjani, and
  Tehranipoor}{Harrison et~al\mbox{.}}{2021}]%
        {HARRISON202112}
\bibfield{author}{\bibinfo{person}{Jacob Harrison}, \bibinfo{person}{Navid
  Asadizanjani}, {and} \bibinfo{person}{Mark Tehranipoor}.}
  \bibinfo{year}{2021}\natexlab{}.
\newblock \showarticletitle{On malicious implants in PCBs throughout the supply
  chain}.
\newblock \bibinfo{journal}{\emph{Integration}}  \bibinfo{volume}{79}
  (\bibinfo{year}{2021}), \bibinfo{pages}{12--22}.
\newblock
\showISSN{0167-9260}
\urldef\tempurl%
\url{https://doi.org/10.1016/j.vlsi.2021.03.002}
\showDOI{\tempurl}


\bibitem[\protect\citeauthoryear{Harsha and Harish}{Harsha and Harish}{2018}]%
        {harsha2018integrated}
\bibfield{author}{\bibinfo{person}{MV Harsha} {and} \bibinfo{person}{BP
  Harish}.} \bibinfo{year}{2018}\natexlab{}.
\newblock \showarticletitle{An integrated maxFit genetic algorithm-SPICE
  framework for 2-Stage Op-Amp design automation}. In
  \bibinfo{booktitle}{\emph{2018 IEEE Computer Society Annual Symposium on VLSI
  (ISVLSI)}}. IEEE, \bibinfo{pages}{170--174}.
\newblock


\bibitem[\protect\citeauthoryear{Hart, Nilsson, and Raphael}{Hart
  et~al\mbox{.}}{1968}]%
        {hart1968formal}
\bibfield{author}{\bibinfo{person}{Peter~E Hart}, \bibinfo{person}{Nils~J
  Nilsson}, {and} \bibinfo{person}{Bertram Raphael}.}
  \bibinfo{year}{1968}\natexlab{}.
\newblock \showarticletitle{A formal basis for the heuristic determination of
  minimum cost paths}.
\newblock \bibinfo{journal}{\emph{IEEE transactions on Systems Science and
  Cybernetics}} \bibinfo{volume}{4}, \bibinfo{number}{2}
  (\bibinfo{year}{1968}), \bibinfo{pages}{100--107}.
\newblock


\bibitem[\protect\citeauthoryear{Hasani, Haerle, Baumgartner, Lomuscio, and
  Grosu}{Hasani et~al\mbox{.}}{2017}]%
        {hasani2017compositional}
\bibfield{author}{\bibinfo{person}{Ramin~M Hasani}, \bibinfo{person}{Dieter
  Haerle}, \bibinfo{person}{Christian~F Baumgartner},
  \bibinfo{person}{Alessio~R Lomuscio}, {and} \bibinfo{person}{Radu Grosu}.}
  \bibinfo{year}{2017}\natexlab{}.
\newblock \showarticletitle{Compositional neural-network modeling of complex
  analog circuits}. In \bibinfo{booktitle}{\emph{2017 International Joint
  Conference on Neural Networks (IJCNN)}}. IEEE, \bibinfo{pages}{2235--2242}.
\newblock


\bibitem[\protect\citeauthoryear{Herder, Yu, Koushanfar, and Devadas}{Herder
  et~al\mbox{.}}{2014}]%
        {herder2014puf}
\bibfield{author}{\bibinfo{person}{Charles Herder}, \bibinfo{person}{Meng-Day
  Yu}, \bibinfo{person}{Farinaz Koushanfar}, {and} \bibinfo{person}{Srinivas
  Devadas}.} \bibinfo{year}{2014}\natexlab{}.
\newblock \showarticletitle{Physical Unclonable Functions and Applications: A
  Tutorial}.
\newblock \bibinfo{journal}{\emph{Proc. IEEE}} \bibinfo{volume}{102},
  \bibinfo{number}{8} (\bibinfo{year}{2014}), \bibinfo{pages}{1126--1141}.
\newblock
\urldef\tempurl%
\url{https://doi.org/10.1109/JPROC.2014.2320516}
\showDOI{\tempurl}


\bibitem[\protect\citeauthoryear{Hiller, K{\"u}rzinger, and Sigl}{Hiller
  et~al\mbox{.}}{2020}]%
        {hiller2020review}
\bibfield{author}{\bibinfo{person}{Matthias Hiller}, \bibinfo{person}{Ludwig
  K{\"u}rzinger}, {and} \bibinfo{person}{Georg Sigl}.}
  \bibinfo{year}{2020}\natexlab{}.
\newblock \showarticletitle{Review of error correction for PUFs and evaluation
  on state-of-the-art FPGAs}.
\newblock \bibinfo{journal}{\emph{Journal of Cryptographic Engineering}}
  \bibinfo{volume}{10}, \bibinfo{number}{3} (\bibinfo{year}{2020}),
  \bibinfo{pages}{229--247}.
\newblock


\bibitem[\protect\citeauthoryear{Hsiao}{Hsiao}{2006}]%
        {HSIAO2006161}
\bibfield{author}{\bibinfo{person}{Michael~S. Hsiao}.}
  \bibinfo{year}{2006}\natexlab{}.
\newblock \showarticletitle{Chapter 4 - Test Generation}.
\newblock In \bibinfo{booktitle}{\emph{VLSI Test Principles and
  Architectures}}, \bibfield{editor}{\bibinfo{person}{Laung-Terng Wang},
  \bibinfo{person}{Cheng-Wen Wu}, {and} \bibinfo{person}{Xiaoqing Wen}} (Eds.).
  \bibinfo{publisher}{Morgan Kaufmann}, \bibinfo{address}{San Francisco},
  \bibinfo{pages}{161--262}.
\newblock
\showISBNx{978-0-12-370597-6}
\urldef\tempurl%
\url{https://doi.org/10.1016/B978-012370597-6/50008-1}
\showDOI{\tempurl}


\bibitem[\protect\citeauthoryear{Hu, Ardeshiricham, and Kastner}{Hu
  et~al\mbox{.}}{2021}]%
        {hu2021}
\bibfield{author}{\bibinfo{person}{Wei Hu}, \bibinfo{person}{Armaiti
  Ardeshiricham}, {and} \bibinfo{person}{Ryan Kastner}.}
  \bibinfo{year}{2021}\natexlab{}.
\newblock \showarticletitle{Hardware Information Flow Tracking}.
\newblock \bibinfo{journal}{\emph{ACM Comput. Surv.}} \bibinfo{volume}{54},
  \bibinfo{number}{4}, Article \bibinfo{articleno}{83} (\bibinfo{date}{may}
  \bibinfo{year}{2021}), \bibinfo{numpages}{39}~pages.
\newblock
\showISSN{0360-0300}
\urldef\tempurl%
\url{https://doi.org/10.1145/3447867}
\showDOI{\tempurl}


\bibitem[\protect\citeauthoryear{Huang, Hu, He, Liu, Ma, Shen, Wu, Xu, Zhang,
  Zhong, et~al\mbox{.}}{Huang et~al\mbox{.}}{2021a}]%
        {huang2021machine}
\bibfield{author}{\bibinfo{person}{Guyue Huang}, \bibinfo{person}{Jingbo Hu},
  \bibinfo{person}{Yifan He}, \bibinfo{person}{Jialong Liu},
  \bibinfo{person}{Mingyuan Ma}, \bibinfo{person}{Zhaoyang Shen},
  \bibinfo{person}{Juejian Wu}, \bibinfo{person}{Yuanfan Xu},
  \bibinfo{person}{Hengrui Zhang}, \bibinfo{person}{Kai Zhong},
  {et~al\mbox{.}}} \bibinfo{year}{2021}\natexlab{a}.
\newblock \showarticletitle{Machine learning for electronic design automation:
  A survey}.
\newblock \bibinfo{journal}{\emph{ACM Transactions on Design Automation of
  Electronic Systems (TODAES)}} \bibinfo{volume}{26}, \bibinfo{number}{5}
  (\bibinfo{year}{2021}), \bibinfo{pages}{1--46}.
\newblock


\bibitem[\protect\citeauthoryear{Huang, Zhang, Tao, Yang, Yan, Zhou, and
  Zeng}{Huang et~al\mbox{.}}{2021b}]%
        {huang2021bayesian}
\bibfield{author}{\bibinfo{person}{Jiangli Huang}, \bibinfo{person}{Shuhan
  Zhang}, \bibinfo{person}{Cong Tao}, \bibinfo{person}{Fan Yang},
  \bibinfo{person}{Changhao Yan}, \bibinfo{person}{Dian Zhou}, {and}
  \bibinfo{person}{Xuan Zeng}.} \bibinfo{year}{2021}\natexlab{b}.
\newblock \showarticletitle{Bayesian Optimization Approach for Analog Circuit
  Design Using Multi-task Gaussian Process}. In \bibinfo{booktitle}{\emph{2021
  IEEE International Symposium on Circuits and Systems (ISCAS)}}. IEEE,
  \bibinfo{pages}{1--5}.
\newblock


\bibitem[\protect\citeauthoryear{Huang, Carulli, and Makris}{Huang
  et~al\mbox{.}}{2012}]%
        {huang2012parametric}
\bibfield{author}{\bibinfo{person}{Ke Huang}, \bibinfo{person}{John~M Carulli},
  {and} \bibinfo{person}{Yiorgos Makris}.} \bibinfo{year}{2012}\natexlab{}.
\newblock \showarticletitle{Parametric counterfeit IC detection via support
  vector machines}. In \bibinfo{booktitle}{\emph{2012 IEEE International
  Symposium on Defect and Fault Tolerance in VLSI and Nanotechnology Systems
  (DFT)}}. IEEE, \bibinfo{pages}{7--12}.
\newblock


\bibitem[\protect\citeauthoryear{Huss, St{\"o}ttinger, and Zohner}{Huss
  et~al\mbox{.}}{2013}]%
        {huss2013amasive}
\bibfield{author}{\bibinfo{person}{Sorin~A Huss}, \bibinfo{person}{Marc
  St{\"o}ttinger}, {and} \bibinfo{person}{Michael Zohner}.}
  \bibinfo{year}{2013}\natexlab{}.
\newblock \showarticletitle{Amasive: an adaptable and modular autonomous
  side-channel vulnerability evaluation framework}.
\newblock In \bibinfo{booktitle}{\emph{Number theory and cryptography}}.
  \bibinfo{publisher}{Springer}, \bibinfo{pages}{151--165}.
\newblock


\bibitem[\protect\citeauthoryear{Imeson, Emtenan, Garg, and Tripunitara}{Imeson
  et~al\mbox{.}}{2013}]%
        {imeson2013securing}
\bibfield{author}{\bibinfo{person}{Frank Imeson}, \bibinfo{person}{Ariq
  Emtenan}, \bibinfo{person}{Siddharth Garg}, {and} \bibinfo{person}{Mahesh
  Tripunitara}.} \bibinfo{year}{2013}\natexlab{}.
\newblock \showarticletitle{{Securing Computer Hardware Using 3D Integrated
  Circuit (IC) Technology and Split Manufacturing for Obfuscation}}. In
  \bibinfo{booktitle}{\emph{22nd USENIX Security Symposium (USENIX Security
  13)}}. \bibinfo{pages}{495--510}.
\newblock


\bibitem[\protect\citeauthoryear{{\.I}slamo{\u{g}}lu, {\c{C}}akici, Afacan, and
  D{\"u}ndar}{{\.I}slamo{\u{g}}lu et~al\mbox{.}}{2019}]%
        {islamouglu2019artificial}
\bibfield{author}{\bibinfo{person}{Gamze {\.I}slamo{\u{g}}lu},
  \bibinfo{person}{Tu{\u{g}}berk~O{\u{g}}ulcan {\c{C}}akici},
  \bibinfo{person}{Engin Afacan}, {and} \bibinfo{person}{G{\"u}nhan
  D{\"u}ndar}.} \bibinfo{year}{2019}\natexlab{}.
\newblock \showarticletitle{Artificial neural network assisted analog IC sizing
  tool}. In \bibinfo{booktitle}{\emph{2019 16th International Conference on
  Synthesis, Modeling, Analysis and Simulation Methods and Applications to
  Circuit Design (SMACD)}}. IEEE, \bibinfo{pages}{9--12}.
\newblock


\bibitem[\protect\citeauthoryear{Jafari, Sadri, and Zekri}{Jafari
  et~al\mbox{.}}{2010}]%
        {jafari2010design}
\bibfield{author}{\bibinfo{person}{Ali Jafari}, \bibinfo{person}{Saeed Sadri},
  {and} \bibinfo{person}{Maryam Zekri}.} \bibinfo{year}{2010}\natexlab{}.
\newblock \showarticletitle{Design optimization of analog integrated circuits
  by using artificial neural networks}. In \bibinfo{booktitle}{\emph{2010
  international conference of soft computing and pattern recognition}}. IEEE,
  \bibinfo{pages}{385--388}.
\newblock


\bibitem[\protect\citeauthoryear{Jayasankaran, Borbon, Abuellil,
  S{\'a}nchez-Sinencio, Hu, and Rajendran}{Jayasankaran et~al\mbox{.}}{2019a}]%
        {jayasankaran2019breaking}
\bibfield{author}{\bibinfo{person}{Nithyashankari~Gummidipoondi Jayasankaran},
  \bibinfo{person}{A~Sanabria Borbon}, \bibinfo{person}{Amr Abuellil},
  \bibinfo{person}{Edgar S{\'a}nchez-Sinencio}, \bibinfo{person}{Jiang Hu},
  {and} \bibinfo{person}{Jeyavijayan Rajendran}.}
  \bibinfo{year}{2019}\natexlab{a}.
\newblock \showarticletitle{Breaking analog locking techniques via
  satisfiability modulo theories}. In \bibinfo{booktitle}{\emph{2019 IEEE
  International Test Conference (ITC)}}. IEEE, \bibinfo{pages}{1--10}.
\newblock


\bibitem[\protect\citeauthoryear{Jayasankaran, Borbon, Abuellil,
  Sánchez-Sinencio, Hu, and Rajendran}{Jayasankaran et~al\mbox{.}}{2019b}]%
        {9000113}
\bibfield{author}{\bibinfo{person}{N.~G. Jayasankaran},
  \bibinfo{person}{A.~Sanabria Borbon}, \bibinfo{person}{A. Abuellil},
  \bibinfo{person}{E. Sánchez-Sinencio}, \bibinfo{person}{J. Hu}, {and}
  \bibinfo{person}{J. Rajendran}.} \bibinfo{year}{2019}\natexlab{b}.
\newblock \showarticletitle{Breaking Analog Locking Techniques via
  Satisfiability Modulo Theories}. In \bibinfo{booktitle}{\emph{2019 IEEE
  International Test Conference (ITC)}}. \bibinfo{pages}{1--10}.
\newblock
\urldef\tempurl%
\url{https://doi.org/10.1109/ITC44170.2019.9000113}
\showDOI{\tempurl}


\bibitem[\protect\citeauthoryear{Jenihhin}{Jenihhin}{[n.d.]}]%
        {jenihhin}
\bibfield{author}{\bibinfo{person}{Maksim Jenihhin}.}
  \bibinfo{year}{[n.d.]}\natexlab{}.
\newblock
\newblock
\urldef\tempurl%
\url{https://pld.ttu.ee/~maksim/benchmarks/}
\showURL{%
\tempurl}


\bibitem[\protect\citeauthoryear{Jiang and Devadas}{Jiang and Devadas}{2009}]%
        {JIANG2009299}
\bibfield{author}{\bibinfo{person}{Jie-Hong~(Roland) Jiang} {and}
  \bibinfo{person}{Srinivas Devadas}.} \bibinfo{year}{2009}\natexlab{}.
\newblock \showarticletitle{CHAPTER 6 - Logic synthesis in a nutshell}.
\newblock In \bibinfo{booktitle}{\emph{Electronic Design Automation}},
  \bibfield{editor}{\bibinfo{person}{Laung-Terng Wang},
  \bibinfo{person}{Yao-Wen Chang}, {and} \bibinfo{person}{Kwang-Ting~(Tim)
  Cheng}} (Eds.). \bibinfo{publisher}{Morgan Kaufmann},
  \bibinfo{address}{Boston}, \bibinfo{pages}{299--404}.
\newblock
\showISBNx{978-0-12-374364-0}
\urldef\tempurl%
\url{https://doi.org/10.1016/B978-0-12-374364-0.50013-8}
\showDOI{\tempurl}


\bibitem[\protect\citeauthoryear{Jiang, Dai, Suh, and Zhang}{Jiang
  et~al\mbox{.}}{2018}]%
        {jiang2018}
\bibfield{author}{\bibinfo{person}{Zhenghong Jiang}, \bibinfo{person}{Steve
  Dai}, \bibinfo{person}{G.~Edward Suh}, {and} \bibinfo{person}{Zhiru Zhang}.}
  \bibinfo{year}{2018}\natexlab{}.
\newblock \showarticletitle{High-Level Synthesis with Timing-Sensitive
  Information Flow Enforcement}. In \bibinfo{booktitle}{\emph{2018 IEEE/ACM
  International Conference on Computer-Aided Design (ICCAD)}}.
  \bibinfo{pages}{1--8}.
\newblock
\urldef\tempurl%
\url{https://doi.org/10.1145/3240765.3240813}
\showDOI{\tempurl}


\bibitem[\protect\citeauthoryear{Jin, Kim, Kim, and Hong}{Jin
  et~al\mbox{.}}{2020}]%
        {jin2020recent}
\bibfield{author}{\bibinfo{person}{Sunghyun Jin}, \bibinfo{person}{Suhri Kim},
  \bibinfo{person}{HeeSeok Kim}, {and} \bibinfo{person}{Seokhie Hong}.}
  \bibinfo{year}{2020}\natexlab{}.
\newblock \showarticletitle{Recent advances in deep learning-based side-channel
  analysis}.
\newblock \bibinfo{journal}{\emph{ETRI Journal}} \bibinfo{volume}{42},
  \bibinfo{number}{2} (\bibinfo{year}{2020}), \bibinfo{pages}{292--304}.
\newblock


\bibitem[\protect\citeauthoryear{Kahng, Varadarajan, and Wang}{Kahng
  et~al\mbox{.}}{[n.d.]}]%
        {kahng_varadarajan_wang_2022}
\bibfield{author}{\bibinfo{person}{Andrew~B Kahng}, \bibinfo{person}{Ravi
  Varadarajan}, {and} \bibinfo{person}{Zhiang Wang}.}
  \bibinfo{year}{[n.d.]}\natexlab{}.
\newblock \showarticletitle{RTL-MP: Toward Practical, Human-Quality Chip
  Planning and Macro Placement}.
\newblock


\bibitem[\protect\citeauthoryear{Karpovsky and Taubin}{Karpovsky and
  Taubin}{2004}]%
        {karpovsky2004new}
\bibfield{author}{\bibinfo{person}{Mark Karpovsky} {and}
  \bibinfo{person}{Alexander Taubin}.} \bibinfo{year}{2004}\natexlab{}.
\newblock \showarticletitle{New class of nonlinear systematic error detecting
  codes}.
\newblock \bibinfo{journal}{\emph{IEEE Transactions on Information Theory}}
  \bibinfo{volume}{50}, \bibinfo{number}{8} (\bibinfo{year}{2004}),
  \bibinfo{pages}{1818--1819}.
\newblock


\bibitem[\protect\citeauthoryear{Kaya, Afacan, and Dundar}{Kaya
  et~al\mbox{.}}{2018}]%
        {kaya2018analog}
\bibfield{author}{\bibinfo{person}{Ezgi Kaya}, \bibinfo{person}{Engin Afacan},
  {and} \bibinfo{person}{Gunhan Dundar}.} \bibinfo{year}{2018}\natexlab{}.
\newblock \showarticletitle{An analog/RF circuit synthesis and design assistant
  tool for analog IP: DATA-IP}. In \bibinfo{booktitle}{\emph{2018 15th
  International Conference on Synthesis, Modeling, Analysis and Simulation
  Methods and Applications to Circuit Design (SMACD)}}. IEEE,
  \bibinfo{pages}{1--9}.
\newblock


\bibitem[\protect\citeauthoryear{Keating and Bricaud}{Keating and
  Bricaud}{2007}]%
        {keating_bricaud_2007}
\bibfield{author}{\bibinfo{person}{Michael Keating} {and}
  \bibinfo{person}{Pierre Bricaud}.} \bibinfo{year}{2007}\natexlab{}.
\newblock \bibinfo{booktitle}{\emph{Reuse methodology manual: For
  system-on-a-chip designs}}.
\newblock \bibinfo{publisher}{Springer}.
\newblock


\bibitem[\protect\citeauthoryear{Khwaja}{Khwaja}{1997}]%
        {khwaja1997edarule}
\bibfield{author}{\bibinfo{person}{Amir~A. Khwaja}.}
  \bibinfo{year}{1997}\natexlab{}.
\newblock \showarticletitle{Enhancing Extensibility of the Design Rule Checker
  of an EDA Tool by Object-Oriented Modeling}. In
  \bibinfo{booktitle}{\emph{Proceedings of the 21st International Computer
  Software and Applications Conference}} \emph{(\bibinfo{series}{COMPSAC
  '97})}. \bibinfo{publisher}{IEEE Computer Society}, \bibinfo{address}{USA},
  \bibinfo{pages}{104–108}.
\newblock
\showISBNx{0818681055}


\bibitem[\protect\citeauthoryear{Kim, Picek, Heuser, Bhasin, and Hanjalic}{Kim
  et~al\mbox{.}}{2019}]%
        {kim2019make}
\bibfield{author}{\bibinfo{person}{Jaehun Kim}, \bibinfo{person}{Stjepan
  Picek}, \bibinfo{person}{Annelie Heuser}, \bibinfo{person}{Shivam Bhasin},
  {and} \bibinfo{person}{Alan Hanjalic}.} \bibinfo{year}{2019}\natexlab{}.
\newblock \showarticletitle{Make Some Noise. Unleashing the Power of
  Convolutional Neural Networks for Profiled Side-channel Analysis}.
\newblock \bibinfo{journal}{\emph{IACR Trans. on Cryptographic Hardware and
  Embedded Systems}} (\bibinfo{year}{2019}), \bibinfo{pages}{148--179}.
\newblock


\bibitem[\protect\citeauthoryear{Kim, Park, Kim, Son, Kim, Son, Choi, Park, and
  Kim}{Kim et~al\mbox{.}}{2020}]%
        {kim2020reinforcement}
\bibfield{author}{\bibinfo{person}{Minsu Kim}, \bibinfo{person}{Hyunwook Park},
  \bibinfo{person}{Seongguk Kim}, \bibinfo{person}{Keeyoung Son},
  \bibinfo{person}{Subin Kim}, \bibinfo{person}{Kyunjune Son},
  \bibinfo{person}{Seonguk Choi}, \bibinfo{person}{Gapyeol Park}, {and}
  \bibinfo{person}{Joungho Kim}.} \bibinfo{year}{2020}\natexlab{}.
\newblock \showarticletitle{Reinforcement learning-based auto-router
  considering signal integrity}. In \bibinfo{booktitle}{\emph{2020 IEEE 29th
  Conference on Electrical Performance of Electronic Packaging and Systems
  (EPEPS)}}. IEEE, \bibinfo{pages}{1--3}.
\newblock


\bibitem[\protect\citeauthoryear{Kison, Awad, Fyrbiak, and Paar}{Kison
  et~al\mbox{.}}{2019}]%
        {kison2019security}
\bibfield{author}{\bibinfo{person}{Christian Kison},
  \bibinfo{person}{Omar~Mohamed Awad}, \bibinfo{person}{Marc Fyrbiak}, {and}
  \bibinfo{person}{Christof Paar}.} \bibinfo{year}{2019}\natexlab{}.
\newblock \showarticletitle{Security implications of intentional capacitive
  crosstalk}.
\newblock \bibinfo{journal}{\emph{IEEE Transactions on Information Forensics
  and Security}} \bibinfo{volume}{14}, \bibinfo{number}{12}
  (\bibinfo{year}{2019}), \bibinfo{pages}{3246--3258}.
\newblock


\bibitem[\protect\citeauthoryear{Kocher, Horn, Fogh, Genkin, Gruss, Haas,
  Hamburg, Lipp, Mangard, Prescher, et~al\mbox{.}}{Kocher
  et~al\mbox{.}}{2019}]%
        {kocher2019spectre}
\bibfield{author}{\bibinfo{person}{Paul Kocher}, \bibinfo{person}{Jann Horn},
  \bibinfo{person}{Anders Fogh}, \bibinfo{person}{Daniel Genkin},
  \bibinfo{person}{Daniel Gruss}, \bibinfo{person}{Werner Haas},
  \bibinfo{person}{Mike Hamburg}, \bibinfo{person}{Moritz Lipp},
  \bibinfo{person}{Stefan Mangard}, \bibinfo{person}{Thomas Prescher},
  {et~al\mbox{.}}} \bibinfo{year}{2019}\natexlab{}.
\newblock \showarticletitle{Spectre attacks: Exploiting speculative execution}.
  In \bibinfo{booktitle}{\emph{2019 IEEE Symposium on Security and Privacy
  (SP)}}. IEEE, \bibinfo{pages}{1--19}.
\newblock


\bibitem[\protect\citeauthoryear{Konigsmark, Chen, and Wong}{Konigsmark
  et~al\mbox{.}}{2017}]%
        {konigsmark2017}
\bibfield{author}{\bibinfo{person}{S.T.~Choden Konigsmark},
  \bibinfo{person}{Deming Chen}, {and} \bibinfo{person}{Martin~D.F. Wong}.}
  \bibinfo{year}{2017}\natexlab{}.
\newblock \showarticletitle{High-Level Synthesis for side-channel defense}. In
  \bibinfo{booktitle}{\emph{2017 IEEE 28th International Conference on
  Application-specific Systems, Architectures and Processors (ASAP)}}.
  \bibinfo{pages}{37--44}.
\newblock
\urldef\tempurl%
\url{https://doi.org/10.1109/ASAP.2017.7995257}
\showDOI{\tempurl}


\bibitem[\protect\citeauthoryear{Krachenfels, Kiyan, Tajik, and
  Seifert}{Krachenfels et~al\mbox{.}}{2021}]%
        {krachenfels2021automatic}
\bibfield{author}{\bibinfo{person}{Thilo Krachenfels}, \bibinfo{person}{Tuba
  Kiyan}, \bibinfo{person}{Shahin Tajik}, {and} \bibinfo{person}{Jean-Pierre
  Seifert}.} \bibinfo{year}{2021}\natexlab{}.
\newblock \showarticletitle{Automatic Extraction of Secrets from the Transistor
  Jungle using Laser-Assisted Side-Channel Attacks}. In
  \bibinfo{booktitle}{\emph{30th USENIX Security Symposium (USENIX Security
  21)}}. \bibinfo{pages}{627--644}.
\newblock


\bibitem[\protect\citeauthoryear{Kumar and Burleson}{Kumar and
  Burleson}{2014}]%
        {kumar2014design}
\bibfield{author}{\bibinfo{person}{Raghavan Kumar} {and} \bibinfo{person}{Wayne
  Burleson}.} \bibinfo{year}{2014}\natexlab{}.
\newblock \showarticletitle{On design of a highly secure PUF based on
  non-linear current mirrors}. In \bibinfo{booktitle}{\emph{2014 IEEE
  international symposium on hardware-oriented security and trust (HOST)}}.
  IEEE, \bibinfo{pages}{38--43}.
\newblock


\bibitem[\protect\citeauthoryear{Kunal, Madhusudan, Sharma, Xu, Burns, Harjani,
  Hu, Kirkpatrick, and Sapatnekar}{Kunal et~al\mbox{.}}{2019}]%
        {kunal2019align}
\bibfield{author}{\bibinfo{person}{Kishor Kunal}, \bibinfo{person}{Meghna
  Madhusudan}, \bibinfo{person}{Arvind~K Sharma}, \bibinfo{person}{Wenbin Xu},
  \bibinfo{person}{Steven~M Burns}, \bibinfo{person}{Ramesh Harjani},
  \bibinfo{person}{Jiang Hu}, \bibinfo{person}{Desmond~A Kirkpatrick}, {and}
  \bibinfo{person}{Sachin~S Sapatnekar}.} \bibinfo{year}{2019}\natexlab{}.
\newblock \showarticletitle{ALIGN: Open-source analog layout automation from
  the ground up}. In \bibinfo{booktitle}{\emph{Proceedings of the 56th Annual
  Design Automation Conference 2019}}. \bibinfo{pages}{1--4}.
\newblock


\bibitem[\protect\citeauthoryear{Kunal, Poojary, Dhar, Madhusudan, Harjani, and
  Sapatnekar}{Kunal et~al\mbox{.}}{2020}]%
        {kunal2020general}
\bibfield{author}{\bibinfo{person}{Kishor Kunal}, \bibinfo{person}{Jitesh
  Poojary}, \bibinfo{person}{Tonmoy Dhar}, \bibinfo{person}{Meghna Madhusudan},
  \bibinfo{person}{Ramesh Harjani}, {and} \bibinfo{person}{Sachin~S
  Sapatnekar}.} \bibinfo{year}{2020}\natexlab{}.
\newblock \showarticletitle{A general approach for identifying hierarchical
  symmetry constraints for analog circuit layout}. In
  \bibinfo{booktitle}{\emph{2020 IEEE/ACM International Conference On Computer
  Aided Design (ICCAD)}}. IEEE, \bibinfo{pages}{1--8}.
\newblock


\bibitem[\protect\citeauthoryear{Lecun, Bengio, and Hinton}{Lecun
  et~al\mbox{.}}{2015}]%
        {lecun2021}
\bibfield{author}{\bibinfo{person}{Yann Lecun}, \bibinfo{person}{Yoshua
  Bengio}, {and} \bibinfo{person}{Geoffrey Hinton}.}
  \bibinfo{year}{2015}\natexlab{}.
\newblock \showarticletitle{Deep learning}.
\newblock \bibinfo{journal}{\emph{Nature Cell Biology}} \bibinfo{volume}{521},
  \bibinfo{number}{7553} (\bibinfo{date}{27 May} \bibinfo{year}{2015}),
  \bibinfo{pages}{436--444}.
\newblock
\showISSN{1465-7392}
\urldef\tempurl%
\url{https://doi.org/10.1038/nature14539}
\showDOI{\tempurl}
\newblock
\shownote{Funding Information: Acknowledgements The authors would like to thank
  the Natural Sciences and Engineering Research Council of Canada, the Canadian
  Institute For Advanced Research (CIFAR), the National Science Foundation and
  Office of Naval Research for support. Y.L. and Y.B. are CIFAR fellows.
  Publisher Copyright: {\textcopyright} 2015 Macmillan Publishers Limited. All
  rights reserved.}


\bibitem[\protect\citeauthoryear{Lee and Touba}{Lee and Touba}{2015}]%
        {lee2015improving}
\bibfield{author}{\bibinfo{person}{Yu-Wei Lee} {and} \bibinfo{person}{Nur~A
  Touba}.} \bibinfo{year}{2015}\natexlab{}.
\newblock \showarticletitle{Improving logic obfuscation via logic cone
  analysis}. In \bibinfo{booktitle}{\emph{2015 16th Latin-American Test
  Symposium (LATS)}}. IEEE, \bibinfo{pages}{1--6}.
\newblock


\bibitem[\protect\citeauthoryear{Leef}{Leef}{[n.d.]}]%
        {darpa}
\bibfield{author}{\bibinfo{person}{Serge Leef}.}
  \bibinfo{year}{[n.d.]}\natexlab{}.
\newblock
\newblock
\urldef\tempurl%
\url{https://www.darpa.mil/program/automatic-implementation-of-secure-silicon}
\showURL{%
\tempurl}


\bibitem[\protect\citeauthoryear{Leiserson, Marson, and Wachs}{Leiserson
  et~al\mbox{.}}{2014}]%
        {Leierson2014}
\bibfield{author}{\bibinfo{person}{Andrew~J. Leiserson},
  \bibinfo{person}{Mark~E. Marson}, {and} \bibinfo{person}{Megan~A. Wachs}.}
  \bibinfo{year}{2014}\natexlab{}.
\newblock \showarticletitle{Gate-Level Masking under a Path-Based Leakage
  Metric}. In \bibinfo{booktitle}{\emph{Cryptographic Hardware and Embedded
  Systems -- CHES 2014}}, \bibfield{editor}{\bibinfo{person}{Lejla Batina}
  {and} \bibinfo{person}{Matthew Robshaw}} (Eds.). \bibinfo{publisher}{Springer
  Berlin Heidelberg}, \bibinfo{address}{Berlin, Heidelberg},
  \bibinfo{pages}{580--597}.
\newblock
\showISBNx{978-3-662-44709-3}


\bibitem[\protect\citeauthoryear{Leprince-Ringuet}{Leprince-Ringuet}{2021}]%
        {leprince-ringuet_2021}
\bibfield{author}{\bibinfo{person}{Daphne Leprince-Ringuet}.}
  \bibinfo{year}{2021}\natexlab{}.
\newblock \bibinfo{title}{The global chip shortage is creating a new problem:
  More fake components}.
\newblock
\newblock
\urldef\tempurl%
\url{https://www.zdnet.com/article/the-global-chip-shortage-is-creating-a-new-problem-more-fake-components-as-fraudsters-cash-in/}
\showURL{%
\tempurl}


\bibitem[\protect\citeauthoryear{Levi, Bellizia, Bol, and Standaert}{Levi
  et~al\mbox{.}}{2020}]%
        {levi2020ask}
\bibfield{author}{\bibinfo{person}{Itamar Levi}, \bibinfo{person}{Davide
  Bellizia}, \bibinfo{person}{David Bol}, {and}
  \bibinfo{person}{Fran{\c{c}}ois-Xavier Standaert}.}
  \bibinfo{year}{2020}\natexlab{}.
\newblock \showarticletitle{Ask less, get more: Side-channel signal hiding,
  revisited}.
\newblock \bibinfo{journal}{\emph{IEEE Transactions on Circuits and Systems I:
  Regular Papers}} \bibinfo{volume}{67}, \bibinfo{number}{12}
  (\bibinfo{year}{2020}), \bibinfo{pages}{4904--4917}.
\newblock


\bibitem[\protect\citeauthoryear{Li, Jiao, and Doboli}{Li
  et~al\mbox{.}}{2016}]%
        {li2016analog}
\bibfield{author}{\bibinfo{person}{Hao Li}, \bibinfo{person}{Fanshu Jiao},
  {and} \bibinfo{person}{Alex Doboli}.} \bibinfo{year}{2016}\natexlab{}.
\newblock \showarticletitle{Analog circuit topological feature extraction with
  unsupervised learning of new sub-structures}. In
  \bibinfo{booktitle}{\emph{2016 Design, Automation \& Test in Europe
  Conference \& Exhibition (DATE)}}. IEEE, \bibinfo{pages}{1509--1512}.
\newblock


\bibitem[\protect\citeauthoryear{Li, Sahu, Talwalkar, and Smith}{Li
  et~al\mbox{.}}{2020c}]%
        {Li2020-federated-challenges}
\bibfield{author}{\bibinfo{person}{Tian Li}, \bibinfo{person}{Anit~Kumar Sahu},
  \bibinfo{person}{Ameet Talwalkar}, {and} \bibinfo{person}{Virginia Smith}.}
  \bibinfo{year}{2020}\natexlab{c}.
\newblock \showarticletitle{{Federated Learning: Challenges, Methods, and
  Future Directions}}.
\newblock \bibinfo{journal}{\emph{IEEE Signal Process. Mag.}}
  \bibinfo{volume}{37}, \bibinfo{number}{3} (\bibinfo{date}{May}
  \bibinfo{year}{2020}), \bibinfo{pages}{50--60}.
\newblock
\showISSN{1558-0792}
\urldef\tempurl%
\url{https://doi.org/10.1109/MSP.2020.2975749}
\showDOI{\tempurl}


\bibitem[\protect\citeauthoryear{Li, Lin, Madhusudan, Sharma, Xu, Sapatnekar,
  Harjani, and Hu}{Li et~al\mbox{.}}{2020a}]%
        {li2020exploring}
\bibfield{author}{\bibinfo{person}{Yaguang Li}, \bibinfo{person}{Yishuang Lin},
  \bibinfo{person}{Meghna Madhusudan}, \bibinfo{person}{Arvind Sharma},
  \bibinfo{person}{Wenbin Xu}, \bibinfo{person}{Sachin Sapatnekar},
  \bibinfo{person}{Ramesh Harjani}, {and} \bibinfo{person}{Jiang Hu}.}
  \bibinfo{year}{2020}\natexlab{a}.
\newblock \showarticletitle{Exploring a machine learning approach to
  performance driven analog IC placement}. In \bibinfo{booktitle}{\emph{2020
  IEEE computer society annual symposium on VLSI (ISVLSI)}}. IEEE,
  \bibinfo{pages}{24--29}.
\newblock


\bibitem[\protect\citeauthoryear{Li, Lin, Madhusudan, Sharma, Xu, Sapatnekar,
  Harjani, and Hu}{Li et~al\mbox{.}}{2020b}]%
        {li2020customized}
\bibfield{author}{\bibinfo{person}{Yaguang Li}, \bibinfo{person}{Yishuang Lin},
  \bibinfo{person}{Meghna Madhusudan}, \bibinfo{person}{Arvind Sharma},
  \bibinfo{person}{Wenbin Xu}, \bibinfo{person}{Sachin~S Sapatnekar},
  \bibinfo{person}{Ramesh Harjani}, {and} \bibinfo{person}{Jiang Hu}.}
  \bibinfo{year}{2020}\natexlab{b}.
\newblock \showarticletitle{A customized graph neural network model for guiding
  analog IC placement}. In \bibinfo{booktitle}{\emph{2020 IEEE/ACM
  International Conference On Computer Aided Design (ICCAD)}}. IEEE,
  \bibinfo{pages}{1--9}.
\newblock


\bibitem[\protect\citeauthoryear{Li, Wang, Li, Zhou, and Lin}{Li
  et~al\mbox{.}}{2019}]%
        {li2019artificial}
\bibfield{author}{\bibinfo{person}{Yaping Li}, \bibinfo{person}{Yong Wang},
  \bibinfo{person}{Yusong Li}, \bibinfo{person}{Ranran Zhou}, {and}
  \bibinfo{person}{Zhaojun Lin}.} \bibinfo{year}{2019}\natexlab{}.
\newblock \showarticletitle{An artificial neural network assisted optimization
  system for analog design space exploration}.
\newblock \bibinfo{journal}{\emph{IEEE Transactions on Computer-Aided Design of
  Integrated Circuits and Systems}} \bibinfo{volume}{39}, \bibinfo{number}{10}
  (\bibinfo{year}{2019}), \bibinfo{pages}{2640--2653}.
\newblock


\bibitem[\protect\citeauthoryear{Li, Zou, Xu, Ou, Jin, Wang, Deng, and
  Zhong}{Li et~al\mbox{.}}{2018}]%
        {li2018vuldeepecker}
\bibfield{author}{\bibinfo{person}{Zhen Li}, \bibinfo{person}{Deqing Zou},
  \bibinfo{person}{Shouhuai Xu}, \bibinfo{person}{Xinyu Ou},
  \bibinfo{person}{Hai Jin}, \bibinfo{person}{Sujuan Wang},
  \bibinfo{person}{Zhijun Deng}, {and} \bibinfo{person}{Yuyi Zhong}.}
  \bibinfo{year}{2018}\natexlab{}.
\newblock \showarticletitle{Vuldeepecker: A deep learning-based system for
  vulnerability detection}.
\newblock \bibinfo{journal}{\emph{arXiv preprint arXiv:1801.01681}}
  (\bibinfo{year}{2018}).
\newblock


\bibitem[\protect\citeauthoryear{Liang, Xie, Jung, Chauha, Chen, Hu, Xiang, and
  Nam}{Liang et~al\mbox{.}}{2020}]%
        {liang2020crosstalk}
\bibfield{author}{\bibinfo{person}{Rongjian Liang}, \bibinfo{person}{Zhiyao
  Xie}, \bibinfo{person}{Jinwook Jung}, \bibinfo{person}{Vishnavi Chauha},
  \bibinfo{person}{Yiran Chen}, \bibinfo{person}{Jiang Hu},
  \bibinfo{person}{Hua Xiang}, {and} \bibinfo{person}{Gi-Joon Nam}.}
  \bibinfo{year}{2020}\natexlab{}.
\newblock \showarticletitle{Routing-Free Crosstalk Prediction}. In
  \bibinfo{booktitle}{\emph{2020 IEEE/ACM International Conference On Computer
  Aided Design (ICCAD)}}. \bibinfo{pages}{1--9}.
\newblock


\bibitem[\protect\citeauthoryear{Liao, Zhang, Dong, Poczos, Shimada, and
  Burak~Kara}{Liao et~al\mbox{.}}{2020}]%
        {liao2020deep}
\bibfield{author}{\bibinfo{person}{Haiguang Liao}, \bibinfo{person}{Wentai
  Zhang}, \bibinfo{person}{Xuliang Dong}, \bibinfo{person}{Barnabas Poczos},
  \bibinfo{person}{Kenji Shimada}, {and} \bibinfo{person}{Levent Burak~Kara}.}
  \bibinfo{year}{2020}\natexlab{}.
\newblock \showarticletitle{A deep reinforcement learning approach for global
  routing}.
\newblock \bibinfo{journal}{\emph{Journal of Mechanical Design}}
  \bibinfo{volume}{142}, \bibinfo{number}{6} (\bibinfo{year}{2020}).
\newblock


\bibitem[\protect\citeauthoryear{Lienig and Scheible}{Lienig and
  Scheible}{2020}]%
        {Lienig2020}
\bibfield{author}{\bibinfo{person}{Jens Lienig} {and} \bibinfo{person}{Juergen
  Scheible}.} \bibinfo{year}{2020}\natexlab{}.
\newblock \bibinfo{booktitle}{\emph{Steps in Physical Design: From Netlist
  Generation to Layout Post Processing}}.
\newblock \bibinfo{publisher}{Springer International Publishing},
  \bibinfo{address}{Cham}, \bibinfo{pages}{165--211}.
\newblock
\showISBNx{978-3-030-39284-0}
\urldef\tempurl%
\url{https://doi.org/10.1007/978-3-030-39284-0_5}
\showDOI{\tempurl}


\bibitem[\protect\citeauthoryear{Lin}{Lin}{2012}]%
        {lin2012binary}
\bibfield{author}{\bibinfo{person}{Kuang~Tsan Lin}.}
  \bibinfo{year}{2012}\natexlab{}.
\newblock \showarticletitle{Based on Binary Encoding Methods and Visual
  Cryptography Schemes to Hide Data}. In \bibinfo{booktitle}{\emph{2012 Eighth
  International Conference on Intelligent Information Hiding and Multimedia
  Signal Processing}}. \bibinfo{pages}{59--62}.
\newblock
\urldef\tempurl%
\url{https://doi.org/10.1109/IIH-MSP.2012.20}
\showDOI{\tempurl}


\bibitem[\protect\citeauthoryear{Lipp, Schwarz, Gruss, Prescher, Haas, Fogh,
  Horn, Mangard, Kocher, Genkin, et~al\mbox{.}}{Lipp et~al\mbox{.}}{2018}]%
        {lipp2018meltdown}
\bibfield{author}{\bibinfo{person}{Moritz Lipp}, \bibinfo{person}{Michael
  Schwarz}, \bibinfo{person}{Daniel Gruss}, \bibinfo{person}{Thomas Prescher},
  \bibinfo{person}{Werner Haas}, \bibinfo{person}{Anders Fogh},
  \bibinfo{person}{Jann Horn}, \bibinfo{person}{Stefan Mangard},
  \bibinfo{person}{Paul Kocher}, \bibinfo{person}{Daniel Genkin},
  {et~al\mbox{.}}} \bibinfo{year}{2018}\natexlab{}.
\newblock \showarticletitle{Meltdown: Reading kernel memory from user space}.
  In \bibinfo{booktitle}{\emph{27th USENIX Security Symposium (USENIX Security
  18)}}. \bibinfo{pages}{973--990}.
\newblock


\bibitem[\protect\citeauthoryear{Liu, Wang, Yu, Liu, Li, Wang, Lu, and
  Fern{\'a}ndez}{Liu et~al\mbox{.}}{2009}]%
        {liu2009analog}
\bibfield{author}{\bibinfo{person}{Bo Liu}, \bibinfo{person}{Yan Wang},
  \bibinfo{person}{Zhiping Yu}, \bibinfo{person}{Leibo Liu},
  \bibinfo{person}{Miao Li}, \bibinfo{person}{Zheng Wang},
  \bibinfo{person}{Jing Lu}, {and} \bibinfo{person}{Francisco~V
  Fern{\'a}ndez}.} \bibinfo{year}{2009}\natexlab{}.
\newblock \showarticletitle{Analog circuit optimization system based on hybrid
  evolutionary algorithms}.
\newblock \bibinfo{journal}{\emph{Integration}} \bibinfo{volume}{42},
  \bibinfo{number}{2} (\bibinfo{year}{2009}), \bibinfo{pages}{137--148}.
\newblock


\bibitem[\protect\citeauthoryear{Liu and Kim}{Liu and Kim}{2019}]%
        {Liu2019}
\bibfield{author}{\bibinfo{person}{Muqing Liu} {and} \bibinfo{person}{Chris~H.
  Kim}.} \bibinfo{year}{2019}\natexlab{}.
\newblock \showarticletitle{Demonstration of a Passive IC Tamper Sensor Based
  on an Exposed Floating Gate Device in a Standard Logic Process}.
\newblock \bibinfo{journal}{\emph{IEEE Transactions on Electron Devices}}
  \bibinfo{volume}{66}, \bibinfo{number}{6} (\bibinfo{year}{2019}),
  \bibinfo{pages}{2735--2740}.
\newblock
\urldef\tempurl%
\url{https://doi.org/10.1109/TED.2019.2909558}
\showDOI{\tempurl}


\bibitem[\protect\citeauthoryear{Liu, Zhu, Gu, Shen, Tang, Sun, and Pan}{Liu
  et~al\mbox{.}}{2020a}]%
        {liu2020towards}
\bibfield{author}{\bibinfo{person}{Mingjie Liu}, \bibinfo{person}{Keren Zhu},
  \bibinfo{person}{Jiaqi Gu}, \bibinfo{person}{Linxiao Shen},
  \bibinfo{person}{Xiyuan Tang}, \bibinfo{person}{Nan Sun}, {and}
  \bibinfo{person}{David~Z Pan}.} \bibinfo{year}{2020}\natexlab{a}.
\newblock \showarticletitle{Towards decrypting the art of analog layout:
  Placement quality prediction via transfer learning}. In
  \bibinfo{booktitle}{\emph{2020 Design, Automation \& Test in Europe
  Conference \& Exhibition (DATE)}}. IEEE, \bibinfo{pages}{496--501}.
\newblock


\bibitem[\protect\citeauthoryear{Liu, Volanis, Huang, and Makris}{Liu
  et~al\mbox{.}}{2015}]%
        {liu2015concurrent}
\bibfield{author}{\bibinfo{person}{Yu Liu}, \bibinfo{person}{Georgios Volanis},
  \bibinfo{person}{Ke Huang}, {and} \bibinfo{person}{Yiorgos Makris}.}
  \bibinfo{year}{2015}\natexlab{}.
\newblock \showarticletitle{Concurrent hardware Trojan detection in wireless
  cryptographic ICs}. In \bibinfo{booktitle}{\emph{2015 IEEE International Test
  Conference (ITC)}}. IEEE, \bibinfo{pages}{1--8}.
\newblock


\bibitem[\protect\citeauthoryear{Liu, Zuzak, Xie, Chakraborty, and
  Srivastava}{Liu et~al\mbox{.}}{2020b}]%
        {liu2020strong}
\bibfield{author}{\bibinfo{person}{Yuntao Liu}, \bibinfo{person}{Michael
  Zuzak}, \bibinfo{person}{Yang Xie}, \bibinfo{person}{Abhishek Chakraborty},
  {and} \bibinfo{person}{Ankur Srivastava}.} \bibinfo{year}{2020}\natexlab{b}.
\newblock \showarticletitle{Strong anti-sat: Secure and effective logic
  locking}. In \bibinfo{booktitle}{\emph{2020 21st International Symposium on
  Quality Electronic Design (ISQED)}}. IEEE, \bibinfo{pages}{199--205}.
\newblock


\bibitem[\protect\citeauthoryear{Liu, Jia, Vong, Bu, Han, and Tang}{Liu
  et~al\mbox{.}}{2017}]%
        {liu2017capturing}
\bibfield{author}{\bibinfo{person}{Zhenbao Liu}, \bibinfo{person}{Zhen Jia},
  \bibinfo{person}{Chi-Man Vong}, \bibinfo{person}{Shuhui Bu},
  \bibinfo{person}{Junwei Han}, {and} \bibinfo{person}{Xiaojun Tang}.}
  \bibinfo{year}{2017}\natexlab{}.
\newblock \showarticletitle{Capturing high-discriminative fault features for
  electronics-rich analog system via deep learning}.
\newblock \bibinfo{journal}{\emph{IEEE Transactions on Industrial Informatics}}
  \bibinfo{volume}{13}, \bibinfo{number}{3} (\bibinfo{year}{2017}),
  \bibinfo{pages}{1213--1226}.
\newblock


\bibitem[\protect\citeauthoryear{Livengood, Tan, Hack, Kane, and
  Greenzweig}{Livengood et~al\mbox{.}}{2011}]%
        {livengood_tan_hack_kane_greenzweig_2011}
\bibfield{author}{\bibinfo{person}{R Livengood}, \bibinfo{person}{S Tan},
  \bibinfo{person}{P Hack}, \bibinfo{person}{M Kane}, {and} \bibinfo{person}{Y
  Greenzweig}.} \bibinfo{year}{2011}\natexlab{}.
\newblock \showarticletitle{Focused Ion Beam Circuit Edit–A Look into the
  Past, Present, and Future}.
\newblock \bibinfo{journal}{\emph{Microscopy and Microanalysis}}
  \bibinfo{volume}{17}, \bibinfo{number}{S2} (\bibinfo{year}{2011}),
  \bibinfo{pages}{672–673}.
\newblock
\urldef\tempurl%
\url{https://doi.org/10.1017/S1431927611004235}
\showDOI{\tempurl}


\bibitem[\protect\citeauthoryear{Louren{\c{c}}o, Afacan, Martins, Passos,
  Canelas, P{\'o}voa, Horta, and Dundar}{Louren{\c{c}}o et~al\mbox{.}}{2019}]%
        {lourencco2019using}
\bibfield{author}{\bibinfo{person}{Nuno Louren{\c{c}}o}, \bibinfo{person}{Engin
  Afacan}, \bibinfo{person}{Ricardo Martins}, \bibinfo{person}{F{\'a}bio
  Passos}, \bibinfo{person}{Ant{\'o}nio Canelas}, \bibinfo{person}{Ricardo
  P{\'o}voa}, \bibinfo{person}{Nuno Horta}, {and} \bibinfo{person}{G Dundar}.}
  \bibinfo{year}{2019}\natexlab{}.
\newblock \showarticletitle{Using polynomial regression and artificial neural
  networks for reusable analog ic sizing}. In \bibinfo{booktitle}{\emph{2019
  16th International Conference on Synthesis, Modeling, Analysis and Simulation
  Methods and Applications to Circuit Design (SMACD)}}. IEEE,
  \bibinfo{pages}{13--16}.
\newblock


\bibitem[\protect\citeauthoryear{Louren{\c{c}}o, Rosa, Martins, Aidos, Canelas,
  P{\'o}voa, and Horta}{Louren{\c{c}}o et~al\mbox{.}}{2018}]%
        {lourencco2018exploration}
\bibfield{author}{\bibinfo{person}{Nuno Louren{\c{c}}o},
  \bibinfo{person}{Jo{\~a}o Rosa}, \bibinfo{person}{Ricardo Martins},
  \bibinfo{person}{Helena Aidos}, \bibinfo{person}{Ant{\'o}nio Canelas},
  \bibinfo{person}{Ricardo P{\'o}voa}, {and} \bibinfo{person}{Nuno Horta}.}
  \bibinfo{year}{2018}\natexlab{}.
\newblock \showarticletitle{On the exploration of promising analog IC designs
  via artificial neural networks}. In \bibinfo{booktitle}{\emph{2018 15th
  International Conference on Synthesis, Modeling, Analysis and Simulation
  Methods and Applications to Circuit Design (SMACD)}}. IEEE,
  \bibinfo{pages}{133--136}.
\newblock


\bibitem[\protect\citeauthoryear{Lu, Nath, Khandelwal, and Lim}{Lu
  et~al\mbox{.}}{2021a}]%
        {lu2021rl}
\bibfield{author}{\bibinfo{person}{Yi-Chen Lu}, \bibinfo{person}{Siddhartha
  Nath}, \bibinfo{person}{Vishal Khandelwal}, {and} \bibinfo{person}{Sung~Kyu
  Lim}.} \bibinfo{year}{2021}\natexlab{a}.
\newblock \showarticletitle{{RL}-sizer: Vlsi gate sizing for timing
  optimization using deep reinforcement learning}. In
  \bibinfo{booktitle}{\emph{2021 58th ACM/IEEE Design Automation Conference
  (DAC)}}. IEEE, \bibinfo{pages}{733--738}.
\newblock


\bibitem[\protect\citeauthoryear{Lu, Pentapati, and Lim}{Lu
  et~al\mbox{.}}{2020}]%
        {lu2020vlsi}
\bibfield{author}{\bibinfo{person}{Yi-Chen Lu}, \bibinfo{person}{Sai
  Pentapati}, {and} \bibinfo{person}{Sung~Kyu Lim}.}
  \bibinfo{year}{2020}\natexlab{}.
\newblock \showarticletitle{VLSI Placement Optimization using Graph Neural
  Networks}. In \bibinfo{booktitle}{\emph{34th Advances in Neural Information
  Processing Systems (NeurIPS) Workshop on ML for Systems}}.
\newblock


\bibitem[\protect\citeauthoryear{Lu, Pentapati, and Lim}{Lu
  et~al\mbox{.}}{2021b}]%
        {lu2021law}
\bibfield{author}{\bibinfo{person}{Yi-Chen Lu}, \bibinfo{person}{Sai
  Pentapati}, {and} \bibinfo{person}{Sung~Kyu Lim}.}
  \bibinfo{year}{2021}\natexlab{b}.
\newblock \showarticletitle{The law of attraction: Affinity-aware placement
  optimization using graph neural networks}. In
  \bibinfo{booktitle}{\emph{Proceedings of the 2021 International Symposium on
  Physical Design}}. \bibinfo{pages}{7--14}.
\newblock


\bibitem[\protect\citeauthoryear{Lyons and Vanderkulk}{Lyons and
  Vanderkulk}{1962}]%
        {lyons1962use}
\bibfield{author}{\bibinfo{person}{Robert~E Lyons} {and}
  \bibinfo{person}{Wouter Vanderkulk}.} \bibinfo{year}{1962}\natexlab{}.
\newblock \showarticletitle{The use of triple-modular redundancy to improve
  computer reliability}.
\newblock \bibinfo{journal}{\emph{IBM journal of research and development}}
  \bibinfo{volume}{6}, \bibinfo{number}{2} (\bibinfo{year}{1962}),
  \bibinfo{pages}{200--209}.
\newblock


\bibitem[\protect\citeauthoryear{Ma, He, Liu, Liu, Zhao, and Jin}{Ma
  et~al\mbox{.}}{2021}]%
        {CADSec2020}
\bibfield{author}{\bibinfo{person}{Haocheng Ma}, \bibinfo{person}{Jiaji He},
  \bibinfo{person}{Yanjiang Liu}, \bibinfo{person}{Leibo Liu},
  \bibinfo{person}{Yiqiang Zhao}, {and} \bibinfo{person}{Yier Jin}.}
  \bibinfo{year}{2021}\natexlab{}.
\newblock \showarticletitle{Security-Driven Placement and Routing Tools for
  Electromagnetic Side-Channel Protection}.
\newblock \bibinfo{journal}{\emph{IEEE Transactions on Computer-Aided Design of
  Integrated Circuits and Systems}} \bibinfo{volume}{40}, \bibinfo{number}{6}
  (\bibinfo{year}{2021}), \bibinfo{pages}{1077--1089}.
\newblock
\urldef\tempurl%
\url{https://doi.org/10.1109/TCAD.2020.3024938}
\showDOI{\tempurl}


\bibitem[\protect\citeauthoryear{Maga{\~n}a, Shi, Melchert, and
  Davoodi}{Maga{\~n}a et~al\mbox{.}}{2017}]%
        {magana2017proximity}
\bibfield{author}{\bibinfo{person}{Jonathon Maga{\~n}a},
  \bibinfo{person}{Daohang Shi}, \bibinfo{person}{Jackson Melchert}, {and}
  \bibinfo{person}{Azadeh Davoodi}.} \bibinfo{year}{2017}\natexlab{}.
\newblock \showarticletitle{Are proximity attacks a threat to the security of
  split manufacturing of integrated circuits?}
\newblock \bibinfo{journal}{\emph{IEEE Transactions on Very Large Scale
  Integration (VLSI) Systems}} \bibinfo{volume}{25}, \bibinfo{number}{12}
  (\bibinfo{year}{2017}), \bibinfo{pages}{3406--3419}.
\newblock


\bibitem[\protect\citeauthoryear{Makrani, Farahmand, Sayadi, Bondi, Dinakarrao,
  Homayoun, and Rafatirad}{Makrani et~al\mbox{.}}{2019}]%
        {Makrani2019}
\bibfield{author}{\bibinfo{person}{Hosein~Mohammadi Makrani},
  \bibinfo{person}{Farnoud Farahmand}, \bibinfo{person}{Hossein Sayadi},
  \bibinfo{person}{Sara Bondi}, \bibinfo{person}{Sai Manoj~Pudukotai
  Dinakarrao}, \bibinfo{person}{Houman Homayoun}, {and}
  \bibinfo{person}{Setareh Rafatirad}.} \bibinfo{year}{2019}\natexlab{}.
\newblock \showarticletitle{Pyramid: Machine learning framework to estimate the
  optimal timing and resource usage of a high-level synthesis design}.
\newblock \bibinfo{journal}{\emph{Proceedings - 29th International Conference
  on Field-Programmable Logic and Applications, FPL 2019}},
  \bibinfo{pages}{397--403}.
\newblock
\showISBNx{9781728148847}
\urldef\tempurl%
\url{https://doi.org/10.1109/FPL.2019.00069}
\showDOI{\tempurl}


\bibitem[\protect\citeauthoryear{Markowitch, Lerman, and Bontempi}{Markowitch
  et~al\mbox{.}}{2011}]%
        {markowitch_2011sca}
\bibfield{author}{\bibinfo{person}{Olivier Markowitch}, \bibinfo{person}{Liran
  Lerman}, {and} \bibinfo{person}{Gianluca Bontempi}.}
  \bibinfo{year}{2011}\natexlab{}.
\newblock \showarticletitle{Side channel attack: An approach based on machine
  learning}.
\newblock \bibinfo{journal}{\emph{Constructive Side-Channel Analysis and Secure
  Design, COSADE}}.
\newblock


\bibitem[\protect\citeauthoryear{Martin, Scheffer, and Lavagno}{Martin
  et~al\mbox{.}}{2016}]%
        {martin_scheffer_lavagno_2016}
\bibfield{author}{\bibinfo{person}{Grant Martin}, \bibinfo{person}{Louis
  Scheffer}, {and} \bibinfo{person}{Luciano Lavagno}.}
  \bibinfo{year}{2016}\natexlab{}.
\newblock \bibinfo{booktitle}{\emph{Electronic Design Automation for Integrated
  Circuits Handbook}}.
\newblock \bibinfo{publisher}{CRC Press}.
\newblock


\bibitem[\protect\citeauthoryear{Massad, Zhang, Garg, and Tripunitara}{Massad
  et~al\mbox{.}}{2017}]%
        {massad2017logic}
\bibfield{author}{\bibinfo{person}{Mohamed~El Massad}, \bibinfo{person}{Jun
  Zhang}, \bibinfo{person}{Siddharth Garg}, {and} \bibinfo{person}{Mahesh~V
  Tripunitara}.} \bibinfo{year}{2017}\natexlab{}.
\newblock \showarticletitle{Logic locking for secure outsourced chip
  fabrication: A new attack and provably secure defense mechanism}.
\newblock \bibinfo{journal}{\emph{arXiv preprint arXiv:1703.10187}}
  (\bibinfo{year}{2017}).
\newblock


\bibitem[\protect\citeauthoryear{Matsuba, Takai, Fukuda, and Kubo}{Matsuba
  et~al\mbox{.}}{2018}]%
        {matsuba2018inference}
\bibfield{author}{\bibinfo{person}{Teruki Matsuba}, \bibinfo{person}{Nobukazu
  Takai}, \bibinfo{person}{Masafumi Fukuda}, {and} \bibinfo{person}{Yusuke
  Kubo}.} \bibinfo{year}{2018}\natexlab{}.
\newblock \showarticletitle{Inference of suitable for required specification
  analog circuit topology using deep learning}. In
  \bibinfo{booktitle}{\emph{2018 International Symposium on Intelligent Signal
  Processing and Communication Systems (ISPACS)}}. IEEE,
  \bibinfo{pages}{131--134}.
\newblock


\bibitem[\protect\citeauthoryear{McMahan and Ramage}{McMahan and
  Ramage}{2017}]%
        {McMahan2017-federated-first-intro}
\bibfield{author}{\bibinfo{person}{Brendan McMahan} {and}
  \bibinfo{person}{Daniel Ramage}.} \bibinfo{year}{2017}\natexlab{}.
\newblock \bibinfo{title}{{Federated Learning: Collaborative Machine Learning
  without Centralized Training Data}}.
\newblock
  \bibinfo{howpublished}{\url{https://ai.googleblog.com/2017/04/federated-learning-collaborative.html}}.
\newblock
\newblock
\shownote{Accessed: 2022-2-1.}


\bibitem[\protect\citeauthoryear{Mina, Jabbour, and Sakr}{Mina
  et~al\mbox{.}}{2022}]%
        {mina2022review}
\bibfield{author}{\bibinfo{person}{Rayan Mina}, \bibinfo{person}{Chadi
  Jabbour}, {and} \bibinfo{person}{George~E Sakr}.}
  \bibinfo{year}{2022}\natexlab{}.
\newblock \showarticletitle{A Review of Machine Learning Techniques in Analog
  Integrated Circuit Design Automation}.
\newblock \bibinfo{journal}{\emph{Electronics}} \bibinfo{volume}{11},
  \bibinfo{number}{3} (\bibinfo{year}{2022}), \bibinfo{pages}{435}.
\newblock


\bibitem[\protect\citeauthoryear{Mirhoseini, Goldie, Yazgan, Jiang, Songhori,
  Wang, Lee, Johnson, Pathak, Bae, et~al\mbox{.}}{Mirhoseini
  et~al\mbox{.}}{2020}]%
        {mirhoseini2020chip}
\bibfield{author}{\bibinfo{person}{Azalia Mirhoseini}, \bibinfo{person}{Anna
  Goldie}, \bibinfo{person}{Mustafa Yazgan}, \bibinfo{person}{Joe Jiang},
  \bibinfo{person}{Ebrahim Songhori}, \bibinfo{person}{Shen Wang},
  \bibinfo{person}{Young-Joon Lee}, \bibinfo{person}{Eric Johnson},
  \bibinfo{person}{Omkar Pathak}, \bibinfo{person}{Sungmin Bae},
  {et~al\mbox{.}}} \bibinfo{year}{2020}\natexlab{}.
\newblock \showarticletitle{Chip placement with deep reinforcement learning}.
\newblock \bibinfo{journal}{\emph{arXiv preprint arXiv:2004.10746}}
  (\bibinfo{year}{2020}).
\newblock


\bibitem[\protect\citeauthoryear{Mirhoseini, Goldie, Yazgan, Jiang, Songhori,
  Wang, Lee, Johnson, Pathak, Nazi, Pak, Tong, Srinivasa, Hang, Tuncer, Le,
  Laudon, Ho, Carpenter, and Dean}{Mirhoseini et~al\mbox{.}}{2021}]%
        {Mirhoseini2021AGP}
\bibfield{author}{\bibinfo{person}{Azalia Mirhoseini}, \bibinfo{person}{Anna
  Goldie}, \bibinfo{person}{Mustafa Yazgan}, \bibinfo{person}{Joe~Wenjie
  Jiang}, \bibinfo{person}{Ebrahim~M. Songhori}, \bibinfo{person}{Shen Wang},
  \bibinfo{person}{Young-Joon Lee}, \bibinfo{person}{Eric Johnson},
  \bibinfo{person}{Omkar Pathak}, \bibinfo{person}{Azade Nazi},
  \bibinfo{person}{Jiwoo Pak}, \bibinfo{person}{Andy Tong},
  \bibinfo{person}{Kavya Srinivasa}, \bibinfo{person}{Will Hang},
  \bibinfo{person}{Emre Tuncer}, \bibinfo{person}{Quoc~V. Le},
  \bibinfo{person}{James Laudon}, \bibinfo{person}{Richard Ho},
  \bibinfo{person}{Roger Carpenter}, {and} \bibinfo{person}{Jeff Dean}.}
  \bibinfo{year}{2021}\natexlab{}.
\newblock \showarticletitle{A graph placement methodology for fast chip
  design.}
\newblock \bibinfo{journal}{\emph{Nature}}  \bibinfo{volume}{594 7862}
  (\bibinfo{year}{2021}), \bibinfo{pages}{207--212}.
\newblock


\bibitem[\protect\citeauthoryear{Moos, Wegener, and Moradi}{Moos
  et~al\mbox{.}}{2021}]%
        {moos2021dl}
\bibfield{author}{\bibinfo{person}{Thorben Moos}, \bibinfo{person}{Felix
  Wegener}, {and} \bibinfo{person}{Amir Moradi}.}
  \bibinfo{year}{2021}\natexlab{}.
\newblock \showarticletitle{{DL-LA}: Deep Learning Leakage Assessment: A modern
  roadmap for SCA evaluations}.
\newblock \bibinfo{journal}{\emph{IACR Transactions on Cryptographic Hardware
  and Embedded Systems}} (\bibinfo{year}{2021}), \bibinfo{pages}{552--598}.
\newblock


\bibitem[\protect\citeauthoryear{Murphy and McCarthy}{Murphy and
  McCarthy}{2021}]%
        {murphy2021automated}
\bibfield{author}{\bibinfo{person}{Se{\'a}n~D Murphy} {and}
  \bibinfo{person}{Kevin~G McCarthy}.} \bibinfo{year}{2021}\natexlab{}.
\newblock \showarticletitle{Automated Design of CMOS Operational Amplifier
  Using a Neural Network}. In \bibinfo{booktitle}{\emph{2021 32nd Irish Signals
  and Systems Conference (ISSC)}}. IEEE, \bibinfo{pages}{1--6}.
\newblock


\bibitem[\protect\citeauthoryear{Nahiyan, Farahmandi, Mishra, Forte, and
  Tehranipoor}{Nahiyan et~al\mbox{.}}{2019}]%
        {nahiyan2019}
\bibfield{author}{\bibinfo{person}{Adib Nahiyan}, \bibinfo{person}{Farimah
  Farahmandi}, \bibinfo{person}{Prabhat Mishra}, \bibinfo{person}{Domenic
  Forte}, {and} \bibinfo{person}{Mark Tehranipoor}.}
  \bibinfo{year}{2019}\natexlab{}.
\newblock \showarticletitle{Security-Aware FSM Design Flow for Identifying and
  Mitigating Vulnerabilities to Fault Attacks}.
\newblock \bibinfo{journal}{\emph{IEEE Transactions on Computer-Aided Design of
  Integrated Circuits and Systems}} \bibinfo{volume}{38}, \bibinfo{number}{6}
  (\bibinfo{year}{2019}), \bibinfo{pages}{1003--1016}.
\newblock
\urldef\tempurl%
\url{https://doi.org/10.1109/TCAD.2018.2834396}
\showDOI{\tempurl}


\bibitem[\protect\citeauthoryear{Nahiyan, Sadi, Vittal, Contreras, Forte, and
  Tehranipoor}{Nahiyan et~al\mbox{.}}{2017}]%
        {nahiyan2017trojandetection}
\bibfield{author}{\bibinfo{person}{Adib Nahiyan}, \bibinfo{person}{Mehdi Sadi},
  \bibinfo{person}{Rahul Vittal}, \bibinfo{person}{Gustavo Contreras},
  \bibinfo{person}{Domenic Forte}, {and} \bibinfo{person}{Mark Tehranipoor}.}
  \bibinfo{year}{2017}\natexlab{}.
\newblock \showarticletitle{Hardware trojan detection through information flow
  security verification}. In \bibinfo{booktitle}{\emph{2017 IEEE International
  Test Conference (ITC)}}. \bibinfo{pages}{1--10}.
\newblock
\urldef\tempurl%
\url{https://doi.org/10.1109/TEST.2017.8242062}
\showDOI{\tempurl}


\bibitem[\protect\citeauthoryear{Narayanan, Chandramohan, Venkatesan, Chen,
  Liu, and Jaiswal}{Narayanan et~al\mbox{.}}{2017}]%
        {narayanan2017graph2vec}
\bibfield{author}{\bibinfo{person}{Annamalai Narayanan},
  \bibinfo{person}{Mahinthan Chandramohan}, \bibinfo{person}{Rajasekar
  Venkatesan}, \bibinfo{person}{Lihui Chen}, \bibinfo{person}{Yang Liu}, {and}
  \bibinfo{person}{Shantanu Jaiswal}.} \bibinfo{year}{2017}\natexlab{}.
\newblock \bibinfo{title}{graph2vec: Learning Distributed Representations of
  Graphs}.
\newblock
\newblock
\showeprint[arxiv]{1707.05005}~[cs.AI]


\bibitem[\protect\citeauthoryear{Neto, Austin, Temple, Amar{\`u}, Tang, and
  Gaillardon}{Neto et~al\mbox{.}}{2019}]%
        {Neto2019LSOracleAL}
\bibfield{author}{\bibinfo{person}{W.~L. Neto}, \bibinfo{person}{Max Austin},
  \bibinfo{person}{Scott Temple}, \bibinfo{person}{L. Amar{\`u}},
  \bibinfo{person}{Xifan Tang}, {and} \bibinfo{person}{P. Gaillardon}.}
  \bibinfo{year}{2019}\natexlab{}.
\newblock \showarticletitle{LSOracle: a Logic Synthesis Framework Driven by
  Artificial Intelligence: Invited Paper}.
\newblock \bibinfo{journal}{\emph{2019 IEEE/ACM International Conference on
  Computer-Aided Design (ICCAD)}} (\bibinfo{year}{2019}),
  \bibinfo{pages}{1--6}.
\newblock


\bibitem[\protect\citeauthoryear{Neumeier and Keren}{Neumeier and
  Keren}{2012}]%
        {neumeier2012punctured}
\bibfield{author}{\bibinfo{person}{Yaara Neumeier} {and} \bibinfo{person}{Osnat
  Keren}.} \bibinfo{year}{2012}\natexlab{}.
\newblock \showarticletitle{Punctured Karpovsky-Taubin binary robust error
  detecting codes for cryptographic devices}. In \bibinfo{booktitle}{\emph{2012
  IEEE 18th International On-Line Testing Symposium (IOLTS)}}. IEEE,
  \bibinfo{pages}{156--161}.
\newblock


\bibitem[\protect\citeauthoryear{Pagliarini, Sweeney, Mai, Blanton, Pileggi,
  and Mitra}{Pagliarini et~al\mbox{.}}{2020}]%
        {pagliarini2020split}
\bibfield{author}{\bibinfo{person}{Samuel Pagliarini}, \bibinfo{person}{Joseph
  Sweeney}, \bibinfo{person}{Ken Mai}, \bibinfo{person}{Shawn Blanton},
  \bibinfo{person}{Larry Pileggi}, {and} \bibinfo{person}{Subhasish Mitra}.}
  \bibinfo{year}{2020}\natexlab{}.
\newblock \showarticletitle{Split-chip design to prevent ip reverse
  engineering}.
\newblock \bibinfo{journal}{\emph{IEEE Design \& Test}} \bibinfo{volume}{38},
  \bibinfo{number}{4} (\bibinfo{year}{2020}), \bibinfo{pages}{109--118}.
\newblock


\bibitem[\protect\citeauthoryear{Pan, Huang, and Chen}{Pan
  et~al\mbox{.}}{2019}]%
        {pan2019late}
\bibfield{author}{\bibinfo{person}{Po-Cheng Pan}, \bibinfo{person}{Chien-Chia
  Huang}, {and} \bibinfo{person}{Hung-Ming Chen}.}
  \bibinfo{year}{2019}\natexlab{}.
\newblock \showarticletitle{Late breaking results: An efficient learning-based
  approach for performance exploration on analog and RF circuit synthesis}. In
  \bibinfo{booktitle}{\emph{2019 56th ACM/IEEE Design Automation Conference
  (DAC)}}. IEEE, \bibinfo{pages}{1--2}.
\newblock


\bibitem[\protect\citeauthoryear{{Pan} and {Yang}}{{Pan} and {Yang}}{2010}]%
        {pan2010transfer}
\bibfield{author}{\bibinfo{person}{S.~J. {Pan}} {and} \bibinfo{person}{Q.
  {Yang}}.} \bibinfo{year}{2010}\natexlab{}.
\newblock \showarticletitle{A Survey on Transfer Learning}.
\newblock \bibinfo{journal}{\emph{IEEE Trans Knowl Data Eng}}
  \bibinfo{volume}{22}, \bibinfo{number}{10} (\bibinfo{year}{2010}),
  \bibinfo{pages}{1345--1359}.
\newblock
\urldef\tempurl%
\url{https://doi.org/10.1109/TKDE.2009.191}
\showDOI{\tempurl}


\bibitem[\protect\citeauthoryear{Pan and Mishra}{Pan and Mishra}{2021}]%
        {pan2021atpg}
\bibfield{author}{\bibinfo{person}{Zhixin Pan} {and} \bibinfo{person}{Prabhat
  Mishra}.} \bibinfo{year}{2021}\natexlab{}.
\newblock \showarticletitle{Automated Test Generation for Hardware Trojan
  Detection Using Reinforcement Learning}. In
  \bibinfo{booktitle}{\emph{Proceedings of the 26th Asia and South Pacific
  Design Automation Conference}} (Tokyo, Japan) \emph{(\bibinfo{series}{ASPDAC
  '21})}. \bibinfo{publisher}{Association for Computing Machinery},
  \bibinfo{address}{New York, NY, USA}, \bibinfo{pages}{408–413}.
\newblock
\showISBNx{9781450379991}
\urldef\tempurl%
\url{https://doi.org/10.1145/3394885.3431595}
\showDOI{\tempurl}


\bibitem[\protect\citeauthoryear{Pasandi, Nazarian, and Pedram}{Pasandi
  et~al\mbox{.}}{2019}]%
        {pasandi2019approximate}
\bibfield{author}{\bibinfo{person}{Ghasem Pasandi}, \bibinfo{person}{Shahin
  Nazarian}, {and} \bibinfo{person}{Massoud Pedram}.}
  \bibinfo{year}{2019}\natexlab{}.
\newblock \showarticletitle{Approximate logic synthesis: A reinforcement
  learning-based technology mapping approach}. In
  \bibinfo{booktitle}{\emph{20th International Symposium on Quality Electronic
  Design (ISQED)}}. IEEE, \bibinfo{pages}{26--32}.
\newblock


\bibitem[\protect\citeauthoryear{Pasandi, Peterson, Herrera, Nazarian, and
  Pedram}{Pasandi et~al\mbox{.}}{2020}]%
        {Pasandi2020}
\bibfield{author}{\bibinfo{person}{Ghasem Pasandi}, \bibinfo{person}{MacKenzie
  Peterson}, \bibinfo{person}{Moises Herrera}, \bibinfo{person}{Shahin
  Nazarian}, {and} \bibinfo{person}{Massoud Pedram}.}
  \bibinfo{year}{2020}\natexlab{}.
\newblock \showarticletitle{Deep-PowerX: A deep learning-based framework for
  low-power approximate logic synthesis}.
\newblock \bibinfo{journal}{\emph{ACM International Conference Proceeding
  Series}}.
\newblock
\showISBNx{9781450370530}
\urldef\tempurl%
\url{https://doi.org/10.1145/3370748.3406555}
\showDOI{\tempurl}


\bibitem[\protect\citeauthoryear{Passos, Silva, and Fernandes}{Passos
  et~al\mbox{.}}{2006}]%
        {passos2006rbf}
\bibfield{author}{\bibinfo{person}{M{\'a}rcio~G Passos},
  \bibinfo{person}{PH~da~F Silva}, {and} \bibinfo{person}{Humberto~CC
  Fernandes}.} \bibinfo{year}{2006}\natexlab{}.
\newblock \showarticletitle{A RBF/MLP modular neural network for microwave
  device modeling}.
\newblock \bibinfo{journal}{\emph{International Journal of computer science and
  network security}} \bibinfo{volume}{6}, \bibinfo{number}{5A}
  (\bibinfo{year}{2006}), \bibinfo{pages}{81--86}.
\newblock


\bibitem[\protect\citeauthoryear{Paszke, Gross, Massa, Lerer, Bradbury, Chanan,
  Killeen, Lin, Gimelshein, Antiga, Desmaison, Kopf, Yang, DeVito, Raison,
  Tejani, Chilamkurthy, Steiner, Fang, Bai, and Chintala}{Paszke
  et~al\mbox{.}}{2019}]%
        {NEURIPS2019_9015}
\bibfield{author}{\bibinfo{person}{Adam Paszke}, \bibinfo{person}{Sam Gross},
  \bibinfo{person}{Francisco Massa}, \bibinfo{person}{Adam Lerer},
  \bibinfo{person}{James Bradbury}, \bibinfo{person}{Gregory Chanan},
  \bibinfo{person}{Trevor Killeen}, \bibinfo{person}{Zeming Lin},
  \bibinfo{person}{Natalia Gimelshein}, \bibinfo{person}{Luca Antiga},
  \bibinfo{person}{Alban Desmaison}, \bibinfo{person}{Andreas Kopf},
  \bibinfo{person}{Edward Yang}, \bibinfo{person}{Zachary DeVito},
  \bibinfo{person}{Martin Raison}, \bibinfo{person}{Alykhan Tejani},
  \bibinfo{person}{Sasank Chilamkurthy}, \bibinfo{person}{Benoit Steiner},
  \bibinfo{person}{Lu Fang}, \bibinfo{person}{Junjie Bai}, {and}
  \bibinfo{person}{Soumith Chintala}.} \bibinfo{year}{2019}\natexlab{}.
\newblock \showarticletitle{PyTorch: An Imperative Style, High-Performance Deep
  Learning Library}.
\newblock In \bibinfo{booktitle}{\emph{Advances in Neural Information
  Processing Systems 32}}, \bibfield{editor}{\bibinfo{person}{H.~Wallach},
  \bibinfo{person}{H.~Larochelle}, \bibinfo{person}{A.~Beygelzimer},
  \bibinfo{person}{F.~d\textquotesingle Alch\'{e}-Buc},
  \bibinfo{person}{E.~Fox}, {and} \bibinfo{person}{R.~Garnett}} (Eds.).
  \bibinfo{publisher}{Curran Associates, Inc.}, \bibinfo{pages}{8024--8035}.
\newblock
\urldef\tempurl%
\url{http://papers.neurips.cc/paper/9015-pytorch-an-imperative-style-high-performance-deep-learning-library.pdf}
\showURL{%
\tempurl}


\bibitem[\protect\citeauthoryear{Pecht and Tiku}{Pecht and Tiku}{2006}]%
        {pecht2006}
\bibfield{author}{\bibinfo{person}{M. Pecht} {and} \bibinfo{person}{S. Tiku}.}
  \bibinfo{year}{2006}\natexlab{}.
\newblock \showarticletitle{Bogus: Electronic Manufacturing and Consumers
  Confront a Rising Tide of Counterfeit Electronics}.
\newblock \bibinfo{journal}{\emph{IEEE Spectr.}} \bibinfo{volume}{43},
  \bibinfo{number}{5} (\bibinfo{date}{may} \bibinfo{year}{2006}),
  \bibinfo{pages}{37–46}.
\newblock
\showISSN{0018-9235}
\urldef\tempurl%
\url{https://doi.org/10.1109/MSPEC.2006.1628506}
\showDOI{\tempurl}


\bibitem[\protect\citeauthoryear{Pedregosa, Varoquaux, Gramfort, Michel,
  Thirion, Grisel, Blondel, Prettenhofer, Weiss, Dubourg, Vanderplas, Passos,
  Cournapeau, Brucher, Perrot, and Duchesnay}{Pedregosa et~al\mbox{.}}{2011}]%
        {scikit-learn}
\bibfield{author}{\bibinfo{person}{F. Pedregosa}, \bibinfo{person}{G.
  Varoquaux}, \bibinfo{person}{A. Gramfort}, \bibinfo{person}{V. Michel},
  \bibinfo{person}{B. Thirion}, \bibinfo{person}{O. Grisel},
  \bibinfo{person}{M. Blondel}, \bibinfo{person}{P. Prettenhofer},
  \bibinfo{person}{R. Weiss}, \bibinfo{person}{V. Dubourg}, \bibinfo{person}{J.
  Vanderplas}, \bibinfo{person}{A. Passos}, \bibinfo{person}{D. Cournapeau},
  \bibinfo{person}{M. Brucher}, \bibinfo{person}{M. Perrot}, {and}
  \bibinfo{person}{E. Duchesnay}.} \bibinfo{year}{2011}\natexlab{}.
\newblock \showarticletitle{Scikit-learn: Machine Learning in {P}ython}.
\newblock \bibinfo{journal}{\emph{Journal of Machine Learning Research}}
  \bibinfo{volume}{12} (\bibinfo{year}{2011}), \bibinfo{pages}{2825--2830}.
\newblock


\bibitem[\protect\citeauthoryear{Perry}{Perry}{2007}]%
        {perry_2007}
\bibfield{author}{\bibinfo{person}{Guy Perry}.}
  \bibinfo{year}{2007}\natexlab{}.
\newblock \bibinfo{booktitle}{\emph{The fundamentals of Digital Semiconductor
  testing}}.
\newblock \bibinfo{publisher}{Soft Test Inc.}
\newblock


\bibitem[\protect\citeauthoryear{Petrovic, Ilic, Schoof, and
  Stamenkovic}{Petrovic et~al\mbox{.}}{2012}]%
        {petrovic2012}
\bibfield{author}{\bibinfo{person}{Vladimir Petrovic}, \bibinfo{person}{Marko
  Ilic}, \bibinfo{person}{Gunter Schoof}, {and} \bibinfo{person}{Zoran
  Stamenkovic}.} \bibinfo{year}{2012}\natexlab{}.
\newblock \showarticletitle{Design methodology for fault tolerant ASICs}. In
  \bibinfo{booktitle}{\emph{2012 IEEE 15th International Symposium on Design
  and Diagnostics of Electronic Circuits Systems (DDECS)}}.
  \bibinfo{pages}{8--11}.
\newblock
\urldef\tempurl%
\url{https://doi.org/10.1109/DDECS.2012.6219014}
\showDOI{\tempurl}


\bibitem[\protect\citeauthoryear{Picek, Heuser, Jovic, Bhasin, and
  Regazzoni}{Picek et~al\mbox{.}}{2019}]%
        {related_works:SCA_metrics}
\bibfield{author}{\bibinfo{person}{Stjepan Picek}, \bibinfo{person}{Annelie
  Heuser}, \bibinfo{person}{Alan Jovic}, \bibinfo{person}{Shivam Bhasin}, {and}
  \bibinfo{person}{Francesco Regazzoni}.} \bibinfo{year}{2019}\natexlab{}.
\newblock \showarticletitle{The Curse of Class Imbalance and Conflicting
  Metrics with Machine Learning for Side-channel Evaluations}.
\newblock \bibinfo{journal}{\emph{IACR Trans. on Cryptographic Hardware and
  Embedded Systems}} \bibinfo{volume}{2019}, \bibinfo{number}{1}
  (\bibinfo{year}{2019}), \bibinfo{pages}{1--29}.
\newblock


\bibitem[\protect\citeauthoryear{Picek, Perin, Mariot, Wu, and Batina}{Picek
  et~al\mbox{.}}{2021}]%
        {picek2021sok}
\bibfield{author}{\bibinfo{person}{Stjepan Picek}, \bibinfo{person}{Guilherme
  Perin}, \bibinfo{person}{Luca Mariot}, \bibinfo{person}{Lichao Wu}, {and}
  \bibinfo{person}{Lejla Batina}.} \bibinfo{year}{2021}\natexlab{}.
\newblock \showarticletitle{{SoK}: Deep Learning-based Physical Side-channel
  Analysis}.
\newblock \bibinfo{journal}{\emph{Cryptology ePrint Archive}}
  (\bibinfo{year}{2021}).
\newblock


\bibitem[\protect\citeauthoryear{Popp and Mangard}{Popp and Mangard}{2005}]%
        {Popp2005}
\bibfield{author}{\bibinfo{person}{Thomas Popp} {and} \bibinfo{person}{Stefan
  Mangard}.} \bibinfo{year}{2005}\natexlab{}.
\newblock \showarticletitle{Masked Dual-Rail Pre-charge Logic: DPA-Resistance
  Without Routing Constraints}. In \bibinfo{booktitle}{\emph{Cryptographic
  Hardware and Embedded Systems -- CHES 2005}},
  \bibfield{editor}{\bibinfo{person}{Josyula~R. Rao} {and}
  \bibinfo{person}{Berk Sunar}} (Eds.). \bibinfo{publisher}{Springer Berlin
  Heidelberg}, \bibinfo{address}{Berlin, Heidelberg},
  \bibinfo{pages}{172--186}.
\newblock
\showISBNx{978-3-540-31940-5}


\bibitem[\protect\citeauthoryear{Prouff and Rivain}{Prouff and Rivain}{2013}]%
        {prouff_2013_masking}
\bibfield{author}{\bibinfo{person}{Emmanuel Prouff} {and}
  \bibinfo{person}{Matthieu Rivain}.} \bibinfo{year}{2013}\natexlab{}.
\newblock \showarticletitle{Masking against Side-Channel Attacks: A Formal
  Security Proof}. In \bibinfo{booktitle}{\emph{Advances in Cryptology --
  EUROCRYPT 2013}}, \bibfield{editor}{\bibinfo{person}{Thomas Johansson} {and}
  \bibinfo{person}{Phong~Q. Nguyen}} (Eds.). \bibinfo{publisher}{Springer
  Berlin Heidelberg}, \bibinfo{address}{Berlin, Heidelberg},
  \bibinfo{pages}{142--159}.
\newblock
\showISBNx{978-3-642-38348-9}


\bibitem[\protect\citeauthoryear{Prouff, Rivain, and Bevan}{Prouff
  et~al\mbox{.}}{2009}]%
        {prouff2009}
\bibfield{author}{\bibinfo{person}{Emmanuel Prouff}, \bibinfo{person}{Matthieu
  Rivain}, {and} \bibinfo{person}{Regis Bevan}.}
  \bibinfo{year}{2009}\natexlab{}.
\newblock \showarticletitle{Statistical Analysis of Second Order Differential
  Power Analysis}.
\newblock \bibinfo{journal}{\emph{IEEE Trans. Comput.}} \bibinfo{volume}{58},
  \bibinfo{number}{6} (\bibinfo{year}{2009}), \bibinfo{pages}{799--811}.
\newblock
\urldef\tempurl%
\url{https://doi.org/10.1109/TC.2009.15}
\showDOI{\tempurl}


\bibitem[\protect\citeauthoryear{Rajendran, Ali, Sinanoglu, and
  Karri}{Rajendran et~al\mbox{.}}{2015a}]%
        {rajendran2015}
\bibfield{author}{\bibinfo{person}{Jeyavijayan Rajendran},
  \bibinfo{person}{Aman Ali}, \bibinfo{person}{Ozgur Sinanoglu}, {and}
  \bibinfo{person}{Ramesh Karri}.} \bibinfo{year}{2015}\natexlab{a}.
\newblock \showarticletitle{Belling the CAD: Toward Security-Centric Electronic
  System Design}.
\newblock \bibinfo{journal}{\emph{IEEE Transactions on Computer-Aided Design of
  Integrated Circuits and Systems}} \bibinfo{volume}{34}, \bibinfo{number}{11}
  (\bibinfo{year}{2015}), \bibinfo{pages}{1756--1769}.
\newblock
\urldef\tempurl%
\url{https://doi.org/10.1109/TCAD.2015.2428707}
\showDOI{\tempurl}


\bibitem[\protect\citeauthoryear{Rajendran, Pino, Sinanoglu, and
  Karri}{Rajendran et~al\mbox{.}}{2012}]%
        {rajendran2012security}
\bibfield{author}{\bibinfo{person}{Jeyavijayan Rajendran},
  \bibinfo{person}{Youngok Pino}, \bibinfo{person}{Ozgur Sinanoglu}, {and}
  \bibinfo{person}{Ramesh Karri}.} \bibinfo{year}{2012}\natexlab{}.
\newblock \showarticletitle{Security analysis of logic obfuscation}. In
  \bibinfo{booktitle}{\emph{Proceedings of the 49th Annual Design Automation
  Conference}}. \bibinfo{pages}{83--89}.
\newblock


\bibitem[\protect\citeauthoryear{Rajendran, Zhang, Zhang, Rose, Pino,
  Sinanoglu, and Karri}{Rajendran et~al\mbox{.}}{2015b}]%
        {6616532}
\bibfield{author}{\bibinfo{person}{Jeyavijayan Rajendran},
  \bibinfo{person}{Huan Zhang}, \bibinfo{person}{Chi Zhang},
  \bibinfo{person}{Garrett~S. Rose}, \bibinfo{person}{Youngok Pino},
  \bibinfo{person}{Ozgur Sinanoglu}, {and} \bibinfo{person}{Ramesh Karri}.}
  \bibinfo{year}{2015}\natexlab{b}.
\newblock \showarticletitle{Fault Analysis-Based Logic Encryption}.
\newblock \bibinfo{journal}{\emph{IEEE Trans. Comput.}} \bibinfo{volume}{64},
  \bibinfo{number}{2} (\bibinfo{year}{2015}), \bibinfo{pages}{410--424}.
\newblock
\urldef\tempurl%
\url{https://doi.org/10.1109/TC.2013.193}
\showDOI{\tempurl}


\bibitem[\protect\citeauthoryear{Razavi}{Razavi}{2000}]%
        {razavi2000design}
\bibfield{author}{\bibinfo{person}{Behzad Razavi}.}
  \bibinfo{year}{2000}\natexlab{}.
\newblock \bibinfo{title}{Design of Analog CMOS Integrated Circuits}.
\newblock
\newblock


\bibitem[\protect\citeauthoryear{Reparaz}{Reparaz}{2016}]%
        {reparaz2016detecting}
\bibfield{author}{\bibinfo{person}{Oscar Reparaz}.}
  \bibinfo{year}{2016}\natexlab{}.
\newblock \showarticletitle{Detecting flawed masking schemes with leakage
  detection tests}. In \bibinfo{booktitle}{\emph{International Conference on
  Fast Software Encryption}}. Springer, \bibinfo{pages}{204--222}.
\newblock


\bibitem[\protect\citeauthoryear{Reparaz, De~Meyer, Bilgin, Arribas, Nikova,
  Nikov, and Smart}{Reparaz et~al\mbox{.}}{2018}]%
        {CAPA}
\bibfield{author}{\bibinfo{person}{Oscar Reparaz}, \bibinfo{person}{Lauren
  De~Meyer}, \bibinfo{person}{Beg{\"u}l Bilgin}, \bibinfo{person}{Victor
  Arribas}, \bibinfo{person}{Svetla Nikova}, \bibinfo{person}{Ventzislav
  Nikov}, {and} \bibinfo{person}{Nigel Smart}.}
  \bibinfo{year}{2018}\natexlab{}.
\newblock \showarticletitle{CAPA: the spirit of beaver against physical
  attacks}. In \bibinfo{booktitle}{\emph{Annual International Cryptology
  Conference}}. Springer, \bibinfo{pages}{121--151}.
\newblock


\bibitem[\protect\citeauthoryear{Rijsdijk, Wu, and Perin}{Rijsdijk
  et~al\mbox{.}}{2021a}]%
        {rijsdijk2021reinforcement}
\bibfield{author}{\bibinfo{person}{Jorai Rijsdijk}, \bibinfo{person}{Lichao
  Wu}, {and} \bibinfo{person}{Guilherme Perin}.}
  \bibinfo{year}{2021}\natexlab{a}.
\newblock \showarticletitle{Reinforcement Learning-based Design of Side-channel
  Countermeasures}.
\newblock \bibinfo{journal}{\emph{Cryptology ePrint Archive}}
  (\bibinfo{year}{2021}).
\newblock


\bibitem[\protect\citeauthoryear{Rijsdijk, Wu, Perin, and Picek}{Rijsdijk
  et~al\mbox{.}}{2021b}]%
        {DBLP:journals/iacr/RijsdijkWPP21}
\bibfield{author}{\bibinfo{person}{Jorai Rijsdijk}, \bibinfo{person}{Lichao
  Wu}, \bibinfo{person}{Guilherme Perin}, {and} \bibinfo{person}{Stjepan
  Picek}.} \bibinfo{year}{2021}\natexlab{b}.
\newblock \showarticletitle{Reinforcement Learning for Hyperparameter Tuning in
  Deep Learning-based Side-channel Analysis}.
\newblock \bibinfo{journal}{\emph{Cryptology ePrint Archive, Report 2021/526}}
  (\bibinfo{year}{2021}), \bibinfo{pages}{71}.
\newblock


\bibitem[\protect\citeauthoryear{Robertson and Riley}{Robertson and
  Riley}{2018}]%
        {robertson2018big}
\bibfield{author}{\bibinfo{person}{Jordan Robertson} {and}
  \bibinfo{person}{Michael Riley}.} \bibinfo{year}{2018}\natexlab{}.
\newblock \showarticletitle{The big hack: How china used a tiny chip to
  infiltrate us companies}.
\newblock \bibinfo{journal}{\emph{Bloomberg Businessweek}} \bibinfo{volume}{4},
  \bibinfo{number}{2018} (\bibinfo{year}{2018}).
\newblock


\bibitem[\protect\citeauthoryear{Rosenbaum, Xiong, Yang, Chen, and
  Raginsky}{Rosenbaum et~al\mbox{.}}{2020}]%
        {rosenbaum2020machine}
\bibfield{author}{\bibinfo{person}{E Rosenbaum}, \bibinfo{person}{J Xiong},
  \bibinfo{person}{A Yang}, \bibinfo{person}{Z Chen}, {and}
  \bibinfo{person}{Maxim Raginsky}.} \bibinfo{year}{2020}\natexlab{}.
\newblock \showarticletitle{Machine learning for circuit aging simulation}. In
  \bibinfo{booktitle}{\emph{2020 IEEE International Electron Devices Meeting
  (IEDM)}}. IEEE, \bibinfo{pages}{39--1}.
\newblock


\bibitem[\protect\citeauthoryear{Rotman and Wolf}{Rotman and Wolf}{2020}]%
        {rotman2020electric}
\bibfield{author}{\bibinfo{person}{Michael Rotman} {and} \bibinfo{person}{Lior
  Wolf}.} \bibinfo{year}{2020}\natexlab{}.
\newblock \showarticletitle{Electric analog circuit design with hypernetworks
  and a differential simulator}. In \bibinfo{booktitle}{\emph{ICASSP 2020-2020
  IEEE International Conference on Acoustics, Speech and Signal Processing
  (ICASSP)}}. IEEE, \bibinfo{pages}{4157--4161}.
\newblock


\bibitem[\protect\citeauthoryear{Rudin}{Rudin}{2019}]%
        {Rudin2019-stop}
\bibfield{author}{\bibinfo{person}{Cynthia Rudin}.}
  \bibinfo{year}{2019}\natexlab{}.
\newblock \showarticletitle{{Stop explaining black box machine learning models
  for high stakes decisions and use interpretable models instead}}.
\newblock \bibinfo{journal}{\emph{Nature Machine Intelligence}}
  \bibinfo{volume}{1}, \bibinfo{number}{5} (\bibinfo{date}{May}
  \bibinfo{year}{2019}), \bibinfo{pages}{206--215}.
\newblock
\showISSN{2522-5839, 2522-5839}
\urldef\tempurl%
\url{https://doi.org/10.1038/s42256-019-0048-x}
\showDOI{\tempurl}


\bibitem[\protect\citeauthoryear{Rutenbar}{Rutenbar}{2006}]%
        {rutenbar2006design}
\bibfield{author}{\bibinfo{person}{Rob~A Rutenbar}.}
  \bibinfo{year}{2006}\natexlab{}.
\newblock \showarticletitle{Design automation for analog: the next generation
  of tool challenges}. In \bibinfo{booktitle}{\emph{2006 IEEE/ACM International
  Conference on Computer Aided Design}}. IEEE, \bibinfo{pages}{458--460}.
\newblock


\bibitem[\protect\citeauthoryear{Salmani, Tehranipoor, and Plusquellic}{Salmani
  et~al\mbox{.}}{2012}]%
        {salmani2012}
\bibfield{author}{\bibinfo{person}{Hassan Salmani}, \bibinfo{person}{Mark
  Tehranipoor}, {and} \bibinfo{person}{J. Plusquellic}.}
  \bibinfo{year}{2012}\natexlab{}.
\newblock \showarticletitle{A Novel Technique for Improving Hardware Trojan
  Detection and Reducing Trojan Activation Time}.
\newblock \bibinfo{journal}{\emph{Very Large Scale Integration (VLSI) Systems,
  IEEE Transactions on}}  \bibinfo{volume}{20} (\bibinfo{date}{02}
  \bibinfo{year}{2012}), \bibinfo{pages}{112 -- 125}.
\newblock
\urldef\tempurl%
\url{https://doi.org/10.1109/TVLSI.2010.2093547}
\showDOI{\tempurl}


\bibitem[\protect\citeauthoryear{Sanabria-Borb{\'o}n, Soto-Aguilar,
  Estrada-L{\'o}pez, Allaire, and S{\'a}nchez-Sinencio}{Sanabria-Borb{\'o}n
  et~al\mbox{.}}{2020}]%
        {sanabria2020gaussian}
\bibfield{author}{\bibinfo{person}{Adriana~C Sanabria-Borb{\'o}n},
  \bibinfo{person}{Sergio Soto-Aguilar}, \bibinfo{person}{Johan~J
  Estrada-L{\'o}pez}, \bibinfo{person}{Douglas Allaire}, {and}
  \bibinfo{person}{Edgar S{\'a}nchez-Sinencio}.}
  \bibinfo{year}{2020}\natexlab{}.
\newblock \showarticletitle{Gaussian-process-based surrogate for
  optimization-aided and process-variations-aware analog circuit design}.
\newblock \bibinfo{journal}{\emph{Electronics}} \bibinfo{volume}{9},
  \bibinfo{number}{4} (\bibinfo{year}{2020}), \bibinfo{pages}{685}.
\newblock


\bibitem[\protect\citeauthoryear{Santhana~Krishnan and
  Palanisamy}{Santhana~Krishnan and Palanisamy}{2021}]%
        {santhana2021recycled}
\bibfield{author}{\bibinfo{person}{Udaya~Shankar Santhana~Krishnan} {and}
  \bibinfo{person}{Kalpana Palanisamy}.} \bibinfo{year}{2021}\natexlab{}.
\newblock \showarticletitle{Recycled integrated circuit detection using
  reliability analysis and machine learning algorithms}.
\newblock \bibinfo{journal}{\emph{IET Computers \& Digital Techniques}}
  \bibinfo{volume}{15}, \bibinfo{number}{1} (\bibinfo{year}{2021}),
  \bibinfo{pages}{20--35}.
\newblock


\bibitem[\protect\citeauthoryear{Schaldenbrand}{Schaldenbrand}{2019}]%
        {schaldenbrand2019analog}
\bibfield{author}{\bibinfo{person}{A Schaldenbrand}.}
  \bibinfo{year}{2019}\natexlab{}.
\newblock \showarticletitle{Analog reliability analysis for mission-critical
  applications white paper}.
\newblock \bibinfo{journal}{\emph{Cadence Design Systems, Inc}}
  (\bibinfo{year}{2019}).
\newblock


\bibitem[\protect\citeauthoryear{Scheible and Lienig}{Scheible and
  Lienig}{2015}]%
        {scheible2015automation}
\bibfield{author}{\bibinfo{person}{Juergen Scheible} {and}
  \bibinfo{person}{Jens Lienig}.} \bibinfo{year}{2015}\natexlab{}.
\newblock \showarticletitle{Automation of analog IC layout: Challenges and
  solutions}. In \bibinfo{booktitle}{\emph{Proceedings of the 2015 Symposium on
  International Symposium on Physical Design}}. \bibinfo{pages}{33--40}.
\newblock


\bibitem[\protect\citeauthoryear{Settaluri, Haj-Ali, Huang, Hakhamaneshi, and
  Nikolic}{Settaluri et~al\mbox{.}}{2020}]%
        {settaluri2020autockt}
\bibfield{author}{\bibinfo{person}{Keertana Settaluri}, \bibinfo{person}{Ameer
  Haj-Ali}, \bibinfo{person}{Qijing Huang}, \bibinfo{person}{Kourosh
  Hakhamaneshi}, {and} \bibinfo{person}{Borivoje Nikolic}.}
  \bibinfo{year}{2020}\natexlab{}.
\newblock \showarticletitle{Autockt: Deep reinforcement learning of analog
  circuit designs}. In \bibinfo{booktitle}{\emph{2020 Design, Automation \&
  Test in Europe Conference \& Exhibition (DATE)}}. IEEE,
  \bibinfo{pages}{490--495}.
\newblock


\bibitem[\protect\citeauthoryear{Shah, Chatterjee, Sapui, Mukhopadhyay, and
  Basu}{Shah et~al\mbox{.}}{2021}]%
        {shah2021introducing}
\bibfield{author}{\bibinfo{person}{Nimesh Shah}, \bibinfo{person}{Durba
  Chatterjee}, \bibinfo{person}{Brojogopal Sapui}, \bibinfo{person}{Debdeep
  Mukhopadhyay}, {and} \bibinfo{person}{Arindam Basu}.}
  \bibinfo{year}{2021}\natexlab{}.
\newblock \showarticletitle{Introducing recurrence in strong PUFs for enhanced
  machine learning attack resistance}.
\newblock \bibinfo{journal}{\emph{IEEE Journal on Emerging and Selected Topics
  in Circuits and Systems}} \bibinfo{volume}{11}, \bibinfo{number}{2}
  (\bibinfo{year}{2021}), \bibinfo{pages}{319--332}.
\newblock


\bibitem[\protect\citeauthoryear{Shakya, Guin, Tehranipoor, and Forte}{Shakya
  et~al\mbox{.}}{2015}]%
        {shakya2015performance}
\bibfield{author}{\bibinfo{person}{Bicky Shakya}, \bibinfo{person}{Ujjwal
  Guin}, \bibinfo{person}{Mark Tehranipoor}, {and} \bibinfo{person}{Domenic
  Forte}.} \bibinfo{year}{2015}\natexlab{}.
\newblock \showarticletitle{Performance optimization for on-chip sensors to
  detect recycled ICs}. In \bibinfo{booktitle}{\emph{2015 33rd IEEE
  International Conference on Computer Design (ICCD)}}. IEEE,
  \bibinfo{pages}{289--295}.
\newblock


\bibitem[\protect\citeauthoryear{Shakya, He, Salmani, Forte, Bhunia, and
  Tehranipoor}{Shakya et~al\mbox{.}}{2017}]%
        {shakya2017benchmarking}
\bibfield{author}{\bibinfo{person}{Bicky Shakya}, \bibinfo{person}{Tony He},
  \bibinfo{person}{Hassan Salmani}, \bibinfo{person}{Domenic Forte},
  \bibinfo{person}{Swarup Bhunia}, {and} \bibinfo{person}{Mark Tehranipoor}.}
  \bibinfo{year}{2017}\natexlab{}.
\newblock \showarticletitle{Benchmarking of hardware trojans and maliciously
  affected circuits}.
\newblock \bibinfo{journal}{\emph{Journal of Hardware and Systems Security}}
  \bibinfo{volume}{1}, \bibinfo{number}{1} (\bibinfo{year}{2017}),
  \bibinfo{pages}{85--102}.
\newblock


\bibitem[\protect\citeauthoryear{Shakya, Shen, Tehranipoor, and Forte}{Shakya
  et~al\mbox{.}}{2019}]%
        {shakya2019covert}
\bibfield{author}{\bibinfo{person}{Bicky Shakya}, \bibinfo{person}{Haoting
  Shen}, \bibinfo{person}{Mark Tehranipoor}, {and} \bibinfo{person}{Domenic
  Forte}.} \bibinfo{year}{2019}\natexlab{}.
\newblock \showarticletitle{Covert gates: Protecting integrated circuits with
  undetectable camouflaging}.
\newblock \bibinfo{journal}{\emph{IACR Transactions on Cryptographic Hardware
  and Embedded Systems}} (\bibinfo{year}{2019}), \bibinfo{pages}{86--118}.
\newblock


\bibitem[\protect\citeauthoryear{Shamsi, Li, Meade, Zhao, Pan, and Jin}{Shamsi
  et~al\mbox{.}}{2017}]%
        {shamsi2017appsat}
\bibfield{author}{\bibinfo{person}{Kaveh Shamsi}, \bibinfo{person}{Meng Li},
  \bibinfo{person}{Travis Meade}, \bibinfo{person}{Zheng Zhao},
  \bibinfo{person}{David~Z Pan}, {and} \bibinfo{person}{Yier Jin}.}
  \bibinfo{year}{2017}\natexlab{}.
\newblock \showarticletitle{AppSAT: Approximately deobfuscating integrated
  circuits}. In \bibinfo{booktitle}{\emph{2017 IEEE International Symposium on
  Hardware Oriented Security and Trust (HOST)}}. IEEE,
  \bibinfo{pages}{95--100}.
\newblock


\bibitem[\protect\citeauthoryear{Shi, Asadizanjani, Forte, and Tehranipoor}{Shi
  et~al\mbox{.}}{2016}]%
        {shi2016layout}
\bibfield{author}{\bibinfo{person}{Qihang Shi}, \bibinfo{person}{Navid
  Asadizanjani}, \bibinfo{person}{Domenic Forte}, {and} \bibinfo{person}{Mark~M
  Tehranipoor}.} \bibinfo{year}{2016}\natexlab{}.
\newblock \showarticletitle{A layout-driven framework to assess vulnerability
  of ICs to microprobing attacks}. In \bibinfo{booktitle}{\emph{2016 IEEE
  International Symposium on Hardware Oriented Security and Trust (HOST)}}.
  IEEE, \bibinfo{pages}{155--160}.
\newblock


\bibitem[\protect\citeauthoryear{Shi, Tehranipoor, and Forte}{Shi
  et~al\mbox{.}}{2018}]%
        {shi2018obfuscated}
\bibfield{author}{\bibinfo{person}{Qihang Shi}, \bibinfo{person}{Mark~M
  Tehranipoor}, {and} \bibinfo{person}{Domenic Forte}.}
  \bibinfo{year}{2018}\natexlab{}.
\newblock \showarticletitle{Obfuscated built-in self-authentication with secure
  and efficient wire-lifting}.
\newblock \bibinfo{journal}{\emph{IEEE Transactions on Computer-Aided Design of
  Integrated Circuits and Systems}} \bibinfo{volume}{38}, \bibinfo{number}{11}
  (\bibinfo{year}{2018}), \bibinfo{pages}{1981--1994}.
\newblock


\bibitem[\protect\citeauthoryear{Sisejkovic, Merchant, Reimann, Srivastava,
  Hallawa, and Leupers}{Sisejkovic et~al\mbox{.}}{2021}]%
        {Sisejkovic2021-challenge-logic-locking}
\bibfield{author}{\bibinfo{person}{Dominik Sisejkovic}, \bibinfo{person}{Farhad
  Merchant}, \bibinfo{person}{Lennart~M Reimann}, \bibinfo{person}{Harshit
  Srivastava}, \bibinfo{person}{Ahmed Hallawa}, {and} \bibinfo{person}{Rainer
  Leupers}.} \bibinfo{year}{2021}\natexlab{}.
\newblock \showarticletitle{{Challenging the Security of Logic Locking Schemes
  in the Era of Deep Learning: A Neuroevolutionary Approach}}.
\newblock \bibinfo{journal}{\emph{J. Emerg. Technol. Comput. Syst.}}
  \bibinfo{volume}{17}, \bibinfo{number}{3} (\bibinfo{date}{May}
  \bibinfo{year}{2021}), \bibinfo{pages}{1--26}.
\newblock
\showISSN{1550-4832}
\urldef\tempurl%
\url{https://doi.org/10.1145/3431389}
\showDOI{\tempurl}


\bibitem[\protect\citeauthoryear{Stratigopoulos and Sunter}{Stratigopoulos and
  Sunter}{2014}]%
        {stratigopoulos2014fast}
\bibfield{author}{\bibinfo{person}{Haralampos-G Stratigopoulos} {and}
  \bibinfo{person}{Stephen Sunter}.} \bibinfo{year}{2014}\natexlab{}.
\newblock \showarticletitle{Fast Monte Carlo-based estimation of analog
  parametric test metrics}.
\newblock \bibinfo{journal}{\emph{IEEE Transactions on Computer-Aided Design of
  Integrated Circuits and Systems}} \bibinfo{volume}{33}, \bibinfo{number}{12}
  (\bibinfo{year}{2014}), \bibinfo{pages}{1977--1990}.
\newblock


\bibitem[\protect\citeauthoryear{Subramani, Antonopoulos, Abotabl, Nosratinia,
  and Makris}{Subramani et~al\mbox{.}}{2017}]%
        {subramani2017ace}
\bibfield{author}{\bibinfo{person}{Kiruba~Sankaran Subramani},
  \bibinfo{person}{Angelos Antonopoulos}, \bibinfo{person}{Ahmed~Attia
  Abotabl}, \bibinfo{person}{Aria Nosratinia}, {and} \bibinfo{person}{Yiorgos
  Makris}.} \bibinfo{year}{2017}\natexlab{}.
\newblock \showarticletitle{ACE: Adaptive channel estimation for detecting
  analog/RF trojans in WLAN transceivers}. In \bibinfo{booktitle}{\emph{2017
  IEEE/ACM International Conference on Computer-Aided Design (ICCAD)}}. IEEE,
  \bibinfo{pages}{722--727}.
\newblock


\bibitem[\protect\citeauthoryear{Subramanyan, Ray, and Malik}{Subramanyan
  et~al\mbox{.}}{2015}]%
        {subramanyan2015evaluating}
\bibfield{author}{\bibinfo{person}{Pramod Subramanyan}, \bibinfo{person}{Sayak
  Ray}, {and} \bibinfo{person}{Sharad Malik}.} \bibinfo{year}{2015}\natexlab{}.
\newblock \showarticletitle{Evaluating the security of logic encryption
  algorithms}. In \bibinfo{booktitle}{\emph{2015 IEEE International Symposium
  on Hardware Oriented Security and Trust (HOST)}}. IEEE,
  \bibinfo{pages}{137--143}.
\newblock


\bibitem[\protect\citeauthoryear{Sugawara, Shoji, Sakiyama, Matsuda, Miura, and
  Nagata}{Sugawara et~al\mbox{.}}{2019}]%
        {SUGAWARA201963}
\bibfield{author}{\bibinfo{person}{Takeshi Sugawara}, \bibinfo{person}{Natsu
  Shoji}, \bibinfo{person}{Kazuo Sakiyama}, \bibinfo{person}{Kohei Matsuda},
  \bibinfo{person}{Noriyuki Miura}, {and} \bibinfo{person}{Makoto Nagata}.}
  \bibinfo{year}{2019}\natexlab{}.
\newblock \showarticletitle{Side-channel leakage from sensor-based
  countermeasures against fault injection attack}.
\newblock \bibinfo{journal}{\emph{Microelectronics Journal}}
  \bibinfo{volume}{90} (\bibinfo{year}{2019}), \bibinfo{pages}{63--71}.
\newblock
\showISSN{0026-2692}
\urldef\tempurl%
\url{https://doi.org/10.1016/j.mejo.2019.05.017}
\showDOI{\tempurl}


\bibitem[\protect\citeauthoryear{Sun, Cao, Zhu, and Zhao}{Sun
  et~al\mbox{.}}{2019}]%
        {sun2019survey}
\bibfield{author}{\bibinfo{person}{Shiliang Sun}, \bibinfo{person}{Zehui Cao},
  \bibinfo{person}{Han Zhu}, {and} \bibinfo{person}{Jing Zhao}.}
  \bibinfo{year}{2019}\natexlab{}.
\newblock \showarticletitle{A survey of optimization methods from a machine
  learning perspective}.
\newblock \bibinfo{journal}{\emph{IEEE transactions on cybernetics}}
  \bibinfo{volume}{50}, \bibinfo{number}{8} (\bibinfo{year}{2019}),
  \bibinfo{pages}{3668--3681}.
\newblock


\bibitem[\protect\citeauthoryear{Sunar, Gaubatz, and Savas}{Sunar
  et~al\mbox{.}}{2007}]%
        {sunar2007sequential}
\bibfield{author}{\bibinfo{person}{Berk Sunar}, \bibinfo{person}{Gunnar
  Gaubatz}, {and} \bibinfo{person}{Erkay Savas}.}
  \bibinfo{year}{2007}\natexlab{}.
\newblock \showarticletitle{Sequential circuit design for embedded
  cryptographic applications resilient to adversarial faults}.
\newblock \bibinfo{journal}{\emph{IEEE Trans. Comput.}} \bibinfo{volume}{57},
  \bibinfo{number}{1} (\bibinfo{year}{2007}), \bibinfo{pages}{126--138}.
\newblock


\bibitem[\protect\citeauthoryear{Svyatkovskiy, Zhao, Fu, and
  Sundaresan}{Svyatkovskiy et~al\mbox{.}}{2019}]%
        {svyatkovskiy2019pythia}
\bibfield{author}{\bibinfo{person}{Alexey Svyatkovskiy}, \bibinfo{person}{Ying
  Zhao}, \bibinfo{person}{Shengyu Fu}, {and} \bibinfo{person}{Neel
  Sundaresan}.} \bibinfo{year}{2019}\natexlab{}.
\newblock \showarticletitle{Pythia: AI-assisted code completion system}. In
  \bibinfo{booktitle}{\emph{Proceedings of the 25th ACM SIGKDD International
  Conference on Knowledge Discovery \& Data Mining}}.
  \bibinfo{pages}{2727--2735}.
\newblock


\bibitem[\protect\citeauthoryear{Swings, Gielen, and Sansen}{Swings
  et~al\mbox{.}}{1990}]%
        {swings1990intelligent}
\bibfield{author}{\bibinfo{person}{Koen Swings}, \bibinfo{person}{Georges
  Gielen}, {and} \bibinfo{person}{Willy Sansen}.}
  \bibinfo{year}{1990}\natexlab{}.
\newblock \showarticletitle{An intelligent analog IC design system based on
  manipulation of design equations}. In \bibinfo{booktitle}{\emph{IEEE
  Proceedings of the Custom Integrated Circuits Conference}}. IEEE,
  \bibinfo{pages}{8--6}.
\newblock


\bibitem[\protect\citeauthoryear{Takai and Fukuda}{Takai and Fukuda}{2017}]%
        {takai2017prediction}
\bibfield{author}{\bibinfo{person}{Nobukazu Takai} {and}
  \bibinfo{person}{Masafumi Fukuda}.} \bibinfo{year}{2017}\natexlab{}.
\newblock \showarticletitle{Prediction of element values of OPAmp for required
  specifications utilizing deep learning}. In \bibinfo{booktitle}{\emph{2017
  International Symposium on Electronics and Smart Devices (ISESD)}}. IEEE,
  \bibinfo{pages}{300--303}.
\newblock


\bibitem[\protect\citeauthoryear{Tan, Karri, Limaye, Sengupta, Sinanoglu,
  Rahman, Bhunia, Duvalsaint, Rezaei, Shen, et~al\mbox{.}}{Tan
  et~al\mbox{.}}{2020}]%
        {tan2020benchmarking}
\bibfield{author}{\bibinfo{person}{Benjamin Tan}, \bibinfo{person}{Ramesh
  Karri}, \bibinfo{person}{Nimisha Limaye}, \bibinfo{person}{Abhrajit
  Sengupta}, \bibinfo{person}{Ozgur Sinanoglu}, \bibinfo{person}{Md~Moshiur
  Rahman}, \bibinfo{person}{Swarup Bhunia}, \bibinfo{person}{Danielle
  Duvalsaint}, \bibinfo{person}{Amin Rezaei}, \bibinfo{person}{Yuanqi Shen},
  {et~al\mbox{.}}} \bibinfo{year}{2020}\natexlab{}.
\newblock \showarticletitle{Benchmarking at the frontier of hardware security:
  Lessons from logic locking}.
\newblock \bibinfo{journal}{\emph{arXiv preprint arXiv:2006.06806}}
  (\bibinfo{year}{2020}).
\newblock


\bibitem[\protect\citeauthoryear{Tao, Liao, Zeng, and Li}{Tao
  et~al\mbox{.}}{2015}]%
        {tao2015harvesting}
\bibfield{author}{\bibinfo{person}{Jun Tao}, \bibinfo{person}{Changhai Liao},
  \bibinfo{person}{Xuan Zeng}, {and} \bibinfo{person}{Xin Li}.}
  \bibinfo{year}{2015}\natexlab{}.
\newblock \showarticletitle{Harvesting design knowledge from the internet:
  High-dimensional performance tradeoff modeling for large-scale analog
  circuits}.
\newblock \bibinfo{journal}{\emph{IEEE Transactions on Computer-Aided Design of
  Integrated Circuits and Systems}} \bibinfo{volume}{35}, \bibinfo{number}{1}
  (\bibinfo{year}{2015}), \bibinfo{pages}{23--36}.
\newblock


\bibitem[\protect\citeauthoryear{Tehranipoor and Koushanfar}{Tehranipoor and
  Koushanfar}{2010}]%
        {tehranipoor2010}
\bibfield{author}{\bibinfo{person}{Mohammad Tehranipoor} {and}
  \bibinfo{person}{Farinaz Koushanfar}.} \bibinfo{year}{2010}\natexlab{}.
\newblock \showarticletitle{A Survey of Hardware Trojan Taxonomy and
  Detection}.
\newblock \bibinfo{journal}{\emph{IEEE Design Test of Computers}}
  \bibinfo{volume}{27}, \bibinfo{number}{1} (\bibinfo{year}{2010}),
  \bibinfo{pages}{10--25}.
\newblock
\urldef\tempurl%
\url{https://doi.org/10.1109/MDT.2010.7}
\showDOI{\tempurl}


\bibitem[\protect\citeauthoryear{Tehranipoor and Wang}{Tehranipoor and
  Wang}{2011}]%
        {hardware2011}
\bibfield{author}{\bibinfo{person}{Mohammad Tehranipoor} {and}
  \bibinfo{person}{Cliff Wang}.} \bibinfo{year}{2011}\natexlab{}.
\newblock \bibinfo{booktitle}{\emph{Introduction to Hardware Security and
  Trust}}.
\newblock \bibinfo{publisher}{Springer Publishing Company, Incorporated}.
\newblock
\showISBNx{1441980792}


\bibitem[\protect\citeauthoryear{Terapasirdsin and
  Wattanapongsakorn}{Terapasirdsin and Wattanapongsakorn}{2010}]%
        {terapasirdsin2010crosstalk}
\bibfield{author}{\bibinfo{person}{Apichat Terapasirdsin} {and}
  \bibinfo{person}{Naruemon Wattanapongsakorn}.}
  \bibinfo{year}{2010}\natexlab{}.
\newblock \showarticletitle{Crosstalk minimization in VLSI design using signal
  transition avoidance}. In \bibinfo{booktitle}{\emph{2010 10th International
  Symposium on Communications and Information Technologies}}.
  \bibinfo{pages}{911--915}.
\newblock
\urldef\tempurl%
\url{https://doi.org/10.1109/ISCIT.2010.5665117}
\showDOI{\tempurl}


\bibitem[\protect\citeauthoryear{Timon}{Timon}{2019}]%
        {timon2019non}
\bibfield{author}{\bibinfo{person}{Benjamin Timon}.}
  \bibinfo{year}{2019}\natexlab{}.
\newblock \showarticletitle{Non-profiled deep learning-based side-channel
  attacks with sensitivity analysis}.
\newblock \bibinfo{journal}{\emph{IACR Transactions on Cryptographic Hardware
  and Embedded Systems}} (\bibinfo{year}{2019}), \bibinfo{pages}{107--131}.
\newblock


\bibitem[\protect\citeauthoryear{Tomashevich, Neumeier, Kumar, Keren, and
  Polian}{Tomashevich et~al\mbox{.}}{2014}]%
        {tomashevich2014protecting}
\bibfield{author}{\bibinfo{person}{Victor Tomashevich}, \bibinfo{person}{Yaara
  Neumeier}, \bibinfo{person}{Raghavan Kumar}, \bibinfo{person}{Osnat Keren},
  {and} \bibinfo{person}{Ilia Polian}.} \bibinfo{year}{2014}\natexlab{}.
\newblock \showarticletitle{Protecting cryptographic hardware against malicious
  attacks by nonlinear robust codes}. In \bibinfo{booktitle}{\emph{2014 IEEE
  International Symposium on Defect and Fault Tolerance in VLSI and
  Nanotechnology Systems (DFT)}}. IEEE, \bibinfo{pages}{40--45}.
\newblock


\bibitem[\protect\citeauthoryear{Trichina}{Trichina}{2003}]%
        {trichina2003combinational}
\bibfield{author}{\bibinfo{person}{Elena Trichina}.}
  \bibinfo{year}{2003}\natexlab{}.
\newblock \showarticletitle{Combinational logic design for AES subbyte
  transformation on masked data}.
\newblock \bibinfo{journal}{\emph{Cryptology EPrint Archive}}
  (\bibinfo{year}{2003}).
\newblock


\bibitem[\protect\citeauthoryear{Tudor, Wang, Liu, and Elhak}{Tudor
  et~al\mbox{.}}{2011}]%
        {tudor2011mos}
\bibfield{author}{\bibinfo{person}{Bogdan Tudor}, \bibinfo{person}{Joddy Wang},
  \bibinfo{person}{Weidong Liu}, {and} \bibinfo{person}{Hany Elhak}.}
  \bibinfo{year}{2011}\natexlab{}.
\newblock \showarticletitle{MOS device aging analysis with HSPICE and
  CustomSim}.
\newblock \bibinfo{journal}{\emph{Synopsys, White Paper}}
  (\bibinfo{year}{2011}).
\newblock


\bibitem[\protect\citeauthoryear{Utyamishev and Partin-Vaisband}{Utyamishev and
  Partin-Vaisband}{2020}]%
        {utyamishev2020power}
\bibfield{author}{\bibinfo{person}{Dmitry Utyamishev} {and}
  \bibinfo{person}{Inna Partin-Vaisband}.} \bibinfo{year}{2020}\natexlab{}.
\newblock \showarticletitle{Real-Time Detection of Power Analysis Attacks by
  Machine Learning of Power Supply Variations On-Chip}.
\newblock \bibinfo{journal}{\emph{IEEE Transactions on Computer-Aided Design of
  Integrated Circuits and Systems}} \bibinfo{volume}{39}, \bibinfo{number}{1}
  (\bibinfo{year}{2020}), \bibinfo{pages}{45--55}.
\newblock
\urldef\tempurl%
\url{https://doi.org/10.1109/TCAD.2018.2883971}
\showDOI{\tempurl}


\bibitem[\protect\citeauthoryear{Vaidyanathan, Das, Sumbul, Liu, and
  Pileggi}{Vaidyanathan et~al\mbox{.}}{2014}]%
        {vaidyanathan2014building}
\bibfield{author}{\bibinfo{person}{Kaushik Vaidyanathan},
  \bibinfo{person}{Bishnu~P Das}, \bibinfo{person}{Ekin Sumbul},
  \bibinfo{person}{Renzhi Liu}, {and} \bibinfo{person}{Larry Pileggi}.}
  \bibinfo{year}{2014}\natexlab{}.
\newblock \showarticletitle{Building trusted ICs using split fabrication}. In
  \bibinfo{booktitle}{\emph{2014 IEEE international symposium on
  hardware-oriented security and trust (HOST)}}. IEEE, \bibinfo{pages}{1--6}.
\newblock


\bibitem[\protect\citeauthoryear{Vijayakumar and Kundu}{Vijayakumar and
  Kundu}{2015}]%
        {vijayakumar2015novel}
\bibfield{author}{\bibinfo{person}{Arunkumar Vijayakumar} {and}
  \bibinfo{person}{Sandip Kundu}.} \bibinfo{year}{2015}\natexlab{}.
\newblock \showarticletitle{A novel modeling attack resistant PUF design based
  on non-linear voltage transfer characteristics}. In
  \bibinfo{booktitle}{\emph{2015 Design, Automation \& Test in Europe
  Conference \& Exhibition (DATE)}}. IEEE, \bibinfo{pages}{653--658}.
\newblock


\bibitem[\protect\citeauthoryear{Vijayakumar, Patil, Prado, and
  Kundu}{Vijayakumar et~al\mbox{.}}{2016}]%
        {vijayakumar2016machine}
\bibfield{author}{\bibinfo{person}{Arunkumar Vijayakumar},
  \bibinfo{person}{Vinay~C Patil}, \bibinfo{person}{Charles~B Prado}, {and}
  \bibinfo{person}{Sandip Kundu}.} \bibinfo{year}{2016}\natexlab{}.
\newblock \showarticletitle{Machine learning resistant strong PUF: Possible or
  a pipe dream?}. In \bibinfo{booktitle}{\emph{2016 IEEE international
  symposium on hardware oriented security and trust (HOST)}}. IEEE,
  \bibinfo{pages}{19--24}.
\newblock


\bibitem[\protect\citeauthoryear{Volanis, Lu, Nimmalapudi, Antonopoulos,
  Marshall, and Makris}{Volanis et~al\mbox{.}}{2019}]%
        {volanis2019analog}
\bibfield{author}{\bibinfo{person}{Georgios Volanis}, \bibinfo{person}{Yichuan
  Lu}, \bibinfo{person}{Sai Govinda~Rao Nimmalapudi}, \bibinfo{person}{Angelos
  Antonopoulos}, \bibinfo{person}{Andrew Marshall}, {and}
  \bibinfo{person}{Yiorgos Makris}.} \bibinfo{year}{2019}\natexlab{}.
\newblock \showarticletitle{Analog performance locking through neural
  network-based biasing}. In \bibinfo{booktitle}{\emph{2019 IEEE 37th VLSI Test
  Symposium (VTS)}}. IEEE, \bibinfo{pages}{1--6}.
\newblock


\bibitem[\protect\citeauthoryear{Wang, Cachecho, Zhang, Sun, Li, Kanj, and
  Gu}{Wang et~al\mbox{.}}{2015a}]%
        {wang2015bayesian}
\bibfield{author}{\bibinfo{person}{Fa Wang}, \bibinfo{person}{Paolo Cachecho},
  \bibinfo{person}{Wangyang Zhang}, \bibinfo{person}{Shupeng Sun},
  \bibinfo{person}{Xin Li}, \bibinfo{person}{Rouwaida Kanj}, {and}
  \bibinfo{person}{Chenjie Gu}.} \bibinfo{year}{2015}\natexlab{a}.
\newblock \showarticletitle{Bayesian model fusion: large-scale performance
  modeling of analog and mixed-signal circuits by reusing early-stage data}.
\newblock \bibinfo{journal}{\emph{IEEE Transactions on Computer-Aided Design of
  Integrated Circuits and Systems}} \bibinfo{volume}{35}, \bibinfo{number}{8}
  (\bibinfo{year}{2015}), \bibinfo{pages}{1255--1268}.
\newblock


\bibitem[\protect\citeauthoryear{Wang, Forte, Tehranipoor, and Shi}{Wang
  et~al\mbox{.}}{2017b}]%
        {wang2017}
\bibfield{author}{\bibinfo{person}{Huanyu Wang}, \bibinfo{person}{Domenic
  Forte}, \bibinfo{person}{Mark~M. Tehranipoor}, {and} \bibinfo{person}{Qihang
  Shi}.} \bibinfo{year}{2017}\natexlab{b}.
\newblock \showarticletitle{Probing Attacks on Integrated Circuits: Challenges
  and Research Opportunities}.
\newblock \bibinfo{journal}{\emph{IEEE Design Test}} \bibinfo{volume}{34},
  \bibinfo{number}{5} (\bibinfo{year}{2017}), \bibinfo{pages}{63--71}.
\newblock
\urldef\tempurl%
\url{https://doi.org/10.1109/MDAT.2017.2729398}
\showDOI{\tempurl}


\bibitem[\protect\citeauthoryear{Wang, Shi, Forte, and Tehranipoor}{Wang
  et~al\mbox{.}}{2019}]%
        {Wang2019}
\bibfield{author}{\bibinfo{person}{Huanyu Wang}, \bibinfo{person}{Qihang Shi},
  \bibinfo{person}{Domenic Forte}, {and} \bibinfo{person}{Mark~M.
  Tehranipoor}.} \bibinfo{year}{2019}\natexlab{}.
\newblock \showarticletitle{Probing Assessment Framework and Evaluation of
  Antiprobing Solutions}.
\newblock \bibinfo{journal}{\emph{IEEE Trans. Very Large Scale Integr. Syst.}}
  \bibinfo{volume}{27}, \bibinfo{number}{6} (\bibinfo{date}{jun}
  \bibinfo{year}{2019}), \bibinfo{pages}{1239–1252}.
\newblock
\showISSN{1063-8210}
\urldef\tempurl%
\url{https://doi.org/10.1109/TVLSI.2019.2901449}
\showDOI{\tempurl}


\bibitem[\protect\citeauthoryear{Wang, Wang, Yang, Shen, Sun, Lee, and
  Han}{Wang et~al\mbox{.}}{2020}]%
        {wang2020gcn}
\bibfield{author}{\bibinfo{person}{Hanrui Wang}, \bibinfo{person}{Kuan Wang},
  \bibinfo{person}{Jiacheng Yang}, \bibinfo{person}{Linxiao Shen},
  \bibinfo{person}{Nan Sun}, \bibinfo{person}{Hae-Seung Lee}, {and}
  \bibinfo{person}{Song Han}.} \bibinfo{year}{2020}\natexlab{}.
\newblock \showarticletitle{GCN-RL circuit designer: Transferable transistor
  sizing with graph neural networks and reinforcement learning}. In
  \bibinfo{booktitle}{\emph{2020 57th ACM/IEEE Design Automation Conference
  (DAC)}}. IEEE, \bibinfo{pages}{1--6}.
\newblock


\bibitem[\protect\citeauthoryear{Wang, Yang, Lee, and Han}{Wang
  et~al\mbox{.}}{2018c}]%
        {wang2018learning}
\bibfield{author}{\bibinfo{person}{Hanrui Wang}, \bibinfo{person}{Jiacheng
  Yang}, \bibinfo{person}{Hae-Seung Lee}, {and} \bibinfo{person}{Song Han}.}
  \bibinfo{year}{2018}\natexlab{c}.
\newblock \showarticletitle{Learning to design circuits}.
\newblock \bibinfo{journal}{\emph{arXiv preprint arXiv:1812.02734}}
  (\bibinfo{year}{2018}).
\newblock


\bibitem[\protect\citeauthoryear{Wang, Shi, Sanchez-Sinencio, and Hu}{Wang
  et~al\mbox{.}}{2015b}]%
        {wang2015built}
\bibfield{author}{\bibinfo{person}{Jiafan Wang}, \bibinfo{person}{Congyin Shi},
  \bibinfo{person}{Edgar Sanchez-Sinencio}, {and} \bibinfo{person}{Jiang Hu}.}
  \bibinfo{year}{2015}\natexlab{b}.
\newblock \showarticletitle{Built-in self optimization for variation resilience
  of analog filters}. In \bibinfo{booktitle}{\emph{2015 IEEE Computer Society
  Annual Symposium on VLSI}}. IEEE, \bibinfo{pages}{656--661}.
\newblock


\bibitem[\protect\citeauthoryear{Wang, Chang, and Cheng}{Wang
  et~al\mbox{.}}{2009}]%
        {wang2009electronic}
\bibfield{author}{\bibinfo{person}{Laung-Terng Wang}, \bibinfo{person}{Yao-Wen
  Chang}, {and} \bibinfo{person}{Kwang-Ting~Tim Cheng}.}
  \bibinfo{year}{2009}\natexlab{}.
\newblock \bibinfo{booktitle}{\emph{Electronic design automation: synthesis,
  verification, and test}}.
\newblock \bibinfo{publisher}{Morgan Kaufmann}.
\newblock


\bibitem[\protect\citeauthoryear{Wang, Chen, Hu, Li, and Rajendran}{Wang
  et~al\mbox{.}}{2018a}]%
        {wang2018cat}
\bibfield{author}{\bibinfo{person}{Yujie Wang}, \bibinfo{person}{Pu Chen},
  \bibinfo{person}{Jiang Hu}, \bibinfo{person}{Guofeng Li}, {and}
  \bibinfo{person}{Jeyavijayan Rajendran}.} \bibinfo{year}{2018}\natexlab{a}.
\newblock \showarticletitle{The cat and mouse in split manufacturing}.
\newblock \bibinfo{journal}{\emph{IEEE Transactions on Very Large Scale
  Integration (VLSI) Systems}} \bibinfo{volume}{26}, \bibinfo{number}{5}
  (\bibinfo{year}{2018}), \bibinfo{pages}{805--817}.
\newblock


\bibitem[\protect\citeauthoryear{Wang and Franzon}{Wang and Franzon}{2018}]%
        {wang2018rfic}
\bibfield{author}{\bibinfo{person}{Yi Wang} {and} \bibinfo{person}{Paul~D
  Franzon}.} \bibinfo{year}{2018}\natexlab{}.
\newblock \showarticletitle{RFIC IP redesign and reuse through surrogate based
  machine learning method}. In \bibinfo{booktitle}{\emph{2018 IEEE MTT-S
  International Conference on Numerical Electromagnetic and Multiphysics
  Modeling and Optimization (NEMO)}}. IEEE, \bibinfo{pages}{1--4}.
\newblock


\bibitem[\protect\citeauthoryear{Wang, Chen, Patil, Jayabalan, Zhang, Chang,
  and Basu}{Wang et~al\mbox{.}}{2017a}]%
        {wang2017current}
\bibfield{author}{\bibinfo{person}{Zheng Wang}, \bibinfo{person}{Yi Chen},
  \bibinfo{person}{Aakash Patil}, \bibinfo{person}{Jayasanker Jayabalan},
  \bibinfo{person}{Xueyong Zhang}, \bibinfo{person}{Chip-Hong Chang}, {and}
  \bibinfo{person}{Arindam Basu}.} \bibinfo{year}{2017}\natexlab{a}.
\newblock \showarticletitle{Current mirror array: A novel circuit topology for
  combining physical unclonable function and machine learning}.
\newblock \bibinfo{journal}{\emph{IEEE Transactions on Circuits and Systems I:
  Regular Papers}} \bibinfo{volume}{65}, \bibinfo{number}{4}
  (\bibinfo{year}{2017}), \bibinfo{pages}{1314--1326}.
\newblock


\bibitem[\protect\citeauthoryear{Wang, Luo, and Gong}{Wang
  et~al\mbox{.}}{2018b}]%
        {wang2018application}
\bibfield{author}{\bibinfo{person}{Zhenyu Wang}, \bibinfo{person}{Xiangzhong
  Luo}, {and} \bibinfo{person}{Zheng Gong}.} \bibinfo{year}{2018}\natexlab{b}.
\newblock \showarticletitle{Application of deep learning in analog circuit
  sizing}. In \bibinfo{booktitle}{\emph{Proceedings of the 2018 2nd
  International Conference on Computer Science and Artificial Intelligence}}.
  \bibinfo{pages}{571--575}.
\newblock


\bibitem[\protect\citeauthoryear{Wolfe and Vemuri}{Wolfe and Vemuri}{2003}]%
        {wolfe2003extraction}
\bibfield{author}{\bibinfo{person}{Glenn Wolfe} {and} \bibinfo{person}{Ranga
  Vemuri}.} \bibinfo{year}{2003}\natexlab{}.
\newblock \showarticletitle{Extraction and use of neural network models in
  automated synthesis of operational amplifiers}.
\newblock \bibinfo{journal}{\emph{IEEE Transactions on Computer-Aided Design of
  Integrated Circuits and Systems}} \bibinfo{volume}{22}, \bibinfo{number}{2}
  (\bibinfo{year}{2003}), \bibinfo{pages}{198--212}.
\newblock


\bibitem[\protect\citeauthoryear{Wu, Perin, and Picek}{Wu
  et~al\mbox{.}}{2020}]%
        {related_works:automated_hyperparameter_tuning}
\bibfield{author}{\bibinfo{person}{Lichao Wu}, \bibinfo{person}{Guilherme
  Perin}, {and} \bibinfo{person}{Stjepan Picek}.}
  \bibinfo{year}{2020}\natexlab{}.
\newblock \showarticletitle{I Choose You: Automated Hyperparameter Tuning for
  Deep Learning-based Side-channel Analysis}.
\newblock \bibinfo{journal}{\emph{IACR Cryptol. ePrint Arch.}}
  \bibinfo{volume}{2020} (\bibinfo{year}{2020}), \bibinfo{pages}{1293}.
\newblock


\bibitem[\protect\citeauthoryear{Wu, Xie, and Hao}{Wu et~al\mbox{.}}{2021}]%
        {wu2021ironman}
\bibfield{author}{\bibinfo{person}{Nan Wu}, \bibinfo{person}{Yuan Xie}, {and}
  \bibinfo{person}{Cong Hao}.} \bibinfo{year}{2021}\natexlab{}.
\newblock \showarticletitle{IronMan: GNN-assisted Design Space Exploration in
  High-Level Synthesis via Reinforcement Learning}.
\newblock \bibinfo{journal}{\emph{CoRR}}  \bibinfo{volume}{abs/2102.08138}
  (\bibinfo{year}{2021}).
\newblock
\showeprint[arXiv]{2102.08138}
\urldef\tempurl%
\url{https://arxiv.org/abs/2102.08138}
\showURL{%
\tempurl}


\bibitem[\protect\citeauthoryear{Xiao, Nahiyan, and Tehranipoor}{Xiao
  et~al\mbox{.}}{2016}]%
        {xiao2016security}
\bibfield{author}{\bibinfo{person}{Kan Xiao}, \bibinfo{person}{Adib Nahiyan},
  {and} \bibinfo{person}{Mark Tehranipoor}.} \bibinfo{year}{2016}\natexlab{}.
\newblock \showarticletitle{Security rule checking in IC design}.
\newblock \bibinfo{journal}{\emph{Computer}} \bibinfo{volume}{49},
  \bibinfo{number}{8} (\bibinfo{year}{2016}), \bibinfo{pages}{54--61}.
\newblock


\bibitem[\protect\citeauthoryear{Xiao and Tehranipoor}{Xiao and
  Tehranipoor}{2013}]%
        {xiao2013bisa}
\bibfield{author}{\bibinfo{person}{Kan Xiao} {and} \bibinfo{person}{Mohammed
  Tehranipoor}.} \bibinfo{year}{2013}\natexlab{}.
\newblock \showarticletitle{BISA: Built-in self-authentication for preventing
  hardware Trojan insertion}.
\newblock \bibinfo{journal}{\emph{Proceedings of the 2013 IEEE International
  Symposium on Hardware-Oriented Security and Trust, HOST 2013}},
  \bibinfo{pages}{45--50}.
\newblock
\showISBNx{978-1-4799-0601-7}
\urldef\tempurl%
\url{https://doi.org/10.1109/HST.2013.6581564}
\showDOI{\tempurl}


\bibitem[\protect\citeauthoryear{Xie, Ren, Khailany, Sheng, Santosh, Hu, and
  Chen}{Xie et~al\mbox{.}}{2020}]%
        {xie2020powernet}
\bibfield{author}{\bibinfo{person}{Zhiyao Xie}, \bibinfo{person}{Haoxing Ren},
  \bibinfo{person}{Brucek Khailany}, \bibinfo{person}{Ye Sheng},
  \bibinfo{person}{Santosh Santosh}, \bibinfo{person}{Jiang Hu}, {and}
  \bibinfo{person}{Yiran Chen}.} \bibinfo{year}{2020}\natexlab{}.
\newblock \showarticletitle{PowerNet: Transferable dynamic IR drop estimation
  via maximum convolutional neural network}. In \bibinfo{booktitle}{\emph{2020
  25th Asia and South Pacific Design Automation Conference (ASP-DAC)}}. IEEE,
  \bibinfo{pages}{13--18}.
\newblock


\bibitem[\protect\citeauthoryear{Xu, Hu, Leskovec, and Jegelka}{Xu
  et~al\mbox{.}}{2018}]%
        {xu2018powerful}
\bibfield{author}{\bibinfo{person}{Keyulu Xu}, \bibinfo{person}{Weihua Hu},
  \bibinfo{person}{Jure Leskovec}, {and} \bibinfo{person}{Stefanie Jegelka}.}
  \bibinfo{year}{2018}\natexlab{}.
\newblock \showarticletitle{How powerful are graph neural networks?}
\newblock \bibinfo{journal}{\emph{arXiv preprint arXiv:1810.00826}}
  (\bibinfo{year}{2018}).
\newblock


\bibitem[\protect\citeauthoryear{Xu, Shakya, Tehranipoor, and Forte}{Xu
  et~al\mbox{.}}{2017}]%
        {xu2017novel}
\bibfield{author}{\bibinfo{person}{Xiaolin Xu}, \bibinfo{person}{Bicky Shakya},
  \bibinfo{person}{Mark~M Tehranipoor}, {and} \bibinfo{person}{Domenic Forte}.}
  \bibinfo{year}{2017}\natexlab{}.
\newblock \showarticletitle{Novel bypass attack and BDD-based tradeoff analysis
  against all known logic locking attacks}. In
  \bibinfo{booktitle}{\emph{International conference on cryptographic hardware
  and embedded systems}}. Springer, \bibinfo{pages}{189--210}.
\newblock


\bibitem[\protect\citeauthoryear{Yasaei, Yu, Naeini, and Faruque}{Yasaei
  et~al\mbox{.}}{2021a}]%
        {rozhin_2021}
\bibfield{author}{\bibinfo{person}{Rozhin Yasaei}, \bibinfo{person}{Shih{-}Yuan
  Yu}, \bibinfo{person}{Emad~Kasaeyan Naeini}, {and} \bibinfo{person}{Mohammad
  Abdullah~Al Faruque}.} \bibinfo{year}{2021}\natexlab{a}.
\newblock \showarticletitle{{GNN4IP:} Graph Neural Network for Hardware
  Intellectual Property Piracy Detection}.
\newblock \bibinfo{journal}{\emph{CoRR}}  \bibinfo{volume}{abs/2107.09130}
  (\bibinfo{year}{2021}).
\newblock
\showeprint[arXiv]{2107.09130}
\urldef\tempurl%
\url{https://arxiv.org/abs/2107.09130}
\showURL{%
\tempurl}


\bibitem[\protect\citeauthoryear{Yasaei, Yu, Naeini, and Faruque}{Yasaei
  et~al\mbox{.}}{2021b}]%
        {Yasaei2021-gnn-piracy}
\bibfield{author}{\bibinfo{person}{Rozhin Yasaei}, \bibinfo{person}{Shih-Yuan
  Yu}, \bibinfo{person}{Emad~Kasaeyan Naeini}, {and} \bibinfo{person}{Mohammad
  Abdullah~Al Faruque}.} \bibinfo{year}{2021}\natexlab{b}.
\newblock \showarticletitle{{GNN4IP: Graph Neural Network for Hardware
  Intellectual Property Piracy Detection}}. In \bibinfo{booktitle}{\emph{{2021
  58th ACM/IEEE Design Automation Conference (DAC)}}}.
  \bibinfo{publisher}{ieeexplore.ieee.org}, \bibinfo{pages}{217--222}.
\newblock
\showISSN{0738-100X}
\urldef\tempurl%
\url{https://doi.org/10.1109/DAC18074.2021.9586150}
\showDOI{\tempurl}


\bibitem[\protect\citeauthoryear{Yasin, Mazumdar, Sinanoglu, and
  Rajendran}{Yasin et~al\mbox{.}}{2017}]%
        {yasin2017security}
\bibfield{author}{\bibinfo{person}{Muhammad Yasin}, \bibinfo{person}{Bodhisatwa
  Mazumdar}, \bibinfo{person}{Ozgur Sinanoglu}, {and}
  \bibinfo{person}{Jeyavijayan Rajendran}.} \bibinfo{year}{2017}\natexlab{}.
\newblock \showarticletitle{Security analysis of anti-sat}. In
  \bibinfo{booktitle}{\emph{2017 22nd Asia and South Pacific Design Automation
  Conference (ASP-DAC)}}. IEEE, \bibinfo{pages}{342--347}.
\newblock


\bibitem[\protect\citeauthoryear{Yasin, Rajendran, and Sinanoglu}{Yasin
  et~al\mbox{.}}{2020}]%
        {Yasin2020}
\bibfield{author}{\bibinfo{person}{Muhammad Yasin},
  \bibinfo{person}{Jeyavijayan~(JV) Rajendran}, {and} \bibinfo{person}{Ozgur
  Sinanoglu}.} \bibinfo{year}{2020}\natexlab{}.
\newblock \bibinfo{booktitle}{\emph{A Brief History of Logic Locking}}.
\newblock \bibinfo{publisher}{Springer International Publishing},
  \bibinfo{address}{Cham}, \bibinfo{pages}{17--31}.
\newblock
\showISBNx{978-3-030-15334-2}
\urldef\tempurl%
\url{https://doi.org/10.1007/978-3-030-15334-2_2}
\showDOI{\tempurl}


\bibitem[\protect\citeauthoryear{Yasin and Sinanoglu}{Yasin and
  Sinanoglu}{2015}]%
        {yasin2015transforming}
\bibfield{author}{\bibinfo{person}{Muhammad Yasin} {and} \bibinfo{person}{Ozgur
  Sinanoglu}.} \bibinfo{year}{2015}\natexlab{}.
\newblock \showarticletitle{Transforming between logic locking and IC
  camouflaging}. In \bibinfo{booktitle}{\emph{2015 10th International Design \&
  Test Symposium (IDT)}}. IEEE, \bibinfo{pages}{1--4}.
\newblock


\bibitem[\protect\citeauthoryear{Ye, Hu, and Li}{Ye et~al\mbox{.}}{2016}]%
        {ye2016poster}
\bibfield{author}{\bibinfo{person}{Jing Ye}, \bibinfo{person}{Yu Hu}, {and}
  \bibinfo{person}{Xiaowei Li}.} \bibinfo{year}{2016}\natexlab{}.
\newblock \showarticletitle{POSTER: Attack on non-linear physical unclonable
  function}. In \bibinfo{booktitle}{\emph{Proceedings of the 2016 ACM SIGSAC
  Conference on Computer and Communications Security}}.
  \bibinfo{pages}{1751--1753}.
\newblock


\bibitem[\protect\citeauthoryear{Yu, Xiao, and De~Micheli}{Yu
  et~al\mbox{.}}{2018}]%
        {yu2018developing}
\bibfield{author}{\bibinfo{person}{Cunxi Yu}, \bibinfo{person}{Houping Xiao},
  {and} \bibinfo{person}{Giovanni De~Micheli}.}
  \bibinfo{year}{2018}\natexlab{}.
\newblock \showarticletitle{Developing synthesis flows without human
  knowledge}. In \bibinfo{booktitle}{\emph{Proceedings of the 55th Annual
  Design Automation Conference}}. \bibinfo{pages}{1--6}.
\newblock


\bibitem[\protect\citeauthoryear{Yu, Swaminathan, Ji, and White}{Yu
  et~al\mbox{.}}{2017}]%
        {yu2017method}
\bibfield{author}{\bibinfo{person}{Huan Yu}, \bibinfo{person}{Madhavan
  Swaminathan}, \bibinfo{person}{Chuanyi Ji}, {and} \bibinfo{person}{David
  White}.} \bibinfo{year}{2017}\natexlab{}.
\newblock \showarticletitle{A method for creating behavioral models of
  oscillators using augmented neural networks}. In
  \bibinfo{booktitle}{\emph{2017 IEEE 26th Conference on Electrical Performance
  of Electronic Packaging and Systems (EPEPS)}}. IEEE, \bibinfo{pages}{1--3}.
\newblock


\bibitem[\protect\citeauthoryear{Yu, Yasaei, Zhou, Nguyen, and Faruque}{Yu
  et~al\mbox{.}}{2021}]%
        {yu2021hw2vec}
\bibfield{author}{\bibinfo{person}{Shih-Yuan Yu}, \bibinfo{person}{Rozhin
  Yasaei}, \bibinfo{person}{Qingrong Zhou}, \bibinfo{person}{Tommy Nguyen},
  {and} \bibinfo{person}{Mohammad Abdullah~Al Faruque}.}
  \bibinfo{year}{2021}\natexlab{}.
\newblock \bibinfo{title}{HW2VEC: A Graph Learning Tool for Automating Hardware
  Security}.
\newblock
\newblock
\showeprint[arxiv]{2107.12328}~[cs.CR]


\bibitem[\protect\citeauthoryear{Zargari, Ashrafiamiri, Seo, Dinakarrao, Fouda,
  and Kurdahi}{Zargari et~al\mbox{.}}{2021}]%
        {zargari2021captive}
\bibfield{author}{\bibinfo{person}{Amir Hosein~Afandizadeh Zargari},
  \bibinfo{person}{Marzieh Ashrafiamiri}, \bibinfo{person}{Minjun Seo},
  \bibinfo{person}{Sai Manoj~Pudukotai Dinakarrao},
  \bibinfo{person}{Mohammed~E. Fouda}, {and} \bibinfo{person}{Fadi~J.
  Kurdahi}.} \bibinfo{year}{2021}\natexlab{}.
\newblock \showarticletitle{{CAPTIVE:} Constrained Adversarial Perturbations to
  Thwart {IC} Reverse Engineering}.
\newblock \bibinfo{journal}{\emph{CoRR}}  \bibinfo{volume}{abs/2110.11459}
  (\bibinfo{year}{2021}).
\newblock
\showeprint[arXiv]{2110.11459}
\urldef\tempurl%
\url{https://arxiv.org/abs/2110.11459}
\showURL{%
\tempurl}


\bibitem[\protect\citeauthoryear{Zebulum, Pacheco, and Vellasco}{Zebulum
  et~al\mbox{.}}{2018}]%
        {zebulum2018evolutionary}
\bibfield{author}{\bibinfo{person}{Ricardo~Salem Zebulum},
  \bibinfo{person}{Marco~Aur{\'e}lio Pacheco}, {and} \bibinfo{person}{Marley
  Maria~Be Vellasco}.} \bibinfo{year}{2018}\natexlab{}.
\newblock \bibinfo{booktitle}{\emph{Evolutionary electronics: automatic design
  of electronic circuits and systems by genetic algorithms}}.
\newblock \bibinfo{publisher}{CRC press}.
\newblock


\bibitem[\protect\citeauthoryear{Zeng, Davoodi, and Topaloglu}{Zeng
  et~al\mbox{.}}{2021}]%
        {zeng2021obfusx}
\bibfield{author}{\bibinfo{person}{Wei Zeng}, \bibinfo{person}{Azadeh Davoodi},
  {and} \bibinfo{person}{Rasit~Onur Topaloglu}.}
  \bibinfo{year}{2021}\natexlab{}.
\newblock \showarticletitle{ObfusX: routing obfuscation with explanatory
  analysis of a machine learning attack}. In \bibinfo{booktitle}{\emph{2021
  26th Asia and South Pacific Design Automation Conference (ASP-DAC)}}. IEEE,
  \bibinfo{pages}{548--554}.
\newblock


\bibitem[\protect\citeauthoryear{Zennaro, Servadei, Devarajegowda, and
  Ecker}{Zennaro et~al\mbox{.}}{2018}]%
        {Zennaro2018}
\bibfield{author}{\bibinfo{person}{Elena Zennaro}, \bibinfo{person}{Lorenzo
  Servadei}, \bibinfo{person}{Keerthikumara Devarajegowda}, {and}
  \bibinfo{person}{Wolfgang Ecker}.} \bibinfo{year}{2018}\natexlab{}.
\newblock \showarticletitle{A machine learning approach for area prediction of
  hardware designs from abstract specifications}.
\newblock \bibinfo{journal}{\emph{Proceedings - 21st Euromicro Conference on
  Digital System Design, DSD 2018}}, \bibinfo{pages}{413--420}.
\newblock
\showISBNx{9781538673768}
\urldef\tempurl%
\url{https://doi.org/10.1109/DSD.2018.00076}
\showDOI{\tempurl}


\bibitem[\protect\citeauthoryear{Zhang, Gupta, and Devabhaktuni}{Zhang
  et~al\mbox{.}}{2003}]%
        {zhang2003artificial}
\bibfield{author}{\bibinfo{person}{Qi-Jun Zhang}, \bibinfo{person}{Kuldip~C
  Gupta}, {and} \bibinfo{person}{Vijay~K Devabhaktuni}.}
  \bibinfo{year}{2003}\natexlab{}.
\newblock \showarticletitle{Artificial neural networks for RF and microwave
  design-from theory to practice}.
\newblock \bibinfo{journal}{\emph{IEEE transactions on microwave theory and
  techniques}} \bibinfo{volume}{51}, \bibinfo{number}{4}
  (\bibinfo{year}{2003}), \bibinfo{pages}{1339--1350}.
\newblock


\bibitem[\protect\citeauthoryear{Zhang and Tehranipoor}{Zhang and
  Tehranipoor}{2014}]%
        {zhang2014}
\bibfield{author}{\bibinfo{person}{Xuehui Zhang} {and}
  \bibinfo{person}{Mohammad Tehranipoor}.} \bibinfo{year}{2014}\natexlab{}.
\newblock \showarticletitle{Design of On-Chip Lightweight Sensors for Effective
  Detection of Recycled ICs}.
\newblock \bibinfo{journal}{\emph{IEEE Transactions on Very Large Scale
  Integration (VLSI) Systems}} \bibinfo{volume}{22}, \bibinfo{number}{5}
  (\bibinfo{year}{2014}), \bibinfo{pages}{1016--1029}.
\newblock
\urldef\tempurl%
\url{https://doi.org/10.1109/TVLSI.2013.2264063}
\showDOI{\tempurl}


\bibitem[\protect\citeauthoryear{Zhang, Dofe, and Yu}{Zhang
  et~al\mbox{.}}{2020}]%
        {zhang2020pdnsca}
\bibfield{author}{\bibinfo{person}{Zhiming Zhang}, \bibinfo{person}{Jaya Dofe},
  {and} \bibinfo{person}{Qiaoyan Yu}.} \bibinfo{year}{2020}\natexlab{}.
\newblock \showarticletitle{Improving power analysis attack resistance using
  intrinsic noise in 3D ICs}.
\newblock \bibinfo{journal}{\emph{Integration}}  \bibinfo{volume}{73}
  (\bibinfo{year}{2020}), \bibinfo{pages}{30--42}.
\newblock
\showISSN{0167-9260}
\urldef\tempurl%
\url{https://doi.org/10.1016/j.vlsi.2020.02.007}
\showDOI{\tempurl}


\bibitem[\protect\citeauthoryear{Zhao, Liu, Zhang, Liu, Niu, and Hu}{Zhao
  et~al\mbox{.}}{2018}]%
        {zhao2018novel}
\bibfield{author}{\bibinfo{person}{Guangquan Zhao}, \bibinfo{person}{Xiaoyong
  Liu}, \bibinfo{person}{Bin Zhang}, \bibinfo{person}{Yuefeng Liu},
  \bibinfo{person}{Guangxing Niu}, {and} \bibinfo{person}{Cong Hu}.}
  \bibinfo{year}{2018}\natexlab{}.
\newblock \showarticletitle{A novel approach for analog circuit fault diagnosis
  based on deep belief network}.
\newblock \bibinfo{journal}{\emph{Measurement}}  \bibinfo{volume}{121}
  (\bibinfo{year}{2018}), \bibinfo{pages}{170--178}.
\newblock


\bibitem[\protect\citeauthoryear{Zhao and Zhang}{Zhao and Zhang}{2020}]%
        {zhao2020deep}
\bibfield{author}{\bibinfo{person}{Zhenxin Zhao} {and} \bibinfo{person}{Lihong
  Zhang}.} \bibinfo{year}{2020}\natexlab{}.
\newblock \showarticletitle{Deep reinforcement learning for analog circuit
  sizing}. In \bibinfo{booktitle}{\emph{2020 IEEE International Symposium on
  Circuits and Systems (ISCAS)}}. IEEE, \bibinfo{pages}{1--5}.
\newblock


\bibitem[\protect\citeauthoryear{Zhou, Jiang, and Kong}{Zhou
  et~al\mbox{.}}{2017}]%
        {zhou2017}
\bibfield{author}{\bibinfo{person}{Hai Zhou}, \bibinfo{person}{Ruifeng Jiang},
  {and} \bibinfo{person}{Shuyu Kong}.} \bibinfo{year}{2017}\natexlab{}.
\newblock \showarticletitle{CycSAT: SAT-based attack on cyclic logic
  encryptions}. In \bibinfo{booktitle}{\emph{2017 IEEE/ACM International
  Conference on Computer-Aided Design (ICCAD)}}. \bibinfo{pages}{49--56}.
\newblock
\urldef\tempurl%
\url{https://doi.org/10.1109/ICCAD.2017.8203759}
\showDOI{\tempurl}


\bibitem[\protect\citeauthoryear{Zhou and Wong}{Zhou and Wong}{1998}]%
        {zhou1998crosstalk}
\bibfield{author}{\bibinfo{person}{H. Zhou} {and} \bibinfo{person}{D.F. Wong}.}
  \bibinfo{year}{1998}\natexlab{}.
\newblock \showarticletitle{Global routing with crosstalk constraints}. In
  \bibinfo{booktitle}{\emph{Proceedings 1998 Design and Automation Conference.
  35th DAC. (Cat. No.98CH36175)}}. \bibinfo{pages}{374--377}.
\newblock
\urldef\tempurl%
\url{https://doi.org/10.1145/277044.277147}
\showDOI{\tempurl}


\bibitem[\protect\citeauthoryear{Zhu, Guo, Jin, and Zhang}{Zhu
  et~al\mbox{.}}{2020}]%
        {zhu2020}
\bibfield{author}{\bibinfo{person}{Huifeng Zhu}, \bibinfo{person}{Xiaolong
  Guo}, \bibinfo{person}{Yier Jin}, {and} \bibinfo{person}{Xuan Zhang}.}
  \bibinfo{year}{2020}\natexlab{}.
\newblock \showarticletitle{PowerScout: A Security-Oriented Power Delivery
  Network Modeling Framework for Cross-Domain Side-Channel Analysis}. In
  \bibinfo{booktitle}{\emph{2020 Asian Hardware Oriented Security and Trust
  Symposium (AsianHOST)}}. \bibinfo{pages}{1--6}.
\newblock
\urldef\tempurl%
\url{https://doi.org/10.1109/AsianHOST51057.2020.9358263}
\showDOI{\tempurl}


\bibitem[\protect\citeauthoryear{Zhu, Liu, Lin, Xu, Li, Tang, Sun, and Pan}{Zhu
  et~al\mbox{.}}{2019}]%
        {zhu2019geniusroute}
\bibfield{author}{\bibinfo{person}{Keren Zhu}, \bibinfo{person}{Mingjie Liu},
  \bibinfo{person}{Yibo Lin}, \bibinfo{person}{Biying Xu},
  \bibinfo{person}{Shaolan Li}, \bibinfo{person}{Xiyuan Tang},
  \bibinfo{person}{Nan Sun}, {and} \bibinfo{person}{David~Z Pan}.}
  \bibinfo{year}{2019}\natexlab{}.
\newblock \showarticletitle{Geniusroute: A new analog routing paradigm using
  generative neural network guidance}. In \bibinfo{booktitle}{\emph{2019
  IEEE/ACM International Conference on Computer-Aided Design (ICCAD)}}. IEEE,
  \bibinfo{pages}{1--8}.
\newblock


\bibitem[\protect\citeauthoryear{Zohner, Stöttinger, Huss, and Stein}{Zohner
  et~al\mbox{.}}{2012}]%
        {zohner2012}
\bibfield{author}{\bibinfo{person}{Michael Zohner}, \bibinfo{person}{Marc
  Stöttinger}, \bibinfo{person}{Sorin~A. Huss}, {and} \bibinfo{person}{Oliver
  Stein}.} \bibinfo{year}{2012}\natexlab{}.
\newblock \showarticletitle{An adaptable, modular, and autonomous side-channel
  vulnerability evaluator}. In \bibinfo{booktitle}{\emph{2012 IEEE
  International Symposium on Hardware-Oriented Security and Trust}}.
  \bibinfo{pages}{43--48}.
\newblock
\urldef\tempurl%
\url{https://doi.org/10.1109/HST.2012.6224317}
\showDOI{\tempurl}


\end{thebibliography}
\end{document}